\documentclass[12pt]{article}

\iffalse
\else

\def\true{{\sf true}}

\def\color[#1]#2{}
\newtheorem{theorem}{Theorem}
\newtheorem{lemma}{Lemma}
\newtheorem{definition}{Definition}
\def\proof{\noindent {\sl Proof.\ \ }}
\def\qed{\hfill{\boxit{}}
  \ifdim\lastskip<\medskipamount \removelastskip\penalty55\medskip\fi}
\long\def\boxit#1{\vbox{\hrule\hbox{\vrule\kern3pt
                  \vbox{\kern3pt#1\kern3pt}\kern3pt\vrule}\hrule}}
\def\l{\langle}
\def\r{\rangle}
\long\def\nop#1{}
\fi

\expandafter\ifx\csname ttytty\endcsname \relax\else\def\footnote#1#2{#2~{\em [footnote: #1}]}\fi

\def\rev{\mathrm{rev}}			
\def\nat{\mathrm{nat}}			
\def\dow{\mathrm{dow}}			
\def\lex{\mathrm{lex}}			
\def\unc{\mathrm{unc}}			

\def\minidx{\mathrm{minidx}}		
\def\maxidx{\mathrm{maxidx}}		
\def\diff{\mathrm{diff}}		
\def\strength{\mathrm{strength}}	

\def\outside{\mbox{\it outside}}
\def\meow{\mbox{\it meow}}
\def\cat{\mbox{\it cat}}
\def\equal{=\!\!}
\def\lequal{\leq\!\!}
\def\greater{>\!\!}

\def\c{\begin{center}}
\def\cc{\end{center}}

\newenvironment{hfigure}{\refstepcounter{figure}\begin{center}}{\end{center}}
\def\hcaption#1{\par Figure \thefigure: #1}

\def\draft{\begingroup\em [draft]}
\def\enddraft{[/draft]\endgroup}
\newif\draft
\let\enddraft=\fi

\title{Natural revision is contingently-conditionalized revision}
\author{Paolo Liberatore\thanks{
DIAG, Sapienza University of Rome.
Via Ariosto 25, 00185 Rome, Italy.
Email: {\tt liberato@diag.uniroma1.it}
}}
\date{}

\begin{document}

\maketitle

\draft

la linea precedente era:

- la revisione naturale funziona o meno a seconda del contesto della revisione;
  quindi serve che tratti anche i contesti; si pone il problema di dove mettere
  i modelli indifferenti al condizionale -P, se al livello di quelli che
  verificano (PA) o di quelli che lo falsificano (P-A)

- ci sono almeno tre motivi per metterli con PA

  * chan-boot-20 fanno cosi' per tenere insieme i modelli di P->A

    ma P->A e' molto diverso da P>A; anzi, lo studio dei condizionali esiste
    proprio perche' sono diversi, e in particolare e' centrato sul fatto che al
    contrario di P->A, tratta diversamente i modelli di PA e di -P

  * systemz mette sempre i modelli al livello piu' basso possibile; non e'
    chiaro il perche'; in generale, sono sistemi orientati alla mutua
    consistenza, che nella revisione non e' cosi' importante

  * recalcitrance, la consistenza massimale; ma e' un obiettivo centrale di
    systemz e simili, non di revisione

  nessuno dei tre sembra avere una valenza generale nel caso della revisione

- guardando bene i meccanismi attuali di revisione si vede che di fatto
  adottano un principio di ritenere i modelli massimalmente plausibili finche'
  non arriva una revisione che li contraddice; questo lo chiamo pricipio di
  ingenuita' (naivety): credo a tutto finche' qualcuno non mi dice che non e'
  cosi'

- la revisione naturale non cambia aggiungendo il contesto attuale, almeno
  sull'esempio; questo suggerisce che la revisione stessa aggiunge il contesto
  attuale

- per verificarlo, creiamo una revisione basata sugli stessi principi (incluso
  quello di ingenuita') ma che non e' condizionale; in questo modo si puo'
  verificare che effettivamente la revisione naturale ha aggiunto il contesto
  attuale, e si vede anche qual e' di preciso

\

la linea attuale la modifica:

- le revisioni naturali e incontingente vengono introdotte nella versione che
  segue ingenuita' senza troppe motivazioni

- solo dopo si analizzano brevemente le motivazioni precedenti,
  inclusi systemz e recalcitrance

- la revisione line-down viene introdotta solo dopo come alternativa

\separator

\

FARE:

fatto, dirlo:

first/second choice -> natural/line-down

line-down non ha poi cosi' tanta importanza ora, viene discussa brevemente e
solo dopo

dato che recalcitrance.tex la confrontava con natural, adesso e' spostata in
appendice, e contiene solo i risultati tecnici

\

FARE: notare dappertutto che il principio di ingenuita' non dice di mettere i
modelli al livello piu' basso possibile; dice che se non sono coinvolti da una
revisione, se una revisione non li riguarda, mantengono il loro livello;
all'inizio sono in alto, ma se una revisione li abbassa rimangono a quel
livello in risposta a una successiva revisione che non li riguarda; non
risalgono solo perche' la revisione non li riguarda; l'ingenuita' non dice che
credo tutto il piu' possibile, dice che credo nei limiti in cui non ho
informazioni contrarie

\

FARE: verificare se il principio di rilevanza e' necessario per ottenere la
revisione lessicografica a partire dai principi

\

FARE: fatte le seguenti formalizzazioni, scriverlo nella risposta:

- definizioni formali di natural e uncontingent (gia' c'erano)

- dimostrazione che implementano i cambiamenti minimi

- dimostrazione di cosa fanno sugli esempi

- definizioni formali di confronto per cambiamento e ingenuita'

- dimostrazione che natural implementa un cambiamento minimo

- dimostrazione che natural e' piu' naive di line-down

- ...

FARE: rivedere le dimostrazioni in appendice

FARE: capisco che e' inusuale avere anche le definizioni tecniche e gli
enunciati dei teoremi in appendice; questo non lo faccio per gli articoli che
riportano principalmente risultati tecnici (citare qualche esempio) dove spesso
cerco di inserire le dimostrazioni nel testo stesso per poterle commentare; in
questo articolo invece i risultati tecnici sono solo di ausilio, il tema
principale e' quello della ricerca dei principi basilari delle revisioni; i
risultati tecnici starebbero nel mezzo; basta che ci siano, basta una nota a
fondo pagina che dica dove e' dimostrata ogni cosa che viene detta nel testo

FARE: il motivo per cui non ci sono postulati che formalizzano i principi e'
che i postulati sono implementazioni specifiche dei principi; ci sono tanti
modi di interpretare il minimo cambiamento, come si vede dal numero di sistemi
di revisione basati su cambiamento minimo; in ognuno lo stesso principio ha una
formalizzazione diversa; invece qui si cercano proprio i principi di base, per
di cui un postulato sarebbe solo un confinamento a un caso specifico

FARE: fra tutti i cambiamenti minimi dell'ordine, natural produce esattamente
quello che colloca i modelli non coinvolti nella revisione piu' in alto
possibile; sono i modelli di -P; questo si dovrebbe poter dimostrare con un
teorema

\

FARE: il principio di indifferenza non e' applicato solo a un ordine iniziale
che riflette la totale ignoranza, ma a qualsiasi coppia di situazioni di cui
non si ha informazioni su come si confrontano

dato che il preordine e' connesso, due modelli non possono mai essere
inconfrontabili; in questo caso I<J sarebbe sicuramente sbagliato, mentre I==J
e' il male minore: in mancanza di informazioni che dicano il contrario, si
assume che i due scenari abbiano la stessa plausibilita'; e' una assunzione
forte, ma e' il massimo che si puo' fare quando non e' ammessa
l'inconfrontabilita'; e' il principio di indifferenza

si applica anche all'ordinamento iniziale, perche' anche li' ignoranza si
traduce in ugualglianza, ma non solo; ma non si applica solo a quello, ne' e'
detto che si applichi a tutti i modelli dello stato iniziale ma solo ad alcuni

dato che il principio di indifferenza e' implicito nell'essere l'ordine
connesso ora non viene piu' esplicitato, tranne dove serve davvero, ossia nella
sezione lexicographic

\

FARE: i postulati sono un inchiodamento di un principio a una realizzazione
specifica; per esempio, non esiste un postulato di "minimal change", che e' una
linea guida che si applica a molte realizzazioni diverse

questo ruolo ce l'ha ora casomai l'enunciato del teorema formale; casomai
mettere delle proprieta', che poi vengono o meno rispettate

\

FARE: la riduzione unc->lex non e' menzionata prima dell'appendice; forse non
e' nemmeno necessaria

\

FARE: risultati in appendice non citati nel testo, verificare se si puo'

down-preserve
	line-down revision preserva i condizionali

	conditional preservation viene introdotto quando gia' si parla di
	naivety, e a questo punto line-down revision e' passata

	casomai andrebbe citato in conditional.tex, dove viene detto che sia
	natural che line-down sono minimi: la nota a fondo pagina che rimanda
	ai teoremi natural-minimal e down-minimal; in questo caso va aggiunto
	anche natural-preserve; ma a questo punto il preservamento condizionale
	ancora non e' stato introdotto

	si potrebbe anche citare in naivety.tex, dove si rimanda a
	natural-preserve; ma questa sezione e' sulla revisions naturale, non si
	parla piu' della revisione line-down; una dei punti guida nel
	modificare l'articolo e' ridurre il ruolo di questa revisione

natural-not-unique
	natural non e' massimalmente naive se si rilassa minimal change

	ridondante, dato che dopo lexicographic-more-naive-natural dimostra che
	lexicographic e' piu' naive e natural-strictly-lesschange-lexicographic
	dimostra indirettamente che lexicographic non e' un cambiamento minimo

uncontingent-lexicographic
	riduzione uncontingent -> lexicographic

	non e' davvero necessaria

\

FARE: (finire) l'ingenuita' e' credere in tutto quello che non conosco; quindi
va misurata rispetto a un insieme di situazioni di ignoranza; nello specifico,
si parla dei modelli di -P, quelli su cui non ho informazione

\

FARE: introduzione

	- EVIDENZIARE:
		- non e' di interesse cercare un nuovo modo
		  piuttosto, capire se c'e' un modo noto che si applica
		- primo = una implementazione di principi validi in questo caso
	  	- altre implementazioni sono possibili
		  esempio numerico, risultato diverso (quello del revisore)
		  weights: in quali casi si applicano numeri e quali no
		  qui: confronti non numerici sembrano validi, forse migliori
	- EVIDENZIARE:
		non si dice che un modo e' l'unica implementazione di principi
		basta che sia una implementazione
		e che sia una implementazione corretta nell'esempio
		lo e' perche' non e' numerica
		vedere weights

\

FARE:

condizionale forte P>>A, che non e' soddisfatto solo da min(PA) < P-A,
ma richiede PA > P-A
verificare se e' uguale a T>P->A, forse no
verificare

\

FARE: rilettura:

- indifference ora e' considerato una conseguenza di minimal change, perche' lo
  e' nel caso dei preordini connessi: dato che natural mantiene l'uguaglianza,
  e che questa viene da ignoranza, di fatto identifica le due cose

- inserire conditional preservation dove puo' servire; non come principio ma
  come conseguenza di minimal change; puo' darsi che non serva, e che gia' sia
  dove serve

\

FARE: uniformare le figure che mostrano il prima e il dopo delle revisioni:
tabular con nat(P>A) in mezzo, e sotto la versione di tutto in un singolo
ascii-art, sempre con nat(P>A) in mezzo

\enddraft

\begin{abstract}

Natural revision seems so natural: it changes beliefs as little as possible to
incorporate new information. Yet, some counterexamples show it wrong. It is so
conservative that it never fully believes. It only believes in the current
conditions. This is right in some cases and wrong in others. Which is which?
The answer requires extending natural revision from simple formulae expressing
universal truths (something holds) to conditionals expressing conditional truth
(something holds in certain conditions). The extension is based on the basic
principles natural revision follows, identified as minimal change and naivety:
change mind as little as possible; believe what not contradicted. The extension
says that natural revision restricts changes to the current conditions. A
comparison with an unrestricting revision shows what exactly the current
conditions are. It is not what currently considered true if it contradicts the
new information. It includes something more and more unlikely until the new
information is at least possible.

\end{abstract}


\noindent {\bf Declaration of interests:} The author declares that he has no
known competing financial interests or personal relationships that could have
appeared to influence the work reported in this paper.  


\section{Introduction}
\label{introduction}

\draft

\begin{itemize}

\item il controesempio centrale del gatto scappato: e' fuori dalla stazione, si
sente miagolare, e' nella stazione

\item la revisione naturale si basa sul principio di minimo cambiamento, che
appare naturale; eppure, nell'esempio la terza revisione cancella la conoscenza
di aver sentito miagolare, che era un fatto sicuro

\item variante del controesempio in invece di sentirlo per certo e' solo stato
visto qualcosa che potrebbe essere un gatto; in questo caso la revisione
naturale funziona

\item entrambi i casi si formalizzano x,y,-x, ma nel primo la y e' una
affermazione assoluta, nel secondo e' valida nel contesto attuale; il risultato
deve essere diverso nei due casi, quindi nessuna revisione puo' essere corretta
in entrambi se partono solo dalla sequenza x,y,-x

\item la revisione naturale e' corretta nel primo caso, o in generale quando si
crede y solo nelle condizioni attuali e non in qualsiasi condizione

\item e' una revisione condizionale: outside>cat, ossia cat e' creduto solo
quando outside e' ritenuto vero

\item la revisione naturale e' stata estesa al caso condizionale in vari modi;
non si tratta di capire quale sia quello giusto, ma quando queste sono giuste;
in quali casi ognuna e' giusta e in quali e' sbagliata; occorre quindi capire
da dove vengono, quali principi le generano

\item la revisione naturale su condizionale si comporta nello stesso modo nelle
due sequenze, perche' aggiunge la condizione attuale outside alla condizione
della revisione; quindi aggiunge outside sia a true>meow che a outside>cat,
rendendole uguali

\item viene quindi definita una revisione che non le renda uguali, una
revisione non contingente

\item questa permette anche di capire quale e' il contesto contingente che la
revisione naturale aggiunge

\item le due revisioni differiscono in questo, ma sono altrimenti basate sugli
stessi principi; in particolare, entrambi credono possibile l'ignoto; questo e'
il principio di ingenuita'

\item riassunto individuale delle sezioni

\end{itemize}

\enddraft

Bruno had a cat.

Bruno had his cat in its cage, hysterically bouncing up and down while he was
waiting for his train, back from the vet. How to calm it down? Nothing had
proven better than scratching the back of its neck. Only takes a moment for
opening the cage and all would be fine again.

Only takes a moment for the cat to explode out of the cage and propel itself
through the platform and disappear down the stairs.

Bruno no longer had a cat.

Out in the dilapidated neighborhood surrounding the station, carrying his empty
cage and calling its cat, Bruno heard meowing from under a rusted pickup.

Not a neighborhood for cats. Maybe opening the cage was not the best move, but
eventually all will be fixed.

His phone ringing stopped him from crawling under the pickup. A feminine voice
asked for the owner of a missing cat. The station employee he implored for help
before venturing in the streets has good news. The cat was lured by a tuna fish
can in one of the station rooms.

The cat was missing.

Bruno heard meowing.

The cat was no longer missing.

Did Bruno hear meowing? The sound was clear. Not his cat, maybe not even the
sound of a cat, but he heard meowing.

From the station and down the street, quite a few things changed in a short
time. The cat out of the station, the sound of meowing, the cat no longer out
of the station. Same cat, same station. What changed was what Bruno believed:
he believed the cat was out of the station, he believed it was not.

How well do belief change formalisms fare with Bruno's story?

Quite naturally, beliefs do not change until disproven. This is natural belief
revision: beliefs stay until contradicted. The cat is outside, meowing is
heard, the cat is not outside:

\

\begin{tabular}{ll}
revision 1:	&	\outside		\\
revision 2:	&	\meow			\\
revision 3:	&	not \outside
\end{tabular}

\

The expected result is that the cat is not outside after all, but meowing was
heard anyway. Because it was, even more surely than the cat's whereabouts.

Quite unnaturally, natural revision denies that.

The result of outside, meow, not outside is that the cat is not outside.
Nothing else. No sound was heard.

Why is this? Why such a gross mistake, even in a very simple case? Why does
natural revision have such unnatural behavior while starting from the very
natural principle of minimally changing beliefs?

Given that natural revision is wrong, what is the correct revision policy?
Maybe lexicographic revision? Maybe restrained revision? What about severe
antiwithdrawal or full meet revision?

In an alternative universe, Bruno did not hear meowing. He saw something moving
like a cat under the rusted pickup. Not a neighbor for cats. But his cat is out
there. It must be his cat.

\

\begin{tabular}{ll}
revision 1:	&	\outside		\\
revision 2:	&	\cat		\\
revision 3:	&	not \outside
\end{tabular}

\

If his cat is out there, what Bruno saw was probably his cat. If his cat is not
out there, what Bruno saw is no longer sure to be a cat. It could have been
something else. This time, the expected result is that the cat is not outside
and nothing else. What is under the rusted pickup could have been a cat, but
could have been not. Maybe a cat, maybe not. Still believing it a cat is
unjustified if Bruno's cat is not around there.

Natural revision is right this time.

Both sequences of revisions are: believe $x$, believe $y$, believe $\neg x$. In
both cases, $x$ stands for the cat being outside. In the first case, $y$ is the
sound of meowing. In the second, is that a cat is under the pickup. Same
sequence, different expected results. Not $x$ in both, but $y$ only in the
first.

What is wrong is not natural revision. Is the question of what the right
revision is.

If two different scenarios are formalized by the same sequence of revisions and
the expected results are different, no revising method could be right in both.
The result of revising by $x$, $y$ and $\neg x$ can only be one. It is $\neg
x,y$. Or it is $\neg x$. Either one, not both.

Natural revision denies $y$. It is wrong if $y$ stands for meowing. It is right
if $y$ stands for a cat being under the pickup.

Still better, it is wrong to apply it in the first scenario. It is right to
apply it in the second. Calling natural revision wrong because it is applied
improperly is unwarranted. It should be applied when appropriate.

The problem is not that natural revision is wrong and needs fixing. The problem
is when it is wrong and what to do instead when it is. In the other way around,
when it is right?

Natural revision is so conservative on beliefs that it only changes them
according to the current situation. When meowing is heard, it is believed to be
heard only in present conditions, only when the cat is believed to be out
there. It is not believed in general. For example, it is not believed if the
cat were in a station office.

This is the right way of revising in the second scenario. That a cat is under
the rusted pickup is only likely when the cat is out there. If it is now
drinking milk in the station, what was seen moving under the pickup may be
something else.

Natural revision is right when the new information is only believed in the
present conditions. A cat is there only if Bruno's cat is outside the station.
Otherwise, it is something else.

Natural revision is wrong with new beliefs that hold in general. Meowing was
heard, period. Whether the cat is outside the station is irrelevant.

This is a conditional: ``if the cat is outside, then a cat is under the rusted
pickup''. It is written ``$\outside > \cat$''. A cat is under the pickup, but
only if Bruno's cat is out there.

The first scenario is different. A sound was heard: ``\meow''. It was heard
regardless of ``\outside''. It is believed regardless of whatever else is
believed. This is ``\meow'', not ``$\outside > \meow$''. Insisting on writing
it as a conditional, it is ``$\true > \meow$'', meaning that ``\meow'' is true
in all possible conditions, all conditions where ``$\true$'' holds.

The two sequences of revisions are now different. The first scenario is:

\

\begin{tabular}{ll}
the cat is out there		&	$\true > \outside$	\\
a sound of meowing was heard	&	$\true > \meow$		\\
the cat is not out there	&	$\true > \neg \outside$	\\
\end{tabular}

\

The second scenario is:

\

\begin{tabular}{ll}
the cat is out there		&	$\true > \outside$	\\
a cat is under the car		&	$outside > \cat$	\\
the cat is not out there 	&	$\true > \neg \outside$
\end{tabular}

\

Two different sequences. The first is $\true > x$, $\true > y$, $\true > \neg
x$, the second is $\true > x$, $x > y$, $\true > \neg x$. Their second revision
differ: $\true > y$ and $x>y$. A revision can now behave differently in the two
cases. It can produce $\neg x,y$ in the first and $\neg x$ in the second.

A revision could do that. Does natural revision do?

Natural revision does not include conditions in its original definition. It
only revises by simple formulae.

Various authors~\cite{hans-92,kern-99,kern-02} extend revision from a formula
$A$ to a conditional $P>A$. Some did it for natural revision:
{} Boutilier and Goldszmidt~\cite{bout-gold-93}
and
{} Chandler and Booth~\cite{chan-boot-20}.
They did in different ways.

Which one is right?

If one is right. Both may. None may.

Their difference reflects their differing founding principles. Uncovering them
is more important than their correctness in specific cases. It allows telling
whether a specific revision policy is the one to use or not in every specific
case.

The final destination of this article is not a way to revise by a conditional
$P>A$ instead of simple formula $A$. It is the understanding of the principles
of natural revision. Extending natural revision in a way that works on specific
cases, or that behaves like natural revision on formulae $\true>A$ is a mean.
What matters is the aim: extending the principles of natural revision from $A$
to $P>A$.

As expected, natural revision equates $\true>cat$ to $outside>cat$: it belives
seeing a cat under the pickup in the current condition where the cat is
outside. It does the same for $\true>meow$: it stubbornly refuses to believe in
the sound of meowing in all circumstances; it only belives it if the cat is
outside, as currently assumed.

Natural revision automatically adds the current situation to the condition $P$
of the conditioned revision $P>A$. Not only $A$ is believed only if $P$ is, it
is only believed if the current conditions are also the case. Their changes
invalidate $A$.

Natural revision cannot believe unconditionally. It cannot believe $P>A$ in all
cases. It cannot believe in $A$ in all cases of $P$, including the ones it
currently excluded. It is contingent: it believe in the current conditions
only. Believing in all conditions requires another mechanism, an uncontingent
revision.

Uncontingent revision uncovers the contingent context of natural revision, the
current conditions that natural revision always assume. It is not always the
set of situations currently considered to be the case. It is not if the
revision contradicts them. It includes situations that are just possible. It
adds less and less believable situations until contradiction disappears.

Natural and uncontingent revision differ on their aim: believing in the current
conditions only and believe in all conditions. Yet, they are based on the same
principles. Comparing natural revision with similar definitions, a difference
emerge: while they hinge on the same principle of minimal change, natural
revision also crucially believes the unknown. A situation remain fully possible
if no revision touches it. Once you eliminate what explicitely negated,
whatever remains, no matter how improbable, is the truth. The plain truth, not
just an unlikely possibility.

This is naivety: believe in something until disproven. Believe in Santa Clause
until told otherwise. Believe in the pot of gold at the end of the rainbow
until chasing it fails. Believe the Holy Graal is around until actually
searching for it.



The principle of naivety is not specific to natural revision. Other revision
mechanisms maintain the likeliness of situations by default when they could
decrease it but are not forced to. Following it when naturally revising by a
conditional is justified by it widespread use in belief revision and
conditional logics.

The addition of the principle of naivety to natural revision restricts it to a
single, specific result. The principles of natural revision, the ones that
generates natural revision and natural revision only are finally revealed:
minimal change, naivety, believe in the current conditions only.

The following sections detail what summarized in this introduction.
Section~\ref{orders} gives preliminary definitions and notations.
Section~\ref{conditional} analyzes natural revision on the examples of this
introduction and extend it to conditionals, showing that it adds the current
context to the conditions of each revision; it also says how to uncontingently
revise, deriving the contingent context for natural revision.
The pattern of believing everything not contradicted emerges: the principle of
naivety, discusses in Section~\ref{naivety}.
Section~\ref{lexicographic-section} examines lexicographic revision under the
lens of the principles of natural revision; while it shares with uncontingent
revision its freedom from the current context, it rather prefers naivety over
mimimal change;
Section~\ref{related} compares to similar works about conditionals;
Section~\ref{conclusions} summarizes what the article does and discusses its
open issues.
The appendixes formalize what presented in the article:
%
%
Appendix~\ref{natural-section} defines natural revision, proves that it
minimally changes the order and shows how it works on the examples;
Appendix~\ref{down-section} shows what it would do if it did non follow the
principle of ingenuity;
Appendix~\ref{morenaive} defines when an ordering is more naive than another
and proves that natural revision is the single most naive minimal-change
ordering;
Appendix~\ref{uncontingent-section} proves the results about uncontingent
revision;
Appendix~\ref{lesschange} proves the results about lexicographic revision.

\section{Preliminaries: orders and conditionals}
\label{orders}

\draft

\begin{itemize}

\item propositional logic

\item connected preorders

\item graphical representation

\item mathematical notations

\end{itemize}

\enddraft

The logic language in this article is propositional logic over a finite
alphabet. Each model represents a possible scenario. Their perceived likeliness
is an order: $I \leq J$ if $I$ is at least as likely as $J$. A connected
preorder, to be precise:
{} reflexive ($I \leq I$),
{} transitive ($I \leq J$ and $J \leq K$ imply $I \leq K$) and
{} connected (either $I \leq J$ or $J \leq I$).

Connected preorders are identified by the sequence of their equivalence classes
$C = [C(0),\ldots,C(m)]$. Every ordered partition of the set of models is a
connected preorder and vice versa. The one-to-one correspondence is lost if
some $C(i)$ are allowed to be empty. Still, sequences of possibly-empty
disjoint sets are used because they allow simplifying the definitions and the
proofs, and the bijectivity of the correspondence can always be recovered by
removing the empty sets. Empty levels are not relevant to connected preorders,
they are on other significant belief revision frameworks not considered
here~\cite{arav-etal-19,schw-etal-22}.

Connected preorders are depicted by drawing the first class on the top and the
others below it, in their order like in Figure~\ref{figure-order}. This way,
the most likely situations are the models at the top. Higher means more likely.
Graphically higher equals higher likeliness. Going down the ladder, credibility
goes down.

\

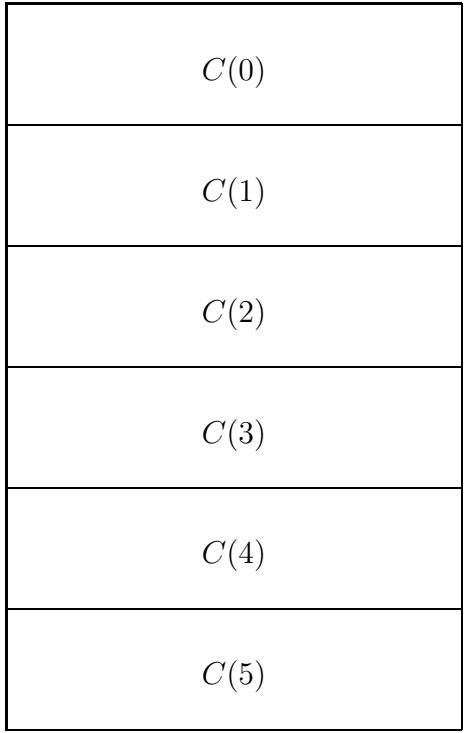
\begin{hfigure}
\setlength{\unitlength}{3750sp}%
\begin{picture}(2274,3624)(4489,-5023)
\thinlines
{\color[rgb]{0,0,0}\put(4501,-2011){\line( 1, 0){2250}}
}%
{\color[rgb]{0,0,0}\put(4501,-2611){\line( 1, 0){2250}}
}%
{\color[rgb]{0,0,0}\put(4501,-3211){\line( 1, 0){2250}}
}%
{\color[rgb]{0,0,0}\put(4501,-3811){\line( 1, 0){2250}}
}%
{\color[rgb]{0,0,0}\put(4501,-4411){\line( 1, 0){2250}}
}%
{\color[rgb]{0,0,0}\put(4501,-5011){\framebox(2250,3600){}}
}%
\put(5626,-1786){\makebox(0,0)[b]{\smash{\fontsize{9}{10.8}
\usefont{T1}{cmr}{m}{n}{\color[rgb]{0,0,0}$C(0)$}%
}}}
\put(5626,-2386){\makebox(0,0)[b]{\smash{\fontsize{9}{10.8}
\usefont{T1}{cmr}{m}{n}{\color[rgb]{0,0,0}$C(1)$}%
}}}
\put(5626,-2986){\makebox(0,0)[b]{\smash{\fontsize{9}{10.8}
\usefont{T1}{cmr}{m}{n}{\color[rgb]{0,0,0}$C(2)$}%
}}}
\put(5626,-3586){\makebox(0,0)[b]{\smash{\fontsize{9}{10.8}
\usefont{T1}{cmr}{m}{n}{\color[rgb]{0,0,0}$C(3)$}%
}}}
\put(5626,-4186){\makebox(0,0)[b]{\smash{\fontsize{9}{10.8}
\usefont{T1}{cmr}{m}{n}{\color[rgb]{0,0,0}$C(4)$}%
}}}
\put(5626,-4786){\makebox(0,0)[b]{\smash{\fontsize{9}{10.8}
\usefont{T1}{cmr}{m}{n}{\color[rgb]{0,0,0}$C(5)$}%
}}}
\end{picture}%
\nop{
 +------+
 | C(0) |
 +------+
 | C(1) |
 +------+
 | C(2) |
 +------+
 | C(3) |
 +------+
 | C(4) |
 +------+
 | C(5) |
 +------+
 | C(6) |
 +------+
 | C(7) |
 +------+
}
\label{figure-order}
\hcaption{Graphical depiction of an order $C$.}
\end{hfigure}

\

Some boxes in the shelf may be empty, but every model is in exactly one box.
Some models satisfy a formula $F$, some others do not. The smallest $i$ such
that $C(i)$ contains some models of $F$ is denoted $\minidx(F)$. Formally, it
should have been $\minidx(C,F)$ since it depends not only on $F$ but also on
$C$, but no confusion derives from omitting $C$. In the same way, $\maxidx(F)$
is the largest $i$ such that $C(i)$ contains models of $F$. The minimal and
maximal models of $F$ are respectively denoted $\min(F)$ and $\max(F)$.

Two specific orderings are relevant:

\begin{description}

\item [flat order]

$C_\epsilon$ sorts $I \leq J$ for all models;
all models are in class zero;

\item [positive order of a formula]

$C_F$ sorts $I \leq J$ if and only if $I \models F$ or $J \models \neg F$;
all models of $F$ are in class zero and all others are in class one.

\end{description}

\

\long\def\ttytex#1#2{#1}
\ttytex{
\begin{tabular}{lll}
notation &  meaning &                        formula \\
\hline
$\minidx(F)$ & minimum index of models of $F$ &
		$\min \{i \mid FC(i) \not= \emptyset\}$ \\
$\maxidx(F)$ & maximum index of models of $F$ &
		$\max \{i \mid FC(i) \not= \emptyset\}$ \\
\hline
\end{tabular}
}{
notation    meaning                         formula
---------------------------------------------------------------
min(F)      minimum index of models of F    min {i | FC(i) != 0}
max(F)      maximum index of models of F    max {i | FC(i) != 0}
---------------------------------------------------------------
}

\

\

The formulae employ some simplifications: first, when a formula occurs in
a context where a set of models is expected, like in $F \cap C(i)$, it stands
for its set of models; second, the intersection symbol $\cap$ is omitted like
the multiplication sign in numerical formulae.

The models in the same classes of $F$ and the classes of lower index are
denoted $(\lequal F)$. The models in the classes of higher index $(\greater
F)$. The models in the same classes are $(\equal F)$.

These functions also depend on the partition $C$, which is omitted because no
confusion will arise.

\

A connected preorder extends from models to sets: $A \leq B$ means that every
model of $A$ is less than or equal to every model of $B$. Equivalently,
$\max(A) \leq J$ for every model $J$ of $B$. Another equivalent form is
$\max(A) \leq \min(B)$.

A propositional formula as an argument of a set operator like intersection,
union or difference stands for its set of models.

A quick brown fox example is $\equal\min(F) < \neg F$: every model in the same
class of the minimal models of $F$ is less than every model not satisfying $F$.

A model verifies, falsifies or is indifferent to a conditional $P>A$ if it
respectively satisfies $PA$, $P \neg A$ or $\neg P$~\cite{lehm-magi-92}. A
conditional $P>A$ is true in an order if all minimal models of $P$ satisfy $A$.
This is equivalent to $\min(PA) < P \neg A$.

\section{Revising by a conditional}
\label{conditional}

\draft

\begin{itemize}

\item revisione naturale

\begin{itemize}

\item cosa fa la revisione naturale su formule non condizionate T>A

\item funziona con outside,cat,-outside, non funziona con
outside,meow,-outside; e' perche' limita ogni revisione al contesto attuale
outside

\item questa e' una revisione con un condizionale P>A;
estensione della revisione naturale ai condizionali sempre seguendo il
principio di cambiamento minimo

\item la revisione naturale crede solo nel contesto attuale: aggiunge il
contesto attuale alla condizione

\item credere in tutti i casi = uncontingent revision;

\end{itemize}

\item revisione non contingente

\begin{itemize}

\item definizione di revisione che crede in tutti i casi, basata su cambiamento
minimo

\item funziona nel caso di esempio, quando si vuole credere in tutte le
condizioni

\item funziona anche nel caso in cui si vuole credere solo nella situazione
corrente, aggiungendo esplicitamente la condizione corrente

\end{itemize}

\item il contesto contingente della revisione naturale

\begin{itemize}

\item non puo' essere solo l'insieme delle condizioni attuali

\item la traduzione natural->uncontingent permette di capire qual e'

\item spiegazione del perche' e' cosi'

\end{itemize}

\item otherwise

\begin{itemize}

\item esistono altri modi di fare revisione di formule condizionali; occorre
capire quali diversi principi applicano

\item l'analisi di meccanismi simili, incluse le semantiche per insiemi di
condizionali invece che di sequenze di revisioni, fornisce sostegni deboli al
sistema adottato qui

\item ma le semantiche per insiemi di condizionali permettono di isolare una
peculiarita' comune: la preferenza per collocare i modelli in alto; gli scenari
sono creduti possibili al massimo finche' non esplicitamente negati

\end{itemize}

\end{itemize}

\enddraft

\

\subsection{Natural revision}

Natural revision makes $A$ believed by a minimal change of beliefs. It makes it
maximally believed: it is true in all of the most believed situations.
Graphically, all models of the top level of the belief strength ordering
satisfy $A$ as exemplified in Figure~\ref{figure-natural}.

It does it by minimally changing how much beliefs are believed.

Models of $\min(A)$ are more believed scenarios than all others of $A$. They
stay at the top, above all others%
{}\footnote{Definition~\ref{natural-propositional}.}.

\begin{hfigure}
\long\def\ttytex#1#2{#1}
\ttytex{%
\begin{tabular}{ccc}
\setlength{\unitlength}{3750sp}%
\begin{picture}(2352,3024)(4489,-4423)
\thinlines
{\color[rgb]{0,0,0}\put(4501,-2011){\line( 1, 0){2250}}
}%
{\color[rgb]{0,0,0}\put(4501,-2611){\line( 1, 0){2250}}
}%
{\color[rgb]{0,0,0}\put(4501,-3211){\line( 1, 0){2250}}
}%
{\color[rgb]{0,0,0}\put(4501,-3811){\line( 1, 0){2250}}
}%
{\color[rgb]{0,0,0}\put(5401,-3661){\framebox(750,1500){}}
}%
{\color[rgb]{0,0,0}\put(4501,-4411){\framebox(2250,3000){}}
}%
\put(6826,-2386){\makebox(0,0)[lb]{\smash{\fontsize{9}{10.8}
\usefont{T1}{cmr}{m}{n}{\color[rgb]{0,0,0}$C(1)$}%
}}}
\put(6826,-1786){\makebox(0,0)[lb]{\smash{\fontsize{9}{10.8}
\usefont{T1}{cmr}{m}{n}{\color[rgb]{0,0,0}$C(0)$}%
}}}
\put(5626,-2986){\makebox(0,0)[b]{\smash{\fontsize{9}{10.8}
\usefont{T1}{cmr}{m}{n}{\color[rgb]{0,0,0}$A$}%
}}}
\put(5776,-2461){\makebox(0,0)[b]{\smash{\fontsize{9}{10.8}
\usefont{T1}{cmr}{m}{n}{\color[rgb]{0,0,0}$\min(A)$}%
}}}
\end{picture}%
&
\setlength{\unitlength}{3750sp}%
\begin{picture}(1104,1824)(5659,-5023)
\thinlines
{\color[rgb]{0,0,0}\multiput(6391,-3391)(-9.47368,4.73684){20}{\makebox(2.1167,14.8167){\tiny.}}
\put(6211,-3301){\line( 0,-1){ 45}}
\put(6211,-3346){\line(-1, 0){180}}
\put(6031,-3346){\line( 0,-1){ 90}}
\put(6031,-3436){\line( 1, 0){180}}
\put(6211,-3436){\line( 0,-1){ 45}}
\multiput(6211,-3481)(9.47368,4.73684){20}{\makebox(2.1167,14.8167){\tiny.}}
}%
\end{picture}%
&
\setlength{\unitlength}{3750sp}%
\begin{picture}(2352,3774)(4489,-4573)
\thinlines
{\color[rgb]{0,0,0}\put(5401,-1261){\framebox(750,450){}}
}%
{\color[rgb]{0,0,0}\put(4501,-3361){\line( 1, 0){2250}}
}%
{\color[rgb]{0,0,0}\put(4501,-3961){\line( 1, 0){2250}}
}%
{\color[rgb]{0,0,0}\put(5401,-3811){\framebox(750,1050){}}
}%
{\color[rgb]{0,0,0}\put(4501,-4561){\framebox(2250,1800){}}
}%
{\color[rgb]{0,0,0}\put(4501,-1861){\framebox(2250,600){}}
}%
{\color[rgb]{0,0,0}\put(4501,-2611){\line( 0, 1){600}}
\put(4501,-2011){\line( 1, 0){2250}}
\put(6751,-2011){\line( 0,-1){600}}
\put(6751,-2611){\line(-1, 0){600}}
\put(6151,-2611){\line( 0, 1){450}}
\put(6151,-2161){\line(-1, 0){750}}
\put(5401,-2161){\line( 0,-1){450}}
\put(5401,-2611){\line(-1, 0){900}}
}%
\put(5776,-1111){\makebox(0,0)[b]{\smash{\fontsize{9}{10.8}
\usefont{T1}{cmr}{m}{n}{\color[rgb]{0,0,0}$\min(A)$}%
}}}
\put(4951,-2386){\makebox(0,0)[b]{\smash{\fontsize{9}{10.8}
\usefont{T1}{cmr}{m}{n}{\color[rgb]{0,0,0}$\equal\min(A)\neg A$}%
}}}
\put(6826,-1111){\makebox(0,0)[lb]{\smash{\fontsize{9}{10.8}
\usefont{T1}{cmr}{m}{n}{\color[rgb]{0,0,0}$C\nat(A)(0)$}%
}}}
\put(6826,-1636){\makebox(0,0)[lb]{\smash{\fontsize{9}{10.8}
\usefont{T1}{cmr}{m}{n}{\color[rgb]{0,0,0}$C\nat(A)(1)$}%
}}}
\put(6826,-2386){\makebox(0,0)[lb]{\smash{\fontsize{9}{10.8}
\usefont{T1}{cmr}{m}{n}{\color[rgb]{0,0,0}$C\nat(A)(2)$}%
}}}
\end{picture}%
\end{tabular}
}{
+----------------------------------+                 +----------------+
| C(0)                             |                 |     min(A)     |
|                                  |                 |                |
|                                  |                 +----------------+
+----------------------------------+      +----------------------------------+
| C(1)     +----------------+      |      | C(0)                             |
|          |     min(A)     |      |      |                                  |
|          |                |      |      |                                  |
+----------+----------------+------+      +----------------------------------+
| C(2)     |                |      |      | C(1)-min(A)---------------+      |
|          |       A        |      |  =>  |          |XXXXXXXXXXXXXXXX|      |
|          |                |      |      |          |XXXXXXXXXXXXXXXX|      |
+----------+----------------+------+      +----------+----------------+------+
|          |                |      |      | C(2)     |                |      |
|          |                |      |      |          |       A        |      |
|          +----------------+      |      |          |                |      |
+----------------------------------+      +----------+----------------+------+
|                                  |      |          |                |      |
|                                  |      |          |                |      |
|                                  |      |          +----------------+      |
+----------------------------------+      +----------------------------------+
                                          |                                  |
                                          |                                  |
                                          |                                  |
                                          +----------------------------------+
}
\label{figure-natural}
\hcaption{How natural revision changes the order\\
(vertical gaps and white spaces emphasize the changes).}
\end{hfigure}

\

Natural revision succeds on $outside$, $cat$, not $outside$ and fails on
$outside$, $meow$, not $outside$.

\

\begin{hfigure}
\begin{tabular}{c}
\setlength{\unitlength}{3750sp}%
\begin{picture}(2424,924)(4339,-3973)
\thinlines
{\color[rgb]{0,0,0}\put(4351,-3961){\framebox(2400,900){}}
}%
{\color[rgb]{0,0,0}\put(4351,-3511){\line( 1, 0){2400}}
}%
\put(4951,-3361){\makebox(0,0)[b]{\smash{\fontsize{9}{10.8}
\usefont{T1}{cmr}{m}{n}{\color[rgb]{0,0,0}$outside,cat$}%
}}}
\put(6001,-3361){\makebox(0,0)[b]{\smash{\fontsize{9}{10.8}
\usefont{T1}{cmr}{m}{n}{\color[rgb]{0,0,0}$outside,\neg cat$}%
}}}
\put(4951,-3811){\makebox(0,0)[b]{\smash{\fontsize{9}{10.8}
\usefont{T1}{cmr}{m}{n}{\color[rgb]{0,0,0}$\neg outside,cat$}%
}}}
\put(6076,-3811){\makebox(0,0)[b]{\smash{\fontsize{9}{10.8}
\usefont{T1}{cmr}{m}{n}{\color[rgb]{0,0,0}$\neg outside,\neg cat$}%
}}}
\end{picture}%
\nop{
      +--------------------------------------------------------+
      | outside,cat  outside,-cat                              |
      +--------------------------------------------------------+
      |                            -outside,cat  -outside,-cat |
      +--------------------------------------------------------+
}
\\
\setlength{\unitlength}{3750sp}%
\begin{picture}(384,204)(6019,-3493)
\thinlines
{\color[rgb]{0,0,0}\multiput(6391,-3391)(-9.47368,4.73684){20}{\makebox(2.1167,14.8167){\tiny.}}
\put(6211,-3301){\line( 0,-1){ 45}}
\put(6211,-3346){\line(-1, 0){180}}
\put(6031,-3346){\line( 0,-1){ 90}}
\put(6031,-3436){\line( 1, 0){180}}
\put(6211,-3436){\line( 0,-1){ 45}}
\multiput(6211,-3481)(9.47368,4.73684){20}{\makebox(2.1167,14.8167){\tiny.}}
}%
\end{picture}%
\nop{
==>
}
\\
\setlength{\unitlength}{3750sp}%
\begin{picture}(4524,1374)(4189,-4273)
\thinlines
{\color[rgb]{0,0,0}\put(4201,-4261){\framebox(4500,1350){}}
}%
{\color[rgb]{0,0,0}\put(4201,-3361){\line( 1, 0){4500}}
}%
{\color[rgb]{0,0,0}\put(4201,-3811){\line( 1, 0){4500}}
}%
\put(5851,-3661){\makebox(0,0)[b]{\smash{\fontsize{9}{10.8}
\usefont{T1}{cmr}{m}{n}{\color[rgb]{0,0,0}$outside,\neg cat$}%
}}}
\put(6901,-4111){\makebox(0,0)[b]{\smash{\fontsize{9}{10.8}
\usefont{T1}{cmr}{m}{n}{\color[rgb]{0,0,0}$\neg outside,cat$}%
}}}
\put(8026,-4111){\makebox(0,0)[b]{\smash{\fontsize{9}{10.8}
\usefont{T1}{cmr}{m}{n}{\color[rgb]{0,0,0}$\neg outside,\neg cat$}%
}}}
\put(4801,-3211){\makebox(0,0)[b]{\smash{\fontsize{9}{10.8}
\usefont{T1}{cmr}{m}{n}{\color[rgb]{0,0,0}$outside,cat$}%
}}}
\end{picture}%
\nop{
      +--------------------------------------------------------+
      | outside,cat                                            |
      +--------------------------------------------------------+
      |              outside,-cat                              |
      +--------------------------------------------------------+
      |                            -outside,cat  -outside,-cat |
      +--------------------------------------------------------+
}
\\
\setlength{\unitlength}{3750sp}%
\begin{picture}(384,204)(6019,-3493)
\thinlines
{\color[rgb]{0,0,0}\multiput(6391,-3391)(-9.47368,4.73684){20}{\makebox(2.1167,14.8167){\tiny.}}
\put(6211,-3301){\line( 0,-1){ 45}}
\put(6211,-3346){\line(-1, 0){180}}
\put(6031,-3346){\line( 0,-1){ 90}}
\put(6031,-3436){\line( 1, 0){180}}
\put(6211,-3436){\line( 0,-1){ 45}}
\multiput(6211,-3481)(9.47368,4.73684){20}{\makebox(2.1167,14.8167){\tiny.}}
}%
\end{picture}%
\nop{
==>
}
\\
\setlength{\unitlength}{3750sp}%
\begin{picture}(4374,1374)(3589,-4123)
\thinlines
{\color[rgb]{0,0,0}\put(3601,-4111){\framebox(4350,1350){}}
}%
{\color[rgb]{0,0,0}\put(3601,-3211){\line( 1, 0){4350}}
}%
{\color[rgb]{0,0,0}\put(3601,-3661){\line( 1, 0){4350}}
}%
\put(6301,-3061){\makebox(0,0)[b]{\smash{\fontsize{9}{10.8}
\usefont{T1}{cmr}{m}{n}{\color[rgb]{0,0,0}$\neg outside,cat$}%
}}}
\put(7351,-3061){\makebox(0,0)[b]{\smash{\fontsize{9}{10.8}
\usefont{T1}{cmr}{m}{n}{\color[rgb]{0,0,0}$\neg outside,\neg cat$}%
}}}
\put(4201,-3511){\makebox(0,0)[b]{\smash{\fontsize{9}{10.8}
\usefont{T1}{cmr}{m}{n}{\color[rgb]{0,0,0}$outside,cat$}%
}}}
\put(5251,-3961){\makebox(0,0)[b]{\smash{\fontsize{9}{10.8}
\usefont{T1}{cmr}{m}{n}{\color[rgb]{0,0,0}$outside,\neg cat$}%
}}}
\end{picture}%
\nop{
      +--------------------------------------------------------+
      |                            -outside,cat  -outside,-cat |
      +--------------------------------------------------------+
      | outside,cat                                            |
      +--------------------------------------------------------+
      |              outside,-cat                              |
      +--------------------------------------------------------+
}
\end{tabular}
\label{figure-bruno-cat1}
\hcaption{Natural revision on $outside$, $cat$, not $outside$.}
\end{hfigure}

\

Natural revision succeds on $outside$, $cat$, not $outside$%
{}\footnote{Theorem~\ref{natural-nocondition} with $x=outside$ and $y=cat$.},
as shown in Figure~\ref{figure-bruno-cat1}. The cat must be under the rusted
pickup in the current conditions, where Bruno's cat is believed to be outside:
$\{outside,cat\}$ is above $\{outside,\neg cat\}$. The cat may not be there if
Bruno's cat is not outside: $\{\neg outside,cat\}$ is not above $\{\neg
outside,\neg cat\}$%
{}\footnote{Theorem~\ref{natural-nocondition} with $x=outside$ and $y=cat$.}.

\

\begin{hfigure}
\begin{tabular}{c}
\setlength{\unitlength}{3750sp}%
\begin{picture}(5574,924)(3139,-3973)
\thinlines
{\color[rgb]{0,0,0}\put(3151,-3961){\framebox(5550,900){}}
}%
{\color[rgb]{0,0,0}\put(3151,-3511){\line( 1, 0){5550}}
}%
\put(3901,-3361){\makebox(0,0)[b]{\smash{\fontsize{9}{10.8}
\usefont{T1}{cmr}{m}{n}{\color[rgb]{0,0,0}$outside,meow$}%
}}}
\put(5251,-3361){\makebox(0,0)[b]{\smash{\fontsize{9}{10.8}
\usefont{T1}{cmr}{m}{n}{\color[rgb]{0,0,0}$outside,\neg meow$}%
}}}
\put(6601,-3811){\makebox(0,0)[b]{\smash{\fontsize{9}{10.8}
\usefont{T1}{cmr}{m}{n}{\color[rgb]{0,0,0}$\neg outside,meow$}%
}}}
\put(7951,-3811){\makebox(0,0)[b]{\smash{\fontsize{9}{10.8}
\usefont{T1}{cmr}{m}{n}{\color[rgb]{0,0,0}$\neg outside,\neg meow$}%
}}}
\end{picture}%
\nop{
      +------------------------------------------------------------+
      | outside,meow  outside,-meow                                |
      +------------------------------------------------------------+
      |                              -outside,meow  -outside,-meow |
      +------------------------------------------------------------+
}
\\
\setlength{\unitlength}{3750sp}%
\begin{picture}(384,204)(6019,-3493)
\thinlines
{\color[rgb]{0,0,0}\multiput(6391,-3391)(-9.47368,4.73684){20}{\makebox(2.1167,14.8167){\tiny.}}
\put(6211,-3301){\line( 0,-1){ 45}}
\put(6211,-3346){\line(-1, 0){180}}
\put(6031,-3346){\line( 0,-1){ 90}}
\put(6031,-3436){\line( 1, 0){180}}
\put(6211,-3436){\line( 0,-1){ 45}}
\multiput(6211,-3481)(9.47368,4.73684){20}{\makebox(2.1167,14.8167){\tiny.}}
}%
\end{picture}%
\nop{
==>
}
\\
\setlength{\unitlength}{3750sp}%
\begin{picture}(5574,1374)(3139,-4423)
\thinlines
{\color[rgb]{0,0,0}\put(3151,-4411){\framebox(5550,1350){}}
}%
{\color[rgb]{0,0,0}\put(3151,-3511){\line( 1, 0){5550}}
}%
{\color[rgb]{0,0,0}\put(3151,-3961){\line( 1, 0){5550}}
}%
\put(3901,-3361){\makebox(0,0)[b]{\smash{\fontsize{9}{10.8}
\usefont{T1}{cmr}{m}{n}{\color[rgb]{0,0,0}$outside,meow$}%
}}}
\put(5251,-3811){\makebox(0,0)[b]{\smash{\fontsize{9}{10.8}
\usefont{T1}{cmr}{m}{n}{\color[rgb]{0,0,0}$outside,\neg meow$}%
}}}
\put(6601,-4261){\makebox(0,0)[b]{\smash{\fontsize{9}{10.8}
\usefont{T1}{cmr}{m}{n}{\color[rgb]{0,0,0}$\neg outside,meow$}%
}}}
\put(7951,-4261){\makebox(0,0)[b]{\smash{\fontsize{9}{10.8}
\usefont{T1}{cmr}{m}{n}{\color[rgb]{0,0,0}$\neg outside,\neg meow$}%
}}}
\end{picture}%
\nop{
      +------------------------------------------------------------+
      | outside,meow                                               |
      +------------------------------------------------------------+
      |               outside,-meow                                |
      +------------------------------------------------------------+
      |                              -outside,meow  -outside,-meow |
      +------------------------------------------------------------+
}
\\
\setlength{\unitlength}{3750sp}%
\begin{picture}(384,204)(6019,-3493)
\thinlines
{\color[rgb]{0,0,0}\multiput(6391,-3391)(-9.47368,4.73684){20}{\makebox(2.1167,14.8167){\tiny.}}
\put(6211,-3301){\line( 0,-1){ 45}}
\put(6211,-3346){\line(-1, 0){180}}
\put(6031,-3346){\line( 0,-1){ 90}}
\put(6031,-3436){\line( 1, 0){180}}
\put(6211,-3436){\line( 0,-1){ 45}}
\multiput(6211,-3481)(9.47368,4.73684){20}{\makebox(2.1167,14.8167){\tiny.}}
}%
\end{picture}%
\nop{
==>
}
\\
\setlength{\unitlength}{3750sp}%
\begin{picture}(5724,1374)(3139,-4423)
\thinlines
{\color[rgb]{0,0,0}\put(3151,-3961){\line( 1, 0){5700}}
}%
{\color[rgb]{0,0,0}\put(3151,-4411){\framebox(5700,1350){}}
}%
{\color[rgb]{0,0,0}\put(3151,-3511){\line( 1, 0){5700}}
}%
\put(6601,-3361){\makebox(0,0)[b]{\smash{\fontsize{9}{10.8}
\usefont{T1}{cmr}{m}{n}{\color[rgb]{0,0,0}$\neg outside,meow$}%
}}}
\put(7951,-3361){\makebox(0,0)[b]{\smash{\fontsize{9}{10.8}
\usefont{T1}{cmr}{m}{n}{\color[rgb]{0,0,0}$\neg outside,\neg meow$}%
}}}
\put(3901,-3811){\makebox(0,0)[b]{\smash{\fontsize{9}{10.8}
\usefont{T1}{cmr}{m}{n}{\color[rgb]{0,0,0}$outside,meow$}%
}}}
\put(5251,-4261){\makebox(0,0)[b]{\smash{\fontsize{9}{10.8}
\usefont{T1}{cmr}{m}{n}{\color[rgb]{0,0,0}$outside,\neg meow$}%
}}}
\end{picture}%
\nop{
      +------------------------------------------------------------+
      |                              -outside,meow  -outside,-meow |
      +------------------------------------------------------------+
      | outside,meow                                               |
      +------------------------------------------------------------+
      |               outside,-meow                                |
      +------------------------------------------------------------+
}
\end{tabular}
\label{figure-bruno-meow1}
\hcaption{Natural revision on $outside$, $meow$, not $outside$.}
\end{hfigure}

\

Natural revision fails on $outside$, $meow$, not $outside$%
{}\footnote{Theorem~\ref{natural-nocondition} with $x=outside$ and $y=meow$.},
as shown in Figure~\ref{figure-bruno-meow1}. That something meows is believed
only if Bruno's cat is outside: $\{outside,meow\}$ may be more believed than
$\{outside,\neg meow\}$, but $\{\neg outside,meow\}$ is not more believed than
$\{\neg outside,\neg meow\}$, neither it is of $\{outside,-meow\}$. This is
wrong since the sound was heard, no matter where Bruno's cat is.

Natural revision fails because it always confines the new belief to the present
conditions: it believes that something meows only if Bruno's cat is outside. It
belives the new fact only in the present conditions.

This is $outside>cat$ or $outside>meow$, not just $cat$ or $meow$.

Revising by $P>A$ is satisfying $P>A$. Is making $A$ true in the most believed
models of $P$.

Several methods achieve this goal%
{}~\cite{hans-92,bout-gold-93,kern-99,kern-02,chan-boot-20}.
Natural revision changes the strength of beliefs as little as possible%
{}\footnote{Definition~\ref{natural}, Theorem~\ref{natural-true},
Definition~\ref{strength}, Definition~\ref{closer} and
Theorem~\ref{natural-minimal}.}.

\begin{hfigure}
\long\def\ttytex#1#2{#1}
\ttytex{
\begin{tabular}{ccc}
%
%
\setlength{\unitlength}{3750sp}%
\begin{picture}(2274,4824)(4489,-5623)
\thinlines
{\color[rgb]{0,0,0}\put(4501,-2011){\line( 1, 0){2250}}
}%
{\color[rgb]{0,0,0}\put(4501,-2611){\line( 1, 0){2250}}
}%
{\color[rgb]{0,0,0}\put(4501,-3211){\line( 1, 0){2250}}
}%
{\color[rgb]{0,0,0}\put(4501,-3811){\line( 1, 0){2250}}
}%
{\color[rgb]{0,0,0}\put(4501,-4411){\line( 1, 0){2250}}
}%
{\color[rgb]{0,0,0}\put(5401,-3661){\framebox(750,900){}}
}%
{\color[rgb]{0,0,0}\put(4501,-1411){\line( 1, 0){2250}}
}%
{\color[rgb]{0,0,0}\put(4501,-5611){\framebox(2250,4800){}}
}%
{\color[rgb]{0,0,0}\put(4501,-5011){\line( 1, 0){2250}}
}%
{\color[rgb]{0,0,0}\put(4951,-4861){\framebox(1350,3300){}}
}%
\put(5626,-3061){\makebox(0,0)[b]{\smash{\fontsize{9}{10.8}
\usefont{T1}{cmr}{m}{n}{\color[rgb]{0,0,0}$PA$}%
}}}
\put(5101,-1861){\makebox(0,0)[b]{\smash{\fontsize{9}{10.8}
\usefont{T1}{cmr}{m}{n}{\color[rgb]{0,0,0}$P$}%
}}}
\put(5776,-2236){\makebox(0,0)[b]{\smash{\fontsize{9}{10.8}
\usefont{T1}{cmr}{m}{n}{\color[rgb]{0,0,0}$P\neg A$}%
}}}
\put(4726,-1711){\makebox(0,0)[b]{\smash{\fontsize{9}{10.8}
\usefont{T1}{cmr}{m}{n}{\color[rgb]{0,0,0}$\neg P$}%
}}}
\end{picture}%
&
\setlength{\unitlength}{3750sp}%
\begin{picture}(1104,2544)(5659,-5743)
\thinlines
{\color[rgb]{0,0,0}\multiput(6391,-3391)(-9.47368,4.73684){20}{\makebox(2.1167,14.8167){\tiny.}}
\put(6211,-3301){\line( 0,-1){ 45}}
\put(6211,-3346){\line(-1, 0){180}}
\put(6031,-3346){\line( 0,-1){ 90}}
\put(6031,-3436){\line( 1, 0){180}}
\put(6211,-3436){\line( 0,-1){ 45}}
\multiput(6211,-3481)(9.47368,4.73684){20}{\makebox(2.1167,14.8167){\tiny.}}
}%
\end{picture}%
&
\setlength{\unitlength}{3750sp}%
\begin{picture}(2274,6774)(4489,-7123)
\thinlines
{\color[rgb]{0,0,0}\put(5401,-2761){\framebox(750,450){}}
}%
{\color[rgb]{0,0,0}\put(4951,-2761){\line(-1, 0){450}}
\put(4501,-2761){\line( 0, 1){2400}}
\put(4501,-361){\line( 1, 0){2250}}
\put(6751,-361){\line( 0,-1){2400}}
\put(6751,-2761){\line(-1, 0){450}}
\put(6301,-2761){\line( 0, 1){1650}}
\put(6301,-1111){\line(-1, 0){1350}}
\put(4951,-1111){\line( 0,-1){1650}}
}%
{\color[rgb]{0,0,0}\put(4501,-2161){\line( 1, 0){450}}
}%
{\color[rgb]{0,0,0}\put(6301,-2161){\line( 1, 0){450}}
}%
{\color[rgb]{0,0,0}\put(4501,-961){\line( 1, 0){2250}}
}%
{\color[rgb]{0,0,0}\put(4501,-1561){\line( 1, 0){450}}
}%
{\color[rgb]{0,0,0}\put(6301,-1561){\line( 1, 0){450}}
}%
\put(5626,-2611){\makebox(0,0)[b]{\smash{\fontsize{9}{10.8}
\usefont{T1}{cmr}{m}{n}{\color[rgb]{0,0,0}$PA$}%
}}}
\put(4726,-1261){\makebox(0,0)[b]{\smash{\fontsize{9}{10.8}
\usefont{T1}{cmr}{m}{n}{\color[rgb]{0,0,0}$\neg P$}%
}}}
{\color[rgb]{0,0,0}\put(4501,-5311){\line( 1, 0){2250}}
}%
{\color[rgb]{0,0,0}\put(5401,-5161){\framebox(750,450){}}
}%
{\color[rgb]{0,0,0}\put(5401,-4561){\line(-1, 0){450}}
\put(4951,-4561){\line( 0, 1){1650}}
\put(4951,-2911){\line( 1, 0){1350}}
\put(6301,-2911){\line( 0,-1){1650}}
\put(6301,-4561){\line(-1, 0){150}}
\put(6151,-4561){\line( 0, 1){450}}
\put(6151,-4111){\line(-1, 0){750}}
\put(5401,-4111){\line( 0,-1){450}}
}%
{\color[rgb]{0,0,0}\put(4951,-3961){\line( 1, 0){1350}}
}%
{\color[rgb]{0,0,0}\put(4951,-3361){\line( 1, 0){1350}}
}%
{\color[rgb]{0,0,0}\put(4501,-7111){\framebox(2250,2400){}}
}%
{\color[rgb]{0,0,0}\put(4501,-6511){\line( 1, 0){2250}}
}%
{\color[rgb]{0,0,0}\put(4501,-5911){\line( 1, 0){2250}}
}%
{\color[rgb]{0,0,0}\put(4951,-6361){\framebox(1350,1650){}}
}%
\put(5776,-3586){\makebox(0,0)[b]{\smash{\fontsize{9}{10.8}
\usefont{T1}{cmr}{m}{n}{\color[rgb]{0,0,0}$P\neg A$}%
}}}
\end{picture}%
\end{tabular}
}{
+--------------------------------+      +--------------------------------+
|                                |      |                                |
|                                |      |                                |
|                                |      |                                |
+--------------------------------+      +--------------------------------+
|       +--------------------+   |      |       +--------------------+   |
| -P    | P-A                |   |      | -P    |                    |   |
|       |                    |   |      |       |                    |   |
+-------|--------------------|---+      +-------+                    +---+
|       | P-A                |   |      |       |                    |   |
|       |        +-------+   |   |      |       |        +-------+   |   |
|       |        |min(PA)|   |   |      |       |        |min(PA)|   |   |
+-------|--------|-------|---|---+  =>  +-------+        +-------+   +---+
|       | P-A  PA|       |   |   |              +--------------------+
|       |        +-------+   |   |              |  P-A               |
|       |                    |   |              |                    |
+-------|--------------------|---+              +--------------------+
|       |                    |   |              |  P-A               |
|       |                    |   |              |        +-------+   |
|       +--------------------+   |              |        |XXXXXXX|   |
+--------------------------------+      +-------+--------+-------+---+---+
|                                |      |       |  P-A   |       |   |   |
|                                |      |       |        +-------+   |   |
|                                |      |       +--------------------+   |
+--------------------------------+      +--------------------------------+
                                        |                                |
                                        |                                |
                                        |                                |
                                        +--------------------------------+
}
\label{figure-natural-conditional}
\hcaption{Natural revision of a conditional.}
\end{hfigure}

Among the models of $P$, the scenarios satisfying the condition $P$, the most
believed $\min(PA)$ of the new belief $A$ are the most believed at all.
Graphically, $\min(PA)$ is above the rest of $P$, as exemplified in
FIgure~\ref{figure-natural-conditional}. The position of $\neg P$ is
unaffected, in accord with minimal change. Its models are models of $\neg P$,
they are unaffected by a revision regarding $P$ only.

\draft

esistono altre scelte, e in particolare altri modi di collocare la parte di -P,
ma verranno discussi dopo

POI VERRA' DETTO CHE SPOSTARE IN BASSO = PRINCIPIO DI INGENUITA'

{\bf qui omettere del tutto}

\enddraft

Natural revision always restricts the new belief to the current situation:
$outside$. It believes that a cat is under the rusted pickup only if Bruno's
cat is outside: $outside>cat$. It believes that something meows only if the cat
is outside: $outside>meow$. Believing $meow$ out of the current conditions is
impossible.

Believing $meow$ in all conditions would be $\true>meow$: believe the sound no
matter of Bruno's cat whereabouts. Natural revision belives it only in the
present condition $outside$. It believes $outside>meow$, never $\true > meow$%
{}\footnote{Theorem~\ref{natural-nocondition} and
Theorem~\ref{natural-condition}.}.

Natural revision succeds in believing in the current conditions. It fails
believing in all conditions.

\subsection{Uncontingent revision}

Natural revision is contingent: it believes only in the current conditions. It
never believes in all conditions.

Believing in all conditions is uncontingent: no matter what the present
conditions are, $A$ is believed. That Bruno's cat is outside is irrelevant:
something meows, anyway.

Uncontingently revising by $P>A$ is believing in $A$ in all conditions meeting
$P$, regardless of which ones are currently the most believed. If $P$ is true,
then $A$ is more believed than $\neg A$. Every scenario of $PA$ is more
believed than any of $P\neg A$%
{}\footnote{Definition~\ref{uncontingent}.}.
Every scenario, not just the currently most believed ones $\min(PA)$.

\begin{hfigure}
\long\def\ttytex#1#2{#1}
\ttytex{
\begin{tabular}{ccc}
%
%
\setlength{\unitlength}{3750sp}%
\begin{picture}(2274,4824)(4489,-5623)
\thinlines
{\color[rgb]{0,0,0}\put(4501,-2011){\line( 1, 0){2250}}
}%
{\color[rgb]{0,0,0}\put(4501,-2611){\line( 1, 0){2250}}
}%
{\color[rgb]{0,0,0}\put(4501,-3211){\line( 1, 0){2250}}
}%
{\color[rgb]{0,0,0}\put(4501,-3811){\line( 1, 0){2250}}
}%
{\color[rgb]{0,0,0}\put(4501,-4411){\line( 1, 0){2250}}
}%
{\color[rgb]{0,0,0}\put(5401,-3661){\framebox(750,900){}}
}%
{\color[rgb]{0,0,0}\put(4501,-1411){\line( 1, 0){2250}}
}%
{\color[rgb]{0,0,0}\put(4501,-5611){\framebox(2250,4800){}}
}%
{\color[rgb]{0,0,0}\put(4501,-5011){\line( 1, 0){2250}}
}%
{\color[rgb]{0,0,0}\put(4951,-4861){\framebox(1350,3300){}}
}%
\put(5626,-3061){\makebox(0,0)[b]{\smash{\fontsize{9}{10.8}
\usefont{T1}{cmr}{m}{n}{\color[rgb]{0,0,0}$PA$}%
}}}
\put(5101,-1861){\makebox(0,0)[b]{\smash{\fontsize{9}{10.8}
\usefont{T1}{cmr}{m}{n}{\color[rgb]{0,0,0}$P$}%
}}}
\put(5776,-2236){\makebox(0,0)[b]{\smash{\fontsize{9}{10.8}
\usefont{T1}{cmr}{m}{n}{\color[rgb]{0,0,0}$P\neg A$}%
}}}
\put(4726,-1711){\makebox(0,0)[b]{\smash{\fontsize{9}{10.8}
\usefont{T1}{cmr}{m}{n}{\color[rgb]{0,0,0}$\neg P$}%
}}}
\end{picture}%
&
\setlength{\unitlength}{3750sp}%
\begin{picture}(1104,2544)(5659,-5743)
\thinlines
{\color[rgb]{0,0,0}\multiput(6391,-3391)(-9.47368,4.73684){20}{\makebox(2.1167,14.8167){\tiny.}}
\put(6211,-3301){\line( 0,-1){ 45}}
\put(6211,-3346){\line(-1, 0){180}}
\put(6031,-3346){\line( 0,-1){ 90}}
\put(6031,-3436){\line( 1, 0){180}}
\put(6211,-3436){\line( 0,-1){ 45}}
\multiput(6211,-3481)(9.47368,4.73684){20}{\makebox(2.1167,14.8167){\tiny.}}
}%
\end{picture}%
&
\setlength{\unitlength}{3750sp}%
\begin{picture}(2274,6624)(4489,-8023)
\thinlines
{\color[rgb]{0,0,0}\put(4501,-2011){\line( 1, 0){2250}}
}%
{\color[rgb]{0,0,0}\put(5251,-3661){\framebox(750,900){}}
}%
{\color[rgb]{0,0,0}\put(4951,-3811){\line( 0, 1){1650}}
\put(4951,-2161){\line( 1, 0){1350}}
\put(6301,-2161){\line( 0,-1){1650}}
\put(6301,-3811){\line( 1, 0){450}}
\put(6751,-3811){\line( 0, 1){2400}}
\put(6751,-1411){\line(-1, 0){2250}}
\put(4501,-1411){\line( 0,-1){2400}}
\put(4501,-3811){\line( 1, 0){450}}
}%
{\color[rgb]{0,0,0}\put(4501,-2611){\line( 1, 0){450}}
}%
{\color[rgb]{0,0,0}\put(4501,-3211){\line( 1, 0){450}}
}%
{\color[rgb]{0,0,0}\put(6301,-3211){\line( 1, 0){450}}
}%
{\color[rgb]{0,0,0}\put(6301,-2611){\line( 1, 0){450}}
}%
{\color[rgb]{0,0,0}\put(5251,-3211){\line( 1, 0){750}}
}%
{\color[rgb]{0,0,0}\put(4951,-5011){\line( 1, 0){1350}}
}%
{\color[rgb]{0,0,0}\put(4951,-6211){\line( 1, 0){1350}}
}%
{\color[rgb]{0,0,0}\put(4501,-8011){\framebox(2250,1800){}}
}%
{\color[rgb]{0,0,0}\put(4501,-7411){\line( 1, 0){2250}}
}%
{\color[rgb]{0,0,0}\put(4501,-6811){\line( 1, 0){2250}}
}%
{\color[rgb]{0,0,0}\put(5251,-6061){\framebox(750,900){}}
}%
{\color[rgb]{0,0,0}\put(4951,-5611){\line( 1, 0){300}}
}%
{\color[rgb]{0,0,0}\put(6001,-5611){\line( 1, 0){300}}
}%
{\color[rgb]{0,0,0}\put(4951,-4411){\line( 1, 0){1350}}
}%
{\color[rgb]{0,0,0}\put(4951,-7261){\framebox(1350,3300){}}
}%
\put(4726,-2311){\makebox(0,0)[b]{\smash{\fontsize{9}{10.8}
\usefont{T1}{cmr}{m}{n}{\color[rgb]{0,0,0}$\neg P$}%
}}}
\put(5776,-4636){\makebox(0,0)[b]{\smash{\fontsize{9}{10.8}
\usefont{T1}{cmr}{m}{n}{\color[rgb]{0,0,0}$P\neg A$}%
}}}
\put(5476,-3061){\makebox(0,0)[b]{\smash{\fontsize{9}{10.8}
\usefont{T1}{cmr}{m}{n}{\color[rgb]{0,0,0}$PA$}%
}}}
\end{picture}%
\end{tabular}
}{
+--------------------------------+      +--------------------------------+
|                                |      |                                |
|                                |      |                                |
|                                |      |                                |
+--------------------------------+      +--------------------------------+
|       +--------------------+   |      |       +--------------------+   |
|       |                    |   |      |       |                    |   |
|       | P                  |   |      |       |                    |   |
+-------|--------------------|---+      +-------|                    |---+
|       |                    |   |      |       |                    |   |
|       |        +-------+   |   |      |       |        +-------+   |   |
|       |        |       |   |   |      |       |        |       |   |   |
+-------|--------|-------|---|---+      +-------|        |-------|   |---+
|       |        |  PA   |   |   |      |       |        |  PA   |   |   |
|       |        |       |   |   |  =>  |       |        |       |   |   |
|       |        +-------+   |   |      |       |        +-------+   |   |
+-------|--------------------|---+      +-------+                    +---+
|       |                    |   |              +--------------------+
|       |                    |   |              |                    |
|       |                    |   |              |                    |
+-------|--------------------|---+              +--------------------+
|       |                    |   |              |                    |
| -P    |                    |   |              |        +-------+   |
|       +--------------------+   |              |        |       |   |
+--------------------------------+              |--------|       |---|
|                                |              |        |       |   |
|                                |              |        |       |   |
|                                |              |        +-------+   |
+--------------------------------+      +-------+--------------------+---+
                                        |       |                    |   |
                                        |       |                    |   |
                                        |       |                    |   |
                                        +-------|--------------------|---+
                                        |       |                    |   |
                                        | -P    |                    |   |
                                        |       +--------------------+   |
                                        +--------------------------------+
                                        |                                |
                                        |                                |
                                        |                                |
                                        +--------------------------------+
}
\label{figure-uncontingent}
\hcaption{Uncontingent revision.}
\end{hfigure}

The conditional formula $P>A$ states that $A$ is believed in all conditions
meeting $P$: all models of $P\neg A$ are below all models of $PA$, not just its
most believed models $\min(PA)$. Coherently with minimal change, the models of
$\neg P$ stand fast. The change is exemplified in
Figure~\ref{figure-uncontingent}.

Coherently with minimal change, two situations compare the same if $P$ and $A$
are the same. Situations of $\neg P$ are unaffected. All cases of $P \neg A$
are below all cases of $PA$. Below all of them, not only the currently most
believed ones. Uncontingent revision is not the minimal change to believe $A$
in some case of $P$%
{}\footnote{Theorem\ref{natural-lesschange-uncontingent}
and Theorem~\ref{natural-strictly-lesschange-uncontingent}.},
it is the minimal change to believe $A$ in all cases of $P$%
{}\footnote{Theorem\ref{uncontingent-minimal}.}.

Uncontingent revision is correct on $\true > outside$, $\true > meow$ and
$\true > \neg outside$. Something meows even if Bruno's cat is not outside:
$\{\neg outside,meow\}$ is more believed than $\{\neg outside,\neg meow\}$. It
is above%
{}\footnote{Theorem~\ref{uncontingent-general}.}.

Uncontingent revision is also correct on $\true > outside$, $outside > cat$,
$\true > \neg outside$, where the current condition that Bruno's cat is
believed to be outside is given explicitely: $outside > cat$, not $\true >
cat$. A cat is under the pickup only if it Bruno's cat is outside:
$\{outside,cat\}$ is more believed than $\{outside,\neg cat\}$ and $\{\neg
outside,cat\}$ is not more believed than $\{\neg outside,\neg cat\}$%
{}\footnote{Theorem~\ref{uncontingent-context}.}.

Uncontingent revision does not restrict the new belief to the currently
believed conditions. The new belief $\true > meow$ is believed in all possible
situations. The new belief $outside > cat$ is believed in all possible
situations of $outside$, no matter if $outside$ is currently believed or not.

Natural revision limits the new beliefs to the currently believed conditions.
If the current belief is $outside$, that a cat is under the pickup $\true >
cat$ is the same as $outside > cat$. Still better, $\true > cat$ in the current
conditions only is $outside > cat$ uncontingently%
{}\footnote{Theorem~\ref{natural-nocondition} and
{} Theorem~\ref{uncontingent-context} with
{} $x=outside$ and $y=cat$ give the same order.}.
Contingently revising by $cat$ is the same as uncontingenty revising by
$outside > cat$ since $outside$ is the contingent context.

\subsection{The contingent context of natural revision}

The contingent context of natural revision is not just the presently most
believed conditions. That would make $\true > \neg x y$ impossible to believe
when $x$ is the case since it would be turned into $x > \neg x y$.

The contingent context of natural revision for $P>A$ is not just $\min(P)$, the
strongest beliefs. It depends on $A$ as well. It is not just $outside$. It
depends on $cat$. Only if $A$ is consistent with $\min(P)$, it is $\min(P)$.

If $A$ is inconsistent with $\min(P)$, the contingent context includes
$\min(P)$, but not only. It includes less and less strongly believed scenarios
of $P$. It does it until reaching $\min(PA)$. It is $P \leq \min(PA)$.

The contingent context of natural revisions is $P \leq \min(PA)$.

Formally, $C \nat(P>A) = C \unc(Q>A)$ with $Q = P \leq \min(PA)$%
{}\footnote{Theorem~\ref{contingent-context}.}.

Graphically, $\min(P)$ is added lower models of $P$ until $A$ is possible,
until a model satisfies $A$, as in Figure~\ref{figure-current}.

\begin{hfigure}
\setlength{\unitlength}{3750sp}%
\begin{picture}(2274,4824)(4489,-5623)
\thinlines
{\color[rgb]{0,0,0}\put(4501,-2011){\line( 1, 0){2250}}
}%
{\color[rgb]{0,0,0}\put(4501,-3811){\line( 1, 0){2250}}
}%
{\color[rgb]{0,0,0}\put(4501,-4411){\line( 1, 0){2250}}
}%
{\color[rgb]{0,0,0}\put(5401,-3661){\framebox(750,900){}}
}%
{\color[rgb]{0,0,0}\put(4501,-3211){\line( 1, 0){2250}}
}%
{\color[rgb]{0,0,0}\put(4501,-2611){\line( 1, 0){2250}}
}%
{\color[rgb]{0,0,0}\multiput(5101,-1561)(0.00000,-39.75904){42}{\line( 0,-1){ 19.880}}
}%
{\color[rgb]{0,0,0}\multiput(5251,-1561)(0.00000,-39.75904){42}{\line( 0,-1){ 19.880}}
}%
{\color[rgb]{0,0,0}\multiput(5401,-1561)(0.00000,-39.75904){42}{\line( 0,-1){ 19.880}}
}%
{\color[rgb]{0,0,0}\multiput(5551,-1561)(0.00000,-39.75904){42}{\line( 0,-1){ 19.880}}
}%
{\color[rgb]{0,0,0}\multiput(5701,-1561)(0.00000,-39.75904){42}{\line( 0,-1){ 19.880}}
}%
{\color[rgb]{0,0,0}\multiput(5851,-1561)(0.00000,-39.75904){42}{\line( 0,-1){ 19.880}}
}%
{\color[rgb]{0,0,0}\multiput(5176,-1561)(0.00000,-39.75904){42}{\line( 0,-1){ 19.880}}
}%
{\color[rgb]{0,0,0}\multiput(5326,-1561)(0.00000,-39.75904){42}{\line( 0,-1){ 19.880}}
}%
{\color[rgb]{0,0,0}\multiput(5476,-1561)(0.00000,-39.75904){42}{\line( 0,-1){ 19.880}}
}%
{\color[rgb]{0,0,0}\multiput(5626,-1561)(0.00000,-39.75904){42}{\line( 0,-1){ 19.880}}
}%
{\color[rgb]{0,0,0}\multiput(6076,-1561)(0.00000,-39.75904){42}{\line( 0,-1){ 19.880}}
}%
{\color[rgb]{0,0,0}\multiput(6226,-1561)(0.00000,-39.75904){42}{\line( 0,-1){ 19.880}}
}%
{\color[rgb]{0,0,0}\multiput(5026,-1561)(0.00000,-39.75904){42}{\line( 0,-1){ 19.880}}
}%
{\color[rgb]{0,0,0}\put(4501,-1411){\line( 1, 0){2250}}
}%
{\color[rgb]{0,0,0}\put(4501,-5611){\framebox(2250,4800){}}
}%
{\color[rgb]{0,0,0}\put(4501,-5011){\line( 1, 0){2250}}
}%
\thicklines
{\color[rgb]{0,0,0}\put(4951,-3211){\framebox(1350,1650){}}
}%
\thinlines
{\color[rgb]{0,0,0}\put(4951,-4861){\framebox(1350,3300){}}
}%
{\color[rgb]{0,0,0}\multiput(6151,-1561)(0.00000,-39.75904){42}{\line( 0,-1){ 19.880}}
}%
{\color[rgb]{0,0,0}\multiput(6001,-1561)(0.00000,-39.75904){42}{\line( 0,-1){ 19.880}}
}%
{\color[rgb]{0,0,0}\multiput(5926,-1561)(0.00000,-39.75904){42}{\line( 0,-1){ 19.880}}
}%
{\color[rgb]{0,0,0}\multiput(5776,-1561)(0.00000,-39.75904){42}{\line( 0,-1){ 19.880}}
}%
\put(5026,-4111){\makebox(0,0)[lb]{\smash{\fontsize{9}{10.8}
\usefont{T1}{cmr}{m}{n}{\color[rgb]{0,0,0}$P$}%
}}}
\put(5476,-3511){\makebox(0,0)[lb]{\smash{\fontsize{9}{10.8}
\usefont{T1}{cmr}{m}{n}{\color[rgb]{0,0,0}$A$}%
}}}
\put(4576,-2461){\makebox(0,0)[lb]{\smash{\fontsize{9}{10.8}
\usefont{T1}{cmr}{m}{n}{\color[rgb]{0,0,0}$\neg P$}%
}}}
\put(5776,-3061){\makebox(0,0)[b]{\smash{\fontsize{9}{10.8}
\usefont{T1}{cmr}{m}{n}{\color[rgb]{0,0,0}$\min(PA)$}%
}}}
\put(6301,-2911){\makebox(0,0)[lb]{\smash{\fontsize{9}{10.8}
\usefont{T1}{cmr}{m}{n}{\color[rgb]{0,0,0}$P(\leq\!\!\min(PA))$}%
}}}
\put(5926,-1861){\makebox(0,0)[b]{\smash{\fontsize{9}{10.8}
\usefont{T1}{cmr}{m}{n}{\color[rgb]{0,0,0}$\min(P)$}%
}}}
\end{picture}%
\nop{
+--------------------------------+
|                                |
|                                |
|                                |
+--------------------------------+
|       +====================+   |
| -P    ||        min(P)    ||   |
|       ||                  ||   |
+-------||------------------||---+
|       ||                  ||   |
|       ||       +-------+  ||P<=min(PA)
|       ||       |min(PA)|  ||   |
+-------|========+=======+===+---+
|       |        | A     |   |   |
|       |        +-------+   |   |
|       |                    |   |
+-------|--------------------|---+
|       |                    |   |
|       | P                  |   |
|       +--------------------+   |
+--------------------------------+
|                                |
|                                |
|                                |
+--------------------------------+
}
\label{figure-current}
\hcaption{The contingent context of natural revision.}
\end{hfigure}

Believing $A$ requires $A$ to be believable. The current context, the
conditions of the revision has to admit $A$ as a possibility. If the most
strongly believed models $\min(P)$ do not, some increasingly weakly believed
models are added until one satisfies $A$.

Even when believing in the current conditions only, believing $A$ implies that
$A$ is at least possible. It may not be possible in the most strongly believed
situations, but it is possible. Insisting on the most strongly believed
situations $\min(P)$ implies believing the impossible $\min(P)A$. Relaxing to
$P \leq \min(PA)$ is necessary to believe $A$.

The addition of $(\lequal\min(PA))$ as a condition links natural revision to
the withdrawal operators, with their bound to $(\lequal\min(A))$ on a revision
$A$. They differ in that the bound extends the range of their change instead of
limiting it.

\subsection{Otherwise}

Summarizing: natural revision accepts a new belief only under the current
conditions. Revising by $A$ is revising by $P>A$, where $P$ are the current
conditions. Since natural revision on propositional formulae hinges on minimal
change, natural revision on conditionals is based on minimal
changes~\cite{chan-boot-20}.

The change is minimal in the sense that it maintains the models in the same
relative order as much as possible: if $I$ is less or equal than $J$, it stays
so unless required to satisfy $P>A$%
{}\footnote{Theorem~\ref{natural-minimal}.}.

Yet, this conditionalized version of natural revision is not the only way to
implement minimal change~\cite{bout-gold-93,chan-boot-20} and not the only way
to implement conditionalized revision%
{}~\cite{hans-92,naya-etal-96,kern-99,kern-02,kern-02-a,kern-04}.
Numerical distances such as the number of comparisons $I \leq J$ changed give
different minima as well. Another alternative is to line $\neg P$ to $P \neg A$
instead of $PA$%
{}\footnote{Definition~\ref{down}.},
as shown in Figure~\ref{figure-down}.

\begin{hfigure}
\begin{tabular}{ccc}

\setlength{\unitlength}{3750sp}%
\begin{picture}(2274,6774)(4489,-7123)
\thinlines
{\color[rgb]{0,0,0}\put(5401,-2761){\framebox(750,450){}}
}%
{\color[rgb]{0,0,0}\put(4951,-2761){\line(-1, 0){450}}
\put(4501,-2761){\line( 0, 1){2400}}
\put(4501,-361){\line( 1, 0){2250}}
\put(6751,-361){\line( 0,-1){2400}}
\put(6751,-2761){\line(-1, 0){450}}
\put(6301,-2761){\line( 0, 1){1650}}
\put(6301,-1111){\line(-1, 0){1350}}
\put(4951,-1111){\line( 0,-1){1650}}
}%
{\color[rgb]{0,0,0}\put(4501,-2161){\line( 1, 0){450}}
}%
{\color[rgb]{0,0,0}\put(6301,-2161){\line( 1, 0){450}}
}%
{\color[rgb]{0,0,0}\put(4501,-961){\line( 1, 0){2250}}
}%
{\color[rgb]{0,0,0}\put(4501,-1561){\line( 1, 0){450}}
}%
{\color[rgb]{0,0,0}\put(6301,-1561){\line( 1, 0){450}}
}%
\put(5626,-2611){\makebox(0,0)[b]{\smash{\fontsize{9}{10.8}
\usefont{T1}{cmr}{m}{n}{\color[rgb]{0,0,0}$PA$}%
}}}
\put(4726,-1261){\makebox(0,0)[b]{\smash{\fontsize{9}{10.8}
\usefont{T1}{cmr}{m}{n}{\color[rgb]{0,0,0}$\neg P$}%
}}}
{\color[rgb]{0,0,0}\put(4501,-5311){\line( 1, 0){2250}}
}%
{\color[rgb]{0,0,0}\put(5401,-5161){\framebox(750,450){}}
}%
{\color[rgb]{0,0,0}\put(5401,-4561){\line(-1, 0){450}}
\put(4951,-4561){\line( 0, 1){1650}}
\put(4951,-2911){\line( 1, 0){1350}}
\put(6301,-2911){\line( 0,-1){1650}}
\put(6301,-4561){\line(-1, 0){150}}
\put(6151,-4561){\line( 0, 1){450}}
\put(6151,-4111){\line(-1, 0){750}}
\put(5401,-4111){\line( 0,-1){450}}
}%
{\color[rgb]{0,0,0}\put(4951,-3961){\line( 1, 0){1350}}
}%
{\color[rgb]{0,0,0}\put(4951,-3361){\line( 1, 0){1350}}
}%
{\color[rgb]{0,0,0}\put(4501,-7111){\framebox(2250,2400){}}
}%
{\color[rgb]{0,0,0}\put(4501,-6511){\line( 1, 0){2250}}
}%
{\color[rgb]{0,0,0}\put(4501,-5911){\line( 1, 0){2250}}
}%
{\color[rgb]{0,0,0}\put(4951,-6361){\framebox(1350,1650){}}
}%
\put(5776,-3586){\makebox(0,0)[b]{\smash{\fontsize{9}{10.8}
\usefont{T1}{cmr}{m}{n}{\color[rgb]{0,0,0}$P\neg A$}%
}}}
\end{picture}%
\nop{
+--------------------------------+
|                                |
|                                |
|                                |
+--------------------------------+
|       +--------------------+   |
| -P    |                    |   |
|       |                    |   |
+-------+                    +---+
|       |                    |   |
|       |        +-------+   |   |
|       |        |min(PA)|   |   |
+-------+        +-------+   +---+
                                                                              .
        +--------------------+
        |  P-A               |
        |                    |
        +--------------------+
        |  P-A               |
        |        +-------+   |
        |        |       |   |
        +--------+       +---+
                                                                              .
+-------+--------+-------+---+---+
|       |  P-A   |       |   |   |
|       |        +-------+   |   |
|       +--------------------+   |
+--------------------------------+
|                                |
|                                |
|                                |
+--------------------------------+
}

& ~ ~ ~ ~ ~ ~ ~ ~ &

%
%
\setlength{\unitlength}{3750sp}%
\begin{picture}(2274,6324)(4489,-7123)
\thinlines
{\color[rgb]{0,0,0}\put(4501,-5311){\line( 1, 0){2250}}
}%
{\color[rgb]{0,0,0}\put(4501,-5911){\line( 1, 0){2250}}
}%
{\color[rgb]{0,0,0}\put(5401,-5161){\framebox(750,450){}}
}%
{\color[rgb]{0,0,0}\put(4501,-3961){\line( 1, 0){450}}
}%
{\color[rgb]{0,0,0}\put(6301,-3961){\line( 1, 0){450}}
}%
{\color[rgb]{0,0,0}\put(4951,-3961){\line( 1, 0){1350}}
}%
{\color[rgb]{0,0,0}\put(4501,-1411){\framebox(2250,600){}}
}%
{\color[rgb]{0,0,0}\put(4501,-3361){\line( 1, 0){2250}}
}%
{\color[rgb]{0,0,0}\put(4951,-4561){\line(-1, 0){450}}
\put(4501,-4561){\line( 0, 1){1800}}
\put(4501,-2761){\line( 1, 0){2250}}
\put(6751,-2761){\line( 0,-1){1800}}
\put(6751,-4561){\line(-1, 0){450}}
\put(6301,-4561){\line( 0, 1){1650}}
\put(6301,-2911){\line(-1, 0){1350}}
\put(4951,-2911){\line( 0,-1){1650}}
}%
{\color[rgb]{0,0,0}\put(5401,-4561){\line(-1, 0){450}}
\put(4951,-4561){\line( 0, 1){1650}}
\put(4951,-2911){\line( 1, 0){1350}}
\put(6301,-2911){\line( 0,-1){1650}}
\put(6301,-4561){\line(-1, 0){150}}
\put(6151,-4561){\line( 0, 1){450}}
\put(6151,-4111){\line(-1, 0){750}}
\put(5401,-4111){\line( 0,-1){450}}
}%
{\color[rgb]{0,0,0}\put(5401,-2611){\framebox(750,450){}}
}%
{\color[rgb]{0,0,0}\put(4501,-7111){\framebox(2250,2400){}}
}%
{\color[rgb]{0,0,0}\put(4951,-6361){\framebox(1350,1650){}}
}%
{\color[rgb]{0,0,0}\put(4501,-6511){\line( 1, 0){2250}}
}%
\put(5626,-2461){\makebox(0,0)[b]{\smash{\fontsize{9}{10.8}
\usefont{T1}{cmr}{m}{n}{\color[rgb]{0,0,0}$PA$}%
}}}
\put(5776,-3586){\makebox(0,0)[b]{\smash{\fontsize{9}{10.8}
\usefont{T1}{cmr}{m}{n}{\color[rgb]{0,0,0}$P\neg A$}%
}}}
\put(4726,-3061){\makebox(0,0)[b]{\smash{\fontsize{9}{10.8}
\usefont{T1}{cmr}{m}{n}{\color[rgb]{0,0,0}$\neg P$}%
}}}
\end{picture}%
\nop{
+--------------------------------+
|                                |
|                                |
|                                |
+--------------------------------+
                                                                              .
                                                                              .
                                                                              .
                                                                              .
                                                                              .
                 +-------+
                 |min(PA)|
                 +-------+
                                                                              .
+--------------------------------+
|       +--------------------+   |
| -P    | P-A                |   |
|       |                    |   |
+-------|--------------------|---+
|       | P-A                |   |
|       |        +-------+   |   |
|       |        |       |   |   |
+-------|--------+       +---|---+
                                                                              .
+-------+--------+-------+---+---+
|       |  P-A   |       |   |   |
|       |        +-------+   |   |
|       +--------------------+   |
+--------------------------------+
|                                |
|                                |
|                                |
+--------------------------------+
}

\\

line $\neg P$ to $PA$

& &

line $\neg P$ to $P \neg A$

\end{tabular}
\label{figure-down}
\hcaption{Alternative ways of revising.}
\end{hfigure}

Both revisions minimally change the order to satisfy $P>A$%
{}\footnote{Theorem~\ref{natural-minimal} and Theorem~\ref{down-minimal}.}.
Which is right? What about revisions employing numerical distances instead?
What about other forms of revision by conditionals%
{}~\cite{hans-92,bout-gold-93,kern-99,kern-02,chan-boot-20}?

Each alternative is based on its own principles. Some hinge on minimal change.
Some evaluate a change in $I \leq J$ the same as a change in $I' \leq J'$. Some
do not. Each may be right in some situations and wrong in others.

Since natural revision comes from minimal change, conditionalizing it hinges on
minimal change. Since natural revisions does not equate changes in $I \leq J$ w
with changes in $I' \leq J'$, its conditionalized version hinges on
non-numerical minimality of change.

What about lining $\neg P$ to $PA$ rather than $P \neg A$? Where does it come
from? When is it right? Which principle justifies it? Some confuting
motivations come from the literature.

\begin{description}

\item[Revising by $P \rightarrow A$ first~\cite{chan-boot-20}]

\

Chandler and Booth~\cite{chan-boot-20} revise by $P>A$ by first revising by the
unconditional formula $P \rightarrow A$ and then minimally modifying the
resulting ordering if it still does not satisfy $P>A$. This procedure hinges
around $P \rightarrow A$. These are the models of $PA$ and the models of $\neg
P$. Maintaining them in the same order is coherent with this idea, if not a
consequence.

While revising by $P>A$ can be done by first revising by $P \rightarrow A$, it
may not. The unconditional formula $P \rightarrow A$ differs quite a lot from
the conditional $P > A$.
The models of $PA$ support $P>A$, the models of $\neg P$ do not, they are
irrelevant. They both satisfy $P \rightarrow A$, they are the same for $P
\rightarrow A$. They differ for $P>A$. The models of $PA$ are confirmed by
$P>A$, the models of $\neg P$ are not. Maintaining the order between the models
$\neg P$ and the models of $PA$ is treating unconfirmed scenarios like
confirmed ones: a model of $\neg P$ is unconfirmed by a revision, yet it
maintains its order with the confirmed ones, the models of $PA$.

\draft

chan-boot-20 also motivate the choice with an example; as told in the
introduction, this proves the first choice sensible in some cases, but not in
general

\enddraft

\item[Recalcitrance~\cite{naya-etal-03}]

\

Recalcitrance is accepting in full a sequence of two consistent revisions%
{}\footnote{Definition~\ref{recalcitrance}.}:
if $A \wedge B$ is consistent, then revising by $A$ by $B$ in sequence is the
same as revising by $A \wedge B$~\cite{naya-etal-03}.

Lining $\neg P$ to $PA$ satisfies it in the very limited case of initial total
ignorance and two revisions only%
{}\footnote{Theorem~\ref{natural-recalcitrant}.}.
Lining $\neg P$ to $P \neg A$ does not
satisfy it even in this case%
{}\footnote{Theorem~\ref{down-recalcitrant-not}.}.
None satisfies it with the ordering resulting
from even a single revision%
{}\footnote{Theorem~\ref{natural-recalcitrant-not}.}.

Recalcitrance motivates the first alignment, but very weakly: not only it holds
only in a very limited case, it has never been a central aim in belief
revision.

\item[Semantics of sets of
conditionals~\cite{pear-90,benf-etal-93,kern-ritt-10,komo-beie-20}]

\

Semantics for sets of conditionals try to produce an ordering of the models
that satisfies all conditional at the same
time~\cite{pear-90,benf-etal-93,kern-ritt-10,komo-beie-20}. They differ from
belief revision in that conditionals are given all at once rather than one at
time.

Yet, they order the conditionals of the set and try to satisfy them in
sequence. Differently from iterated revision, the order is not given but
produced from the set of conditionals themselves. They are sorted according to
how difficult they are to be satisfied: the easier first, the hardest last.

Disregarding how the sequence of conditionals is obtained, of whether it is
given or is derived, belief revision and semantics of conditions both try to
satisfy conditionals in some order.

Semantics of conditionals start with the flat initial order, where all
situations compare the same. All situations are at the top of the order.
Satisfying a conditional $P>A$ is achieved by lowering its contradicting
models, those satisfying $P$ but not $A$. Only these models change position:
models of $\neg P$ stay where they were. If they were at the top, they stay at
the top. If they were at some level of the ordering, they stay at that level.

This is the change to the order that satisfies the conditional while leaving
the largest room for the further changes needed for satisfying the following
formulae. Other changes satisfy the conditional, but would need to invalidate
it when later trying to satisfy another conditional.

\draft

The goal is to produce an ordering that satisfies all conditionals, if any
exists. A conditional $P > A$ is satisfied if the minimal models of $P$ all
satisfy $A$ and none satisfies $\neg A$. The initial, flat order does not have
this property unless $P \models A$, since the minimal models of $P$ in the
initial ordering are all models of $P$, including all satisfying $\neg A$. To
satisfy $P > A$, the models of $P \wedge \neg A$ are lowered, leaving only the
models of $P \wedge A$ at the top.

Lowering these models leaves the models of $\min(PA)$ above all models of $P
\neg A$, as required to satisfy $P > A$.

Why do the models of $\neg P$ stay at the top?

Moving them down is irrelevant to the satisfaction of $P > A$, but this is only
one step of the process. The following satisfy other conditionals $P' > A'$. If
satisfying them invalidates $P > A$, the process fails. Therefore, satisfying
$P' > A'$ cannot swap a strict comparison of two models: since the initial
order is flat, if $I$ is above $J$ it is because $J$ was moved down to satisfy
$P > A$. Swapping these two models invalidates $P > A$.

The models of $\neg P$ may include a model of $P'A'$. Leaving it at the top may
itself satisfy $P' > A'$. If not, the models of $P' \neg A$ at the top are
moved down. This only turns some equal comparisons $I \equiv J$ into a strict
order $I < J$: a model at the top with another is moved down. No strict order
is changed; models are never swapped. Satisfaction of $P > A$ is unharmed.

\enddraft

\end{description}

Three supports of the choice of lining the models of $\neg P$ with $PA$ come
from the literature. Yet, all three of them are weak: satisfing $P \rightarrow
A$ as a first step is a way to satisfy $P > A$ but not the only way;
recalcitrance is only met in a very limited case and is not a central goal of
belief revision anyway; not invalidating a previous revision has never been
either.

Lining $\neg P$ to $PA$ leaves models as low as possible in the ordering. Low
in the ordering means strongly believed. Natural revision leaves models low, it
maintains the strength of their belief. Models are disbelived only by
conflicting information. The next section discusses this guideline and its
rationale.

\section{The principle of naivety}
\label{naivety}

\draft

\begin{itemize}

\item le motivazioni date prima sono deboli, ma ne esiste una che viene da
belief revision: molte revisioni iniziano con i modelli al massimo livello di
credibilita' e usano le revisioni per escludere casi piu' che affermarli

\item questo si puo' chiamare principio di ingenuita', perche' quando non so
niente credo a tutto; quando non ho informazioni su una possibilita', ci credo
fino in fondo; in termini di revisione, metto i modelli piu' in alto possibile,
compatibilmente con le revisioni e con il principio di minimo cambiamento; non
si sta affermando che il principio di ingenuita' vale sempre, e ci sono infatti
dei controesempi in cui il risultato e' assurdo; viene invece detto che viene
di fatto usato da molte revisioni

\item piu' del principio di ingenuita', la revisione naturale segue il
principio di cambiamento minimo, con le due conseguenze di preservamento
condizionale e indifferenza;

\item alcune note su ingenuita' e indifferenza:

sono diversi: il secondo porta a confrontare certi modelli come uguali, il
primo a considerarne certi modelli piu' credibili possibile

\end{itemize}

\enddraft

\subsection{Previously}

Minimally changing an order to satisfy a conditional $P>A$ is achieved by
lining $(\equal\min(PA)) \neg P$ to $\min(PA)$. Three motivations come from the
literature, but they are weak. A stronger reason come from belief revision
itself:

\begin{quote}
\em

revisions exclude situations, rather than supporting them.

\end{quote}



A revision contradicting a model lowers the strength of its belief: this is a
move down in the order. The other models maintain their strength. If they are
strongly believed, they remain so. They do not go down in the order. If they
are the most strongly believed, they remain so.


\

\begin{tabular}{l}
$s$		=	Santa Claus exists \\
$\neg s$ 	=	Santa Claus does not exist
\end{tabular}

\draft
\enddraft

\

That Santa exists is as likely as he does not. The situation where he exists is
as likely as the situation where he does not. The models are equally likely.
After repeatedly spying my parents placing gifts under the Christmas tree, that
Santa exists becomes less and less likely. Yet, the belief that it does not
exist is as likely as before. Not more. It was minimal, it is minimal. It was
initially in class $0$, it is still in class $0$.

A similar example compares two contemporary situations. Having lost so many
years spying my parents, I could not check whether a pot of gold is at the end
of the rainbow.

\

\begin{tabular}{ll}
$s$		=	Santa Claus exists \\
$p$		=	there is a pot of gold at the end of the rainbow
\end{tabular}

\draft
\enddraft

\

The belief in $s$ decreases at each Christmas. The belief in $p$ does not. The
situation $\neg sp$ is as likely as $\neg s \neg p$. They are both maximally
likely. The pot of gold exists, this is as likely as that it does not. That it
exists is one of the most likely scenarios. Nothing is more likely than the
existence of the pot of gold.

This approach is not wrong in general. Some statements are simply unknown. They
may be, they may not. Maybe a gas station is within ten kilometers, maybe it
is not. Maybe the shop down the road is open, maybe it is not. Maybe it will
rain tomorrow, maybe it will not. Complete ignorance about something means that
it may or may not be. It is as likely as its negation. Both scenarios are fully
possible. The Santa and pot of gold examples do not invalidate this approach in
general, but show it is a form naivety: a naive approach to the unknown follows
the same line of translating ignorance into maximal likeliness.

Another example: I left my parents' home to become a mythical treasure hunter.
I am informed that, to cover my debts and not go bankrupt, I need to succeed in
my project to find the pot of gold at the end of the rainbow. Later, I am also
told that I need to find the Holy Grail. Finally, I learned that the amount of
my debt is so large I also need to receive my straitjacket made of gold fabric
I was promised by Santa Claus. I know everything will go as I expected, except
that some of these things may not exist.

\

\begin{tabular}l
Santa does not exist
	$\rightarrow$
	I go broken
\\
no pot of gold at the end of the rainbow
	$\rightarrow$
	I go broken
\\
the Holy Grail is a myth
	$\rightarrow$
	I go broken
\end{tabular}

\

The variables $g$ and $b$ respectively stand for the reality of the Holy Grail
and for me getting broken.

\begin{eqnarray*}
\neg s \rightarrow b \\
\neg p \rightarrow b \\
\neg g \rightarrow b
\end{eqnarray*}

I strongly believe I will not get broken. I am perfectly rational. It is a
perfectly rational opinion. The scenario $\{s, p, g, \neg b\}$ is likely:
Santa, the pot of gold and the Holy Grail all exist and I do not get broken.
Not only it is likely, it is maximally likely. It is as likely as $\{\neg s, p,
g, b\}$, where Santa does not exist, the other two things do and I get broken.
It is as likely as the Holy Grail being a myth or the pot of gold being a
legend. It is as likely as every other scenario, if not more. It may not be
certain, but nothing suggest it is not possible. No scenario is more likely.

Three revisions suggest I will get broken. Yet, none negate I will not if the
three things exist. That scenario remains maximally plausible. It does because
no revision directly contradicts it, not because a revision supports it. Not
even a single revision supports this conclusion. Not contradicting it is
tantamount to support it.

Like the previous examples, this one does not invalidate this naive approach of
believing everything until contradicted. No example does if naivety works in
other cases. It only shows that naivety is used by natural revision.

This is the main theme of this article: choices may not be just right or just
wrong; sometimes, they simply apply in different contexts. Rather than chasing
the perfect approach by following an infinite wake of counterexamples, the goal
is to uncover the applicability of each approach.

\subsection{The principle of naivety}

Accepting everything until contradicted is not an explicit principle of belief
revision, but many methods for revising follow it. The strength of belief in
situations do not decrease in lack of contrary information. An explicit
formulation of this principle:

\begin{quote}
\em

things are as likely as allowed by the available information.

\end{quote}

If strength of belief is formalized by an ordering between models, the
principle dictates that:

\begin{quote}
\em

models are as low in the order as allowed by the previous revisions.

\end{quote}

Connected preorders do not allow incomparability: in lack of any information
about their likeliness, all situations are equally likely: the rain tomorrow,
the gas station close, the shop open, but even Santa Claus, the pot of gold and
the Holy Grail.

Information contradicts some of them. The sun contradicts the rain. Seeing the
gas station closed contradicts it being open. Parents leaving gifts under the
Christmas tree contradict Santa Claus. Information excludes situations. It
weakens how much they are believed possible. Maybe the gas station closure sign
has been left by mistake. Maybe parents are just moving gifts left by Santa in
front of the fireplace. Still, the belief in a gas station weakens. The
the belief in Santa weakens. Revisions weakens the strength of beliefs.

No scenario will ever be more strongly believed than every one in case of total
ignorance. No matter how much information arrives, no matter how many revisions
support a situation, it will never be more strongly believed than that. It
never goes above level zero. Ignorance gives the strongest beliefs. This is the
principle of naivety. Stay naive. Believe everything. Doubt only when told.

It is not a general principle, and is not suggested to apply it recklessly.
Still, it is de facto adopted by many belief revision semantics, including
natural revision. Adopting it when extending from formulae to conditionals is a
natural choice.


\subsection{The guidelines of natural revisions}

Natural revision comes from the following principles, by definition or as a
matter of fact.

\begin{quote}

\begin{itemize}

\item Natural revision changes the ordering only as much as needed to make the
formula true%
{}\footnote{Theorem~\ref{natural-minimal}.}.

This is the {\em principle of minimal change}:
do not change mind if not necessary.

A consequence is {\em conditional preservation}%
{}\footnote{Definition~\ref{conditional-preservation}.}%
{}~\cite{kern-02,kern-02-a,kern-04,kern-etal-23}:
do not change mind on what unrelated to the revision%
{}\footnote{Theorem~\ref{natural-preserve}.}.

Another consequence is that natural revision equates models in absence of
information. Natural revision works on connected preorders, and models cannot
be incomparable in connected preorders. In lack of information supporting one
model or negating another, they cannot be strictly sorted; therefore, they can
only be equivalent: the strength of belief in them is the same. Natural
revision does not differentiate equivalent models if not necessary. It adheres
to the {\em principle of indifference} of probability
theory~\cite{keyn-21,shac-07} and belief merging~\cite{libe-22}: ignorance
gives equal likeliness. The rain is as likely as the sun. Santa is as likely as
no Santa. It differs from naivety in that it states equality, not maximal
likeliness. Revisions that may or may not differentiate equivalent models do
not. They respect the principle of equating unknown and equal.


\item Natural revision keeps models not contradicted by a revision low in the
ordering%
{}\footnote{Definition~\ref{strength}, Definition~\ref{more-naive} and
Theorem~\ref{unique}}.

Contradicting information lowers the strength of belief. Non-contradicting
information does not change it, exactly like confirming information.

This is the {\em principle of naivety}: believe everything, keep believing
everything until contradicted. Still believe everything as much as possible.

\end{itemize}

\end{quote}

\subsection{Natural revision on conditionals, from principles}

Minimal change keeps models in their order unless necessary to satisfy $P>A$.
Conditional preservation keeps models in their order if unrelated to $P>A$.
Two models that evaluate both $P$ and $A$ the same stay in the same order: two
models of $\neg P$, two models of $PA$ and two models of $P \neg A$. What
changes is the order between models of different groups: one of $\neg P$ and
one of $PA$, for example.

Satisfying $P > A$ requires $\min(PA)$ to be more likely than $P \neg A$.

The position of $\neg P$ is dictated by naivety: $P > A$ does not say anything
about the models of $\neg P$. It is all about the models of $P$. By naivety,
the models of $\neg P$ maintain their position in the order. A model of $\neg
P$ and a model of $PA$ stay in the same order by naivety: the first because it
is uninvolved by $P > A$, the second because it does not negate $P > A$.

The resulting order is fully determined by these principles. Natural revision
is the only maximally naive minimal change that preserve the conditionals%
{}\footnote{Theorem~\ref{unique}.}.

\subsection{Naivety vs. indifference}

Naivety may look similar to indifference, if not identical. Yet, they differ.

Indifference is equating the strength of belief in models in lack of support in
one. Naivety is believing in models as strongly as allowed. Believed the same
vs. maximally believed.

Before receiving any information regarding the models $I$, $J$, $K$ and $L$,
how much they are likely is unknown. Indifference says they are equally likely,
but does not say whether they are all likely or all unlikely. Naivety says
they are likely.

If $I$ is found out to be more likely than $J$, the order changes: $I < J$.
Indifference still says that $K$ is as likely as $L$, as before: $K=L$. Yet,
it does not say whether they are as likely as $I$ or as likely as $J$. It
allows both $I \equiv K \equiv L < J$ and $I < J \equiv K \equiv L$. Naivety
says to keep $K$ and $L$ low in the order, with $I$ and not with $J$.

\section{Lexicographic revision on conditionals}
\label{lexicographic-section}

\draft

\begin{itemize}

\item esistono almeno tre motivazioni per implementare la revisione con P>A
come revisione lessicografica con P->A

\item sembra un modo sensato, ma le motivazioni non sono generali

\item occorre capire quando si applicano, e cosa fare altrimenti

\item si puo' spiegare la revisione lessicografica con P->A come il risultato
di mettere il principio di ingenuita' prima di cambiamento minim

\item se non applico il principio di indifferenza ottengo un ulteriore tipo di
revisione, una forma intermedia fra uncontingent e lex(P->A)

\end{itemize}

\enddraft

An order $C$ verifies a conditional $P>A$ if lexicographically revised by $P
\rightarrow A$. This revision is supported by Nayak et al.~\cite{naya-etal-96}
and by Chandler and Booth~\cite{chan-boot-20}. Its rationale is:

\begin{enumerate}

\item it works on certain examples;

\item always makes $P>A$ true;

\item it preserves the order between every two models of $P \rightarrow A$.

\end{enumerate}

They are valid motivations in specific cases, but not in general.

That a revision mechanism works on certain examples proves that it works in
certain cases. The history of belief revision, and nonmonotonic reasoning in
general, is plagued by counterexamples disproving the general validity of
previous mechanisms and showing how well others work. An example by itself
proves that something works in certain cases, but does not even say which ones
they are.

Making $P>A$ true is the final destination of revision, but many paths lead to
it. Several just do not make much sense. For example, all orders verify $P>A$
if turned into the fixed order $[\neg P, PA, P\neg A]$. As well as $[PA, \neg
P, P\neg A]$ and $[PAx, PA\neg x, \neg P, xP\neg A, \neg xP \neg A]$ where $x$
is an arbitrary variable that is consistent with $PA$. All these ``revisions''
verify $P>A$. None is sensible. Arbitrarily meeting a certain condition is not
enough. This is the main theme of this article: do not ``prove'' that a
mechanism works. Show {\em when} it works. {\em Why} it works when it works.
Which principles it follows.

Lexicographically revising by $P \rightarrow A$ maintains the order between any
two models of $P \rightarrow A$. By the equivalence with to $\neg P \vee PA$,
it maintains the order within $\neg P$, within $PA$ and between them. Since
$P>A$ does not differentiate between two models of $\neg P$, revising by $P>A$
has no reason to change the order between them. Same for two models of $PA$.
What about a model of $\neg P$ and a model of $PA$? A conditional $P>A$
supports the latter; it does not support the former, just does not oppose it
either. Equating them is equating support and indifference. Is equating
confirmation and unrelatedness. Differentiang them, differentiating $\neg P$
from $PA$, is not only crucial to conditionals $P>A$, is the very reason why
conditionals they exist in the first place: $P \rightarrow A$ fails on
counterfactual statements because it fails to recognize that a false $P$ does
not say anything about $P>A$. A false $P$ supports $P \rightarrow A$ but not $P
> A$. It does not support ``if $P$ were the case, then $A$ would be as well''
at all. It is just irrelevant. The choice of keeping models of $\neg P$ in the
same order with the models of $PA$ makes sense when revising by $P>A$ hinges
around the revision by $P \rightarrow A$. The centrality of $P \rightarrow A$
justifies it. Yet, this is only a way to revise by $P>A$. Validity for a
mechanism does not prove validity in general.

That three motivations for revising $P>A$ by lexicographically revising by $P
\rightarrow A$ are not general may suggest that this mechanism is wrong. It is
not. The three motivations prove without doubt that the mechanism is valid. The
discussion above only shows that it is not general. It works in certain cases.
Which cases? The problem is not whether it works, is when it works.

\subsection{What lexicographic revision does}

A lexicographic revision $P \rightarrow A$ separates the models of $P
\rightarrow A$ from the models of $\neg(P \rightarrow A)$%
{}\footnote{Definition\ref{lexicographic}.}.

Lexicographic revision shifts all models of $P \rightarrow A$ above the others.
Equivalently, it shifts the models of $\neg (P \rightarrow A)$ below the
others, as shown in Figure~\ref{figure-lexicographic}.

\

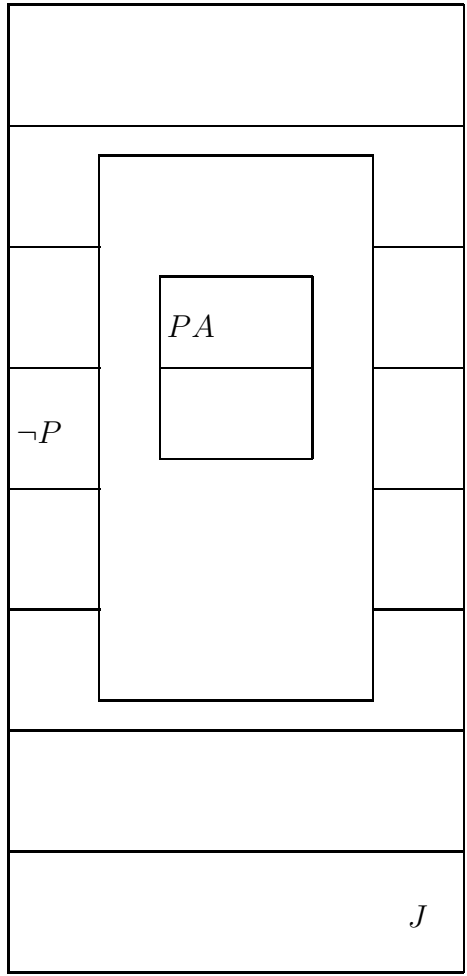
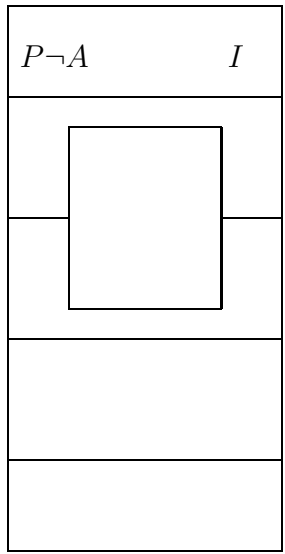
\begin{hfigure}
\long\def\ttytex#1#2{#1}
\ttytex{
\begin{tabular}{ccc}
%
%
\setlength{\unitlength}{3750sp}%
\begin{picture}(2274,4824)(4489,-5623)
\thinlines
{\color[rgb]{0,0,0}\put(4501,-2011){\line( 1, 0){2250}}
}%
{\color[rgb]{0,0,0}\put(4501,-2611){\line( 1, 0){2250}}
}%
{\color[rgb]{0,0,0}\put(4501,-3211){\line( 1, 0){2250}}
}%
{\color[rgb]{0,0,0}\put(4501,-3811){\line( 1, 0){2250}}
}%
{\color[rgb]{0,0,0}\put(4501,-4411){\line( 1, 0){2250}}
}%
{\color[rgb]{0,0,0}\put(5401,-3661){\framebox(750,900){}}
}%
{\color[rgb]{0,0,0}\put(4501,-1411){\line( 1, 0){2250}}
}%
{\color[rgb]{0,0,0}\put(4501,-5611){\framebox(2250,4800){}}
}%
{\color[rgb]{0,0,0}\put(4501,-5011){\line( 1, 0){2250}}
}%
{\color[rgb]{0,0,0}\put(4951,-4861){\framebox(1350,3300){}}
}%
\put(5626,-3061){\makebox(0,0)[b]{\smash{\fontsize{9}{10.8}
\usefont{T1}{cmr}{m}{n}{\color[rgb]{0,0,0}$PA$}%
}}}
\put(5101,-1861){\makebox(0,0)[b]{\smash{\fontsize{9}{10.8}
\usefont{T1}{cmr}{m}{n}{\color[rgb]{0,0,0}$P$}%
}}}
\put(5776,-2236){\makebox(0,0)[b]{\smash{\fontsize{9}{10.8}
\usefont{T1}{cmr}{m}{n}{\color[rgb]{0,0,0}$P\neg A$}%
}}}
\put(4726,-1711){\makebox(0,0)[b]{\smash{\fontsize{9}{10.8}
\usefont{T1}{cmr}{m}{n}{\color[rgb]{0,0,0}$\neg P$}%
}}}
\end{picture}%
&
\setlength{\unitlength}{3750sp}%
\begin{picture}(1104,2544)(5659,-5743)
\thinlines
{\color[rgb]{0,0,0}\multiput(6391,-3391)(-9.47368,4.73684){20}{\makebox(2.1167,14.8167){\tiny.}}
\put(6211,-3301){\line( 0,-1){ 45}}
\put(6211,-3346){\line(-1, 0){180}}
\put(6031,-3346){\line( 0,-1){ 90}}
\put(6031,-3436){\line( 1, 0){180}}
\put(6211,-3436){\line( 0,-1){ 45}}
\multiput(6211,-3481)(9.47368,4.73684){20}{\makebox(2.1167,14.8167){\tiny.}}
}%
\end{picture}%
&
\setlength{\unitlength}{3750sp}%
\begin{picture}(2274,8274)(4489,-9073)
\thinlines
{\color[rgb]{0,0,0}\put(4501,-5011){\line( 1, 0){2250}}
}%
{\color[rgb]{0,0,0}\put(5251,-7861){\framebox(750,900){}}
}%
{\color[rgb]{0,0,0}\put(4951,-6811){\line( 1, 0){1350}}
}%
{\color[rgb]{0,0,0}\put(4951,-8011){\line( 1, 0){1350}}
}%
{\color[rgb]{0,0,0}\put(4951,-8611){\line( 1, 0){1350}}
}%
{\color[rgb]{0,0,0}\put(5251,-3661){\framebox(750,900){}}
}%
{\color[rgb]{0,0,0}\put(5251,-3211){\line( 1, 0){750}}
}%
{\color[rgb]{0,0,0}\put(4501,-2611){\line( 1, 0){450}}
}%
{\color[rgb]{0,0,0}\put(4501,-3211){\line( 1, 0){450}}
}%
{\color[rgb]{0,0,0}\put(4501,-3811){\line( 1, 0){450}}
}%
{\color[rgb]{0,0,0}\put(4501,-4411){\line( 1, 0){450}}
}%
{\color[rgb]{0,0,0}\put(6301,-4411){\line( 1, 0){450}}
}%
{\color[rgb]{0,0,0}\put(6301,-3811){\line( 1, 0){450}}
}%
{\color[rgb]{0,0,0}\put(6301,-3211){\line( 1, 0){450}}
}%
{\color[rgb]{0,0,0}\put(6301,-2611){\line( 1, 0){450}}
}%
{\color[rgb]{0,0,0}\put(4951,-7411){\line( 1, 0){300}}
}%
{\color[rgb]{0,0,0}\put(6001,-7411){\line( 1, 0){300}}
}%
{\color[rgb]{0,0,0}\put(4501,-1411){\line( 1, 0){2250}}
}%
{\color[rgb]{0,0,0}\put(4951,-4861){\framebox(1350,3300){}}
}%
{\color[rgb]{0,0,0}\put(4501,-2011){\line( 1, 0){450}}
}%
{\color[rgb]{0,0,0}\put(6301,-2011){\line( 1, 0){450}}
}%
{\color[rgb]{0,0,0}\put(4501,-5611){\framebox(2250,4800){}}
}%
{\color[rgb]{0,0,0}\put(4951,-9061){\framebox(1350,3300){}}
}%
{\color[rgb]{0,0,0}\put(4951,-6211){\line( 1, 0){1350}}
}%
\put(6076,-6661){\makebox(0,0)[b]{\smash{\fontsize{9}{10.8}
\usefont{T1}{cmr}{m}{n}{\color[rgb]{0,0,0}$I$}%
}}}
\put(6526,-5386){\makebox(0,0)[b]{\smash{\fontsize{9}{10.8}
\usefont{T1}{cmr}{m}{n}{\color[rgb]{0,0,0}$J$}%
}}}
\put(5701,-6436){\makebox(0,0)[b]{\smash{\fontsize{9}{10.8}
\usefont{T1}{cmr}{m}{n}{\color[rgb]{0,0,0}$P\neg A$}%
}}}
\put(4726,-1711){\makebox(0,0)[b]{\smash{\fontsize{9}{10.8}
\usefont{T1}{cmr}{m}{n}{\color[rgb]{0,0,0}$\neg P$}%
}}}
\put(5476,-3061){\makebox(0,0)[b]{\smash{\fontsize{9}{10.8}
\usefont{T1}{cmr}{m}{n}{\color[rgb]{0,0,0}$PA$}%
}}}
\end{picture}%
\end{tabular}
}{
+--------------------------------+           +--------------------------------+
|                                |           |                                |
|                                |           |                                |
|                                |           |                                |
+--------------------------------+           +--------------------------------+
|       +--------------------+   |           |       +--------------------+   |
|       |                  I |   |           |       |                    |   |
|       | P-A                |   |           |       |                    |   |
+-------|--------------------|---+           +-------+                    +---+
|       |                    |   |           |       |                    |   |
|       |        +-------+   |   |           |       |        +-------+   |   |
|       |        |       |   |   |           |       |        |       |   |   |
+-------|--------|-------|---|---+           +-------+        +-------|   +---+
|       |        |  PA   |   |   |           |       |        |  PA   |   |   |
|       |        |       |   |   |           |       |        |       |   |   |
|       |        +-------+   |   |           |       |        +-------+   |   |
+-------|--------------------|---+ lex(PA) = +-------+                    +---+
|       |                    |   |           |       |                    |   |
|       |                    |   |           |       |                    |   |
|       |                    |   |           |       |                    |   |
+-------|--------------------|---+           +-------+                    +---+
|       |                    |   |           |       |                    |   |
| -P    |                    |   |           | -P    |                    |   |
|       +--------------------+   |           |       +--------------------+   |
+--------------------------------+           +--------------------------------+
|                                |           |                                |
|                                |           |                                |
|                                |           |                                |
+--------------------------------+           +--------------------------------+
|                          J     |           |                          J     |
|                                |           |                                |
|                                |           |                                |
+--------------------------------+           +--------------------------------+
                                                     +--------------------+
                                                     |                  I |
                                                     | P-A                |
                                                     +--------------------+
                                                     |                    |
                                                     |        +-------+   |
                                                     |        |       |   |
                                                     +--------|       |---+
                                                     |        |       |   |
                                                     |        |       |   |
                                                     |        +-------+   |
                                                     +--------------------+
                                                     |                    |
                                                     |                    |
                                                     |                    |
                                                     +--------------------+
                                                     |                    |
                                                     |                    |
                                                     +--------------------+
}
\label{figure-lexicographic}
\hcaption{Lexicographic revision.}
\end{hfigure}

\

Every model of $PA$ is above every model of $P \neg A$. In every situation
where $P$ holds, $A$ is more likely than $\neg A$. Every situation, not just
the currently likely ones. Lexicographic revision is uncontingent.

The figure shows more than meets the eye. The models of $\neg (P \rightarrow
A)$ go down while the models of $P \rightarrow A \equiv \neg P \vee A$ stay
where they are. The models of $\neg P$ stay where they are.

This is necessary to ensure that $P \rightarrow A$ is more likely than its
negation in every possible situation, but is not necessary to $P>A$. The models
of $\neg P$ are irrelevant to $P>A$. They could go anywhere. They could stay
where they are. They could move down with $\neg (P \rightarrow A)$.
Lexicographic revision keeps them high.

Keeping irrelevant models high is naivety. Believe what undenied. Believe it as
much as what confirmed. Believe it as much as possible. Do not decrease the
strength of beliefs in situations unless told. Do not drop the models that are
unrelated to the revision $P>A$.

Lexicographic revision follows the principle of naivety.

Uncontingent revision follows it as well%
{}\footnote{It even extends lexicographic revision and reduces to it:
{} Theorem~\ref{lexicographic-uncontingent} and
{} Theorem~\ref{uncontingent-lexicographic}.}
Why does it differ, then?

Shifting the models of $\neg (P \rightarrow A)$ under all models of $\neg P$
violates the minimal change principle. In the figure, $I$ drops below $J$. This
is unnecessary to satisfy $P>A$ since $J$ is irrelevant to $P>A$. It is not
even useful. Yet, its position relative to $I$ changes. Minimal change forbids
gratuitous changes. Lexicographic revision does not minimally change the order%
{}\footnote{Theorem~\ref{natural-strictly-lesschange-lexicographic}.}.

Keeping $\neg P$ where it is adhere to naivety and violates minimal change.
Lexicographic revision prefers naivety over minimal change%
{}\footnote{Theorem~\ref{naive-chain}.}.
It prefers believing over not changing mind. The situation $J$ is less believed
than $I$. A revision $P>A$ does not imply anything about $J$. Minimal change
dictates it is still more strongly believed than $I$. Lexicographic revision
does not listen. It retains the strength of belief in $I$, even at the cost of
making it stronger than $J$. It keeps believing $I$. This is naivety over
minimal change.

\subsection{How to be naive}

Naivety wins over minimal change. Believing wins over not changing mind. If
this is where lexicographic revision comes from implementing naivety and
minimal change in this order should result in lexicographic revision.

The revision $P>A$ justifies $PA$ over $P \neg A$. Naivety justifies $\neg P$
over $P \neg A$: unconfirmed situations are firmly believed. So much so to
being treated as confirmed: they are more strongly believed by the excluded
situations $P \neg A$. Minimal change justifies the maintained order between
the models of $\neg P$, between those of $PA$ and between those of $P \neg A$.
Lexicographic revision comes from naivety first and minimal change later.

\section{Related work}
\label{related}

Various authors studied revision by conditionals rather than simple formulae,
and some did for natural revision in particular.

\subsection{Nayak et al., learning from conditionals, 1996}

By a detour from formulae and conditionals to probability distributions,
through the application of criteria on probability distributions and extension
of their revision mechanisms, Nayak et al.~\cite{naya-etal-96} derive a
revision by a conditional $P > A$. It coincides with lexicographically revising
by the simple formula $P \rightarrow A$.

Like in the present article, rather than trying to directly solve the problem,
they derive a solution from basic principles (coarsest distributions) and known
methods from related fields (updates of probability distributions). Unlike in
this article, they only arrive at lexicographic revision, which is only one
way to revise by a conditional.

\draft

naya-etal-96

nayak-etal-learning-from-conditionals.pdf

aggiornano un ordinamento con un altro

su un esempio, fanno vedere gli ordinamenti piu' grossolani che caratterizzano
-x e x>y, che sono [-x,x] e [x->y,x-y]; con la revisione lessicografica si
ottiene il risultato voluto

era gia' un problema noto quello di aggiornare una distribuzione di
probabilita' per rendere vera una certa probabilita' condizionale Pr(A|P); nel
caso Pr(P) != 0 e' semplicemente la condizionalizzazione bayesiana; il problema
si pone nel caso Pr(P) = 0

sempre nel caso Pr(P) != 0, la condizionalizzazione bayesiana si estende alla
cinematica probabilistica di Jeffrey, ma ci sono dei casi in cui non esiste una
soluzione unica come l'esempio di judy benjamin citato; di questi interessano i
principi piu' che i metodi o gli esempi, dato che quello che viene analizzato
qui e' il caso Pr(P) = 0 che non viene considerato da questi

nayak et al. trasportano il problema su belief revision

il problema e' analogo a revisione iterata; viene riassunto il meccanismo
lessicografico di aggiornamento di un ordinamento con un altro ordinamento:
x<=y e' definito come x<'y oppure x='y e x<=''y (definizione 2)

nel caso probabilistico era stato impiegato il principio di scegliere le
distribuzioni piu' grossolane, che corrisponde al principio di indifferenza:
probabilita' uguali se non c'e' un motivo per differenziarle

fanno solo vedere l'approccio sull'esempio di rivedere -x con x>y

- stabiliscono gli ordinamenti piu' grossolani che soddisfano -x e x>y

- scelgono quelli che meglio si adatta al significato di -x e x>y: viene fatto
  prima sui trinceramenti epistemici e poi sui sistemi di sfere

  * per -x e' facile: [-x,x]

  * per x>y non e' ovvio, perche' potrebbe essere [x->y,x-y] ma anche
    [xy,-xv-y]; entrambi gli ordinamenti garantiscono che y sia vero quando si
    rivede con x; pero' il secondo funziona soltanto quando rivedo con x;
    quindi la scelta ricade sul primo: [x->y,x-y]

  * esiste un altro ordinamento, ma non e' il piu' grossolano

  * conclusione: il problema e' quello di rivedere [-x,x] con [x->y,x-y]

- applicano la revisione lessicografica e verificano che il risultato e'
  corretto:

  * non cambia la base, che rimane -x

  * ma xy e' ora considerato piu' plausibile

- tutto questo viene tradotto sui sistemi di sfere; e' tutto uguale, salvo che
  vengono considerati anche altri due ordinamenti possibili per x>y, ma vengono
  scartati subito e comunque non sono i piu' grossolani

tutto il discorso viene fatto sull'esempio

fanno poi il confronto con boutilier e goldszmidt

\

confronto:

qui gia' si parte da un ordinamento, quindi tutto il problema di esprimere -x
come un condizionale non si pone

si puo' presumere che P>A venga espresso da [P->A,P-A] per qualsiasi
condizionale; e' l'estensione ovvia di quello che naya-etal fanno esprimento
x>y con [x->y,x-y]

il risultato e' quindi C lex(P->A)

non e' la stessa cosa di C unc(P->A):

per esempio, se alcuni modelli massimi soddisfano -P, soddisfano anche -PvA;
quindi diventano minori di tutti i modelli di P-A anche se sono fuori dal
contesto della revisione P e il cambiamento non serve per soddisfare P>A,
nemmeno restringendo l'insieme dei modelli di P dato che soddisfano -P

in generale, e' vero che l'ordine fra modelli di -P non cambia, dato che tutti
loro soddisfano -PvA; pero' vengono vengono alterati dei confronti < fra loro e
alcuni modelli di P senza che sia necessario per assicurare la validita' di
P>A, nemmeno in un sottocontesto (dato che sono modelli di -P, lo sono anche
restringendo P)

\enddraft

\subsection{Boutilier and Goldszmidt, Revision By Conditional Beliefs, 1993}

The policy of minimally modifying the ordering is extended from revision of
simple formulae to contraction and expansion by conditionals. Revision by a
conditional $P>A$ is the contraction of $P>\neg A$ followed by an expansion by
$P>A$~\cite{bout-gold-93}.

\begin{itemize}

\item The contraction of $P > \neg A$ is the minimal change in the order that
makes $P > \neg A$ false: $\min(PA)$ is raised to $\min(P \neg A)$ if not
already.

\item Expansion is commonly defined as the plain incorporation of a new belief
regardless of consistency; in this extension of natural revision it is actually
an adaptation that drops $\min(P \neg A)$ just under $\min(PA)$. Since it does
not move any other part of $P \neg A$, it may not make $P>A$ true. It does if
preceded by a contraction of $P > \neg A$.

\end{itemize}

This procedure is graphically exemplified as follows. The initial order is in
Figure~\ref{figure-initial}.

\

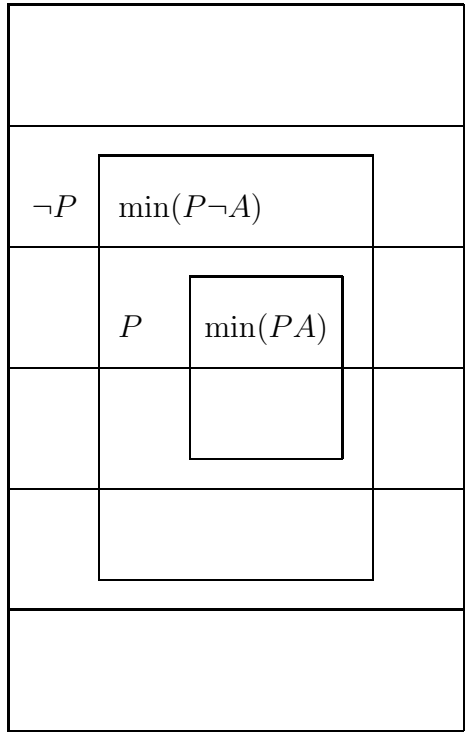
\begin{hfigure}
\setlength{\unitlength}{3750sp}%
\begin{picture}(2274,3624)(4489,-5023)
\thinlines
{\color[rgb]{0,0,0}\put(4501,-2011){\line( 1, 0){2250}}
}%
{\color[rgb]{0,0,0}\put(4501,-2611){\line( 1, 0){2250}}
}%
{\color[rgb]{0,0,0}\put(4501,-3211){\line( 1, 0){2250}}
}%
{\color[rgb]{0,0,0}\put(4501,-3811){\line( 1, 0){2250}}
}%
{\color[rgb]{0,0,0}\put(4501,-4411){\line( 1, 0){2250}}
}%
{\color[rgb]{0,0,0}\put(4951,-4261){\framebox(1350,2100){}}
}%
{\color[rgb]{0,0,0}\put(4501,-5011){\framebox(2250,3600){}}
}%
{\color[rgb]{0,0,0}\put(5401,-3661){\framebox(750,900){}}
}%
\put(4726,-2461){\makebox(0,0)[b]{\smash{\fontsize{9}{10.8}
\usefont{T1}{cmr}{m}{n}{\color[rgb]{0,0,0}$\neg P$}%
}}}
\put(5101,-3061){\makebox(0,0)[b]{\smash{\fontsize{9}{10.8}
\usefont{T1}{cmr}{m}{n}{\color[rgb]{0,0,0}$P$}%
}}}
\put(5776,-3061){\makebox(0,0)[b]{\smash{\fontsize{9}{10.8}
\usefont{T1}{cmr}{m}{n}{\color[rgb]{0,0,0}$\min(PA)$}%
}}}
\put(5401,-2461){\makebox(0,0)[b]{\smash{\fontsize{9}{10.8}
\usefont{T1}{cmr}{m}{n}{\color[rgb]{0,0,0}$\min(P\neg A)$}%
}}}
\end{picture}%
\nop{
+--------------------------------+
|                                |
|                                |
|                                |
+--------------------------------+
|       +--------------------+   |
|       |    min(P-A)        |   |
|       | P                  |   |
+-------|--------------------|---+
|       |                    |   |
|       |        +-------+   |   |
|       |        |min(PA)|   |   |
+-------|--------|-------|---|---+
|       |        |  PA   |   |   |
|       |        |       |   |   |
|       |        +-------+   |   |
+-------|--------------------|---+
|       |                    |   | 
|       |                    |   |
|       +--------------------+   | 
+--------------------------------+
|                                |
|                                |
|                                |
+--------------------------------+
}
\label{figure-initial}
\hcaption{The initial order.}
\end{hfigure}

\

Contracting $P > \neg A$ raises $\min(PA)$ to $\min(P \neg A)$ if not already,
as shown in Figure~\ref{figure-contraction}.

\

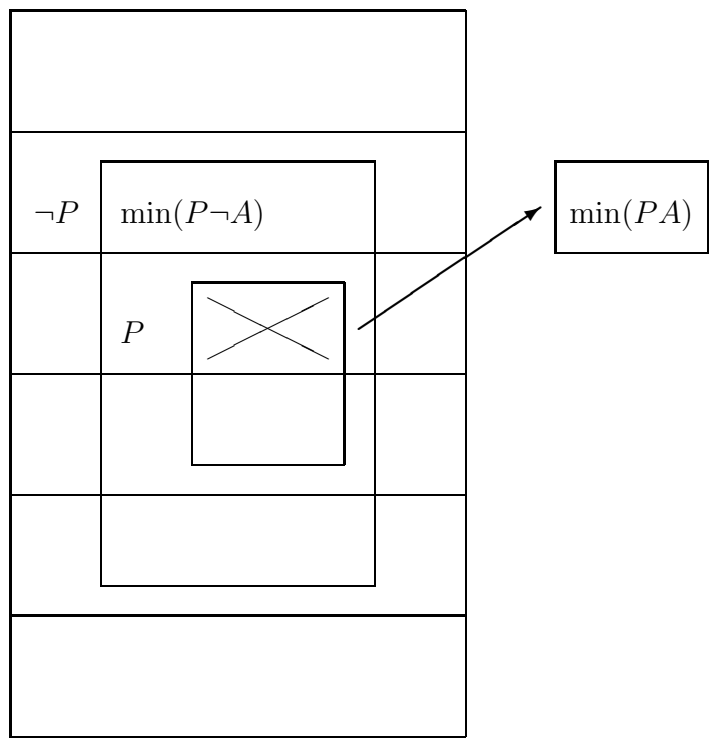
\begin{hfigure}
\setlength{\unitlength}{3750sp}%
\begin{picture}(3474,3624)(4489,-5023)
\thinlines
{\color[rgb]{0,0,0}\put(4501,-2011){\line( 1, 0){2250}}
}%
{\color[rgb]{0,0,0}\put(4501,-2611){\line( 1, 0){2250}}
}%
{\color[rgb]{0,0,0}\put(4501,-3211){\line( 1, 0){2250}}
}%
{\color[rgb]{0,0,0}\put(4501,-3811){\line( 1, 0){2250}}
}%
{\color[rgb]{0,0,0}\put(4501,-4411){\line( 1, 0){2250}}
}%
{\color[rgb]{0,0,0}\put(4951,-4261){\framebox(1350,2100){}}
}%
{\color[rgb]{0,0,0}\put(4501,-5011){\framebox(2250,3600){}}
}%
{\color[rgb]{0,0,0}\put(7201,-2611){\framebox(750,450){}}
}%
{\color[rgb]{0,0,0}\put(5401,-3661){\framebox(750,900){}}
}%
{\color[rgb]{0,0,0}\put(5476,-2836){\line( 2,-1){600}}
}%
{\color[rgb]{0,0,0}\put(5476,-3136){\line( 2, 1){600}}
}%
\thicklines
{\color[rgb]{0,0,0}\put(6226,-2986){\vector( 3, 2){900}}
}%
\put(4726,-2461){\makebox(0,0)[b]{\smash{\fontsize{9}{10.8}
\usefont{T1}{cmr}{m}{n}{\color[rgb]{0,0,0}$\neg P$}%
}}}
\put(5101,-3061){\makebox(0,0)[b]{\smash{\fontsize{9}{10.8}
\usefont{T1}{cmr}{m}{n}{\color[rgb]{0,0,0}$P$}%
}}}
\put(5401,-2461){\makebox(0,0)[b]{\smash{\fontsize{9}{10.8}
\usefont{T1}{cmr}{m}{n}{\color[rgb]{0,0,0}$\min(P\neg A)$}%
}}}
\put(7576,-2461){\makebox(0,0)[b]{\smash{\fontsize{9}{10.8}
\usefont{T1}{cmr}{m}{n}{\color[rgb]{0,0,0}$\min(PA)$}%
}}}
\end{picture}%
\nop{
+--------------------------------+
|                                |
|                                |
|                                |
+--------------------------------+
|       +--------------------+   |
|       |    min(P-A)        |   |  +-------+
|       | P                  |   |  |min(PA)|
+-------|--------------------|---+  +-------+
|       |                    |   |
|       |        +-------+   |   |
|       |        |XXXXXXX|   |   |
+-------|--------+-------+---|---+
|       |        |  PA   |   |   |
|       |        |       |   |   |
|       |        +-------+   |   |
+-------|--------------------|---+
|       |                    |   | 
|       |                    |   |
|       +--------------------+   | 
+--------------------------------+
|                                |
|                                |
|                                |
+--------------------------------+
}
\label{figure-contraction}
\hcaption{The contracted order.}
\end{hfigure}

\

Expansion by $P>A$ drops $\min(P \neg A)$, as in
Figure~\ref{figure-contraction}.

\

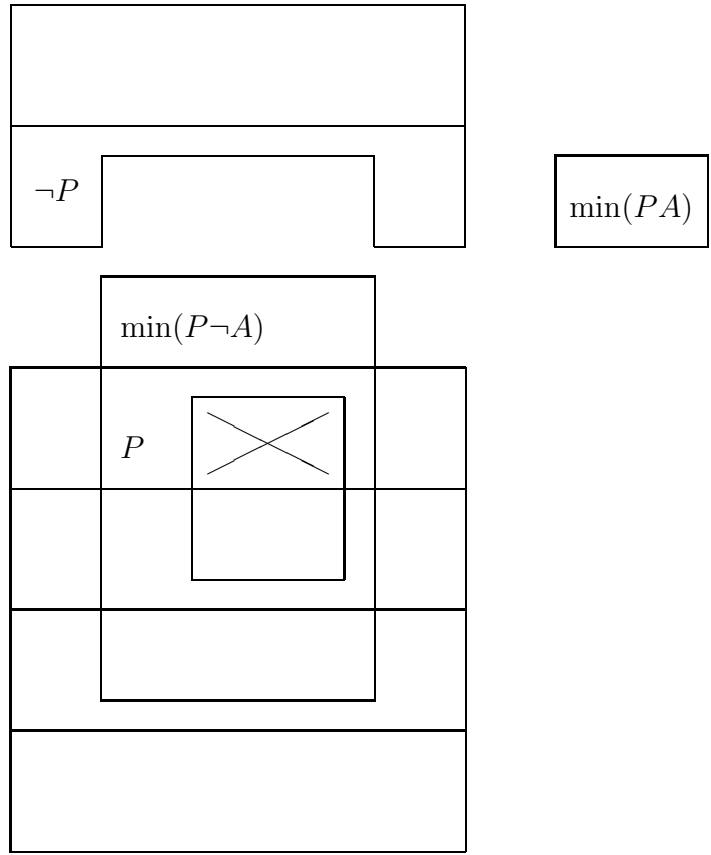
\begin{hfigure}
\setlength{\unitlength}{3750sp}%
\begin{picture}(3474,4224)(4489,-5023)
\thinlines
{\color[rgb]{0,0,0}\put(4501,-2611){\line( 1, 0){2250}}
}%
{\color[rgb]{0,0,0}\put(4501,-3211){\line( 1, 0){2250}}
}%
{\color[rgb]{0,0,0}\put(4501,-3811){\line( 1, 0){2250}}
}%
{\color[rgb]{0,0,0}\put(4501,-4411){\line( 1, 0){2250}}
}%
{\color[rgb]{0,0,0}\put(5401,-3661){\framebox(750,900){}}
}%
{\color[rgb]{0,0,0}\put(5476,-2836){\line( 2,-1){600}}
}%
{\color[rgb]{0,0,0}\put(5476,-3136){\line( 2, 1){600}}
}%
{\color[rgb]{0,0,0}\put(4501,-5011){\framebox(2250,2400){}}
}%
{\color[rgb]{0,0,0}\put(4951,-4261){\framebox(1350,2100){}}
}%
{\color[rgb]{0,0,0}\put(4951,-1561){\line( 0,-1){450}}
\put(4951,-2011){\line(-1, 0){450}}
\put(4501,-2011){\line( 0, 1){1200}}
\put(4501,-811){\line( 1, 0){2250}}
\put(6751,-811){\line( 0,-1){1200}}
\put(6751,-2011){\line(-1, 0){450}}
\put(6301,-2011){\line( 0, 1){450}}
\put(6301,-1561){\line(-1, 0){1350}}
}%
{\color[rgb]{0,0,0}\put(4501,-1411){\line( 1, 0){2250}}
}%
{\color[rgb]{0,0,0}\put(7201,-2011){\framebox(750,450){}}
}%
\put(5101,-3061){\makebox(0,0)[b]{\smash{\fontsize{9}{10.8}
\usefont{T1}{cmr}{m}{n}{\color[rgb]{0,0,0}$P$}%
}}}
\put(5401,-2461){\makebox(0,0)[b]{\smash{\fontsize{9}{10.8}
\usefont{T1}{cmr}{m}{n}{\color[rgb]{0,0,0}$\min(P\neg A)$}%
}}}
\put(7576,-1861){\makebox(0,0)[b]{\smash{\fontsize{9}{10.8}
\usefont{T1}{cmr}{m}{n}{\color[rgb]{0,0,0}$\min(PA)$}%
}}}
\put(4726,-1786){\makebox(0,0)[b]{\smash{\fontsize{9}{10.8}
\usefont{T1}{cmr}{m}{n}{\color[rgb]{0,0,0}$\neg P$}%
}}}
\end{picture}%
\nop{
+--------------------------------+
|                                |
|                                |
|                                |
+--------------------------------+
|       +--------------------+   |
|       |                    |   |  +-------+
|       |                    |   |  |min(PA)|
+-------+                    +---+  +-------+
                                                                             .
        +--------------------+
        |    min(P-A)        |
        |                    |
        +--------------------+
                                                                             .
+-------|--------------------|---+ 
|       |                    |   |
|       |        +-------+   |   |
|       |        |XXXXXXX|   |   |
+-------|--------|-------|---|---+
|       |        |  PA   |   |   |
|       |        |       |   |   |
|       |        +-------+   |   |
+-------|--------------------|---+
|       |                    |   | 
|       |                    |   |
|       +--------------------+   | 
+--------------------------------+
|                                |
|                                |
|                                |
+--------------------------------+
}
\label{figure-expansion}
\hcaption{The contracted and expanded order.}
\end{hfigure}

\

Overall, $\min(PA)$ raises to $\min(P \neg A)$, which drops just below it.

This is very different from all other revisions by conditionals analyzed in
this and in the other related articles. The principles it is based upon are
clear: minimal change, revision by contraction and adaptation. The cases where
it applies to are not, and deserve further investigation.

\draft

bout-gold-93

boutilier-goldszmidt-revision-by-conditional-beliefs.pdf

rivedono un ordine con un condizionale estendendo la revisione naturale, che e'
simile a quello che viene fatto in questo articolo; qui pero' ricostruiamo il
meccanismo partendo dai principi di base, perche' questo ci consentira' poi di
capire e correggere il problema (quando serve correggerlo perche' e' un
problema)

- contrazione con P>-A:
  si alza min(PA) al livello di min(P-A)
- espansione con P>A:
  si abbassa min(P-A) subito sotto, a un nuovo livello

come visto qui, non e' unico lo spostamento minimo che rende vero P>A

potrebbe essere lo spostamento che mantiene il piu' possibile la soddisfazione
di tutti gli altri condizionali P>A, ma questo non sembra essere dimostrato;
c'e' pero' un teorema che riguarda min(A)<=min(B) nell'ordine rivisto

e' diverso da come si estende qui la revisione naturale, dove i modelli di
min(PA) non cambiano il loro ordine con nessun altro modello tranne quelli di
P-A; sono solo alcuni modelli di P-A a spostarsi

la differenza potrebbe derivare dal fatto che questo sistema di revisione non
da' preferenza al mantenere < rispetto a mantenere = e/o non cerca di mantenere
l'ordine con i modelli di -P

---

dettagli tecnici:

contrazione C - P>A:
	- due modelli che non sono in min(P-A) mantengono il loro ordine
	- i modelli di min(P-A):
	  * diventano minori o uguali a tutti quelli maggiori o uguali a
	    un modello di min(PA)
	  * diventano maggiori o uguali a tutti quelli minori o uguali a
	    un modello di min(PA)

	riassumendo:
	i modelli di min(P-A) si spostano al livello di min(PA),
	il resto non cambia

	questo altera l'ordine anche se P>A gia' non e' vero

espansione C + P>A:
	- i modelli che non sono in min(P-A) mantengono l'ordine con gli altri
	- i modelli di min(P-A):
	  * rimangono uguali agli altri di min(P-A)
	  * se non sono maggiori o uguali a min(PA),
	    rimangono minori o uguali ai modelli che gia' lo erano

	riassumendo:
	alcuni confronti I<=J con I in min(P-A) si perdono, e basta
	min(P-A) diventa strettamente superiore al livello di min(PA),
	ma solo se prima min(P-A) era maggiore o uguale a min(PA)

	i modelli di min(P-A) si spostano subito sopra min(PA) se erano allo
	stesso livello

revisione C * P>A:
	- contrazione - P>-A
	  i modelli di min(PA) scendono al livello di min(P-A)
	- espansione + P>A
	  i modelli di min(P-A) salgono di un livello

non sembra venire dimostrato che i condizionali si mantengono il piu' possibile

pero' il teorema 6 caratterizza i condizionali nell'ordine rivisto

dimostra che il risultato e' univoco partendo da un insieme di condizionali

questo permette di definire una revisione di una base condizionale
{A>B, C>D} * E>F

dato che una base condizionale {A>B, C>D} puo' essere soddisfatta da vari
ordinamenti, si rivede ognuno; questo pero' crea il problema che per esempio
{A>B, C>D} * A>-B non e' {A>-B, C>D}, puo' non implicare C>D

per questo si divide la base in una parte consistente con il nuovo
condizionale, e questa viene preservata; viene definita la consistenza come
l'esistenza di un ordinamento in cui min(A) c= B per ogni A>B; in sostanza,
questo rende A>B almeno possibile; vengono poi definite delle condizioni
necessarie e sufficenti

\enddraft

\subsection{Hansson, In Defense of the Ramsey Test, 1992}

The core of this syntax-based mechanism is that a deductively-unclosed set of
formulae fully represents the doxastic state~\cite{hans-92}.

A mathematical structure generates a function from such a set. This function
selects maximal subsets consistent with the new, simple formula. For
conditionals, a similarity measure selects among the sets that satisfy $P>A$
the most similar to the original.

Contrary to all other revision mechanisms analyzed in this and related
articles, this one works on sets of formulae rather than semantical structures.
Equivalent sets are not the same. The doxastic state is the set itself, not any
of its semantically equivalent forms such as its set of models. This is very
different from the other approaches, which derive the set of formulae believed
true from a semantical representation.

\draft

hans-92

Hansson, Sven Ove "In Defense of the Ramsey Test," Journal of Philosophy, vol.
89, pp. 522-540, 1992

lo stato doxastico e' un insieme di formule K, non chiuso deduttivamente

le definizioni vengono date prima nella loro massima generalita', ma un esempio
e' questo: una struttura <B,f> permette di ottenere da K una funzione di
selezione g() da usare per selezionare un insieme di formule da K.-P, l'insieme
degli insiemi massimali di K che non implicano -P; intersecandoli e unendoli
con {P} si ottiene un insieme KoP

uso un simbolo diverso o per la revisione per evitare una ambiguita'; dato che
K e' si' un insieme di formule ma al tempo stesso uno stato doxastico, K*P
indica la revisione dello stato doxastico K con la formula P; questo viene
definito come KoP, dove K e' visto come un insieme di formule e quindi KoP e'
l'unione detta sopra

tutto questo definisce K*P quando P e' una formula indicativa, quindi senza >

il calcolo di K*(P>A) si fa con un sistema di similarita': si prendono gli
insiemi di formule K' che soddisfano P>A e che sono piu' simili possibile a K
secondo la misura |K'-K| - |K'\&K|

- il sistema permette quindi la revisione condizionale anche annidata
- e' sintattico, quindi diverso da quelli considerati qui

FARE: vedere se c'e' altro in:

- Iterated Descriptor Revision and the Logic of Ramsey Test Conditionals. J.  
  Philos. Log. 45(4): 429-450 (2016)

- Alternatives to the Ramsey Test. Computational Models of Rationality 2016:
  84-97
  INACCESSIBILE

\enddraft

\subsection{Chandler and Booth, Revision by Conditionals: From Hook to Arrow,
2020}

Revising by $P>A$ is revising by $P \rightarrow A$ followed by a minimal change
that makes $P>A$ true if not already~\cite{chan-boot-20}. The first step can be
every iterated revision mechanism.

For natural revision, the first step can be skipped without changing the final
result. This raises the question of why revising by $P \rightarrow A$ at all.
The answer is that this first step sets the direction of the revision, the
second merely completes it.

The minimal change that makes $P>A$ true is not unique. It does not identify
the position of the models of $\neg P$ that are in the class $\minidx(PA)$ or
above. Since $P \rightarrow A$ is central in this procedure, the choice is to
maintain the internal order of $P \rightarrow A$. Since $P \rightarrow A$ has
the models of $\neg P$ and the models of $PA$, they are kept together. A
concrete example shows a case where this choice works.


An example proves the correctness in some cases, not in general. Revising by
conditionals can be done in other ways than turning $P>A$ into $P \rightarrow
A$. The simple formula $P \rightarrow A$ is very different from $P>A$. The
difference is exactly the difference between the models of $\neg P$ and the
models of $PA$: unconfirmed but not denied vs. confirmed.

\draft

chan-boot-20

chandler-booth-revision-by-conditionals.pdf

riassume altre forme gia' esistenti

il risultato principale e' che si puo' ottenere un sistema di revisione con
condizionali a partire da un sistema di revisione normale, in due passi:

\begin{itemize}

\item si rivede l'ordinamento con P->A

\item si modifica minimalmente l'ordinamento perche' soddisfi P>A

\end{itemize}

si usa natural o lexi o qualsiasi altro modo per il primo passo

per il secondo si minimizza il numero di coppie diverse negli ordinamenti

si pone sempre il problema di dove collocare i modelli di -P; usando la
cardinalita' della differenza potrebbe venire univocamente il contrario, cioe'
che -P=min(PA) resti con con P-A=min(A) invece che con min(PA)

un caso in cui C nat(P->A) non soddisfa P>A e' quando min(P) = 0 < min(PA), per
esempio questo:

\

\setlength{\unitlength}{3750sp}%
\begin{picture}(2274,3024)(4489,-5023)
\thinlines
{\color[rgb]{0,0,0}\put(4501,-2611){\line( 1, 0){2250}}
}%
{\color[rgb]{0,0,0}\put(4501,-3211){\line( 1, 0){2250}}
}%
{\color[rgb]{0,0,0}\put(4501,-3811){\line( 1, 0){2250}}
}%
{\color[rgb]{0,0,0}\put(4501,-4411){\line( 1, 0){2250}}
}%
{\color[rgb]{0,0,0}\put(4951,-4261){\framebox(1350,2100){}}
}%
{\color[rgb]{0,0,0}\put(5401,-3661){\framebox(750,900){}}
}%
{\color[rgb]{0,0,0}\put(4501,-5011){\framebox(2250,3000){}}
}%
\put(4726,-2986){\makebox(0,0)[b]{\smash{\fontsize{9}{10.8}
\usefont{T1}{cmr}{m}{n}{\color[rgb]{0,0,0}$\neg P$}%
}}}
\put(4726,-2386){\makebox(0,0)[b]{\smash{\fontsize{9}{10.8}
\usefont{T1}{cmr}{m}{n}{\color[rgb]{0,0,0}$\neg P$}%
}}}
\put(5776,-3061){\makebox(0,0)[b]{\smash{\fontsize{9}{10.8}
\usefont{T1}{cmr}{m}{n}{\color[rgb]{0,0,0}$\min(PA)$}%
}}}
\put(5176,-2461){\makebox(0,0)[b]{\smash{\fontsize{9}{10.8}
\usefont{T1}{cmr}{m}{n}{\color[rgb]{0,0,0}$P\neg A$}%
}}}
\put(5176,-2986){\makebox(0,0)[b]{\smash{\fontsize{9}{10.8}
\usefont{T1}{cmr}{m}{n}{\color[rgb]{0,0,0}$P\neg A$}%
}}}
\end{picture}%
\nop{
+--------------------------------+
|       +--------------------+   |
|  -P   | P-A                |   |
|       |                    |   |
+-------|--------------------|---+
|       | P-A                |   |
|  -P   |        +-------+   |   |=min(PA)
|       |        |min(PA)|   |   |
+-------|--------|-------|---|---+
|       | P-A    |       |   |   |
|  -P   |        +-------+   |   |
|       |                    |   |
+-------|--------------------|---+
|       | P-A                |   |
|  -P   |                    |   |
|       +--------------------+   |
+--------------------------------+
|                                |
|  -P                            |
|                                |
+--------------------------------+
}

\

dato che P->A equivale a -Pv(PA), la revisione naturale nat(P->A) porta i
modelli di -P del livello 0 sopra quelli di P-A:

\

\setlength{\unitlength}{3750sp}%
\begin{picture}(2274,3624)(4489,-5023)
\thinlines
{\color[rgb]{0,0,0}\put(4501,-3211){\line( 1, 0){2250}}
}%
{\color[rgb]{0,0,0}\put(4501,-3811){\line( 1, 0){2250}}
}%
{\color[rgb]{0,0,0}\put(4501,-4411){\line( 1, 0){2250}}
}%
{\color[rgb]{0,0,0}\put(4951,-4261){\framebox(1350,2100){}}
}%
{\color[rgb]{0,0,0}\put(5401,-3661){\framebox(750,900){}}
}%
{\color[rgb]{0,0,0}\put(4501,-5011){\framebox(2250,2400){}}
}%
{\color[rgb]{0,0,0}\put(4951,-2011){\line(-1, 0){450}}
\put(4501,-2011){\line( 0, 1){600}}
\put(4501,-1411){\line( 1, 0){2250}}
\put(6751,-1411){\line( 0,-1){600}}
\put(6751,-2011){\line(-1, 0){450}}
\put(6301,-2011){\line( 0, 1){450}}
\put(6301,-1561){\line(-1, 0){1350}}
\put(4951,-1561){\line( 0,-1){450}}
}%
\put(5176,-2986){\makebox(0,0)[b]{\smash{\fontsize{9}{10.8}
\usefont{T1}{cmr}{m}{n}{\color[rgb]{0,0,0}$P\neg A$}%
}}}
\put(4726,-1786){\makebox(0,0)[b]{\smash{\fontsize{9}{10.8}
\usefont{T1}{cmr}{m}{n}{\color[rgb]{0,0,0}$\neg P$}%
}}}
\put(4726,-2986){\makebox(0,0)[b]{\smash{\fontsize{9}{10.8}
\usefont{T1}{cmr}{m}{n}{\color[rgb]{0,0,0}$\neg P$}%
}}}
\put(5776,-3061){\makebox(0,0)[b]{\smash{\fontsize{9}{10.8}
\usefont{T1}{cmr}{m}{n}{\color[rgb]{0,0,0}$\min(PA)$}%
}}}
\put(5176,-2461){\makebox(0,0)[b]{\smash{\fontsize{9}{10.8}
\usefont{T1}{cmr}{m}{n}{\color[rgb]{0,0,0}$P\neg A$}%
}}}
\end{picture}%
\nop{
+--------------------------------+
|       +--------------------+   |
|  -P   |                    |   |
|       |                    |   |
+-------+                    +---+
        +--------------------+
        | P-A                |
        |                    |
+-------|--------------------|---+
|       | P-A                |   |
|  -P   |        +-------+   |   |=min(PA)
|       |        |min(PA)|   |   |
+-------|--------|-------|---|---+
|       | P-A    |       |   |   |
|  -P   |        +-------+   |   |
|       |                    |   |
+-------|--------------------|---+
|       | P-A                |   |
|  -P   |                    |   |
|       +--------------------+   |
+--------------------------------+
|                                |
|  -P                            |
|                                |
+--------------------------------+
}

\

per verificare P>A serve che i modelli minimi di P soddisfino tutti PA

qui invece e' l'esatto contrario

i modelli di min(PA) vanno spostati sopra tutti quelli di P-A

ma come come si confrontano i modelli di -P dei livelli fra min(P) escluso e
min(PA) incluso non e' univoco:

\

\setlength{\unitlength}{3750sp}%
\begin{picture}(5124,4224)(4489,-5023)
\thinlines
{\color[rgb]{0,0,0}\put(4501,-3811){\line( 1, 0){2250}}
}%
{\color[rgb]{0,0,0}\put(4501,-4411){\line( 1, 0){2250}}
}%
{\color[rgb]{0,0,0}\put(4951,-4261){\framebox(1350,2100){}}
}%
{\color[rgb]{0,0,0}\put(5401,-3661){\framebox(750,900){}}
}%
{\color[rgb]{0,0,0}\put(4501,-5011){\framebox(2250,1800){}}
}%
{\color[rgb]{0,0,0}\put(4951,-2611){\line( 1, 0){1350}}
}%
{\color[rgb]{0,0,0}\put(4951,-1411){\line(-1, 0){450}}
\put(4501,-1411){\line( 0, 1){600}}
\put(4501,-811){\line( 1, 0){2250}}
\put(6751,-811){\line( 0,-1){600}}
\put(6751,-1411){\line(-1, 0){450}}
\put(6301,-1411){\line( 0, 1){450}}
\put(6301,-961){\line(-1, 0){1350}}
\put(4951,-961){\line( 0,-1){450}}
}%
{\color[rgb]{0,0,0}\put(5476,-2836){\line( 2,-1){600}}
}%
{\color[rgb]{0,0,0}\put(5476,-3136){\line( 2, 1){600}}
}%
{\color[rgb]{0,0,0}\put(5401,-2011){\framebox(750,450){}}
}%
{\color[rgb]{0,0,0}\put(7351,-2611){\framebox(450,600){}}
}%
{\color[rgb]{0,0,0}\put(7801,-2311){\line( 1, 0){1350}}
}%
{\color[rgb]{0,0,0}\put(9151,-2611){\framebox(450,600){}}
}%
\put(5176,-2986){\makebox(0,0)[b]{\smash{\fontsize{9}{10.8}
\usefont{T1}{cmr}{m}{n}{\color[rgb]{0,0,0}$P\neg A$}%
}}}
\put(5776,-1861){\makebox(0,0)[b]{\smash{\fontsize{9}{10.8}
\usefont{T1}{cmr}{m}{n}{\color[rgb]{0,0,0}$\min(PA)$}%
}}}
\put(4726,-1186){\makebox(0,0)[b]{\smash{\fontsize{9}{10.8}
\usefont{T1}{cmr}{m}{n}{\color[rgb]{0,0,0}$\neg P$}%
}}}
\put(5176,-2461){\makebox(0,0)[b]{\smash{\fontsize{9}{10.8}
\usefont{T1}{cmr}{m}{n}{\color[rgb]{0,0,0}$P\neg A$}%
}}}
\put(7576,-2386){\makebox(0,0)[b]{\smash{\fontsize{9}{10.8}
\usefont{T1}{cmr}{m}{n}{\color[rgb]{0,0,0}$\neg P$}%
}}}
\end{picture}%
\nop{
+--------------------------------+
|       +--------------------+   |
|  -P   |                    |   |
|       |                    |   |
+-------+                    +---+
                                                                             .
                 +-------+                  -P=min(PA)
                 |min(PA)|               +-------+                    +---+
                 +-------+               |       |                    |   |
        +--------------------+           |  -P   |....................|   |
        | P-A                |           |       |                    |   |
        |                    |           +-------+                    +---+
        +--------------------+
        | P-A=min(PA)        |
        |        +-------+   |
        |        |.......|   |
+-------|--------|-------|---|---+
|       | P-A    |       |   |   |
|  -P   |        +-------+   |   |
|       |                    |   |
+-------|--------------------|---+
|       | P-A                |   |
|  -P   |                    |   |
|       +--------------------+   |
+--------------------------------+
|                                |
|  -P                            |
|                                |
+--------------------------------+
}

\

le due strisce verticali di -P potrebbero restare con min(PA) o restare con
P-A=min(PA); entrambe le scelte mantengono massimamente le coppie dell'ordine

se pero' si guarda la cardinalita' potrebbe anche venire univocamente preferita
la scelta di tenere -P=min(PA) con P-A=min(PA); succede se per esempio in
P-A=min(PA) ci sono due modelli, in min(PA) uno e in -P=min(PA) uno

per ottenere lo stesso risultato ottenuto qui serve l'ulteriore vincolo che
l'ordine interno dei modelli di P->A non cambi; dato che P->A e' equivalente a
-P v PA, questo forza -P=min(PA) a restare con min(PA) dato che entrambi i
gruppi soddisfano -P v PA; questo vincolo di non cambiare l'ordine fra i
modelli di P->A non viene giustificata in generale:

"Plausibly, for instance, it should not occur at the expense of the worlds in
[[-A]]. In fact, it seems quite reasonable that, more broadly, the internal
ordering of [[A -> B]] should be left untouched"

avrei detto anch'io che l'ordine interno a -P va mantenuto perche' la revisione
e' limitata al contesto P, quindi tutto quello che e' fuori dal contesto non
deve cambiare

ma perche' P->A? il contesto e' P, perche' non deve cambiare l'ordine interno a
P->A? l'unica motivazione potrebbe essere e' che e' stata fatta prima una
revisione per P->A, ma e' stata fatta solo come passo intermedio

"Our principle adds to these the constraint that conditional revision by A>B
does not affect the relative standing of worlds in [[-A]] in relation to worlds
in [[A \& B]]. This further restriction yields the correct verdict in the
following scenario: [example]"

e' vero che funziona in quell'esempio, ma perche' non viene detto;
e non viene motivato in generale

anche perche' come si e' visto prima un singolo esempio non invalida niente, a
meno che non siamo sicuri di usare la revisione giusta per quell'esempio; meno
che mai un esempio positivo

quello che e' stato fatto qui invece e' ripartire dai principi di base e da
questi concludere che -P=min(PA) va con min(PA) e non con P-A=min(PA); questo
in termini di formule corrisponde a mantenere l'ordine di interno a P->A
perche' questo e' -PvA, che e' equivalente a -P v (PA); ma e' stato ottenuto
dai principi, non solo estendendo l'uso del contesto o su un esempio; dice che
il criterio proviene dall'applicare il principio di ingenuita' (naivety) quando
i due precedenti non danno una indicazione; questo chiarisce anche quando il
criterio non si applica, per esempio quando:

- l'uguaglianza I=J fra un modello I di -P=min(PA) e uno J di P-A=min(PA)
  deriva da una affermazione esplicita

- la analoga affermazione I=Z con Z in min(PA) e' per indifferenza

in questo caso I=J e' piu' forte di I=Z, quindi almeno questo modello I di
-P=min(PA) va con P-A=min(PA) e non con min(PA)

fra gli ordinamenti che tengono i modelli di -P con quelli di min(PA), la
definizione seleziona quelli che richiedono un minimo cambiamento come numero
di coppie I<=J diverse:

"But unfortunately, (S) [=ordine finale <=P>A soddisfa P>A] and (Ret1)
[=mantenimento ordine interno a P->A] aren't jointly sufficient to have <=P->A
determine <=P>A. Our suggestion is to close the gap by means of distance
minimisation. More specifically, we propose to consider the closest TPO that
satisfies--or, in the case of a tie, some aggregation of the closest TPOs that
satisfy--our two constraints. In terms of measuring the distance between TPOs,
a natural choice is the so-called Kemeny distance: [distanza uguale numero di
coppie differenti nei due ordinamenti]."

dovrebbe venire la stessa cosa che e' stata fatta qui

i casi in cui l'ordinamento dopo la revisione per P->A non soddisfa P>A e
quindi richiede la modifica e quindi la scelta sono due:

- 0 = min(P) = min(P-A) < min(PA)

- 0 < min(P) = min(P-A)

la differenza e' che nel primo caso i modelli minimi di P soddisfano tutti -A,
mentre per nel secondo lo fanno alcuni, ma non e' richiesto che lo facciano
tutti

il primo caso e' quello analizzato graficamente sopra

vediamo ora il secondo caso, per esempio proprio nel caso in cui ci siano anche
dei modelli di PA al minimo di P:

\

%
%
\setlength{\unitlength}{3750sp}%
\begin{picture}(2274,6324)(4489,-7123)
\thinlines
{\color[rgb]{0,0,0}\put(4501,-5311){\line( 1, 0){2250}}
}%
{\color[rgb]{0,0,0}\put(4501,-5911){\line( 1, 0){2250}}
}%
{\color[rgb]{0,0,0}\put(5401,-5161){\framebox(750,450){}}
}%
{\color[rgb]{0,0,0}\put(4501,-3961){\line( 1, 0){450}}
}%
{\color[rgb]{0,0,0}\put(6301,-3961){\line( 1, 0){450}}
}%
{\color[rgb]{0,0,0}\put(4951,-3961){\line( 1, 0){1350}}
}%
{\color[rgb]{0,0,0}\put(4501,-1411){\framebox(2250,600){}}
}%
{\color[rgb]{0,0,0}\put(4501,-3361){\line( 1, 0){2250}}
}%
{\color[rgb]{0,0,0}\put(4951,-4561){\line(-1, 0){450}}
\put(4501,-4561){\line( 0, 1){1800}}
\put(4501,-2761){\line( 1, 0){2250}}
\put(6751,-2761){\line( 0,-1){1800}}
\put(6751,-4561){\line(-1, 0){450}}
\put(6301,-4561){\line( 0, 1){1650}}
\put(6301,-2911){\line(-1, 0){1350}}
\put(4951,-2911){\line( 0,-1){1650}}
}%
{\color[rgb]{0,0,0}\put(5401,-4561){\line(-1, 0){450}}
\put(4951,-4561){\line( 0, 1){1650}}
\put(4951,-2911){\line( 1, 0){1350}}
\put(6301,-2911){\line( 0,-1){1650}}
\put(6301,-4561){\line(-1, 0){150}}
\put(6151,-4561){\line( 0, 1){450}}
\put(6151,-4111){\line(-1, 0){750}}
\put(5401,-4111){\line( 0,-1){450}}
}%
{\color[rgb]{0,0,0}\put(5401,-2611){\framebox(750,450){}}
}%
{\color[rgb]{0,0,0}\put(4501,-7111){\framebox(2250,2400){}}
}%
{\color[rgb]{0,0,0}\put(4951,-6361){\framebox(1350,1650){}}
}%
{\color[rgb]{0,0,0}\put(4501,-6511){\line( 1, 0){2250}}
}%
\put(5626,-2461){\makebox(0,0)[b]{\smash{\fontsize{9}{10.8}
\usefont{T1}{cmr}{m}{n}{\color[rgb]{0,0,0}$PA$}%
}}}
\put(5776,-3586){\makebox(0,0)[b]{\smash{\fontsize{9}{10.8}
\usefont{T1}{cmr}{m}{n}{\color[rgb]{0,0,0}$P\neg A$}%
}}}
\put(4726,-3061){\makebox(0,0)[b]{\smash{\fontsize{9}{10.8}
\usefont{T1}{cmr}{m}{n}{\color[rgb]{0,0,0}$\neg P$}%
}}}
\put(6001,-2536){\makebox(0,0)[b]{\smash{\fontsize{9}{10.8}
\usefont{T1}{cmr}{m}{n}{\color[rgb]{0,0,0}$J$}%
}}}
\put(4801,-3811){\makebox(0,0)[b]{\smash{\fontsize{9}{10.8}
\usefont{T1}{cmr}{m}{n}{\color[rgb]{0,0,0}$I$}%
}}}
\put(6076,-3211){\makebox(0,0)[b]{\smash{\fontsize{9}{10.8}
\usefont{T1}{cmr}{m}{n}{\color[rgb]{0,0,0}$K$}%
}}}
\end{picture}%
\nop{
+--------------------------------+
|                                |
|  -P                            |
|                                |
+--------------------------------+
|       +--------------------+   |
|  -P   | P-A    +-------+   |   |=min(PA)
|       |        |min(PA)|   |   |
+-------|--------|-------|---|---+
|       | P-A    |       |   |   |
|  -P   |        +-------+   |   |
|       |                    |   |
+-------|--------------------|---+
|       | P-A                |   |
|  -P   |                    |   |
|       +--------------------+   |
+--------------------------------+
|                                |
|  -P                            |
|                                |
+--------------------------------+
}

\

l'ordinamento non viene modificato dalla revisione per P->A, dato che 0 <
min(P) implica che tutti i modelli minimi dell'ordinamento gia' soddisfano -P e
quindi -PvA, che e' P->A

pero' il risultato continua ad avere dei modelli di P-A al livello minimo di P,
quindi non soddisfa P>A

per verificare P>A serve spostare i modelli di min(PA) sopra quelli di
P-A=min(PA), e questo pone ancora il problema di dove tenere i modelli di
-P=min(PA)

questo caso e' anche piu' critico del precedente per il sistema di chandler e
booth, dato che questa volta il cambiamento minimo e il tenere -Pv(PA)
inalterato devono fare tutta la revisione dell'ordine da soli, dato che la
precedente revisione per P->A non cambia niente

\

esiste anche un'altra differenza con quello fatto qui, ed e' la seconda scelta:
i modelli di -P restano con min(P-A) invece che con min(PA); nel loro sistema,
il primo passo solleva min(-P) sopra a tutto; il secondo passo solleva min(PA)
sopra P-A, ma lascia tutto il resto di -P in basso

\

%
%
\setlength{\unitlength}{3750sp}%
\begin{picture}(2274,6324)(4489,-7123)
\thinlines
{\color[rgb]{0,0,0}\put(4501,-5311){\line( 1, 0){2250}}
}%
{\color[rgb]{0,0,0}\put(4501,-5911){\line( 1, 0){2250}}
}%
{\color[rgb]{0,0,0}\put(5401,-5161){\framebox(750,450){}}
}%
{\color[rgb]{0,0,0}\put(4501,-3961){\line( 1, 0){450}}
}%
{\color[rgb]{0,0,0}\put(6301,-3961){\line( 1, 0){450}}
}%
{\color[rgb]{0,0,0}\put(4951,-3961){\line( 1, 0){1350}}
}%
{\color[rgb]{0,0,0}\put(4501,-1411){\framebox(2250,600){}}
}%
{\color[rgb]{0,0,0}\put(4501,-3361){\line( 1, 0){2250}}
}%
{\color[rgb]{0,0,0}\put(4951,-4561){\line(-1, 0){450}}
\put(4501,-4561){\line( 0, 1){1800}}
\put(4501,-2761){\line( 1, 0){2250}}
\put(6751,-2761){\line( 0,-1){1800}}
\put(6751,-4561){\line(-1, 0){450}}
\put(6301,-4561){\line( 0, 1){1650}}
\put(6301,-2911){\line(-1, 0){1350}}
\put(4951,-2911){\line( 0,-1){1650}}
}%
{\color[rgb]{0,0,0}\put(5401,-4561){\line(-1, 0){450}}
\put(4951,-4561){\line( 0, 1){1650}}
\put(4951,-2911){\line( 1, 0){1350}}
\put(6301,-2911){\line( 0,-1){1650}}
\put(6301,-4561){\line(-1, 0){150}}
\put(6151,-4561){\line( 0, 1){450}}
\put(6151,-4111){\line(-1, 0){750}}
\put(5401,-4111){\line( 0,-1){450}}
}%
{\color[rgb]{0,0,0}\put(5401,-2611){\framebox(750,450){}}
}%
{\color[rgb]{0,0,0}\put(4501,-7111){\framebox(2250,2400){}}
}%
{\color[rgb]{0,0,0}\put(4951,-6361){\framebox(1350,1650){}}
}%
{\color[rgb]{0,0,0}\put(4501,-6511){\line( 1, 0){2250}}
}%
\put(5626,-2461){\makebox(0,0)[b]{\smash{\fontsize{9}{10.8}
\usefont{T1}{cmr}{m}{n}{\color[rgb]{0,0,0}$PA$}%
}}}
\put(5776,-3586){\makebox(0,0)[b]{\smash{\fontsize{9}{10.8}
\usefont{T1}{cmr}{m}{n}{\color[rgb]{0,0,0}$P\neg A$}%
}}}
\put(4726,-3061){\makebox(0,0)[b]{\smash{\fontsize{9}{10.8}
\usefont{T1}{cmr}{m}{n}{\color[rgb]{0,0,0}$\neg P$}%
}}}
\put(6001,-2536){\makebox(0,0)[b]{\smash{\fontsize{9}{10.8}
\usefont{T1}{cmr}{m}{n}{\color[rgb]{0,0,0}$J$}%
}}}
\put(4801,-3811){\makebox(0,0)[b]{\smash{\fontsize{9}{10.8}
\usefont{T1}{cmr}{m}{n}{\color[rgb]{0,0,0}$I$}%
}}}
\put(6076,-3211){\makebox(0,0)[b]{\smash{\fontsize{9}{10.8}
\usefont{T1}{cmr}{m}{n}{\color[rgb]{0,0,0}$K$}%
}}}
\end{picture}%
\nop{
+--------------------------------+
|       +--------------------+   |
|  -P   |                    |   |
|       |                    |   |
+-------+                    +---+
                                                                             .
                 +-------+
                 |min(PA)|
                 +-------+
        +--------------------+
        | P-A                |
        |                    |
+-------+--------------------+---+
|       | P-A=min(PA)        |   |
|  -P   |        +-------+   |   |
|       |        |.......|   |   |
+-------|--------|-------|---|---+
|       | P-A    |       |   |   |
|  -P   |        +-------+   |   |
|       |                    |   |
+-------|--------------------|---+
|       | P-A                |   |
|  -P   |                    |   |
|       +--------------------+   |
+--------------------------------+
|                                |
|  -P                            |
|                                |
+--------------------------------+
}

\

la seconda scelta fatta qui non solleva alla posizione zero i modelli di
min(-P), che restano allineati con P-A; fra l'altro, in questo modo non si
capisce nemmeno perche' quella singola striscia di -P dovrebbe andare sopra,
specie considerato che sono una parte dei modelli indifferenti per P>A, l'unica
motivazione plausibile e' il mantenimento delle formule non condizionate in
questo caso in cui la condizione e' falsa

\

forse il risultato che la revisione naturale per P>A e' come una non
contingente per P<=min(PA) > A, ossia che la revisione naturale aggiunge
<=min(PA) al contesto si potrebbe anche ottenere dal loro teorema 1, che in
generale afferma questo:

- downset D(<=,PA) e' definito come <=min(PA)

- la revisione con P>A si ottiene anche rivedendo prima l'ordine iniziale <=
  con P->A usando la revisione considerata, e poi rivedendo con revisione
  lessicografica il risultato <=' con la formula D(<=',PA) \& (P->A)

nello specifico della revisione naturale, serve prima la revisione naturale di
<= con P->A e poi una revisione lessicografica 

questo non corrisponde alla loro revisione lessicografica estesa ai
condizionali, che rivede l'ordine iniziale con P->A e basta dato che il
risultato soddisfa P>A

nella sezione 3.4 viene dimostrato che:

- per la revisione naturale il primo passo della definizione non serve: non
  serve rivedere l'ordinamento iniziale per P->A, basta anche soltanto adattare
  quello per soddisfare P>A

- per la revisione lessicografica il secondo passo della definizione non serve,
  dato che il risultato della prima gia' soddisfa P>A

mettendo insieme tutti i pezzi forse si potrebbe ottenere lo stesso risultato,
ma non e' diretto; in particolare, comunque il teorema richede la prima
revisione naturale per P->A, anche se e' ridondante nella definizione di
revisione naturale per P>A

\

notare anche che chan-boot-20 non spiegano perche' effettuare prima la
revisione per P->A quando si sa benissimo che P->A e' molto diverso da P>A; nel
caso specifico della revisione naturale questo passo nemmeno serve, ma la loro
definizione comunque lo include

VERIFICARE: lo spiegano?

\

FARE: la revisione naturale puo' anche venire uguale, ma quella
uncontingente/lessicografica no

la loro revisione lessicografica coincide con la revisione con P->A; questo e'
dimostrato dall'equazione (V*), dal momento che per la revisione lessicografica
rivedere con P->A gia' soddisfa P>A

questo suggerisce che di fatto privilegiano il principio di ingenuita' a quello
di minimo cambiamento, visto che la revisione con P->A e' esattamente quello
che viene facendo questa scelta sempre: dato che -P e' irrilevante per rendere
vero P>A, lo metto in basso; e' quello che succede rivedendo con P->A: i
modelli di -P soddisfano P->A e quindi vanno sotto

\

FARE: vedere articolo seguente, dove ci sono controesempi all'uso di un
semplice ordinamento come stato doxastico

Booth, R., and Chandler, J. 2017. The irreducibility of iterated to single
revision. J. Philosophical Logic 46(4):405- 418.

FARE: anche questo:

"Accordingly, (Booth and Chandler 2018) propose a strengthening of the DP
postulates that is weak enough to avoid an identification of states with TPOs
and is consistent with the characteristic postulates of both *R and *L (albeit
not of *N ). They suggest associating states with structures that are richer
than TPOs: `proper ordinal interval (POI) as signments'.

Booth, R., and Chandler, J. 2018. On strengthening the logic of iterated belief
revision: Proper ordinal interval operators. KR-2018

\enddraft

\subsection{Kern-Isberner, Postulates for conditional belief revision, 1999}

The titular postulates hinge on specificity, based on the three classes a
conditional distinguishes models: verifying, falsifying and indifferent. The
first definition of specificity is that each class of a conditional is
contained in a class of another. The second is that the condition of one
entails one of the classes of the other~\cite{kern-99}.

A specific revision operator satisfying the postulates is given. It shifts all
models of $PA$ and all models of $P \neg A$, not only the most likely ones. It
is a form of uncontingent revision since it does not affect only the currently
most likely situations. Yet, the models of $PA$ may not all be more likely than
all of $P \neg A$. Natural revision also satisfies the postulates%
{}\footnote{Theorem~\ref{natural-postulates}.};
uncontingent revision does not%
{}\footnote{Theorem~\ref{uncontingent-postulates}.}.

The largest difference with the other revisions by conditionals is that OCFs
are quantitative, not qualitative like connected preorders. They do not just
say that a situation is more likely than another. They also say how much more.

\draft

postulates-for-conditional-revision-kern-isbner.pdf

P>A e' scritto A|P

Q>B <<= P>A	definito come	 QB c= PA
				Q-B c= P-A
	significa che il primo e' piu' specifico del secondo
			
Q>B ! P>A	definito come	Q c= PA or
				Q c= P-A or
				Q c= -P
	significa che il primo ha un contesto interno a un caso del secondo
		
postulati, riformulati sull'ordinamento:

(CR5): non cambia l'ordine interno a PA, interno a P-A, interno a -P

se I,J sono in PA,P-A:
	(CR6): I<J implica I<'J
	(CR7): I>'J solo se I>J
		ossia
	se sono gia' nell'ordine giusto, ci rimangono;
	se finiscono nell'ordine opposto, e' perche' gia' lo erano

viene dato un operatore specifico che soddisfa tutti i postulati
	- e' su OCF
	- diminuisce i modelli di PA di min(PA)
	- aumenta di uno i modelli di P-A se P-A<=PA

postulato ulteriore (CR8): se il condizionale Q>B e' piu' specifico di P>A e
non sono veri ne' P>A ne' Q>-B, allora e' Q>B e' vero dopo la revisione per
P>A; significa che non solo rimangono veri i condizionali piu' specifici, ma
che lo diventano se possibile

FARE: citare per il fatto che vogliamo mantenere gli ordini interni a PA e a
P-A, e che vogliamo alterare il resto solo se serve

FARE: citare in relazione a quando il principio di indifferenza NON si applica:
questa revisione afferma l'uguaglianza di livello, per esempio pone -P(=0)
uguale a min(PA)

\enddraft

\subsection{Kern-Isberner et al., Conditional preservation, 2002}

Kern-Isberner and others~\cite{kern-02,kern-02-a,kern-04,kern-etal-23} bound
conditional revision to move the models of $PA$ together, and the models of $P
\neg A$ together. This constraint emanates from the general principle of
conditional preservation. Shifting makes these revisions not bounded to the
current conditions but not uncontingent either. Like the revision satisfying
the postulates mentioned above, they are quantitative instead of qualitative
like the ones analyzed in the present and other articles. Yet, they are still
somehow an expression of minimal change, since they forbid revision to change
the order between certain models.

\draft

serie di articoli di Kern-Isberner e altri

\

{\em revisione singola non condizionata}

per rendere vero A, il livello di qualche modello I di A deve diminuire e/o il
livello di qualche modello J di -A aumentare; in generale, i livelli di I e J
possono cambiare, e possono cambiare in modo diverso:

-I c  A cambia di li livelli					\newline
-J c -A cambia di lj livelli

lo stesso vale per altri due modelli di A,-A:
possono cambiare di numero diverso di livelli li', lj'

il principio di preservamento condizionale dice che la differenza e' la stessa:

	li - lj = li' - lj'

i modelli di A translano rigidamente, e cosi' anche quelli di -A; per questo
vale una condizione equivalente che dice che il livello di un modello cambia di
una quota fissa che dipende solo dal valore di A nel modello

\

{\em revisione multipla}

nel caso di revisione con un insieme di formule C * {A1..Am}, il preservamento
condizionale e' uguale per ogni Ai, ma solo per i modelli che si comportano
nello stesso modo per tutte le altre formule Aj

la differenza di spostamento deve essere la stessa solo per coppie di modelli
di Ai,-Ai che valutano nello stesso modo tutte le altre formule Aj

si generalizza al numero di formule soddisfatte

la condizione equivalente diventa che il cambiamento di livello di un modello
e' lineare e dipendente solo dalla verita' delle formule Ai nel modello

\

{\em revisione condizionata}

nel caso condizionale non considero piu' modelli di A,-A ma modelli di PA,P-A

nel caso condizionale multiplo, si guardano solo le coppie di modelli che
valutano nello stesso modo gli altri condizionali, quindi le formule P'A'/P'-A'

[verificare]

\

{\em c-representation, c-revision}

la c-representation e' un ocf che rispetta l'indifferenza: il livello di un
modello e' una combinazione lineare delle PiAi e delle Pi-Ai che soddisfa

la c-revision modifica l'ocf di una combinazione lineare delle PiAi e delle
Pi-Ai che soddisfa

nel caso di formula singola, il livello di un modello diventa k(w) - k(A) + x,
dove x=0 se w soddisfa A, altrimenti e' un valore costante maggiore di
k(A)-k(-A)

\

{\em articoli}

\begin{itemize}

\item kern-02

Kern-Isberner02aJANCL02.pdf

Handling conditionals adequately in uncertain reasoning and belief revision

indifferenza condizionale: due modelli hanno lo stesso ranking se valutano
nello stesso modo un insieme di condizionali

si dimostra essere equivalente a: il rango di un modello w e' una combinazione
lineare di w(PiAi) e w(Pi-Ai)

preservamento condizionale: se due modelli si comportano nello stesso modo
rispetto ai condizionali, allora si spostano dello stesso numero di livelli

\

\item kern-02-a Kern-Isberner01fFOIKS02.pdf,
kern-04 A-1026110129951.pdf

The principle of conditional preservation in belief revision

A Thorough Axiomatization of a Principle of Conditional Preservation in
Belief Revision. Ann. Math. Artif. Intell. 40(1-2): 127-164 (2004)

revisione con insieme di condizionali

lo stato epistemico e' rappresentato da una funzione chiamata funzione di
valutazione condizionale, che associa a un insieme di formule e condizionali
un'algebra; e' una generalizzazione degli ocf, delle valutazioni
possibilistiche e delle distribuzione di probabilita'

i postulati generici della revisione iterata e quello di mantenimento della
indifferenza condizionale vengono espressi su questa rappresentazione astratta
dello stato epistemico

viene formulata la c-revision in questi termini

\

\item 1911.08833.pdf

A Conditional Perspective for Iterated Belief Contraction

la contrazione e' sempre per una formula, ma le sue proprieta' vengono espresse
in termini di contrazionali, che sono l'equivalente dei condizionali su
contrazione: A>>B significa che B e' vero dopo la contrazione di A, ossia che B
e' vero anche se A non fosse vero

\

\item 1-s2.0-S1570868316300921-main.pdf

What kind of independence do we need for multiple iterated belief change?

applicazione del principio a revisione multipla con formule semplici, relazione
con le altre condizioni di indipendenza che sono state proposte

\

\item kern-etal-22

Conditional Independence for Iterated Belief Revision

ijcai22-br-cond-independence-final.pdf

il preservamento della indipendenza condizionale viene espresso in termini dei
preordini totali, che non e' ovvio perche' non usano direttamente valori
numerici come le funzioni di ranking

la c-revision soddisfa questa condizione

si rivede per una formula, non per un condizionale

\

\item kern-etal-23

A kinematics principle for iterated revision

1-s2.0-S0004370222001679-main.pdf

principio di separazione, per cui revisioni su un argomento non devono cambiare
i condizionali su argomenti diversi

\

\end{itemize}

\

{\em come citare}

due modi: perche' si rivede per un condizionale, e come esempio di revisione
che pone I=J

il principio di preservamento condizionale si applica anche a revisione singola
non condizionale; pero' non e' un principio seguito da revisione naturale, dato
che questa cambia diversamente il livello di min(A) e del resto di A

dato che kern-02 rivede un ocf con un condizionale, e'
correlato; qui si segue la stessa linea di partire dalla semantica dei
condizionali (variante di system-z*) per ottenere dei principi da applicare a
revisione

non si ottengono le stesse cose perche' la revisione naturale sembra seguire
fedelemente gli stessi principi di system-z e non di system-z*, per questo la
similitudine non prosegue oltre

gli articoli della serie in cui la revisione e' per un condizionale o un
insieme di condizionali sono				\newline
- kern-02 (Kern-Isberner02aJANCL02.pdf)			\newline
- kern-02-a (Kern-Isberner01fFOIKS02.pdf)		\newline
- kern-04 (A-026110129951.pdf)

c'e' un articolo precedente dove vengono stabiliti dei postulati per la
revisione per un condizionale: kern-99

\

dato che la c-revision puo' porre I=J, e' un esempio in cui l'uguaglianza puo'
essere affermata e non discendere dal principio di indifferenza; su questo
l'articolo piu' rilevante sembra essere kern-etal-22,
ijcai22-br-cond-independence-final.pdf, in cui c'e' la definizione semplificata
di c-revision come k(w) - k(A) + x

\enddraft

\draft

\subsection{estensioni di system-z}

questi si possono citare per dire che esistono varianti di system-z; ma in ogni
caso sono semantiche per insiemi di condizionali; una condizione aggiuntiva che
viene imposta in alcuni e' il syntax splitting; non vale per revisione
naturale, ne' dovrebbe valere, perche' il contesto attuale puo' correlare due
variabili che non sono correlate da nessuna revisione; un altro e' il principio
di preservamento condizionale, che li pone vicini agli operatori incontingenti

{\em massima entropia}

system-z* (Goldszmidt, Morris, Pearl, 1993) e' una estensione di system-z che
permette di gestire certe situazioni come la ereditarieta' eccezionale e altro;
e' basato sul principio di massima entropia; viene poi generalizzato (Bourne,
Parsons, 1999); vedere per esempio:

"Of those which handle the more complex default interactions, such as
exceptional inheritance, correctly, the maximum entropy approach (me) has,
arguably, the clearest objective justification being derived from a well
understood principle of indifference" (Bourne, Parsons, 1999)

"The main weakness of z-entailment is its inability to sanction property
inheritance from classes to exceptional subclasses. For example, it will not
conclude that penguins have wings (p->w) by virtue of being birds (albeit
exceptional birds). The reason is that the Z-label 1 assigned to all rules
emanating from p amounts to proclaiming penguins an exceptional type of birds
in all respects, barred from inheriting any birdlike properties (e.g., laying
eggs, having beaks, etc.). This is a drawback that cannot be remedied by
methods based solely on the Z-ordering of defaults; a more refined ordering is
required which also takes into account the number of rules tolerating a
formula, not merely their rank orders. One such refinement is provided by the
maximum entropy approach [31] where each model is ranked by the sum of weights
on the rules falsified" (Goldszmidt, Pearl 1996)

"The main focus of this paper lies in augmenting system-Z with the capability
of handling richer types of input information, including variable-strength
rules (Section 3), indirect evidence (Section 5) and causal rules (Section 7)."
(Goldszmidt, Pearl 1996)

{\em generalizzazione}

l'insieme di condizionali p>a, p>b, pa>b ha soluzioni multiple in system-z* e
nella sua estensione (non in system-z, che fornisce solo [pab, resto])

questa cosa viene risolta nell'articolo: il rango di un condizionale non e'
specificato univocamente ma solo attraverso una disuguaglianza; questo fa
perdere l'unicita' della soluzione, ma viene dimostrato avere una sua
giustificazione teorica, il principio di indifferenza condizionale

{\em applicazione a revisione}

viene definito il principio di preservamento condizionale che la revisione
dovrebbe soddisfare

e' definito in termini generali, ma esiste una semplice condizione equivalente
per un ocf k() rivisto con un insieme di condizionali Pi>Ai: il rango di un
modello diminuisce/aumenta in proporzione ai condizionali soddisfatti/violati

tecnicamente, esistono un valore k0 e due valori ki+ e ki- per ogni
condizionale Pi>Ai e il rango di ogni modello diventa:

	k(I) + k0 + sum(ki+ | I |= PiAi) + sum(ki- | I |= Pi-Ai) 

a cosa serve:

"This example shows impressingly how the problem of irrelevant information can
be solved by c-revisions. But the principle of conditional preservation means
more than simply respecting irrelevance: By balancing numerical values
according to structural information, highly complex interactions of conditional
knowledge can be taken into regard and preserved under revision."

"the principle of conditional preservation, as defined here, can be shown to
cover the postulates for iterative belief revision of [Darwiche, Pearl, 1997]"

{\em articoli}

\begin{itemize}

\item 
	Goldszmidt, Morris, Pearl "A maximum entropy approach to non monotonic
	reasoning", IEEE Transactions on Pattern Analysis and Machine
	Intelligence, vol. 15, num. 3, 1993, p. 220-232.

\item 
	Goldszmidt, Pearl "Qualitative probabilities for default reasoning,
	belief revision, and causal modeling", Artificial Intelligence, 1996
		il sistema Z* viene usato per belief revision, ma la revisione
		e' sempre con una formula semplice, non con una formula
		condizionale

\item 
	Bourne, Parsons, "Maximum entropy and variable strength defaults",
	IJCAI'99

\end{itemize}

{\em syntax splitting}

un'altra ragione per cui il sistema z viene esteso e' che non soddisfa syntax
splitting; questo si aggiunge all'effetto di annegamento come motivazione alle
sue varianti

in particolare, partendo da T>x, T>y si ottiene un ordinamento [xy;resto];
quindi vale T>y ma non vale T-x>y: i due sottoalfabeti {x} e {y} partizionano i
condizionali in {T>x} e {T>y}, ma nonostante questo l'aggiunta della premessa
-x al condizionale soddisfatto T>y lo rende falso; questo controesempio e' in
letteratura: Example 4 in Lexicographic Entailment, Syntax Splitting and the
Drowning Problem, Heyninck, Kern-Isberner, Meyer

la condizione di syntax splitting si potrebbe tradurre sulla revisione nel modo
seguente: se rivedo con formule su alfabeti partizionati, le conclusioni sono
anche loro partizionate; stessa regola anche se i condizionali arrivano uno per
volta

revisione naturale non ha questa proprieta', come mostrato ancora dalla
sequenza di revisioni x, y, -x:

\begin{verbatim}
                  x    -xy -x-y    y    -xy -x-y   -x
-x-y -xy x-y xy  --->   x-y xy    --->    x-y     ---->    x-y
                                           xy               xy
                                                         -xy -x-y
\end{verbatim}

i due sottoalfabeti {x} e {y} partizionano le revisioni; syntax splitting
richiede che siano partizionate anche le conclusioni; ma T>y non vale mentre
Tx>y si'; l'aggiunta di x dovrebbe essere irrilevante per concludere y, ma non
lo e'

il fatto e' che syntax splitting non deve valere in questo caso; non deve
valere perche' e' erronea la premessa che {x} sia irrilevante per la revisione
y; al contrario, dato che y la affermo in un contesto in cui e' vero x, e' come
se affermassi x>y; quindi syntax splitting non deve valere perche' a livello
concettuale non e' valida la sua premessa

non importa se e' computazionalmente vantaggioso; se non si applica, non si
applica; non deve valere

probabilmente vale per uncontingent, verificare

\enddraft

\draft

\subsection{Liberatore, Mixed iterated revisions, 2023}

The idea that natural revision changes only the situations currently belived
possible is stated with not much details~\cite{libe-23}. The context is the
mutual presence of different kind of revisions on simple formulae.
Lexicographic revisions give general statements, natural revisions give
statements only contingently belived. The question what exactly a contingently
believed statement is not further analyzed. This is the done in the present
article.

\enddraft

\section{Conclusions}
\label{conclusions}


Natural revision does not believe unconditionally. It does not just believe
that something is true; it does only in the current conditions. It does not
just believe that a cat is under the rusted pickup; it does only because the
cat of Bruno is around there. It does not believe in the sound of meowing; it
does only when the cat of Bruno is around there. It is right on the cat under
the pickup; it is wrong on meowing. Still better, it applies to the former case
and does not apply to the latter.

Natural revision believes conditionally. It accepts that new information is
correct, but is still skeptical. That information may be correct, but it is
only in the current conditions~\cite{libe-23}. What believing in the current
conditions exactly means is the outcome of this article:

\begin{itemize}

\item natural revision applies when the new information is about the current
conditions only;

\item the current conditions are not what currently true; they are what
currently likely enough to accept the new information.

\end{itemize}

The second point is obvious in hindsight: information cannot be true in
situations contradicting it; it can only in situations consistent with it.
Consistency is a minimal requirement. Natural revision strengthens it by
further bounding to the most likely situations. New information does not apply
to all cases, it only applies to the most likely cases where it could apply,
the most likely cases consistent with it.

Technically, revising by $A$ is bounded by the condition $(\lequal\min(A))$:
the new piece of information $A$ does not apply all situations; it applies to
the most likely situations, but not only; it also applies to situations that
are unlikely enough to accept $A$.

Formalizing the condition $(\lequal\min(A))$ requires a detour to conditional
revision, where the new information is not ``$A$ is more likely than $\neg A$''
but ``whenever $P$ holds, $A$ is more likely than $\neg A$'', written $P>A$.
Natural revision further bounds such a statement to the most likely cases where
it could apply: $(\lequal\min(PA))$. It automatically switches from $P>A$ to $P
(\lequal\min(PA)) > A$.

Two side results of this detour stand by themselves.

When extending revision from simple formulae $A$ to conditionals $P>A$, the
principle of minimal change does not completely dictate how to revise. It does
not say how to change the strength of belief in certain situations
contradicting $P$; it changes, but it can change in various ways.

Natural revision is supported by the previous literature, but the motivations
are not general. A stronger support comes from a principle implicitely followed
by many revision mechanisms: believe until contradicted. Situations are
strongly believed until revisions negate them. This is naivety: believe the
unknown.

The principle of naivety, together with minimal change, says how to revise by a
conditional $P>A$. It stands as a side outcome of this article.

Another is a form of revision that does not limit the new information to the
current conditions only. The uncontingent revision defined by minimal change
and naivety allows to find the contingent context that natural revision
automatically adds to the condition of revisions.

This revision looks like lexicographic revision, but not quite. Lexicographic
revision still comes from minimal change and naivety, but in the opposite
order: first naivety, then minimal change.

\


This is where the analisys arrived. Where it could go is next.

Lexicographic revision differs from natural revisions in two ways. First, it
always favours $PA$ over $P \neg A$, not just in the current conditions.
Second, it favours naivety at the expenses of minimal change. Two steps from
natural revision: always, naivety. The first only leaves in uncontingent
revision. The second only? It would be a fourth possible way of revising:
favour naivety at the expenses of minimal change only in the current
conditions.

Natural revision was the first (or the second, depending on how revisions are
counted) iterated revision mechanism to be defined. Many others followed. Many
others differ in their aim, such as
withdrawal~\cite{maki-87,rott-pagn-99,ferm-rodr-98}. The same kind of analysis
applies to withdrawals: they come from certain basic principles, which could be
extended to conditionals.

Multiple revisions receive pieces of information in groups, as sets of
formulae. This is especially important for contingent-depending revisions such
as natural revision, since incorporating a formula at time changes the
contingent conditions at each step. Multiple conditional revision is not much
studied in the literature~\cite{kern-02-a,kern-04}.

Non-prioritized revision may not accept new information, or accept it only in
part~\cite{hans-99}. It could be extended from simple formulae to conditionals
as well.

The properties of natural and lexicographic revisions such as their
complexity~\cite{eite-gott-96,libe-97-c}, postulates%
~\cite{alch-gard-maki-85,darw-pear-97,pari-99,jin-thie-07,kern-99,kern-04},
syntactic characterization~\cite{alch-gard-maki-85,bout-96-a} and
expressiveness~\cite{libe-23} may extend from simple formulae to conditionals.

Connected preorders are a way to represent the likeliness of the possible
scenarios, but are not the only one. Another is Ordinal Conditional Functions,
functions from models to naturals~\cite{spoh-88}. Numbers make them
quantitative rather than qualitative like connected preorders. Rankings of
models roughly gauge their unlikeliness. A model ranked 1 is more likely than
one ranked 2, which is much more likely than one ranked 6. Connected preorders
can be expressed as sequences of non-intersecting sets, which are like OCFs,
but this is not a two-way road. The opposite direction loses the quantitative
information.
Revisions of OCFs may sensibly equate ranks of scenarios. Numbers allow them.
The difference between 1 and 2 is double that from 3 to 5; if revision raises a
3-model to 0, it may also raise the 5-model to 2. Equating it to the previous
2-model creates a new equality from new information.

\appendix

\section{Natural revision}
\label{natural-section}

\draft

\begin{itemize}

\item propositional natural revision

\item conditional natural revision

\item natural revision on the example

\item difference between two orders

\item closeness

\item natural revision minimally changes the ordering, in terms of closeness

\item conditional preservation

\item natural revision preserves the conditionals

\end{itemize}

\enddraft

%

The first step in the analysis of natural revision is its formal definition on
simple formulae~\cite{bout-96-a}.

\begin{definition}
\label{natural-propositional}

The {\em natural revision} of the order $[C(0),\ldots,C(m)]$ by a propositional
formula $A$ is as follows.

\long\def\ttytex#1#2{#1}
\ttytex{
\begin{eqnarray*}
\lefteqn{[C(0),\ldots,C(m)] \nat(A) = }			\\
&& [
\min(A),
C(0),
\ldots,
C(\minidx(A)-1),
C(\minidx(A)) \backslash A,
C(\minidx(A)+1),
\ldots,
C(m)
]							\\
\end{eqnarray*}
}{
[C(0) .. C(m)] nat(A) =
[
        C(min(A)) A
        C(0)          .. C(min(A)-1)
        C(min(A)) - A
        C(min(A)+1)   .. C(m)
]
}

\end{definition}

Natural revision extends to conditionals.

\begin{definition}
\label{natural}

The {\em natural revision} of the order $[C(0),\ldots,C(m)]$ by a conditional
formula $P>A$ is as follows.

\long\def\ttytex#1#2{#1}
\ttytex{
\begin{eqnarray*}
\lefteqn{[C(0),\ldots,C(m)] \nat(P>A) = }				\\
&& [ C(0),                 \ldots C(\minidx(P)-1),				\\
&& C(\minidx(P)) \backslash (P \neg A),
                           \ldots C(\minidx(PA)) \backslash (P \neg A),	\\
&& C(\minidx(P)) \cap (P \neg A), \ldots C(\minidx(PA)) \cap (P \neg A),\\
&& C(\minidx(PA)+1),       \ldots C(m)]
\end{eqnarray*}
}{
[C(0) .. C(m)] nat(P>A) =
[
        C(0)              .. C(min(P)-1)
        C(min(P)) - (P-A) .. C(min(PA)) - (P-A)
        C(min(P)) (P-A)   .. C(min(PA)) (P-A)
        C(min(PA)+1)      .. C(m)
]
}

\end{definition}

Natural revision only changes the ordering between $\minidx(P)$ and
$\minidx(PA)$, between the models of
{} $L = (\equal\min(P)) \cup \cdots \cup (\equal\min(PA))$.
Among these models, it sets $J<K$ and $I<K$ for all models $I$, $J$ and $K$ of
respectively $\neg PL$, $PAL$ and $P \neg AL$ since the classes of $P \neg A$
are after those of $\neg(P \neg A) = \neg P \vee PA$.

\

%

Natural revision on conditionals extends propositional natural revision:
$\true > A$ is the same as $A$.

\begin{theorem}
\label{natural-true}

For every connected preorder $C$ and propositional formula $A$, it holds
$C \nat(\true > A) = C \nat(A)$.

\end{theorem}

\proof

The claim is proved by the following chain of transformations.

\long\def\ttytex#1#2{#1}
\ttytex{
\begin{eqnarray*}
\lefteqn{[C(0) \ldots C(m)] \nat(\true>A) = }				\\
&& [
        C(0)
                                              \ldots C(\minidx(\true)-1)\\
&&      C(\minidx(\true)) \backslash (\true\neg A)
                       \ldots C(\minidx(\true A)) \backslash (\true\neg A) \\
&&      C(\minidx(\true)) (\true\neg A)
                       \ldots C(\minidx(\true A)) (\true\neg A)		\\
&&      C(\minidx(\true A)+1)                    \ldots C(m)
] =									\\
&& [
        C(0)            \ldots C(0-1)					\\
&&      C(0) \backslash (\neg A) \ldots C(\minidx(A)) \backslash (\neg A)\\
&&      C(0) (\neg A)   \ldots C(\minidx(A)) (\neg A)			\\
&&      C(\minidx(A)+1)    \ldots C(m)
] =									\\
&& [
        C(0) A        \ldots C(\minidx(A)) A				\\
&&      C(0) \neg A \ldots C(\minidx(A)) \neg A				\\
&&      C(\minidx(A)+1)  \ldots C(m)
] =									\\
&& [
        C(0) A        \ldots C(\minidx(A)-1) A, C(\minidx(A)) A		\\
&&      C(0) (\neg A) \ldots C(\minidx(A)-1) (\neg A), C(\minidx(A)) (\neg A)\\
&&      C(\minidx(A)+1)  \ldots C(m)
] =									\\
&& [
        0            \ldots 0, C(\minidx(A)) A				\\
&&      C(0)         \ldots C(\minidx(A)-1), C(\minidx(A)) (\neg A)	\\
&&      C(\minidx(A)+1) \ldots C(m)
] =									\\
&& [
        C(\minidx(A)) A,						\\
&&      C(0)           \ldots C(\minidx(A)-1),				\\
&&      C(\minidx(A)) - A,						\\
&&      C(\minidx(A)+1)   \ldots C(m)
] =									\\
&& [C(0) \ldots C(m)] \nat(A)
\end{eqnarray*}
}{
[C(0) .. C(m)] nat(true>A) =
[
        C(0)                    .. C(min(true)-1)
        C(min(true)) - (true-A) .. C(min(true A)) - (true-A)
        C(min(true)) (true-A)   .. C(min(true A)) (true-A)
        C(min(true A)+1)        .. C(m)
] =
[
        C(0)        .. C(0-1)
        C(0) - (-A) .. C(min(A)) - (-A)
        C(0) (-A)   .. C(min(A)) (-A)
        C(min(A)+1) .. C(m)
] =
[
        C(0) A      .. C(min(A)) A
        C(0) (-A)   .. C(min(A)) (-A)
        C(min(A)+1) .. C(m)
] =
[
        C(0) A      .. C(min(A)-1) A, C(min(A)) A
        C(0) (-A)   .. C(min(A)-1) (-A), C(min(A)) (-A)
        C(min(A)+1) .. C(m)
] =
[
        0           .. 0, C(min(A)) A                 no A < min(A)
        C(0)        .. C(min(A)-1), C(min(A)) (-A)    all -A < min(A)
        C(min(A)+1) .. C(m)
] =
        C(min(A)) A
        C(0)          .. C(min(A)-1)
        C(min(A)) - A
        C(min(A)+1)   .. C(m)
] =
[C(0) .. C(m)] nat(A)
}

\qed

\

%

The following two theorems prove that natural revision restricts the revision
$cat$ to the current situation $outside$: revising by $\true > cat$ is the same
as revising by $outside > cat$. The same for $meow$, meaning that really
revising by $\true > meow$ is impossible.

\begin{theorem}
\label{natural-nocondition}

Naturally revising
{} $[
{}	\{\{x,y\},\{x,\neg y\}\},
{}	\{\{\neg x,y\},\{\neg x,\neg y\}\}
{} ]$
by $y$ gives
{} $[
{}	\{\{x,y\}\},
{}	\{\{x,\neg y\}\},
{}	\{\{\neg x,y\},\{\neg x,\neg y\}\}
{} ]$.

\end{theorem}

\proof The order to be revised is named $C$. Its equivalence classes are the
following ones.

\begin{itemize}

\item $C(0) = \{\{x,y\},\{x,\neg y\}\}$;

\item $C(1) = \{\{\neg x,y\},\{\neg x,\neg y\}\}$.

\end{itemize}

Natural revision depends on $\minidx(A)$. Since $A=y$, a model of $C(0)$
satisfies it. Therefore, $\minidx(A) = 0$.

The result of natural revision is:

\long\def\ttytex#1#2{#1}
\ttytex{
\begin{eqnarray*}
\lefteqn{[C(0),\ldots,C(m)] \nat(A) = }			\\
&& [
\min(A),
C(0),                     \ldots C(\minidx(A)-1),
C(\minidx(A)) \backslash A,
C(\minidx(A)+1),          \ldots C(m)
] =							\\
&& [
C(0) y,
C(0),                     \ldots C(0-1),
C(0) \backslash y,
C(0+1),                   \ldots C(1)
] =							\\
&& [
\{\{x,y\},\{x,\neg y\}\} y,
\{\{x,y\},\{x,\neg y\}\} \backslash y,
\{\{\neg x,y\},\{\neg x,\neg y\}\}
] =							\\
&& [
\{\{x,y\}\},
\{\{x,\neg y\}\},
\{\{\neg x,y\},\{\neg x,\neg y\}\}
]
\end{eqnarray*}
}{
[C(0) .. C(m)] nat(A) =
[
        C(min(A)) A
        C(0)          .. C(min(A)-1)
        C(min(A)) - A
        C(min(A)+1)   .. C(m)
] =
[
        C(0) y
        C(0) C(0-1)
        C(0) - y
        C(1)
] =
[
        {{x,y},{x,-y}} y
        {{x,y},{x,-y}} - y
        {{-x,y},{-x,-y}}
] =
[
        {{x,y}}
        {{x,-y}}
        {{-x,y},{-x,-y}}
]
}
\qed

\begin{theorem}
\label{natural-condition}

Naturally revising
{} $[
{}	\{\{x,y\},\{x,\neg y\}\},
{}	\{\{\neg x,y\},\{\neg x,\neg y\}\}
{} ]$
by $x>y$ gives
{} $[
{}	\{\{x,y\}\},
{}	\{\{x,\neg y\}\},
{}	\{\{\neg x,y\},\{\neg x,\neg y\}\}
{} ]$.

\end{theorem}

\proof The order to be revised is named $C$. Its equivalence classes are the
following ones.

\begin{itemize}

\item $C(0) = \{\{x,y\},\{x,\neg y\}\}$;

\item $C(1) = \{\{\neg x,y\},\{\neg x,\neg y\}\}$.

\end{itemize}

Natural revision depends on $\minidx(P)$ and $\minidx(PA)$. They are both $0$
since $C(0)$ contains a model satisfying both.

\begin{itemize}

\item $\minidx(P) = \minidx(x) = 0$

\item $\minidx(PA) = \minidx(x \wedge y) = 0$

\end{itemize}

The result of natural revision is:

\long\def\ttytex#1#2{#1}
\ttytex{
\begin{eqnarray*}
\lefteqn{[C(0),\ldots,C(m)] \nat(P>A) = }				\\
&& [ C(0),                     \ldots, C(\minidx(P)-1),			\\
&& C(\minidx(P)) \backslash (P \neg A), \ldots, C(\minidx(PA)) \backslash (P \neg A),	\\
&& C(\minidx(P)) \cap (P \neg A), \ldots, C(\minidx(PA)) \cap (P \neg A),	\\
&& C(\minidx(PA)+1),              \ldots, C(m)] =				\\
&& [ C(0),                     \ldots, C(0-1),				\\
&& C(0) \backslash (P \neg A), \ldots, C(0) \backslash  (P \neg A),	\\
&& C(0) \cap (P \neg A),            \ldots, C(0) \cap (P \neg A),	\\
&& C(1),                       \ldots, C(1)] =				\\
&& [
   C(0) \backslash (P \neg A),						\\
&& C(0) \cap (P \neg A),						\\
&& C(1)] =								\\
&& [
   \{\{x,y\},\{x,\neg y\}\} \backslash (x \wedge \neg y),		\\
&& \{\{x,y\},\{x,\neg y\}\} \cap (x \wedge \neg y),			\\
&& \{\{\neg x,y\},\{\neg x,\neg y\}\}] =				\\
&& [
   \{\{x,y\}\}								\\
&& \{\{x,\neg y\}\}							\\
&& \{\{\neg x,y\},\{\neg x,\neg y\}\}]
\end{eqnarray*}
}{
[C(0) .. C(m)] nat(P>A) =
[
        C(0)              .. C(min(P)-1)
        C(min(P)) - (P-A) .. C(min(PA)) - (P-A)
        C(min(P)) (P-A)   .. C(min(PA)) (P-A)
        C(min(PA)+1)      .. C(m)
] =
[
        C(0)         .. C(0-1)
        C(0) - (P-A) .. C(0) - (P-A)
        C(0) (P-A)   .. C(0) (P-A)
        C(0+1)       .. C(1)
] =
[
        C(0) - (P-A)
        C(0) (P-A)
        C(1)
] =
[
        {{x,y},{x,-y}} - (x-y)
        {{x,y},{x,-y}} (x-y)
        {{-x,y},{-x,-y}}
] =
[
        {{x,y}}
        {{x,-y}}
        {{-x,y},{-x,-y}}
]
}
\qed

\

%

Revisions change a connected preorder, a set of pairs $I \leq J$. The set of
changed pairs gauges the amount of change.

\begin{definition}
\label{difference}

The difference between two orderings $C$ and $C'$ is the set of pairs of
models they compare differently:

\[
\diff(C,C') = (C \backslash C') \cup (C' \backslash C)
\]

\end{definition}

The distance between orderings is gauged by their difference, the set of pairs
they compare differently. The smaller this set, the closer the orderings. Set
containment measures this smallness, the vicinity between orderings%
{}~\cite{alch-gard-maki-85,sato-88,bout-96-a,delg-etal-13}.

\begin{definition}
\label{closer}

A connected preorder $C'$ is at least as close to $C$ than $C''$ if its
difference from $C$ is contained in that of $C''$:

\[
\diff(C',C) \subseteq \diff(C'',C)
\]

\end{definition}

Natural revision follows the principle of minimal change by changing the order
into one of its closest one among the ones satisfying $P>A$.

\begin{theorem}
\label{natural-minimal}

No connected preorder satisfying $P>A$ is strictly closer to $C$ than natural
revision $C \nat(P>A)$.

\end{theorem}

\draft

\[
\forall C, P>A ~.~
\not\exists C' ~.~
	C' \mbox{ satisfies } P>A ,~
	\diff(C',C) \subset \diff(C \nat(P>A),C)
\]

\enddraft

\proof The proof shows that every connected preorder $C'$ that satisfies $P>A$
is either $C \nat(P>A)$ or disagrees with $C$ on a comparison agreed by $C$ and
$C \nat(P>A)$. Such a comparison $I \leq J$ implies that the difference between
$C'$ and $C$ contains an element that is not in the difference between $C
\nat(P>A)$ and $C$. Therefore, $C'$ is no closer to $C$ than $C \nat(P>A)$.

The claim is that every preorder $C'$ that satisfies $P>A$ either:

\begin{itemize}

\item is $C \nat(P>A)$;

\item contains a pair $I \leq J$
that is neither in $C$ nor in $C \nat(P>A)$; or

\item does not contain a pair $I \leq J$
that is both in $C$ and in $C \nat(P>A)$.

\end{itemize}

If $C'$ is not $C \nat(P>A)$, then it either contains a pair of models that is
not in $C \nat(P>A)$, or it does not contain a pair of models that is in $C
\nat(P>A)$. If $C$ agrees with $C \nat(P>A)$ on this pair, either the second or
the third part of the claim is proved.

The other case is that $C$ and $C \nat(P>A)$ disagree on this pair: natural
revision changes the order between the two models.

Natural revision only changes the order between $\minidx(P \neg A)$ and
$\minidx(PA)$.

\[
L = (\equal\min(P \neg A)) \cup \cdots \cup (\equal\min(PA))
\]

Natural revision only changes the order between some models of $L$. Namely, it
only changes how to compare the models of $\neg P$ with the models of $P \neg
A$ and how to compare the models of $\min(PA)$ with the models of $P \neg A$.
Arbitrary such models are denoted as follows.

\begin{eqnarray*}
I	& \in &		\neg P L			\\
J	& \in &		\min(PA)			\\
K	& \in &		P \neg A L
\end{eqnarray*}

Natural revision forces $I < K$ and $J < K$ and changes nothing else. It does
not change any other comparison, as shown in
Figure~\ref{figure-natural-changes}.

\begin{hfigure}
\long\def\ttytex#1#2{#1}
\ttytex{
\begin{tabular}{ccc}
%
%
\setlength{\unitlength}{3750sp}%
\begin{picture}(2274,4824)(4489,-5623)
\thinlines
{\color[rgb]{0,0,0}\put(4501,-2011){\line( 1, 0){2250}}
}%
{\color[rgb]{0,0,0}\put(4501,-2611){\line( 1, 0){2250}}
}%
{\color[rgb]{0,0,0}\put(4501,-3211){\line( 1, 0){2250}}
}%
{\color[rgb]{0,0,0}\put(4501,-3811){\line( 1, 0){2250}}
}%
{\color[rgb]{0,0,0}\put(4501,-4411){\line( 1, 0){2250}}
}%
{\color[rgb]{0,0,0}\put(5401,-3661){\framebox(750,900){}}
}%
{\color[rgb]{0,0,0}\put(4501,-1411){\line( 1, 0){2250}}
}%
{\color[rgb]{0,0,0}\put(4501,-5611){\framebox(2250,4800){}}
}%
{\color[rgb]{0,0,0}\put(4501,-5011){\line( 1, 0){2250}}
}%
{\color[rgb]{0,0,0}\put(4951,-4861){\framebox(1350,3300){}}
}%
\put(5626,-3061){\makebox(0,0)[b]{\smash{\fontsize{9}{10.8}
\usefont{T1}{cmr}{m}{n}{\color[rgb]{0,0,0}$PA$}%
}}}
\put(5101,-1861){\makebox(0,0)[b]{\smash{\fontsize{9}{10.8}
\usefont{T1}{cmr}{m}{n}{\color[rgb]{0,0,0}$P$}%
}}}
\put(5776,-2236){\makebox(0,0)[b]{\smash{\fontsize{9}{10.8}
\usefont{T1}{cmr}{m}{n}{\color[rgb]{0,0,0}$P\neg A$}%
}}}
\put(4726,-1711){\makebox(0,0)[b]{\smash{\fontsize{9}{10.8}
\usefont{T1}{cmr}{m}{n}{\color[rgb]{0,0,0}$\neg P$}%
}}}
\put(6001,-3136){\makebox(0,0)[b]{\smash{\fontsize{9}{10.8}
\usefont{T1}{cmr}{m}{n}{\color[rgb]{0,0,0}$J$}%
}}}
\put(6076,-1861){\makebox(0,0)[b]{\smash{\fontsize{9}{10.8}
\usefont{T1}{cmr}{m}{n}{\color[rgb]{0,0,0}$K$}%
}}}
\put(4801,-2461){\makebox(0,0)[b]{\smash{\fontsize{9}{10.8}
\usefont{T1}{cmr}{m}{n}{\color[rgb]{0,0,0}$I$}%
}}}
\end{picture}%
&
\setlength{\unitlength}{3750sp}%
\begin{picture}(1104,2544)(5659,-5743)
\thinlines
{\color[rgb]{0,0,0}\multiput(6391,-3391)(-9.47368,4.73684){20}{\makebox(2.1167,14.8167){\tiny.}}
\put(6211,-3301){\line( 0,-1){ 45}}
\put(6211,-3346){\line(-1, 0){180}}
\put(6031,-3346){\line( 0,-1){ 90}}
\put(6031,-3436){\line( 1, 0){180}}
\put(6211,-3436){\line( 0,-1){ 45}}
\multiput(6211,-3481)(9.47368,4.73684){20}{\makebox(2.1167,14.8167){\tiny.}}
}%
\end{picture}%
&
\setlength{\unitlength}{3750sp}%
\begin{picture}(2274,6774)(4489,-7123)
\thinlines
{\color[rgb]{0,0,0}\put(5401,-2761){\framebox(750,450){}}
}%
{\color[rgb]{0,0,0}\put(4951,-2761){\line(-1, 0){450}}
\put(4501,-2761){\line( 0, 1){2400}}
\put(4501,-361){\line( 1, 0){2250}}
\put(6751,-361){\line( 0,-1){2400}}
\put(6751,-2761){\line(-1, 0){450}}
\put(6301,-2761){\line( 0, 1){1650}}
\put(6301,-1111){\line(-1, 0){1350}}
\put(4951,-1111){\line( 0,-1){1650}}
}%
{\color[rgb]{0,0,0}\put(4501,-2161){\line( 1, 0){450}}
}%
{\color[rgb]{0,0,0}\put(6301,-2161){\line( 1, 0){450}}
}%
{\color[rgb]{0,0,0}\put(4501,-961){\line( 1, 0){2250}}
}%
{\color[rgb]{0,0,0}\put(4501,-1561){\line( 1, 0){450}}
}%
{\color[rgb]{0,0,0}\put(6301,-1561){\line( 1, 0){450}}
}%
\put(5626,-2611){\makebox(0,0)[b]{\smash{\fontsize{9}{10.8}
\usefont{T1}{cmr}{m}{n}{\color[rgb]{0,0,0}$PA$}%
}}}
\put(4726,-1261){\makebox(0,0)[b]{\smash{\fontsize{9}{10.8}
\usefont{T1}{cmr}{m}{n}{\color[rgb]{0,0,0}$\neg P$}%
}}}
{\color[rgb]{0,0,0}\put(4501,-5311){\line( 1, 0){2250}}
}%
{\color[rgb]{0,0,0}\put(5401,-5161){\framebox(750,450){}}
}%
{\color[rgb]{0,0,0}\put(5401,-4561){\line(-1, 0){450}}
\put(4951,-4561){\line( 0, 1){1650}}
\put(4951,-2911){\line( 1, 0){1350}}
\put(6301,-2911){\line( 0,-1){1650}}
\put(6301,-4561){\line(-1, 0){150}}
\put(6151,-4561){\line( 0, 1){450}}
\put(6151,-4111){\line(-1, 0){750}}
\put(5401,-4111){\line( 0,-1){450}}
}%
{\color[rgb]{0,0,0}\put(4951,-3961){\line( 1, 0){1350}}
}%
{\color[rgb]{0,0,0}\put(4951,-3361){\line( 1, 0){1350}}
}%
{\color[rgb]{0,0,0}\put(4501,-7111){\framebox(2250,2400){}}
}%
{\color[rgb]{0,0,0}\put(4501,-6511){\line( 1, 0){2250}}
}%
{\color[rgb]{0,0,0}\put(4501,-5911){\line( 1, 0){2250}}
}%
{\color[rgb]{0,0,0}\put(4951,-6361){\framebox(1350,1650){}}
}%
\put(5776,-3586){\makebox(0,0)[b]{\smash{\fontsize{9}{10.8}
\usefont{T1}{cmr}{m}{n}{\color[rgb]{0,0,0}$P\neg A$}%
}}}
\put(6001,-2686){\makebox(0,0)[b]{\smash{\fontsize{9}{10.8}
\usefont{T1}{cmr}{m}{n}{\color[rgb]{0,0,0}$J$}%
}}}
\put(4801,-2011){\makebox(0,0)[b]{\smash{\fontsize{9}{10.8}
\usefont{T1}{cmr}{m}{n}{\color[rgb]{0,0,0}$I$}%
}}}
\put(6076,-3211){\makebox(0,0)[b]{\smash{\fontsize{9}{10.8}
\usefont{T1}{cmr}{m}{n}{\color[rgb]{0,0,0}$K$}%
}}}
\end{picture}%
\end{tabular}
}{
+--------------------------------+      +--------------------------------+
|                                |      |                                |
|                                |      |                                |
|                                |      |                                |
+--------------------------------+      +--------------------------------+
|       +--------------------+   |      |       +--------------------+   |
| -P    |           P-A      |   |      | -P    |                    |   |
|       |  K                 |   |      |       |                    |   |
+-------|--------------------|---+      +-------+                    +---+
|       |                    |   |      |       |                    |   |
|       |                    |   |      |       |                    |   |
|    I  |                    |   |      |    I  |                    |   |
+-------|--------------------|---+      +-------+                    +---+
|       |                    |   |      |       |                    |   |
|       |        +-------+   |   |      |       |        +-------+   |   |
|       |        |m(PA) J|   |   |      |       |        |m(PA) J|   |   |
+-------|--------|-------|---|---+  =>  +-------+        +-------+   +---+
|       |        |       |   |   |                                           .
|       |        +-------+   |   |              +--------------------+
|       |                    |   |              |           P-A      |
+-------|--------------------|---+              |   K                |
|       |                    |   |              +--------------------+
|       |                    |   |              |                    |
|       +--------------------+   |              |        +-------+   |
+--------------------------------+              |        |       |   |
|                                |              +--------+       +---+
|                                |                                           .
|                                |      +-------+--------+-------+---+---+
+--------------------------------+      |       |        |       |   |   |
                                        |       |        +-------+   |   |
                                        |       +--------------------+   |
                                        +--------------------------------+
                                        |                                |
                                        |                                |
                                        |                                |
                                        +--------------------------------+
}
\label{figure-natural-changes}
\hcaption{Natural revision.}
\end{hfigure}


Natural revision does not change the order between a model of $\neg P$ and a
model of $\min(PA)$. Since $I$ is less than or equal to $\min(PA)$, it is less
than or equal to $J$. This pair $I \leq J$ is therefore also in $C \nat(P>A)$.
If $C'$ does not contain it, it falls in the third part of the claim. The rest
of the proof assumes that $I \leq J$ is in $C'$.

The only possible differences between $C$ and $C \nat(P>A)$ are:

\begin{itemize}

\item

$I \equiv K$ becomes $I < K$;

in terms of pairs: $I \leq K$ and $K \leq I$ become $I \leq K$

the pair $K \leq I$ is removed;

the same for $J$ in place of $I$;

\item

$K < I$ becomes $I < K$;

in terms of pairs: $K \leq I$ becomes $I \leq K$

the pair $K \leq I$ is removed;

the pair $I \leq K$ is added;

the same for $J$ in place of $I$.

\end{itemize}

The result is the same: $C \nat(P>A)$ contains $I \leq K$ and not $K \leq I$.
Since $C'$ differs from $C$, either it does not contain a pair $I \leq K$ or it
contains a pair $K \leq I$.

\begin{description}

\item[$K \leq I$ is in $C'$]

since $I \leq J$ is in $C'$, transitivity implies $K \leq J$: a model of $P
\neg A$ is less than or equal to a model of $\min(PA)$; this is a violation of
$P > A$;

the same holds for $J$ in place of $I$, since $J \leq J$ follows from
reflexivity;

\item[$I \leq K$ is not in $C'$]

if $C'$ does not contain $K \leq I$, it is not a connected preorder;

otherwise, $C'$ contains $K \leq I$;

this is the first case, which also holds for $J$ in place of $I$.

\end{description}\qed

\draft

this is not a consequnce of minimality of natural revision as defined in
[CB20], in spite of the coincidence of the result of natural revision; [CB20]
first revise by P->A and then minimally change the order to satisfy P>A;
minimality in [CB20] is between the result and Cnat(true>P->A), not between the
result and C

\enddraft

\

%

The previous theorem shows that natural revision minimally changes the order to
satisfy $P>A$; it implements the principle of minimal change. It also
implements conditional preservation: it does not change the order between
models that evaluate $P>A$ in the same way.

\begin{definition}
\label{conditional-preservation}

An ordering $C'$ {\em preserves the conditionals} of another order $C$ over
$P>A$ if $I \leq J \in C$ is the same as $I \leq J \in C'$ for all pair of
models $I$ and $J$ that either both satisfy $\neg P$, both satisfy $PA$ or both
satisfy $P \neg A$.

\end{definition}

\begin{theorem}
\label{natural-preserve}

Natural revision preserves the conditionals.

\end{theorem}

\proof Natural revision only realizes $J < K$ and $I < K$, where $I$, $J$ and
$K$ are respectively models of $\neg P$, of $PA$ and of $P \neg A$. It does not
change the order between any other two models.~\qed

\draft

[formal proof]

The claim is proved by showing that intersecting every class of $C$ and every
class of $C \nat(P>A)$ with $\neg P$ results in the same order. In the proof,
$m$ and $m'$ are respectively the number of equivalence classes of $C$ and $C
\nat(P>A)$.

\long\def\ttytex#1#2{#1}
\ttytex{
\[
\begin{array}{lll}
[	C \nat(P>A)(0) \cap \neg P,			\ldots ,
	C \nat(P>A)(m') \cap \neg P
]									\\
& = & \\
\relax[	C(0) \cap \neg P,				\ldots ,
	C(\minidx(P)-1) \cap \neg P,					\\
	C(\minidx(P)) \backslash (P \neg A) \cap \neg P,	\ldots ,
	C(\minidx(PA)) \backslash (P \neg A) \cap \neg P,			\\
	C(\minidx(P)) \cap (P \neg A) \cap \neg P,		\ldots ,
	C(\minidx(PA)) \cap (P \neg A) \cap \neg P,			\\
	C(\minidx(PA)+1) \cap \neg P,			\ldots ,
	C(m) \cap \neg P
]									\\
& = & \\
\relax[	C(0) \cap \neg P,				\ldots ,
	C(\minidx(P)-1) \cap \neg P,					\\
	C(\minidx(P)) \backslash (P \neg A) \backslash P,	\ldots ,
	C(\minidx(PA)) \backslash (P \neg A) \backslash P,			\\
	\emptyset , \ldots ,
	\emptyset,	
&
		\mbox{ since } P\neg A \cap \neg P = \emptyset		\\
	C(\minidx(PA)+1) \cap \neg P,			\ldots ,
	C(m) \cap \neg P]						\\
& = & \\
\relax[	C(0) \cap \neg P,				\ldots ,
	C(\minidx(P)-1) \cap \neg P,					\\
	C(\minidx(P)) \backslash P,				\ldots ,
	C(\minidx(PA)) \backslash P,
&		\mbox{ since } P\neg A \subseteq P			\\
	C(\minidx(PA)+1) \cap \neg P,			\ldots ,
	C(m) \cap \neg P]						\\
& = & \\
\relax[	C(0) \cap \neg P,				\ldots ,
	C(\minidx(P)-1) \cap \neg P,					\\
	C(\minidx(P)) \cap \neg P, \ldots ,
	C(\minidx(PA)) \cap \neg P,					\\
	C(\minidx(PA)+1) \cap \neg P,			\ldots ,
	C(m) \cap \neg P]
\end{array}
\]
}{
[
        C(0)-P                  .. C(min(P)-1)-P
        (C(min(P)) - (P-A))-P   .. (C(min(PA)) - (P-A))-P
        (C(min(P)) (P-A))-P     .. (C(min(PA)) (P-A))-P
        C(min(PA)+1)-P          .. C(m)-P
]
=
[
        C(0)-P                  .. C(min(P)-1)-P
        (C(min(P)) - (P-A)) - P .. (C(min(PA)) - (P-A)) - P
        0                       .. 0				P-A -P = 0
        C(min(PA)+1)-P          .. C(m)-P
]
=
[
        C(0)-P                  .. C(min(P)-1)-P
        C(min(P))-P             .. C(min(PA))-P			P-A c= P
        C(min(PA)+1)-P          .. C(m)-P
]
=
[
        C(0)-P                  .. C(min(P)-1)-P
        C(min(P))-P             .. C(min(PA))-P
        C(min(PA)+1)-P          .. C(m)-P
]
}

The same applies intersecting with $PA$ and with $P \neg A$.

\enddraft

\

Since natural revision is a minimal change among all orders that satisfy $P>A$
and it preserves the conditionals, it is also a minimal change among the orders
that satisfy $P>A$ and preserve the conditionals.

\

%

Natural revision satisfies the Kern-Isberner postulates for iterated belief
revision~\cite{kern-99}.

\begin{theorem}
\label{natural-postulates}

Natural revision satisfies postulates CR0-CR7~\cite{kern-99}.

\end{theorem}

\proof
\begin{description}

\item[CR0] $C \nat(P>A)$ is a connected preorder.

Holds by definition.

\item[CR1] $C \nat(P>A) \models P>A$.

Holds because natural revision establishes $J < K$ for all models $J \in
\min(PA)$ and $K \in P \neg A$.

\item[CR2] $C \nat(P>A) = C$ if and only if $C \models P>A$.

The premise $C \nat(P>A) = C$ with $C \nat(P>A) \models P>A$ proved above as
CR1 implies the claim $C \models P>A$.

The converse is that $C \models P>A$ implies $C \nat(P>A) = C$: natural
revision does not touch an order that already entails $P>A$.

Natural revision establishes $J <_{C \nat(P>A)} K$ and $I <_{C \nat(P>A)} K$
for all models $I$, $J$ and $K$ as follows, where
{} $L = (\equal\min(P \neg A)) \cup \cdots \cup (\equal\min(PA))$.

\begin{eqnarray*}
I	& \in &		\neg P L			\\
J	& \in &		\min(PA)			\\
K	& \in &		P \neg A L
\end{eqnarray*}

Since $C$ entails $P>A$, it compares all models of $\min(PA)$ less than all
models of $P \neg A$. The first established comparison $J <_{C \nat(P>A)} K$
also holds in $C$.

Since $I$ is in $\neg P L$, it is in $L$, and is therefore less than or equal
to the models of $\min(PA)$ according to $C$. Therefore, it is less than all
models of $P \neg A$ since $C \models P>A$. The second established comparison
$I <_{C \nat(P>A)} K$ also holds in $C$.

\item[CR3] $C \nat(\true>A)$ is an AGM revision operator.

Consequence of Theorem~\ref{natural-true}: $C \nat(\true>A) = C \nat(A)$, and
propositional natural revision is an AGM revision operator.

\item[CR4] If $P>A \equiv Q>B$ then $C \nat(P>A) = C \nat(Q>B)$.

Equivalence $P>A \equiv Q>B$ is $P \equiv Q$ and $PA \equiv QB$. It implies
$\neg P \equiv \neg Q$ and $P \neg A \equiv Q \neg B$. Natural revision is
defined from these formulae only, beside $C$.

\item[CR5] If $Q \subseteq PA$
then $C \models Q>B$ if and only if $C \nat(P>A) \models Q>B$.

Theorem~8 by Kern--Isberner~\cite{kern-99} prove that this postulate is the
same as conditional preservation, which natural revision obeys by
Theorem~\ref{natural-preserve}.

\item[CR6]
If $QB \subseteq PA$, $Q \neg B \subseteq P \neg A$ and $C \models Q>B$
then $C \nat(P>A) \models Q>B$.

Theorem~11 by Kern--Isberner~\cite{kern-99} gives an equivalent formulation:
$J' <_C K'$ implies $J' <_{C \nat(P>A)} K'$ for all $J' \in PA$ and $K' \in P
\neg A$. This is the case because natural revision only changes the order
between a model $J' \in PA$ and a model $K \in P \neg A$ to establish $J' <_{C
\nat(P>A)} K'$.

\item[CR7]
If $QB \subseteq P\neg A$, $Q \neg B \subseteq PA$ and $C \models Q>B$
then $C \nat(P>A) \models Q>B$.

Theorem~11 by Kern--Isberner~\cite{kern-99} gives an equivalent formulation:
$K' <_{C \nat(P>A)} J'$ implies $K' <_C J'$ for all $J' \in PA$ and $K' \in P
\neg A$. The conditions $C \nat(P>A) \models P>A$ and $K' <_{C \nat(P>A)} J'$
are incompatible with $J' \in \min(PA)$. As a result, $J'$ is a models of $PA$
but not a minimal model of it. Therefore, natural revision does not change its
comparison with any other models.

\end{description}
\qed

\section{Line-down revision}
\label{down-section}

\draft

down.tex:

\begin{itemize}

\item line-down revision

\item line-down revision minimally changes the ordering, in terms of closeness

\item line-down revision preserves the conditionals

\end{itemize}

recalcitrance.tex:

\begin{itemize}

\item recalcitrance

\item natural revision is recalcitrant for two revisions on the flat ordering

\item line-down revision is not recalcitrant for two revisions on the flat ordering

\item natural revision is recalcitrant for two revisions on the order x<-x,
same as the flat order revised by x: the example

\end{itemize}

\enddraft

%

Line-down revision lines the models of $\neg P$ with the models of $P \neg A$.
It lifts $\min(PA)$ over them.

\begin{definition}
\label{down}

The {\em line-down revision} of the order $[C(0),\ldots,C(m)]$ by a conditional
formula $P>A$ is as follows.

\long\def\ttytex#1#2{#1}
\ttytex{
\begin{eqnarray*}
\lefteqn{[C(0),\ldots,C(m)] \dow(P>A) = }				\\
&& [ C(0),                        \ldots C(\minidx(P \neg A)-1),		\\
&& \min(PA) \backslash C(0) \backslash \cdots \backslash C(\minidx(P \neg A)-1), \\
&& C(\minidx(P \neg A)) \backslash \min(PA) \ldots C(m) \backslash \min(PA)]
\end{eqnarray*}
}{
[C(0) .. C(m)] dow(P>A) =
[
        C(0)                                 .. C(min(P-A)-1)
        min(PA) - C(0) - .. - C(min(P-A)-1)
        C(min(P-A)) - min(PA)                .. C(m) - min(PA)
]
}

\end{definition}

The ordering contains
{} $\min(PA) \backslash C(0) \backslash \cdots
{}  \backslash C(\minidx(P \neg A)-1)$
instead of just $\min(PA)$ because $\minidx(P)$ may be equal to $\minidx(PA)$.
In this case, $\min(PA)$ is between $0$ and $\minidx(P \neg A) - 1$; the set
difference avoids it being included twice. It is only included within
{} $C(0) \ldots C(\minidx(P \neg A) - 1)$
and not as
{} $\min(PA) \backslash \cdots$
because $C$ satisfies $P>A$ if $\minidx(P)=\minidx(PA)$. This way, $C$ is not
changed.

Line-down revision only changes the ordering between the equivalence classes
$\minidx(PA)$ and $\minidx(P \neg A)$, between the models of
{} $L = (\equal\min(P)) \cup \cdots \cup (\equal\min(PA))$.
Among these models, it sets $J<K$ and $J<I$ for all models $I$, $J$ and $K$ of
respectively $\neg PL$, $PAL$ and $P \neg AL$ since the classes of $P
\backslash \min(PA)$ and $P \neg A$ are below those of $\min(PA)$.

\

%

Line-down revision implements minimal change: it minimally changes the order to
satisfy $P>A$ according to Definition~\ref{closer}.

\begin{theorem}
\label{down-minimal}

No connected preorder satisfying $P>A$ is strictly closer to $C$ than line-down
revision $C \dow(P>A)$.

\end{theorem}

\draft

\[
\forall C, P>A ~.~
\not\exists C' ~.~
	C' \mbox{ satisfies } P>A ,~
	\diff(C',C) \subset \diff(C \dow(P>A),C)
\]

\enddraft

\proof The claim is equivalent to: every every connected preorder $C'$ that
satisfies $P>A$ is either $C \dow(P>A)$ or disagrees with $C$ on a comparison
agreed by $C$ and $C \dow(P>A)$. Such a comparison $I \leq J$ implies that the
difference from $C'$ to $C$ contains an element that is not the difference from
$C \dow(P>A)$ to $C$. Therefore, $C'$ is not strictly closer to $C$ than $C
\dow(P>A)$.

The claim is that if $C'$ is a connected preorder satisfying $P>A$, then
it either:

\begin{itemize}

\item is $C \dow(P>A)$;

\item contains a pair $I \leq K$
that is neither in $C$ nor in $C \dow(P>A)$; or

\item does not contain a pair $I \leq K$
that is both in $C$ and in $C \dow(P>A)$.

\end{itemize}

If $C'$ is not $C \dow(P>A)$, then it either contains a pair of models that is
not in $C \dow(P>A)$, or it does not contain a pair of models that is in $C
\dow(P>A)$. If $C$ agrees with $C \dow(P>A)$ on this pair, either the second or
the third part of the claim is proved.

The other case is that $C$ and $C \dow(P>A)$ disagree on this pair: line-down 
revision changes the order between the two models.

Line-down revision only changes the order between $\minidx(P \neg A)$ and
$\minidx(PA)$.

\[
L = (\equal\min(P \neg A)) \cup \cdots \cup (\equal\min(PA))
\]

Line-down revision only changes the order between some models of $L$. Namely,
it only changes how to compare the models of $\min(PA)$ with the models of
$\neg P$ and the models of $\min(PA)$ with the models of $P \neg A$. Arbitrary
such models are denoted as follows.

\begin{eqnarray*}
I	& \in &		\neg P L			\\
J	& \in &		\min(PA)			\\
K	& \in &		P \neg A L
\end{eqnarray*}

\begin{hfigure}
\long\def\ttytex#1#2{#1}
\ttytex{
\begin{tabular}{ccc}
%
%
\setlength{\unitlength}{3750sp}%
\begin{picture}(2274,4824)(4489,-5623)
\thinlines
{\color[rgb]{0,0,0}\put(4501,-2011){\line( 1, 0){2250}}
}%
{\color[rgb]{0,0,0}\put(4501,-2611){\line( 1, 0){2250}}
}%
{\color[rgb]{0,0,0}\put(4501,-3211){\line( 1, 0){2250}}
}%
{\color[rgb]{0,0,0}\put(4501,-3811){\line( 1, 0){2250}}
}%
{\color[rgb]{0,0,0}\put(4501,-4411){\line( 1, 0){2250}}
}%
{\color[rgb]{0,0,0}\put(5401,-3661){\framebox(750,900){}}
}%
{\color[rgb]{0,0,0}\put(4501,-1411){\line( 1, 0){2250}}
}%
{\color[rgb]{0,0,0}\put(4501,-5611){\framebox(2250,4800){}}
}%
{\color[rgb]{0,0,0}\put(4501,-5011){\line( 1, 0){2250}}
}%
{\color[rgb]{0,0,0}\put(4951,-4861){\framebox(1350,3300){}}
}%
\put(5626,-3061){\makebox(0,0)[b]{\smash{\fontsize{9}{10.8}
\usefont{T1}{cmr}{m}{n}{\color[rgb]{0,0,0}$PA$}%
}}}
\put(5101,-1861){\makebox(0,0)[b]{\smash{\fontsize{9}{10.8}
\usefont{T1}{cmr}{m}{n}{\color[rgb]{0,0,0}$P$}%
}}}
\put(5776,-2236){\makebox(0,0)[b]{\smash{\fontsize{9}{10.8}
\usefont{T1}{cmr}{m}{n}{\color[rgb]{0,0,0}$P\neg A$}%
}}}
\put(4726,-1711){\makebox(0,0)[b]{\smash{\fontsize{9}{10.8}
\usefont{T1}{cmr}{m}{n}{\color[rgb]{0,0,0}$\neg P$}%
}}}
\put(6001,-3136){\makebox(0,0)[b]{\smash{\fontsize{9}{10.8}
\usefont{T1}{cmr}{m}{n}{\color[rgb]{0,0,0}$J$}%
}}}
\put(6076,-1861){\makebox(0,0)[b]{\smash{\fontsize{9}{10.8}
\usefont{T1}{cmr}{m}{n}{\color[rgb]{0,0,0}$K$}%
}}}
\put(4801,-2461){\makebox(0,0)[b]{\smash{\fontsize{9}{10.8}
\usefont{T1}{cmr}{m}{n}{\color[rgb]{0,0,0}$I$}%
}}}
\end{picture}%
&
\setlength{\unitlength}{3750sp}%
\begin{picture}(1104,2544)(5659,-5743)
\thinlines
{\color[rgb]{0,0,0}\multiput(6391,-3391)(-9.47368,4.73684){20}{\makebox(2.1167,14.8167){\tiny.}}
\put(6211,-3301){\line( 0,-1){ 45}}
\put(6211,-3346){\line(-1, 0){180}}
\put(6031,-3346){\line( 0,-1){ 90}}
\put(6031,-3436){\line( 1, 0){180}}
\put(6211,-3436){\line( 0,-1){ 45}}
\multiput(6211,-3481)(9.47368,4.73684){20}{\makebox(2.1167,14.8167){\tiny.}}
}%
\end{picture}%
&
%
%
\setlength{\unitlength}{3750sp}%
\begin{picture}(2274,6324)(4489,-7123)
\thinlines
{\color[rgb]{0,0,0}\put(4501,-5311){\line( 1, 0){2250}}
}%
{\color[rgb]{0,0,0}\put(4501,-5911){\line( 1, 0){2250}}
}%
{\color[rgb]{0,0,0}\put(5401,-5161){\framebox(750,450){}}
}%
{\color[rgb]{0,0,0}\put(4501,-3961){\line( 1, 0){450}}
}%
{\color[rgb]{0,0,0}\put(6301,-3961){\line( 1, 0){450}}
}%
{\color[rgb]{0,0,0}\put(4951,-3961){\line( 1, 0){1350}}
}%
{\color[rgb]{0,0,0}\put(4501,-1411){\framebox(2250,600){}}
}%
{\color[rgb]{0,0,0}\put(4501,-3361){\line( 1, 0){2250}}
}%
{\color[rgb]{0,0,0}\put(4951,-4561){\line(-1, 0){450}}
\put(4501,-4561){\line( 0, 1){1800}}
\put(4501,-2761){\line( 1, 0){2250}}
\put(6751,-2761){\line( 0,-1){1800}}
\put(6751,-4561){\line(-1, 0){450}}
\put(6301,-4561){\line( 0, 1){1650}}
\put(6301,-2911){\line(-1, 0){1350}}
\put(4951,-2911){\line( 0,-1){1650}}
}%
{\color[rgb]{0,0,0}\put(5401,-4561){\line(-1, 0){450}}
\put(4951,-4561){\line( 0, 1){1650}}
\put(4951,-2911){\line( 1, 0){1350}}
\put(6301,-2911){\line( 0,-1){1650}}
\put(6301,-4561){\line(-1, 0){150}}
\put(6151,-4561){\line( 0, 1){450}}
\put(6151,-4111){\line(-1, 0){750}}
\put(5401,-4111){\line( 0,-1){450}}
}%
{\color[rgb]{0,0,0}\put(5401,-2611){\framebox(750,450){}}
}%
{\color[rgb]{0,0,0}\put(4501,-7111){\framebox(2250,2400){}}
}%
{\color[rgb]{0,0,0}\put(4951,-6361){\framebox(1350,1650){}}
}%
{\color[rgb]{0,0,0}\put(4501,-6511){\line( 1, 0){2250}}
}%
\put(5626,-2461){\makebox(0,0)[b]{\smash{\fontsize{9}{10.8}
\usefont{T1}{cmr}{m}{n}{\color[rgb]{0,0,0}$PA$}%
}}}
\put(5776,-3586){\makebox(0,0)[b]{\smash{\fontsize{9}{10.8}
\usefont{T1}{cmr}{m}{n}{\color[rgb]{0,0,0}$P\neg A$}%
}}}
\put(4726,-3061){\makebox(0,0)[b]{\smash{\fontsize{9}{10.8}
\usefont{T1}{cmr}{m}{n}{\color[rgb]{0,0,0}$\neg P$}%
}}}
\put(6001,-2536){\makebox(0,0)[b]{\smash{\fontsize{9}{10.8}
\usefont{T1}{cmr}{m}{n}{\color[rgb]{0,0,0}$J$}%
}}}
\put(4801,-3811){\makebox(0,0)[b]{\smash{\fontsize{9}{10.8}
\usefont{T1}{cmr}{m}{n}{\color[rgb]{0,0,0}$I$}%
}}}
\put(6076,-3211){\makebox(0,0)[b]{\smash{\fontsize{9}{10.8}
\usefont{T1}{cmr}{m}{n}{\color[rgb]{0,0,0}$K$}%
}}}
\end{picture}%
\end{tabular}
}{
+--------------------------------+      +--------------------------------+
|                                |      |                                |
|                                |      |                                |
|                                |      |                                |
+--------------------------------+      +--------------------------------+
|       +--------------------+   |                                        
| -P    |           P-A      |   |                       +-------+
|       |  K                 |   |                       |m(PA) J|
+-------|--------------------|---+      +----------------+-------+-------+
|       |                    |   |      |       +--------------------+   |
|       |                    |   |      | -P    |           P-A      |   |
|    I  |                    |   |      |       |  K                 |   |
+-------|--------------------|---+      +-------|--------------------|---+
|       |                    |   |      |       |                    |   |
|       |        +-------+   |   |      |       |                    |   |
|       |        |m(PA) J|   |   |      |    I  |                    |   |
+-------|--------|-------|---|---+  =>  +-------|--------------------|---+
|       |        |       |   |   |      |       |                    |   |
|       |        +-------+   |   |      |       |        +-------+   |   |
|       |                    |   |      |       |        |XXXXXXX|   |   |
+-------|--------------------|---+      +-------+--------+-------+---+---+
|       |                    |   |      |       |        |       |   |   |
|       |                    |   |      |       |        +-------+   |   |
|       +--------------------+   |      |       |                    |   |
+--------------------------------+      +-------|--------------------|---+
|                                |      |       |                    |   |
|                                |      |       |                    |   |
|                                |      |       +--------------------+   |
+--------------------------------+      +--------------------------------+
                                        |                                |
                                        |                                |
                                        |                                |
                                        +--------------------------------+
}
\label{figure-down-changes}
\hcaption{Line-down revision.}
\end{hfigure}


Line-down revision does not change the order between $I$ and $K$. If $I \leq K$
is in $C$ for some pair of these models, then it is also in $C \dow(P>A)$. If
$C'$ does not contain $I \leq K$, it falls under the third case of the
statement. Otherwise, $C'$ contains $I \leq K$ if $C$ does. The change is shown
in Figure~\ref{figure-down-changes}.

Line down revision $C \dow(P>A)$ only changes the order between a model $J$ and
a model $I$ or a model $K$. It lifts $J$ over $I$ and $K$. Strictly: $J < I$
and $J < K$. The following analysis shows that $C'$ does the same, as otherwise
it does not satisfy $P>A$:

\begin{description}

\item[$J < K$:]

the minimal models of $PA$ according to $C$ are also the minimal models of $PA$
according to $C'$, as $C$ and $C \dow(P>A)$ agree on all pairs of models of
$PA$; all models $J$ are minimal among $PA$ in $C'$.

if a model $J$ of $\min(PA)$ is not strictly above all models $K$ of $P \neg
A$, then $\min(PA)$ is not above $P \neg A$, violating $P>A$;

\item[$J < I$:]

since $C$ locates $I$ at $\minidx(P \neg A)$ or below, it contains $K \leq I$
for some model of $\min(P \neg A)$; since line-down revision does not change
the relative order between these models, $C \dow(P>A)$ does the same;
therefore, $C'$ does as well: it contains $K \leq I$, where $K$ is a model of
$P \neg A$;

if $C'$ does not establish $J < I$, then either it does not contain $J \leq I$
or it contains $I \leq J$; in the former case, connectivity implies $I \leq J$;
either way, $C'$ contains $I \leq J$; transitivity with $K \leq I$ implies $K
\leq J$: a model of $P \neg A$ is not strictly under a model of $PA$, violating
$P>A$.

\end{description}
\qed

\

%

Line-down revision implements minimal change, like natural revision. Like
natural revision, it also preserves the conditionals: it does not change the
order between models that evaluate $P>A$ in the same way.

\begin{theorem}
\label{down-preserve}

Line-down revision preserves the conditionals.

\end{theorem}

\proof Line-down revision only realizes $J < K$ and $J < I$, where $I$, $J$ and
$K$ are respectively models of $\neg P$, of $PA$ and of $P \neg A$. It does not
change the order between any other two models.~\qed

%

\draft

\separator

[no section]

\begin{itemize}

\item recalcitrance

\item natural revision is recalcitrant for two revisions on the flat ordering

\item line-down revision is not recalcitrant for two revisions on the flat
ordering

\item natural revision is not recalcitrant for two revisions on the order x<-x,
the same as the flat order revised by x, the example of the introduction

\end{itemize}

\enddraft

Line-down revision is compared with natural revision on a specific property:
recalcitrance.

Recalcitrance on propositional formulae is $K \rev(A) \rev(B) = K \rev(A \wedge
B)$ whenever $A \wedge B$ is consistent. Extending this definition to
conditionals requires extending revision to sets of conditionals $C \rev(\{P>A,
Q>B\})$~\cite{kern-04}. An equivalent definition is used instead: $A$ is
implied by $K \rev(A) \rev(B)$ whenever $\{A,B\}$ is consistent. This is an
equivalent definition of recalcitrance for all revisions that do not change the
orders that satisfy the revision, like natural and line-down revision.

\begin{definition}
\label{recalcitrance}

A revision $\rev$ is {\em recalcitrant} if $C \rev(P>A) \rev(Q>B)$ satifies
$P>A$ whenever an order satisfying both $P>A$ and $Q>B$ exists.

\end{definition}

Natural revision is recalcitrant on the flat ordering, the one that compares
all models the same, denoted $C_{\epsilon}$. The proof comprises two steps. The
first shows what happens when the second revision invalidates the first.

\begin{lemma}
\label{first-unsatisfied-inconsistent}

If $C_{\epsilon} \nat(P>A) \nat(Q>B)$ falsifies $P>A$, then
{} $PA \models Q \neg B$
and
{} $QB \models P \neg A$.

\end{lemma}

\proof The order $C_{\epsilon} \nat(P>A) \nat(Q>B)$ is computed.

The first revision $P>A$ leaves $PA$ and $\neg P$ where they are and drops $P
\neg A$ just below $\min(PA)$. The models of $PA$ and $\neg P$ form the top
class, the models $P \neg A$ the bottom one, as shown in
Figure~\ref{figure-two-general1}.

\

\begin{hfigure}
\setlength{\unitlength}{3750sp}%
\begin{picture}(3024,1374)(4489,-3973)
\thinlines
{\color[rgb]{0,0,0}\put(4501,-3211){\framebox(900,600){}}
}%
{\color[rgb]{0,0,0}\put(5551,-3961){\framebox(900,600){}}
}%
{\color[rgb]{0,0,0}\put(6601,-3211){\framebox(900,600){}}
}%
\put(4951,-2986){\makebox(0,0)[b]{\smash{\fontsize{9}{10.8}
\usefont{T1}{cmr}{m}{n}{\color[rgb]{0,0,0}$PA$}%
}}}
\put(6001,-3736){\makebox(0,0)[b]{\smash{\fontsize{9}{10.8}
\usefont{T1}{cmr}{m}{n}{\color[rgb]{0,0,0}$P\neg A$}%
}}}
\put(7051,-2986){\makebox(0,0)[b]{\smash{\fontsize{9}{10.8}
\usefont{T1}{cmr}{m}{n}{\color[rgb]{0,0,0}$\neg P$}%
}}}
\end{picture}%
\nop{
+-------+   +-------+
|  PA   |   |   -P  |
+-------+   +-------+
                                                      .
      +-------+
      |  P-A  |
      +-------+
}
\label{figure-two-general1}
\hcaption{The order $C_{\epsilon} \nat(P>A)$.}
\end{hfigure}

\

The second revision $Q>B$ depends on whether $\minidx(Q)$ is $0$ or $1$.


The first case is $\minidx(Q)=1$: all models of $Q$ are in $P \neg A$. Natural
revision drops the $Q \neg B$ part of $P \neg A$, as show in
Figure~\ref{figure-two-general2}.

\

\begin{hfigure}
\setlength{\unitlength}{3750sp}%
\begin{picture}(3024,2124)(4489,-4723)
\thinlines
{\color[rgb]{0,0,0}\put(4501,-3211){\framebox(900,600){}}
}%
{\color[rgb]{0,0,0}\put(6601,-3211){\framebox(900,600){}}
}%
{\color[rgb]{0,0,0}\put(4501,-3961){\framebox(900,600){}}
}%
{\color[rgb]{0,0,0}\put(6601,-3961){\framebox(900,600){}}
}%
{\color[rgb]{0,0,0}\put(5551,-4711){\framebox(900,600){}}
}%
\put(4951,-2986){\makebox(0,0)[b]{\smash{\fontsize{9}{10.8}
\usefont{T1}{cmr}{m}{n}{\color[rgb]{0,0,0}$PA$}%
}}}
\put(7051,-2986){\makebox(0,0)[b]{\smash{\fontsize{9}{10.8}
\usefont{T1}{cmr}{m}{n}{\color[rgb]{0,0,0}$\neg P$}%
}}}
\put(4951,-3736){\makebox(0,0)[b]{\smash{\fontsize{9}{10.8}
\usefont{T1}{cmr}{m}{n}{\color[rgb]{0,0,0}$P\neg AQB$}%
}}}
\put(7051,-3736){\makebox(0,0)[b]{\smash{\fontsize{9}{10.8}
\usefont{T1}{cmr}{m}{n}{\color[rgb]{0,0,0}$P\neg A\neg Q$}%
}}}
\put(6001,-4486){\makebox(0,0)[b]{\smash{\fontsize{9}{10.8}
\usefont{T1}{cmr}{m}{n}{\color[rgb]{0,0,0}$P\neg AQ\neg B$}%
}}}
\end{picture}%
\nop{
+-------+   +-------+
|  PA   |   |   -P  |
+-------+   +-------+
                                                      .
+-------+   +-------+
| P-AQB |   | P-A-Q |
+-------+   +-------+
                                                      .
      +-------+
      | P-AQ-B|
      +-------+
}
\label{figure-two-general2}
\hcaption{The order $C_{\epsilon} \nat(P>A) \nat(Q>B)$ when $\minidx(Q)=1$.}
\end{hfigure}

\

The minimal models of $PA$ are still above all models of $P \neg A$. The first
conditional $P>A$ is still satisfied, contradicting the premise of the lemma.


The second case is $\minidx(Q)=0$: either $PA$ or $\neg P$ contain some models
of $Q$. Either way, the second revision drops the part of $Q \neg B$ that is
above $\min(QB)$ or at the same level. Two subcases are considered: either $PA$
contains some models of $\neg Q \vee B$, or it does not.


The first subcase is that $PA$ contains some models of $\neg Q \vee B$.

The second revision only drops models of $Q \neg B$. It moves no model of its
negation $\neg (Q \neg B) = \neg Q \vee B$. If some of them are in $PA$, they
do not move. They remain in class $0$. Natural revision raises no model,
including those of $P \neg A$. As a result, the models of $PA (\neg Q \vee B)$
are still minimal models of $PA$ and are still above all models of $P \neg A$.
The first conditional $P>A$ is satisfied, contradicting the premise of the
lemma.


The second subcase is that $PA$ contains no model of $\neg Q \vee B$.

It only comprises models of its negation $\neg (\neg Q \vee B) = Q \neg B$, as
shown in Figure~\ref{figure-two-general3}. This is the first part of the claim:
$PA \models Q \neg B$.

\

\begin{hfigure}
\setlength{\unitlength}{3750sp}%
\begin{picture}(3024,1374)(4489,-3973)
\thinlines
{\color[rgb]{0,0,0}\put(4501,-3211){\framebox(900,600){}}
}%
{\color[rgb]{0,0,0}\put(5551,-3961){\framebox(900,600){}}
}%
{\color[rgb]{0,0,0}\put(6601,-3211){\framebox(900,600){}}
}%
\put(4951,-2986){\makebox(0,0)[b]{\smash{\fontsize{9}{10.8}
\usefont{T1}{cmr}{m}{n}{\color[rgb]{0,0,0}$PA$}%
}}}
\put(6001,-3736){\makebox(0,0)[b]{\smash{\fontsize{9}{10.8}
\usefont{T1}{cmr}{m}{n}{\color[rgb]{0,0,0}$P\neg A$}%
}}}
\put(7051,-2986){\makebox(0,0)[b]{\smash{\fontsize{9}{10.8}
\usefont{T1}{cmr}{m}{n}{\color[rgb]{0,0,0}$\neg P$}%
}}}
\put(5476,-2986){\makebox(0,0)[lb]{\smash{\fontsize{9}{10.8}
\usefont{T1}{cmr}{m}{n}{\color[rgb]{0,0,0}$\subseteq Q\neg B$}%
}}}
\end{picture}%
\nop{
+-------+        +-------+
|  PA   |c=Q-B   |   -P  |
+-------+        +-------+
                                                      .
      +-------+
      |  P-A  |
      +-------+
}
\label{figure-two-general3}
\hcaption
{The order $C_{\epsilon} \nat(P>A) \nat(Q>B)$
when $\minidx(B)=0$ and $PA \subseteq Q \neg B$.}
\end{hfigure}

\

The second revision $Q>B$ drops the models of $Q \neg B$ either by one or two,
depending on where the models of $\min(QB)$ are. Two sub-subcases: either $\neg
P$ contains some models of $QB$ or it does not.


The first sub-subcase is that $\neg P$ contains some models of $QB$. A
consequence is $\minidx(QB) = 0$. Natural revision drops all models of $Q \neg
B$ at level $\min(QB) = 0$ or above under it. They are all models of $PA$ and
the models of $\neg PQ\neg B$.

\

\begin{hfigure}
\setlength{\unitlength}{3750sp}%
\begin{picture}(3774,2124)(4489,-3973)
\thinlines
{\color[rgb]{0,0,0}\put(4501,-3211){\framebox(900,600){}}
}%
{\color[rgb]{0,0,0}\put(5551,-3961){\framebox(900,600){}}
}%
{\color[rgb]{0,0,0}\put(6601,-3211){\framebox(900,600){}}
}%
{\color[rgb]{0,0,0}\put(7351,-2461){\framebox(900,600){}}
}%
{\color[rgb]{0,0,0}\put(5851,-2461){\framebox(900,600){}}
}%
\put(4951,-2986){\makebox(0,0)[b]{\smash{\fontsize{9}{10.8}
\usefont{T1}{cmr}{m}{n}{\color[rgb]{0,0,0}$PA$}%
}}}
\put(6001,-3736){\makebox(0,0)[b]{\smash{\fontsize{9}{10.8}
\usefont{T1}{cmr}{m}{n}{\color[rgb]{0,0,0}$P\neg A$}%
}}}
\put(6301,-2236){\makebox(0,0)[b]{\smash{\fontsize{9}{10.8}
\usefont{T1}{cmr}{m}{n}{\color[rgb]{0,0,0}$\neg PQB$}%
}}}
\put(7801,-2236){\makebox(0,0)[b]{\smash{\fontsize{9}{10.8}
\usefont{T1}{cmr}{m}{n}{\color[rgb]{0,0,0}$\neg P\neg Q$}%
}}}
\put(7051,-2986){\makebox(0,0)[b]{\smash{\fontsize{9}{10.8}
\usefont{T1}{cmr}{m}{n}{\color[rgb]{0,0,0}$\neg PQ\neg B$}%
}}}
\put(5476,-2986){\makebox(0,0)[lb]{\smash{\fontsize{9}{10.8}
\usefont{T1}{cmr}{m}{n}{\color[rgb]{0,0,0}$\subseteq Q\neg B$}%
}}}
\end{picture}%
\nop{
                 +--------+       +--------+
                 | -P-Q   |       |  -PQB  |
                 +--------+       +--------+
                                                      .
+-------+                 +-------+
|  PA   |c=Q-B            | -PQ-B |
+-------+                 +-------+
                                                      .
      +-------+
      |  P-A  |
      +-------+
}
\label{figure-two-general4}
\hcaption
{The order $C_{\epsilon} \nat(P>A) \nat(Q>B)$
when $\minidx(B)=0$, $PA \not\subseteq Q \neg B$
and $\neg P \cap Q \neg B \not= \emptyset$.}
\end{hfigure}

\

The minimal models of $PA$ are still above all models of $P \neg A$, as seen in
Figure~\ref{figure-two-general4}, contradicting the premise that the first
revision $P>A$ is falsified.


The second sub-subcase is that $QB$ has no model in $\neg P$. It is therefore
all contained in $P \neg A$, since the other part $PA$ only comprises models of
$Q \neg B$.

The conclusion $QB \subseteq P \neg A$ is equivalent to $QB \models P \neg A$,
the second part of the claim.~\qed

An easy consequence of this lemma is that natural revision meets recalcitrance
on the flat order.

\begin{theorem}
\label{natural-recalcitrant}

Natural revision is recalcitrant on the flat order $C_\epsilon$.

\end{theorem}

\proof The converse of the claim is disproved: a second revision invalidates a
first while an order satisfying both exists.

Lemma~\ref{first-unsatisfied-inconsistent} says that if
{} $C_{\epsilon} \nat(P>A) \nat(Q>B)$
falsifies $P>A$, then $PA \models Q \neg B$ and $QB \models P \neg A$.

An order satisfying both conditionals satisfies $P>A$. Some models of $PA$ are
above all models of $P \neg A$. Since all models of $PA$ are models of $Q \neg
B$, this means that some models of $Q \neg A$ are above all models of $P \neg
A$. Since $QB$ is included in $P \neg A$, this means that some models of $Q
\neg A$ are above all models of $QB$. The second conditional is therefore
falsified, contrary to the assumption that the order satisfies both
conditionals.~\qed

Line-down revision is instead not recalcitrant even on this simple case.

\begin{theorem}
\label{down-recalcitrant-not}

Line-down revision is not recalcitrant on the flat ordering $C_\epsilon$.

\end{theorem}

\proof The counterexample is $C_{\epsilon} \dow(x>y) \dow(\true > \neg y \neg
z)$. The revisions $x>y$ and $\true > \neg y \neg z$ are both satisfied by the
order $C=[C(0), C(1), C(2)]$ depicted in Figure~\ref{figure-dow-consistent}.

\begin{eqnarray*}
C(0) &=&
	\{					
		\{\neg x, \neg y, \neg z\}
	\}
\\
C(1) &=&
	\{					
		\{x, y, \neg z\},		
		\{x, y, z\},
		\{\neg x, y, \neg z\},		
		\{\neg x, y, z\},
		\{\neg x, \neg y, z\}
	\},
\\
C(2) &=&
	\{					
		\{x, \neg y, \neg z\},		
		\{x, \neg y, z\}
	\}
\end{eqnarray*}

\begin{hfigure}
\setlength{\unitlength}{3750sp}%
\begin{picture}(2274,1824)(4489,-2623)
\thinlines
{\color[rgb]{0,0,0}\put(4501,-2011){\line( 1, 0){2250}}
}%
{\color[rgb]{0,0,0}\put(4501,-2611){\framebox(2250,1800){}}
}%
{\color[rgb]{0,0,0}\put(4501,-1411){\line( 1, 0){2250}}
}%
\put(4951,-1786){\makebox(0,0)[b]{\smash{\fontsize{9}{10.8}
\usefont{T1}{cmr}{m}{n}{\color[rgb]{0,0,0}$xy$}%
}}}
\put(5401,-2386){\makebox(0,0)[b]{\smash{\fontsize{9}{10.8}
\usefont{T1}{cmr}{m}{n}{\color[rgb]{0,0,0}$x\neg y$}%
}}}
\put(5851,-1786){\makebox(0,0)[b]{\smash{\fontsize{9}{10.8}
\usefont{T1}{cmr}{m}{n}{\color[rgb]{0,0,0}$\neg xy$}%
}}}
\put(6301,-1186){\makebox(0,0)[b]{\smash{\fontsize{9}{10.8}
\usefont{T1}{cmr}{m}{n}{\color[rgb]{0,0,0}$\neg x\neg y\neg z$}%
}}}
\put(6301,-1786){\makebox(0,0)[b]{\smash{\fontsize{9}{10.8}
\usefont{T1}{cmr}{m}{n}{\color[rgb]{0,0,0}$\neg x\neg yz$}%
}}}
\end{picture}%
\nop{
                                  +--------+
                                  | -x-y-z |
                                  +--------+
.
                +----+      +-----+--------+
                | xy |      | -xy | -x-yz  |
                +----+      +-----+--------+
.
                     +-----+
                     | x-y |
                     +-----+
}
\label{figure-dow-consistent}
\hcaption{An order satisfying both $x>y$ and $\true>\neg y\neg z$.}
\end{hfigure}

This order satisfies the first revision $x>y$. The models of $x \wedge y$ are
in $C(1)$, the models of $x \wedge \neg y$ are in $C(2)$. Therefore, $\min(xy)
< x \neg y$ holds. This is the definition of the the order satisfying $x>y$.

This order also satisfies the second revision $\true > \neg x \neg y$. The
conjuction $\true \wedge \neg x \wedge \neg y$ is satisfied by the single model
in $C(0)$. Therefore, its minimal models are $C(0)$, which does not comprise
any other model. As a result, $\min(\true \wedge \neg x \wedge \neg y)$ is
strictly above all models of $\true \wedge \neg(\neg x \wedge \neg y)$

\

The initial order is flat, as in Figure~\ref{figure-flat}: all models are in
the first and only class $C_{\epsilon}(0)$.

\

\begin{hfigure}
\setlength{\unitlength}{3750sp}%
\begin{picture}(1899,399)(4489,-1348)
\thinlines
{\color[rgb]{0,0,0}\put(4501,-1336){\framebox(1875,375){}}
}%
\put(4726,-1186){\makebox(0,0)[b]{\smash{\fontsize{9}{10.8}
\usefont{T1}{cmr}{m}{n}{\color[rgb]{0,0,0}$xy$}%
}}}
\put(5176,-1186){\makebox(0,0)[b]{\smash{\fontsize{9}{10.8}
\usefont{T1}{cmr}{m}{n}{\color[rgb]{0,0,0}$x\neg y$}%
}}}
\put(5626,-1186){\makebox(0,0)[b]{\smash{\fontsize{9}{10.8}
\usefont{T1}{cmr}{m}{n}{\color[rgb]{0,0,0}$\neg xy$}%
}}}
\put(6076,-1186){\makebox(0,0)[b]{\smash{\fontsize{9}{10.8}
\usefont{T1}{cmr}{m}{n}{\color[rgb]{0,0,0}$\neg x\neg y$}%
}}}
\end{picture}%
\nop{
                +----+-----+-----+------+
                | xy | x-y | -xy | -x-y |
                +----+-----+-----+------+
}
\label{figure-flat}
\hcaption{The flat order $C_\epsilon$.}
\end{hfigure}

\

The first revision $x>y$ raises the models of $x \wedge y$ above the others, as
in Figure~\ref{figure-dow-first}.

\

\begin{hfigure}
\setlength{\unitlength}{3750sp}%
\begin{picture}(2274,1224)(4489,-2623)
\thinlines
{\color[rgb]{0,0,0}\put(4501,-2611){\framebox(2250,1200){}}
}%
{\color[rgb]{0,0,0}\put(4501,-2011){\line( 1, 0){2250}}
}%
{\color[rgb]{0,0,0}\put(4951,-2311){\vector( 0, 1){375}}
}%
\put(4951,-1786){\makebox(0,0)[b]{\smash{\fontsize{9}{10.8}
\usefont{T1}{cmr}{m}{n}{\color[rgb]{0,0,0}$xy$}%
}}}
\put(5401,-2386){\makebox(0,0)[b]{\smash{\fontsize{9}{10.8}
\usefont{T1}{cmr}{m}{n}{\color[rgb]{0,0,0}$x\neg y$}%
}}}
\put(5851,-2386){\makebox(0,0)[b]{\smash{\fontsize{9}{10.8}
\usefont{T1}{cmr}{m}{n}{\color[rgb]{0,0,0}$\neg xy$}%
}}}
\put(6301,-2386){\makebox(0,0)[b]{\smash{\fontsize{9}{10.8}
\usefont{T1}{cmr}{m}{n}{\color[rgb]{0,0,0}$\neg x\neg y$}%
}}}
\end{picture}%
\nop{
                +----+
                | xy |
                +----+
.
                     +-----+-----+------+
                     | x-y | -xy | -x-y |
                     +-----+-----+------+
}
\label{figure-dow-first}
\hcaption{The order $C_\epsilon \dow(x>y)$.}
\end{hfigure}

\

The second revision $\true > \neg y \neg z$ raises the models of $\min(\neg y
\neg z)$. The two models of $\neg y \neg z$ are in $x \neg y$ and in $\neg x
\neg y$. Both are in the bottom class. They both make $\min(\neg y \neg z)$.
They both jump at the top, as shown by Figure~\ref{figure-two-case2-2}.

\

\begin{hfigure}
\setlength{\unitlength}{3750sp}%
\begin{picture}(2274,1824)(4489,-2623)
\thinlines
{\color[rgb]{0,0,0}\put(4501,-2011){\line( 1, 0){2250}}
}%
{\color[rgb]{0,0,0}\put(4501,-2611){\framebox(2250,1800){}}
}%
{\color[rgb]{0,0,0}\put(4501,-1411){\line( 1, 0){2250}}
}%
{\color[rgb]{0,0,0}\put(5401,-2236){\vector( 0, 1){975}}
}%
{\color[rgb]{0,0,0}\put(6301,-2161){\vector( 0, 1){900}}
}%
\put(4951,-1786){\makebox(0,0)[b]{\smash{\fontsize{9}{10.8}
\usefont{T1}{cmr}{m}{n}{\color[rgb]{0,0,0}$xy$}%
}}}
\put(5851,-2386){\makebox(0,0)[b]{\smash{\fontsize{9}{10.8}
\usefont{T1}{cmr}{m}{n}{\color[rgb]{0,0,0}$\neg xy$}%
}}}
\put(5401,-1186){\makebox(0,0)[b]{\smash{\fontsize{9}{10.8}
\usefont{T1}{cmr}{m}{n}{\color[rgb]{0,0,0}$x\neg y\neg z$}%
}}}
\put(6301,-1186){\makebox(0,0)[b]{\smash{\fontsize{9}{10.8}
\usefont{T1}{cmr}{m}{n}{\color[rgb]{0,0,0}$\neg x\neg y\neg z$}%
}}}
\put(5401,-2386){\makebox(0,0)[b]{\smash{\fontsize{9}{10.8}
\usefont{T1}{cmr}{m}{n}{\color[rgb]{0,0,0}$x\neg yz$}%
}}}
\put(6301,-2386){\makebox(0,0)[b]{\smash{\fontsize{9}{10.8}
\usefont{T1}{cmr}{m}{n}{\color[rgb]{0,0,0}$\neg x\neg yz$}%
}}}
\end{picture}%
\nop{
                     +-------+     +--------+
                     | x-y-z |     | -x-y-z |
                     +-------+     +--------+
                         ^              ^ 
                +----+   |              |
                | xy |   |              |
                +----+   |              |
                         |              |
                     +-------+-----+--------+
                     | x-yz  | -xy | -x-yz  |
                     +-------+-----+--------+
}
\label{figure-two-case2-2}
\hcaption{The order $C_\epsilon \dow(x>y) \dow(\true > \neg y \neg z)$.}
\end{hfigure}

\

This order falsifies $x>y$ since $x \wedge \neg y$ has some models in class
zero.~\qed

While natural revision is recalcitrant on the flat ordering, it is not in
general. A counterexample with two classes only is shown.

\begin{theorem}
\label{natural-recalcitrant-not}

Natural revision is not recalcitrant on $C_x$, the positive order of $x$.

\end{theorem}

\proof The order $C_x$ compares every model satisfying $x$ strictly less than
every model falsifying it: $C_x = [x, \neg x]$. The revisions are
$\true > y$ and $\true > \neg x$. The first revision lowers the models of $\neg
y$ under the minimal models of $\min(\true \wedge y) = y$. The second lowers
the models of $\neg x$ under the minimal models of $\min(\true \wedge x) = x$.
Figure~\ref{figure-three} shows the sequence of orders.

\

\vbox{
\begin{hfigure}
\noindent
\setlength{\unitlength}{3750sp}%
\begin{picture}(6174,1824)(4489,-2923)
\thinlines
{\color[rgb]{0,0,0}\put(9601,-2911){\framebox(1050,1800){}}
}%
{\color[rgb]{0,0,0}\put(9601,-1711){\line( 1, 0){1050}}
}%
{\color[rgb]{0,0,0}\put(9601,-2311){\line( 1, 0){1050}}
}%
{\color[rgb]{0,0,0}\put(4501,-2611){\framebox(1050,1200){}}
}%
{\color[rgb]{0,0,0}\put(4501,-2011){\line( 1, 0){1050}}
}%
{\color[rgb]{0,0,0}\put(7051,-2911){\framebox(1050,1800){}}
}%
{\color[rgb]{0,0,0}\put(7051,-1711){\line( 1, 0){1050}}
}%
{\color[rgb]{0,0,0}\put(7051,-2311){\line( 1, 0){1050}}
}%
\put(9901,-1486){\makebox(0,0)[b]{\smash{\fontsize{9}{10.8}
\usefont{T1}{cmr}{m}{n}{\color[rgb]{0,0,0}$\neg xy$}%
}}}
\put(10351,-1486){\makebox(0,0)[b]{\smash{\fontsize{9}{10.8}
\usefont{T1}{cmr}{m}{n}{\color[rgb]{0,0,0}$\neg x\neg y$}%
}}}
\put(9901,-2086){\makebox(0,0)[b]{\smash{\fontsize{9}{10.8}
\usefont{T1}{cmr}{m}{n}{\color[rgb]{0,0,0}$xy$}%
}}}
\put(10351,-2686){\makebox(0,0)[b]{\smash{\fontsize{9}{10.8}
\usefont{T1}{cmr}{m}{n}{\color[rgb]{0,0,0}$x\neg y$}%
}}}
\put(8851,-2086){\makebox(0,0)[b]{\smash{\fontsize{9}{10.8}
\usefont{T1}{cmr}{m}{n}{\color[rgb]{0,0,0}$\nat(\true>\neg x)=$}%
}}}
\put(6301,-2086){\makebox(0,0)[b]{\smash{\fontsize{9}{10.8}
\usefont{T1}{cmr}{m}{n}{\color[rgb]{0,0,0}$\nat(\true>y)=$}%
}}}
\put(4801,-1786){\makebox(0,0)[b]{\smash{\fontsize{9}{10.8}
\usefont{T1}{cmr}{m}{n}{\color[rgb]{0,0,0}$xy$}%
}}}
\put(5251,-1786){\makebox(0,0)[b]{\smash{\fontsize{9}{10.8}
\usefont{T1}{cmr}{m}{n}{\color[rgb]{0,0,0}$x\neg y$}%
}}}
\put(4801,-2386){\makebox(0,0)[b]{\smash{\fontsize{9}{10.8}
\usefont{T1}{cmr}{m}{n}{\color[rgb]{0,0,0}$\neg xy$}%
}}}
\put(5251,-2386){\makebox(0,0)[b]{\smash{\fontsize{9}{10.8}
\usefont{T1}{cmr}{m}{n}{\color[rgb]{0,0,0}$\neg x\neg y$}%
}}}
\put(7351,-1486){\makebox(0,0)[b]{\smash{\fontsize{9}{10.8}
\usefont{T1}{cmr}{m}{n}{\color[rgb]{0,0,0}$xy$}%
}}}
\put(7801,-2086){\makebox(0,0)[b]{\smash{\fontsize{9}{10.8}
\usefont{T1}{cmr}{m}{n}{\color[rgb]{0,0,0}$x\neg y$}%
}}}
\put(7351,-2686){\makebox(0,0)[b]{\smash{\fontsize{9}{10.8}
\usefont{T1}{cmr}{m}{n}{\color[rgb]{0,0,0}$\neg xy$}%
}}}
\put(7801,-2686){\makebox(0,0)[b]{\smash{\fontsize{9}{10.8}
\usefont{T1}{cmr}{m}{n}{\color[rgb]{0,0,0}$\neg x\neg y$}%
}}}
\end{picture}%
\nop{
start:
     +-----+   +------+
     | xy  |   |  x-y |
     +-----+   +------+
                                                      .
     +-----+   +------+
     | -xy |   | -x-y |
     +-----+   +------+
                                                      .
T>y:
     +-----+
     | xy  |
     +-----+
                                                      .
               +------+
               |  x-y |
               +------+
                                                      .
     +-----+   +------+
     | -xy |   | -x-y |
     +-----+   +------+
                                                      .
T>-x:
     +-----+   +------+
     | -xy |   | -x-y |
     +-----+   +------+
                                                      .
     +-----+
     |  xy |
     +-----+
                                                      .
               +------+
               | x-y  |
               +------+
}
\label{figure-three}
\hcaption
{The orders $C_x$, $C_x \nat(\true > y)$
and $C_x \nat(\true > y) \nat(\true > \neg x)$.}
\end{hfigure}
}

\

The conditional $\true > y$ requires the minimal models of $\true \wedge y = y$
to be over all models of $\true \wedge \neg y = \neg y$. It is not satisfied
because the minimal models of $y$ are $\neg x y$ alone, which is at the same
level of $\neg x \neg y$, a model of $\neg y$.

The mutual consistency of $\true > y$ and $\true > \neg x$ is proved by the
order $C = [\neg x \wedge y, x \vee \neg y]$, since its top class $C(0)$ only
comprises only the model of $\neg x$ and $y$.

Natural revision does not satisfy the first conditional in spite of mutual
consistency. It fails recalcitrance.~\qed

Incidentally, the order in the proof is the same as that resulting from
revising the flat order by $\true > x$. The order is the same as produced by
$\true > x$, $\true > y$ and $\true > \neg x$ from the flat order. Bruno's cat
again.

\section{Naivety}
\label{morenaive}

\draft

\begin{itemize}

\item strength of a model

\item naivety of orderings

\item natural revision is the unique naivest minimal change that satisfy the
revision and preserve the conditionals

\item does not hold if change minimality is lifted

\end{itemize}

\enddraft

%

Naivety is believing what unconfirmed by the available information. The
available information comes from the revisions $P>A$. A situation is
unconfirmed by $P>A$, but not denied either, if it satisfies $\neg P$. Naivety
is the strength of belief in the models of $\neg P$.

At a first sight, the strength of belief in a model is given by the index of
its class: the lower the index, the stronger the model is believed. This
definition does not work, because a model may compare the same to all other
models in two orderings while belonging to different equivalence classes.

An example is
{} $C = [\{J,J'\}, \{K\}, \{I\}]$
and
{} $C' = [\{J\}, \{J'\}, \{K\}, \{I\}]$
with $P$ being falsified only by $I$. This model $I$ is in the third
equivalence class of the $C$ and the fourth of $C'$. The former ordering is
more naive than the latter, according to class index. Yet, $I$ compares exactly
the same to all other models in the two orderings.

This counterexample suggests to measure the strength of a model by how it
compares to the other models. However, measuring strength by
{} $\{ J \mid I \leq J\}$
does not work because it does not distinguish the strenght of $I$ in
{} $C = [\{I,J\}]$ and $C' = [\{I\}, \{J\}]$
since $I \leq J$ is in both. Measuring strength from
{} $\{ J \mid I < J\}$
does not work either because it does not distinguish the strength of $I$ in
{} $C = [\{I,J\}]$ and $C'' = [\{J\}, \{I\}]$
since $I < J$ holds in neither. Figure~\ref{figure-distinguish} shows how two
models can compare.

\

\begin{hfigure}
\noindent
\begin{tabular}{ccccc}
\setlength{\unitlength}{3750sp}%
\begin{picture}(702,924)(4339,-4423)
\thinlines
{\color[rgb]{0,0,0}\put(4351,-4111){\framebox(600,300){}}
}%
\put(4501,-4036){\makebox(0,0)[b]{\smash{\fontsize{9}{10.8}
\usefont{T1}{cmr}{m}{n}{\color[rgb]{0,0,0}$I$}%
}}}
\put(4801,-4036){\makebox(0,0)[b]{\smash{\fontsize{9}{10.8}
\usefont{T1}{cmr}{m}{n}{\color[rgb]{0,0,0}$J$}%
}}}
\put(5026,-4036){\makebox(0,0)[lb]{\smash{\fontsize{9}{10.8}
\usefont{T1}{cmr}{m}{n}{\color[rgb]{0,0,0}$C$}%
}}}
\end{picture}%
\nop{
+-----+
| I J | C
+-----+
}
& ~~~~~~~~~ &
\setlength{\unitlength}{3750sp}%
\begin{picture}(702,924)(4339,-4423)
\thinlines
{\color[rgb]{0,0,0}\put(4351,-4111){\framebox(600,300){}}
}%
\put(4501,-4036){\makebox(0,0)[b]{\smash{\fontsize{9}{10.8}
\usefont{T1}{cmr}{m}{n}{\color[rgb]{0,0,0}$I$}%
}}}
{\color[rgb]{0,0,0}\put(4351,-4411){\framebox(600,300){}}
}%
\put(4801,-4336){\makebox(0,0)[b]{\smash{\fontsize{9}{10.8}
\usefont{T1}{cmr}{m}{n}{\color[rgb]{0,0,0}$J$}%
}}}
\put(5026,-4036){\makebox(0,0)[lb]{\smash{\fontsize{9}{10.8}
\usefont{T1}{cmr}{m}{n}{\color[rgb]{0,0,0}$C'$}%
}}}
\end{picture}%
\nop{
+-----+
| I   |
+-----+ C'
|   J |
+-----+
}
& ~~~~~~~~~ &
\setlength{\unitlength}{3750sp}%
\begin{picture}(702,924)(4339,-4423)
\thinlines
{\color[rgb]{0,0,0}\put(4351,-4111){\framebox(600,300){}}
}%
\put(4501,-4036){\makebox(0,0)[b]{\smash{\fontsize{9}{10.8}
\usefont{T1}{cmr}{m}{n}{\color[rgb]{0,0,0}$I$}%
}}}
{\color[rgb]{0,0,0}\put(4351,-3811){\framebox(600,300){}}
}%
\put(4801,-3736){\makebox(0,0)[b]{\smash{\fontsize{9}{10.8}
\usefont{T1}{cmr}{m}{n}{\color[rgb]{0,0,0}$J$}%
}}}
\put(5026,-4036){\makebox(0,0)[lb]{\smash{\fontsize{9}{10.8}
\usefont{T1}{cmr}{m}{n}{\color[rgb]{0,0,0}$C''$}%
}}}
\end{picture}%
\nop{
+-----+
|   J |
+-----+ C''
| I   |
+-----+
}
\\
\long\def\ttytex#1#2{#1}
\ttytex{$I \leq J$}{}	&& \ttytex{$I \leq J$}{}&&			\\
\long\def\ttytex#1#2{#1}
\ttytex{$I \not< J$}{}	&& 			&& \ttytex{$I \not< J$}{}
\end{tabular}
\label{figure-distinguish}
\hcaption{The three ways models compare.}
\end{hfigure}

\

The strength of $I$ is given by both
{} $\{ J \mid I \leq J\}$
and
{} $\{ J \mid I < J\}$.
Both sets are necessary to know how $I$ compares to all other models.

\begin{definition}
\label{strength}

The strength of a model $I$ in an ordering $C$ is:

\begin{eqnarray*}
\strength(I,C)	&=&	\l \strength_P(I,C), \strength_E(I,C) \r	\\
\strength_P(I,C)&=&	\{ J \mid I < J \in C \}			\\
\strength_E(I,C)&=&	\{ J \mid I \leq J \in C \}
\end{eqnarray*}

\end{definition}

An ordering is more naive than another if all models of $\neg P$ are stronger
in it than they are in the other. By an abuse of notation, containment by pairs
denotes pairwise containement.

\begin{definition}
\label{more-naive}

An ordering $C$ is at least as naive as another $C'$ on a formula $F$ if
{} $\strength(I,C') \subseteq \strength(I,C)$
holds for all models $I \in F$, where
{} $\l A,B \r \subseteq \l C,D \r$
is
{}
$A \subseteq C$ and $B \subseteq D$.

\end{definition}

In the context of a revision by $P>A$, the formula $F$ is implicitely $\neg P$:
the strength of an order is the strength of the models that are indifferent to
the revision. Naivety is believing what unconfirmed, and what unconfirmed by
$P>A$ are the situations where $P$ is false.

\

%

Natural revision is the unique most naive among the minimally-changing
conditional-preserving revisions. Minimal change is a necessary precondition:
releasing it invalidates not only uniqueness, but even maximality.

\begin{theorem}
\label{natural-not-unique}

Some ordering satisfies $P>A$, preserves the conditionals of $C$ and is
strictly more naive than $C \nat (P>A)$.

\end{theorem}

\proof The ordering $C'$ is like natural revision, but also realizes $I < J$
for all models $I$ of $\neg P L$ and $J \in \min(PA)$, where
{} $L = (\equal\min(P)) \cup \cdots \cup (\equal\min(PA))$.

\long\def\ttytex#1#2{#1}
\ttytex{
\begin{eqnarray*}
\lefteqn{C'= }								\\
&& [ C(0),		\ldots,		C(\minidx(P) - 1),		\\
&& C(\minidx(P)) (-P),	\ldots,		C(\minidx(PA)) (-P),		\\
&& \min(PA),								\\
&& C(\minidx(P)) (-P \cup \neg\min(PA)),
		\ldots,	C(\minidx(PA) (-P \cup \neg\min(PA)),  		\\
&& C(\minidx(P)) \backslash (\neg P \cup \neg\min(PA)),	\ldots,
		C(\minidx(PA) \backslash (\neg P \cup \neg \min(PA)),	\\
&& C(\minidx(PA)+1),	\ldots,		C(m)
]									\\
&=&									\\
&& [ C(0),		\ldots,		C(\minidx(P) - 1),		\\
&& C(\minidx(P)) (-P),	\ldots,		C(\minidx(PA)) (-P),		\\
&& \min(PA),								\\
&& C(\minidx(P)) \backslash (P \neg A \cup \min(PA)),
		\ldots,	C(\minidx(PA) (P \neg A \cup \min(PA)), \\
&& C(\minidx(P)) (P \neg A \cup \neg \min(PA)),	\ldots,
		C(\minidx(PA) (P \neg A \cup \neg \min(PA)),\\
&& C(\minidx(PA)+1),	\ldots,		C(m)]
\end{eqnarray*}
}{
C' = [
        C(0)                         .. C(min(P) - 1)
        C(min(P)) (-P)               .. C(min(PA)) (-P)
        min(PA)
        C(min(P)) (-P u -min(PA))    .. C(min(PA) (-P u -min(PA))
        C(min(P)) - (-P u -min(PA))  .. C(min(PA) - (-P u -min(PA))
        C(min(PA)+1)                 .. C(m)
]
=
[
        C(0)                         .. C(min(P)-1)
        C(min(P)) (-P)               .. C(min(PA)) (-P)
        min(PA)
        C(min(P)) - (P-A u min(PA))  .. C(min(PA)) - (P-A u min(PA))
        C(min(P)) (P-A u min(PA))    .. C(min(PA)) (P-A u min(PA))
        C(min(PA)+1)                 .. C(m)
]
}

\begin{hfigure}
\long\def\ttytex#1#2{#1}
\ttytex{
\begin{tabular}{ccc}
\setlength{\unitlength}{3750sp}%
\begin{picture}(2274,6774)(4489,-7123)
\thinlines
{\color[rgb]{0,0,0}\put(5401,-2761){\framebox(750,450){}}
}%
{\color[rgb]{0,0,0}\put(4951,-2761){\line(-1, 0){450}}
\put(4501,-2761){\line( 0, 1){2400}}
\put(4501,-361){\line( 1, 0){2250}}
\put(6751,-361){\line( 0,-1){2400}}
\put(6751,-2761){\line(-1, 0){450}}
\put(6301,-2761){\line( 0, 1){1650}}
\put(6301,-1111){\line(-1, 0){1350}}
\put(4951,-1111){\line( 0,-1){1650}}
}%
{\color[rgb]{0,0,0}\put(4501,-2161){\line( 1, 0){450}}
}%
{\color[rgb]{0,0,0}\put(6301,-2161){\line( 1, 0){450}}
}%
{\color[rgb]{0,0,0}\put(4501,-961){\line( 1, 0){2250}}
}%
{\color[rgb]{0,0,0}\put(4501,-1561){\line( 1, 0){450}}
}%
{\color[rgb]{0,0,0}\put(6301,-1561){\line( 1, 0){450}}
}%
\put(5626,-2611){\makebox(0,0)[b]{\smash{\fontsize{9}{10.8}
\usefont{T1}{cmr}{m}{n}{\color[rgb]{0,0,0}$PA$}%
}}}
\put(4726,-1261){\makebox(0,0)[b]{\smash{\fontsize{9}{10.8}
\usefont{T1}{cmr}{m}{n}{\color[rgb]{0,0,0}$\neg P$}%
}}}
{\color[rgb]{0,0,0}\put(4501,-5311){\line( 1, 0){2250}}
}%
{\color[rgb]{0,0,0}\put(5401,-5161){\framebox(750,450){}}
}%
{\color[rgb]{0,0,0}\put(5401,-4561){\line(-1, 0){450}}
\put(4951,-4561){\line( 0, 1){1650}}
\put(4951,-2911){\line( 1, 0){1350}}
\put(6301,-2911){\line( 0,-1){1650}}
\put(6301,-4561){\line(-1, 0){150}}
\put(6151,-4561){\line( 0, 1){450}}
\put(6151,-4111){\line(-1, 0){750}}
\put(5401,-4111){\line( 0,-1){450}}
}%
{\color[rgb]{0,0,0}\put(4951,-3961){\line( 1, 0){1350}}
}%
{\color[rgb]{0,0,0}\put(4951,-3361){\line( 1, 0){1350}}
}%
{\color[rgb]{0,0,0}\put(4501,-7111){\framebox(2250,2400){}}
}%
{\color[rgb]{0,0,0}\put(4501,-6511){\line( 1, 0){2250}}
}%
{\color[rgb]{0,0,0}\put(4501,-5911){\line( 1, 0){2250}}
}%
{\color[rgb]{0,0,0}\put(4951,-6361){\framebox(1350,1650){}}
}%
\put(5776,-3586){\makebox(0,0)[b]{\smash{\fontsize{9}{10.8}
\usefont{T1}{cmr}{m}{n}{\color[rgb]{0,0,0}$P\neg A$}%
}}}
\put(6001,-2686){\makebox(0,0)[b]{\smash{\fontsize{9}{10.8}
\usefont{T1}{cmr}{m}{n}{\color[rgb]{0,0,0}$J$}%
}}}
\put(4801,-2011){\makebox(0,0)[b]{\smash{\fontsize{9}{10.8}
\usefont{T1}{cmr}{m}{n}{\color[rgb]{0,0,0}$I$}%
}}}
\put(6076,-3211){\makebox(0,0)[b]{\smash{\fontsize{9}{10.8}
\usefont{T1}{cmr}{m}{n}{\color[rgb]{0,0,0}$K$}%
}}}
\end{picture}%
& ~ ~ ~ ~ ~ &
\setlength{\unitlength}{3750sp}%
\begin{picture}(2274,6924)(4489,-7123)
\thinlines
{\color[rgb]{0,0,0}\put(5401,-3211){\framebox(750,450){}}
}%
{\color[rgb]{0,0,0}\put(4501,-811){\line( 1, 0){2250}}
}%
{\color[rgb]{0,0,0}\put(4501,-1411){\line( 1, 0){450}}
}%
{\color[rgb]{0,0,0}\put(6301,-1411){\line( 1, 0){450}}
}%
{\color[rgb]{0,0,0}\put(6301,-2011){\line( 1, 0){450}}
}%
{\color[rgb]{0,0,0}\put(4501,-2011){\line( 1, 0){450}}
}%
{\color[rgb]{0,0,0}\put(4951,-2611){\line(-1, 0){450}}
\put(4501,-2611){\line( 0, 1){2400}}
\put(4501,-211){\line( 1, 0){2250}}
\put(6751,-211){\line( 0,-1){2400}}
\put(6751,-2611){\line(-1, 0){450}}
\put(6301,-2611){\line( 0, 1){1650}}
\put(6301,-961){\line(-1, 0){1350}}
\put(4951,-961){\line( 0,-1){1650}}
}%
{\color[rgb]{0,0,0}\put(4501,-5911){\line( 1, 0){2250}}
}%
{\color[rgb]{0,0,0}\put(4501,-6511){\line( 1, 0){2250}}
}%
{\color[rgb]{0,0,0}\put(4501,-7111){\framebox(2250,1800){}}
}%
{\color[rgb]{0,0,0}\put(4951,-6361){\framebox(1350,1050){}}
}%
{\color[rgb]{0,0,0}\put(5401,-5761){\framebox(750,450){}}
}%
{\color[rgb]{0,0,0}\put(5401,-5161){\line(-1, 0){450}}
\put(4951,-5161){\line( 0, 1){1650}}
\put(4951,-3511){\line( 1, 0){1350}}
\put(6301,-3511){\line( 0,-1){1650}}
\put(6301,-5161){\line(-1, 0){150}}
\put(6151,-5161){\line( 0, 1){450}}
\put(6151,-4711){\line(-1, 0){750}}
\put(5401,-4711){\line( 0,-1){450}}
}%
{\color[rgb]{0,0,0}\put(4951,-4561){\line( 1, 0){1350}}
}%
{\color[rgb]{0,0,0}\put(4951,-3961){\line( 1, 0){1350}}
}%
\put(5626,-3061){\makebox(0,0)[b]{\smash{\fontsize{9}{10.8}
\usefont{T1}{cmr}{m}{n}{\color[rgb]{0,0,0}$PA$}%
}}}
\put(4726,-1111){\makebox(0,0)[b]{\smash{\fontsize{9}{10.8}
\usefont{T1}{cmr}{m}{n}{\color[rgb]{0,0,0}$\neg P$}%
}}}
\put(5851,-4186){\makebox(0,0)[b]{\smash{\fontsize{9}{10.8}
\usefont{T1}{cmr}{m}{n}{\color[rgb]{0,0,0}$P\neg A$}%
}}}
\put(6001,-3136){\makebox(0,0)[b]{\smash{\fontsize{9}{10.8}
\usefont{T1}{cmr}{m}{n}{\color[rgb]{0,0,0}$J$}%
}}}
\put(4801,-1861){\makebox(0,0)[b]{\smash{\fontsize{9}{10.8}
\usefont{T1}{cmr}{m}{n}{\color[rgb]{0,0,0}$I$}%
}}}
\put(6076,-3811){\makebox(0,0)[b]{\smash{\fontsize{9}{10.8}
\usefont{T1}{cmr}{m}{n}{\color[rgb]{0,0,0}$K$}%
}}}
\end{picture}%
\\
natural revision & & supernaive revision
\end{tabular}
}{
               natural                                supernaive
 +--------------------------------+      +--------------------------------+  
 |                                |      |                                |
 |                                |      |                                |
 |                                |      |                                |
 +--------------------------------+      +--------------------------------+
 |       +--------------------+   |      |       +--------------------+   |
 | -P    |                    |   |      | -P    |                    |   |
 |       |                    |   |      |       |                    |   |
 +-------+                    +---+      +-------+                    +---+
 |       |                    |   |      |       |                    |   |
 |    I  |                    |   |      |    I  |                    |   |
 |       |                    |   |      |       |                    |   |
 +-------+                    +---+      +-------+                    +---+
 |       |                    |   |      |       |                    |   |
 |       |        +-------+   |   |      |       |                    |   |
 |       |        |m(PA) J|   |   |      |       |                    |   |
 +-------+        +-------+   +---+      +-------+                    +---+
.
         +--------------------+                           +-------+
         |           P-A      |                           |m(PA) J|
         |                    |                           +-------+
         +--------------------+                  +--------------------+
         |                    |                  |           P-A      |
         |        +-------+   |                  |                    |
         |   K    |       |   |                  +--------------------+
         +--------+       +---+                  |                    |
                                                 |        +-------+   |
 +-------+--------+-------+---+---+              |   K    |       |   |
 |       |        |       |   |   |              +--------+       +---+
 |       |        +-------+   |   |                                        
 |       +--------------------+   |      +-------+--------+-------+---+---+
 +--------------------------------+      |       |        |       |   |   |
 |                                |      |       |        +-------+   |   |
 |                                |      |       +--------------------+   |
 |                                |      +--------------------------------+
 +--------------------------------+      |                                |
                                         |                                |
                                         |                                |
                                         +--------------------------------+
}
\label{figure-supernatural}
\hcaption{Natural and supernaive revisions.}
\end{hfigure}

This supernaive ordering $C'$ depicted in Figure~\ref{figure-supernatural}
satisfies $P>A$ because it realizes $J < K$ like natural revision does for all
models $K$ of $P \neg A L$. It preserves the conditionals because natural
revision does and its only additional changes are in the comparison between
some models of $\neg P$ and some of $PA$.

The strength of the models $I \in \neg P$ is unchanged except for those in
$(\equal\min(PA))$. The only difference is that $C$ compares them the same to
the models of $\min(PA)$ while $C'$ compares them less. As a result,
{} $\strength_P(I,C) \subset \strength_P(I,C')$.
Since natural revision does not change the comparison between models of $\neg
P$ and models of $PA$, is the same as $C$ on them. Therefore,
{} $\strength_P(I,C) = \strength_P(I,N)$
where $N = C \nat(P>A)$. A consequence is
{} $\strength_P(I,N) \subset \strength_P(I,C')$.

All other strengths are unchanged. As a result, natural revision is strictly
less naive than $C'$. This ordering is not a minimal change because it realizes
$I < J$ in addition to the changes of natural revision. It removes some
comparisons $J \leq I$ that natural revision does not.~\qed

\

%

\draft

technical result, irrelevant because naivey is believing what unconfirmed (-P),
not believing what confirmed (PA)

\

Natural revision is not unique when including the strength of the models of
$PA$ in the comparison of naivety. It is still minimal, but not unique. It is
incomparable with line-down revision:

\begin{itemize}

\item the strength of all models $I$ of $\neg P L$ includes all models of
$\min(PA)$ in natural revision; it does not in line-down revision;

\item the strength of all models of $\min(PA)$ includes the models of $\neg P
(\equal\min(PA))$ in line-down revision; it does in natural revision.

\end{itemize}

\enddraft

%

The previous theorem shows a revision that is strictly more naive than natural
revision. Not a revision of minimal change, however. The minimal distance
requirement makes natural revision the maximally naive ordering.

\begin{theorem}
\label{unique}

Natural revision is the only maximally naive, minimal-distance and
conditional-preserving revision.

\end{theorem}

\proof The claim is proved by assuming that another connected preorder $C'$

\begin{enumerate}

\item verifies $P>A$;

\item preserves the conditionals of $C$ on $P>A$;

\item is one of the closest such connected preordes to $C$;

\item is more naive than natural revision on $P>A$ or equally so.

\end{enumerate}

The proof progressively excludes every difference between $C'$ and natural
revision by showing that every such difference implies that either $C'$ is
strictly less naive than natural revision or that another preorder $C''$

\begin{enumerate}

\item verifies $P>A$;

\item preserves the conditionals of $C$ on $P>A$;

\item is strictly closer than $C'$ to $C$.

\end{enumerate}

The naivety of $C''$ is irrelevant because contradiction derives from being
closer to $C$ than $C'$.

These connected preorders obey the following two rules.

\begin{enumerate}

\item Since $C'$ and $C''$ preserve the conditionals of $C$ on $P>A$, they sort
the models of $\neg P$ like $C$. The first consequence is that models of $\neg
P$ that are equivalent in $C$ are also equivalent in $C'$ and $C''$. Classes of
equivalence of $C$ intersected by $\neg P$ are the same in $C'$ and $C''$.
These intersections are denoted $X$, $X'$, $X''$. The same for $PA$ and $P \neg
A$.

\[
\begin{array}{llll}
X 	&=&	 C(i) \neg P	& \mbox{for some index } i		\\
X' 	&=&	 C(i') \neg P	& \mbox{for some index } i'		\\
X'' 	&=&	 C(i'') \neg P	& \mbox{for some index } i''		\\
Y 	&=&	 C(i) PA						\\
Z 	&=&	 C(i) P\neg A
\end{array}
\]

Connected preorders are tables of such subsets. Irrelevant subsets are spaces,
empty subsets are dots.

\begin{hfigure}
\setlength{\unitlength}{3750sp}%
\begin{picture}(1374,1818)(5239,-4276)
\thinlines
{\color[rgb]{0,0,0}\put(5251,-3061){\line( 1, 0){1350}}
}%
\put(5926,-2611){\makebox(0,0)[b]{\smash{\fontsize{9}{10.8}
\usefont{T1}{cmr}{m}{n}{\color[rgb]{0,0,0}$C$}%
}}}
\put(5476,-2911){\makebox(0,0)[b]{\smash{\fontsize{9}{10.8}
\usefont{T1}{cmr}{m}{n}{\color[rgb]{0,0,0}$\neg P$}%
}}}
\put(5926,-2911){\makebox(0,0)[b]{\smash{\fontsize{9}{10.8}
\usefont{T1}{cmr}{m}{n}{\color[rgb]{0,0,0}$PA$}%
}}}
\put(6376,-2911){\makebox(0,0)[b]{\smash{\fontsize{9}{10.8}
\usefont{T1}{cmr}{m}{n}{\color[rgb]{0,0,0}$P\neg A$}%
}}}
\put(5476,-3361){\makebox(0,0)[b]{\smash{\fontsize{9}{10.8}
\usefont{T1}{cmr}{m}{n}{\color[rgb]{0,0,0}$X$}%
}}}
\put(5926,-3361){\makebox(0,0)[b]{\smash{\fontsize{9}{10.8}
\usefont{T1}{cmr}{m}{n}{\color[rgb]{0,0,0}$.$}%
}}}
\put(5476,-3661){\makebox(0,0)[b]{\smash{\fontsize{9}{10.8}
\usefont{T1}{cmr}{m}{n}{\color[rgb]{0,0,0}$X'$}%
}}}
\put(5926,-3661){\makebox(0,0)[b]{\smash{\fontsize{9}{10.8}
\usefont{T1}{cmr}{m}{n}{\color[rgb]{0,0,0}$Y$}%
}}}
\put(6376,-3661){\makebox(0,0)[b]{\smash{\fontsize{9}{10.8}
\usefont{T1}{cmr}{m}{n}{\color[rgb]{0,0,0}$Z$}%
}}}
\put(5926,-3961){\makebox(0,0)[b]{\smash{\fontsize{9}{10.8}
\usefont{T1}{cmr}{m}{n}{\color[rgb]{0,0,0}$.$}%
}}}
\put(6376,-3961){\makebox(0,0)[b]{\smash{\fontsize{9}{10.8}
\usefont{T1}{cmr}{m}{n}{\color[rgb]{0,0,0}$Z'$}%
}}}
\put(5926,-4261){\makebox(0,0)[b]{\smash{\fontsize{9}{10.8}
\usefont{T1}{cmr}{m}{n}{\color[rgb]{0,0,0}$Y'$}%
}}}
\put(5476,-4261){\makebox(0,0)[b]{\smash{\fontsize{9}{10.8}
\usefont{T1}{cmr}{m}{n}{\color[rgb]{0,0,0}$X''$}%
}}}
\end{picture}%
\nop{
      C
 -P   PA   P-A
---------------
  X    .    
  X'   Y    Z
       .    Z'
  X''  Y'
}
\end{hfigure}

\item Conditional preservation implies that $C'$ and $C''$ sort models of $X$
less than models of $X'$, and models of $X'$ less than models of $X''$, if so
does $C$. Graphically, $C'$ and $C''$ sort each column like $C$. Columns move
up and down with no reshuffle, with the possible insertion and removal of dots,
empty subsets.

\begin{hfigure}
\setlength{\unitlength}{3750sp}%
\begin{picture}(1374,2718)(5239,-5176)
\thinlines
{\color[rgb]{0,0,0}\put(5251,-3061){\line( 1, 0){1350}}
}%
{\color[rgb]{0,0,0}\put(5551,-3586){\vector( 0,-1){225}}
}%
{\color[rgb]{0,0,0}\put(6001,-4036){\vector( 0,-1){600}}
}%
{\color[rgb]{0,0,0}\put(6451,-4036){\vector( 0,-1){900}}
}%
\put(5476,-2911){\makebox(0,0)[b]{\smash{\fontsize{9}{10.8}
\usefont{T1}{cmr}{m}{n}{\color[rgb]{0,0,0}$\neg P$}%
}}}
\put(5926,-2911){\makebox(0,0)[b]{\smash{\fontsize{9}{10.8}
\usefont{T1}{cmr}{m}{n}{\color[rgb]{0,0,0}$PA$}%
}}}
\put(6376,-2911){\makebox(0,0)[b]{\smash{\fontsize{9}{10.8}
\usefont{T1}{cmr}{m}{n}{\color[rgb]{0,0,0}$P\neg A$}%
}}}
\put(5476,-3361){\makebox(0,0)[b]{\smash{\fontsize{9}{10.8}
\usefont{T1}{cmr}{m}{n}{\color[rgb]{0,0,0}$X$}%
}}}
\put(5926,-3361){\makebox(0,0)[b]{\smash{\fontsize{9}{10.8}
\usefont{T1}{cmr}{m}{n}{\color[rgb]{0,0,0}$.$}%
}}}
\put(5926,-3661){\makebox(0,0)[b]{\smash{\fontsize{9}{10.8}
\usefont{T1}{cmr}{m}{n}{\color[rgb]{0,0,0}$Y$}%
}}}
\put(6376,-3661){\makebox(0,0)[b]{\smash{\fontsize{9}{10.8}
\usefont{T1}{cmr}{m}{n}{\color[rgb]{0,0,0}$Z$}%
}}}
\put(5926,-3961){\makebox(0,0)[b]{\smash{\fontsize{9}{10.8}
\usefont{T1}{cmr}{m}{n}{\color[rgb]{0,0,0}$.$}%
}}}
\put(5476,-3961){\makebox(0,0)[b]{\smash{\fontsize{9}{10.8}
\usefont{T1}{cmr}{m}{n}{\color[rgb]{0,0,0}$X'$}%
}}}
\put(5476,-4561){\makebox(0,0)[b]{\smash{\fontsize{9}{10.8}
\usefont{T1}{cmr}{m}{n}{\color[rgb]{0,0,0}$X''$}%
}}}
\put(5476,-3661){\makebox(0,0)[b]{\smash{\fontsize{9}{10.8}
\usefont{T1}{cmr}{m}{n}{\color[rgb]{0,0,0}$.$}%
}}}
\put(5926,-4861){\makebox(0,0)[b]{\smash{\fontsize{9}{10.8}
\usefont{T1}{cmr}{m}{n}{\color[rgb]{0,0,0}$Y'$}%
}}}
\put(6376,-5161){\makebox(0,0)[b]{\smash{\fontsize{9}{10.8}
\usefont{T1}{cmr}{m}{n}{\color[rgb]{0,0,0}$Z'$}%
}}}
\put(5926,-4261){\makebox(0,0)[b]{\smash{\fontsize{9}{10.8}
\usefont{T1}{cmr}{m}{n}{\color[rgb]{0,0,0}$.$}%
}}}
\put(6376,-3961){\makebox(0,0)[b]{\smash{\fontsize{9}{10.8}
\usefont{T1}{cmr}{m}{n}{\color[rgb]{0,0,0}$.$}%
}}}
\put(6376,-4261){\makebox(0,0)[b]{\smash{\fontsize{9}{10.8}
\usefont{T1}{cmr}{m}{n}{\color[rgb]{0,0,0}$.$}%
}}}
\put(5926,-4561){\makebox(0,0)[b]{\smash{\fontsize{9}{10.8}
\usefont{T1}{cmr}{m}{n}{\color[rgb]{0,0,0}$.$}%
}}}
\put(6376,-4561){\makebox(0,0)[b]{\smash{\fontsize{9}{10.8}
\usefont{T1}{cmr}{m}{n}{\color[rgb]{0,0,0}$.$}%
}}}
\put(6376,-4861){\makebox(0,0)[b]{\smash{\fontsize{9}{10.8}
\usefont{T1}{cmr}{m}{n}{\color[rgb]{0,0,0}$.$}%
}}}
\put(5926,-2611){\makebox(0,0)[b]{\smash{\fontsize{9}{10.8}
\usefont{T1}{cmr}{m}{n}{\color[rgb]{0,0,0}$C'$}%
}}}
\end{picture}%
\nop{
      C
 -P   PA   P-A
---------------
  X    .    
  .v   Y    Z
  X'   .    .|
       .|   .|
  X''  .|   .|
       .v   .|
       Y'   .v
            Z'
}
\end{hfigure}

\end{enumerate}

The following terms are used when comparing $C$ and $C'$:

\begin{description}

\item[increase] $X$ increases compared to $Y$ if either:

\begin{itemize}

\item $X$ is strictly less than $Y$ in $C$ and greater than or equivalent to it
in $C'$;

\item $X$ is less than or equal to $Y$ in $C$ and strictly greater than it in
$C'$.

\end{itemize}

\item[decrease] $X$ decreases compared to $Y$ if $Y$ increases compared to $X$.

\end{description}

The same terminology is also applied when comparing $C'$ with $C''$.

The proof progressively excludes every difference between $C'$ and natural
revision.

\begin{enumerate}

\item No $Y$ and no $Z$ decreases compared to any $X$.

Any such decrease would decrease the stregth of $X$. It contradicts the
assumption that $C'$ is more naive than natural revision or equally so, since
natural revision does not decrease the strength of any $X$.

\item No $Y$ increases over any $X$ and any $Z$.

By contradiction, some $Y$ increases compared to some $X$, to some $Z$ or both.

\begin{hfigure}
%
%
\setlength{\unitlength}{3750sp}%
\begin{picture}(1374,1887)(5239,-4348)
\thinlines
{\color[rgb]{0,0,0}\put(5251,-3061){\line( 1, 0){1350}}
}%
{\color[rgb]{0,0,0}\put(5926,-3436){\vector( 0,-1){900}}
}%
\put(5926,-2611){\makebox(0,0)[b]{\smash{\fontsize{9}{10.8}
\usefont{T1}{cmr}{m}{n}{\color[rgb]{0,0,0}$C$}%
}}}
\put(5476,-2911){\makebox(0,0)[b]{\smash{\fontsize{9}{10.8}
\usefont{T1}{cmr}{m}{n}{\color[rgb]{0,0,0}$\neg P$}%
}}}
\put(5926,-2911){\makebox(0,0)[b]{\smash{\fontsize{9}{10.8}
\usefont{T1}{cmr}{m}{n}{\color[rgb]{0,0,0}$PA$}%
}}}
\put(6376,-2911){\makebox(0,0)[b]{\smash{\fontsize{9}{10.8}
\usefont{T1}{cmr}{m}{n}{\color[rgb]{0,0,0}$P\neg A$}%
}}}
\put(5926,-3361){\makebox(0,0)[b]{\smash{\fontsize{9}{10.8}
\usefont{T1}{cmr}{m}{n}{\color[rgb]{0,0,0}$Y$}%
}}}
\put(5476,-3961){\makebox(0,0)[b]{\smash{\fontsize{9}{10.8}
\usefont{T1}{cmr}{m}{n}{\color[rgb]{0,0,0}$X$}%
}}}
\put(6376,-3961){\makebox(0,0)[b]{\smash{\fontsize{9}{10.8}
\usefont{T1}{cmr}{m}{n}{\color[rgb]{0,0,0}$Z$}%
}}}
\end{picture}%
\nop{
     C
------------
     Y
     |
     |     
X    |    Z
     v
                             .
}
\end{hfigure}

Among all such $X$, $Y$ and $Z$, a pair $X/Y$ or $Y/Z$ of minimal interdistance
in $C'$ is selected, two in case of tie on the same $Y$ as in the example
depiction. This selection ensures that no other subset is in between in $C'$.

\begin{hfigure}
%
%
\setlength{\unitlength}{3750sp}%
\begin{picture}(1374,2115)(5239,-4576)
\thinlines
{\color[rgb]{0,0,0}\put(5251,-3061){\line( 1, 0){1350}}
}%
{\color[rgb]{0,0,0}\put(6001,-3436){\vector( 0,-1){900}}
}%
\put(5926,-2611){\makebox(0,0)[b]{\smash{\fontsize{9}{10.8}
\usefont{T1}{cmr}{m}{n}{\color[rgb]{0,0,0}$C'$}%
}}}
\put(5476,-2911){\makebox(0,0)[b]{\smash{\fontsize{9}{10.8}
\usefont{T1}{cmr}{m}{n}{\color[rgb]{0,0,0}$\neg P$}%
}}}
\put(5926,-2911){\makebox(0,0)[b]{\smash{\fontsize{9}{10.8}
\usefont{T1}{cmr}{m}{n}{\color[rgb]{0,0,0}$PA$}%
}}}
\put(6376,-2911){\makebox(0,0)[b]{\smash{\fontsize{9}{10.8}
\usefont{T1}{cmr}{m}{n}{\color[rgb]{0,0,0}$P\neg A$}%
}}}
\put(5476,-3961){\makebox(0,0)[b]{\smash{\fontsize{9}{10.8}
\usefont{T1}{cmr}{m}{n}{\color[rgb]{0,0,0}$X$}%
}}}
\put(6376,-3961){\makebox(0,0)[b]{\smash{\fontsize{9}{10.8}
\usefont{T1}{cmr}{m}{n}{\color[rgb]{0,0,0}$Z$}%
}}}
\put(5926,-4561){\makebox(0,0)[b]{\smash{\fontsize{9}{10.8}
\usefont{T1}{cmr}{m}{n}{\color[rgb]{0,0,0}$Y$}%
}}}
\put(5476,-4261){\makebox(0,0)[b]{\smash{\fontsize{9}{10.8}
\usefont{T1}{cmr}{m}{n}{\color[rgb]{0,0,0}$.$}%
}}}
\put(5476,-4561){\makebox(0,0)[b]{\smash{\fontsize{9}{10.8}
\usefont{T1}{cmr}{m}{n}{\color[rgb]{0,0,0}$.$}%
}}}
\put(5926,-3961){\makebox(0,0)[b]{\smash{\fontsize{9}{10.8}
\usefont{T1}{cmr}{m}{n}{\color[rgb]{0,0,0}$.$}%
}}}
\put(5926,-4261){\makebox(0,0)[b]{\smash{\fontsize{9}{10.8}
\usefont{T1}{cmr}{m}{n}{\color[rgb]{0,0,0}$.$}%
}}}
\put(6376,-4261){\makebox(0,0)[b]{\smash{\fontsize{9}{10.8}
\usefont{T1}{cmr}{m}{n}{\color[rgb]{0,0,0}$.$}%
}}}
\put(6376,-4561){\makebox(0,0)[b]{\smash{\fontsize{9}{10.8}
\usefont{T1}{cmr}{m}{n}{\color[rgb]{0,0,0}$.$}%
}}}
\end{picture}%
\nop{
     C'
------------
                             .
    |
    | 
X   |.    Z
.   v.    .
.    Y    .
}
\end{hfigure}

The connected preorder $C''$ sorts $X$, $Y$ and $Z$ like $C$ and all other
subsets like $C'$.

Graphically, all subsets over $X$, $Y$, $Z$ in $C'$ remain over them and all
under remain under. The figure shows the case of a tie, where $X$ is equivalent
to $Z$; consequently, $Y$ raises just over $X$ and $Z$.

\begin{hfigure}
%
%
\setlength{\unitlength}{3750sp}%
\begin{picture}(1374,2118)(5239,-4576)
\thinlines
{\color[rgb]{0,0,0}\put(5251,-3061){\line( 1, 0){1350}}
}%
{\color[rgb]{0,0,0}\put(6001,-4336){\vector( 0, 1){600}}
}%
\put(5926,-2611){\makebox(0,0)[b]{\smash{\fontsize{9}{10.8}
\usefont{T1}{cmr}{m}{n}{\color[rgb]{0,0,0}$C''$}%
}}}
\put(5476,-2911){\makebox(0,0)[b]{\smash{\fontsize{9}{10.8}
\usefont{T1}{cmr}{m}{n}{\color[rgb]{0,0,0}$\neg P$}%
}}}
\put(5926,-2911){\makebox(0,0)[b]{\smash{\fontsize{9}{10.8}
\usefont{T1}{cmr}{m}{n}{\color[rgb]{0,0,0}$PA$}%
}}}
\put(6376,-2911){\makebox(0,0)[b]{\smash{\fontsize{9}{10.8}
\usefont{T1}{cmr}{m}{n}{\color[rgb]{0,0,0}$P\neg A$}%
}}}
\put(5476,-3961){\makebox(0,0)[b]{\smash{\fontsize{9}{10.8}
\usefont{T1}{cmr}{m}{n}{\color[rgb]{0,0,0}$X$}%
}}}
\put(6376,-3961){\makebox(0,0)[b]{\smash{\fontsize{9}{10.8}
\usefont{T1}{cmr}{m}{n}{\color[rgb]{0,0,0}$Z$}%
}}}
\put(5476,-4261){\makebox(0,0)[b]{\smash{\fontsize{9}{10.8}
\usefont{T1}{cmr}{m}{n}{\color[rgb]{0,0,0}$.$}%
}}}
\put(5476,-4561){\makebox(0,0)[b]{\smash{\fontsize{9}{10.8}
\usefont{T1}{cmr}{m}{n}{\color[rgb]{0,0,0}$.$}%
}}}
\put(5926,-3961){\makebox(0,0)[b]{\smash{\fontsize{9}{10.8}
\usefont{T1}{cmr}{m}{n}{\color[rgb]{0,0,0}$.$}%
}}}
\put(5926,-4261){\makebox(0,0)[b]{\smash{\fontsize{9}{10.8}
\usefont{T1}{cmr}{m}{n}{\color[rgb]{0,0,0}$.$}%
}}}
\put(6376,-4261){\makebox(0,0)[b]{\smash{\fontsize{9}{10.8}
\usefont{T1}{cmr}{m}{n}{\color[rgb]{0,0,0}$.$}%
}}}
\put(6376,-4561){\makebox(0,0)[b]{\smash{\fontsize{9}{10.8}
\usefont{T1}{cmr}{m}{n}{\color[rgb]{0,0,0}$.$}%
}}}
\put(5926,-3661){\makebox(0,0)[b]{\smash{\fontsize{9}{10.8}
\usefont{T1}{cmr}{m}{n}{\color[rgb]{0,0,0}$Y$}%
}}}
\put(5476,-3661){\makebox(0,0)[b]{\smash{\fontsize{9}{10.8}
\usefont{T1}{cmr}{m}{n}{\color[rgb]{0,0,0}$.$}%
}}}
\put(6376,-3661){\makebox(0,0)[b]{\smash{\fontsize{9}{10.8}
\usefont{T1}{cmr}{m}{n}{\color[rgb]{0,0,0}$.$}%
}}}
\end{picture}%
\nop{
     C''
------------
                             .
                             .
                             .
.    Y    .
X   ^.    Z
.   |.    .
.    .    .
}
\end{hfigure}

Since $C''$ sorts some subsets like $C$ and all others like $C'$, it is
strictly closer to $C$ than $C'$. Since $Y$ is greater than or equal to $Z$ in
$C'$, it is not in $\min(PA)$ as otherwise $C'$ would falsify $P>A$. Since $C''$
sorts $\min(PA)$ like $C'$, it also verifies $P>A$.

Overall, $C''$ verifies $P>A$ while preserving its conditionals, and is
strictly closer to $C$ than $C'$. This contradicts the assumption of minimal
distance of $C'$. The assumption that $Y$ increases compared to any $X$ or any
$Z$ is therefore false: no $Y$ increases compared to any $X$ or to any $Z$.

\

The first part of the conclusion is that no $Y$ increases compared to any $X$.
The previous point proves it does not decrease either. A consequence is that
$C$ and $C'$ sort $X$ and $Y$ the same.

\[
X \mbox{ \bf and } Y \mbox{ \bf maintain their order}
\]

The second part of the conclusion is that no $Y$ increases compared to any $Z$.
In the other way around, no $Z$ decreases compared to any $Y$. The previous
point proves that no $Z$ does not decrease compared to any $X$. Overall, no $Z$
decreases over any $X$ and any $Y$.

\[
Z \mbox{ \bf does not decrease compared to } X \mbox{ \bf or } Y
\]

The only change left is some $Z$ increasing over some $X$ or some $Y$.

\item No $Z$ increases compared to some $X$ and not to any $Y$.

By contradiction, $Z$ increases compared to some $X$ and to no $Y$. The figure
shows $Z$ turning from strictly less to strictly more than $X$. The other cases
are analogous.

\begin{hfigure}
%
%
\setlength{\unitlength}{3750sp}%
\begin{picture}(1374,1887)(5239,-4348)
\thinlines
{\color[rgb]{0,0,0}\put(5251,-3061){\line( 1, 0){1350}}
}%
{\color[rgb]{0,0,0}\put(6376,-3436){\vector( 0,-1){900}}
}%
\put(5926,-2611){\makebox(0,0)[b]{\smash{\fontsize{9}{10.8}
\usefont{T1}{cmr}{m}{n}{\color[rgb]{0,0,0}$C$}%
}}}
\put(5476,-2911){\makebox(0,0)[b]{\smash{\fontsize{9}{10.8}
\usefont{T1}{cmr}{m}{n}{\color[rgb]{0,0,0}$\neg P$}%
}}}
\put(5926,-2911){\makebox(0,0)[b]{\smash{\fontsize{9}{10.8}
\usefont{T1}{cmr}{m}{n}{\color[rgb]{0,0,0}$PA$}%
}}}
\put(6376,-2911){\makebox(0,0)[b]{\smash{\fontsize{9}{10.8}
\usefont{T1}{cmr}{m}{n}{\color[rgb]{0,0,0}$P\neg A$}%
}}}
\put(5476,-3961){\makebox(0,0)[b]{\smash{\fontsize{9}{10.8}
\usefont{T1}{cmr}{m}{n}{\color[rgb]{0,0,0}$X$}%
}}}
\put(6376,-3361){\makebox(0,0)[b]{\smash{\fontsize{9}{10.8}
\usefont{T1}{cmr}{m}{n}{\color[rgb]{0,0,0}$Z$}%
}}}
\end{picture}%
\nop{
     C
------------
          Z
          |
X         |
          |
          v
                                    .
}
\end{hfigure}

Among all such pairs $X/Z$, one of minimal interdistance in $C'$ is selected.
This ensures that no subset $X'$ and $Z'$ is between $X$ and $Z$ in $C'$.

No $Y$ is between them as well. Depending on how it compares to $Z$ in $C$,
every such $Y$ either increases compared to $X$ or decreases compared to $Z$.
This is a contradiction with either the first conclusion of the previous point
or with the assumption that $Z$ increases compared to $Y$.

\begin{hfigure}
%
%
\setlength{\unitlength}{3750sp}%
\begin{picture}(1374,2115)(5239,-4576)
\thinlines
{\color[rgb]{0,0,0}\put(5251,-3061){\line( 1, 0){1350}}
}%
{\color[rgb]{0,0,0}\put(6451,-3436){\vector( 0,-1){900}}
}%
\put(5926,-2611){\makebox(0,0)[b]{\smash{\fontsize{9}{10.8}
\usefont{T1}{cmr}{m}{n}{\color[rgb]{0,0,0}$C'$}%
}}}
\put(5476,-2911){\makebox(0,0)[b]{\smash{\fontsize{9}{10.8}
\usefont{T1}{cmr}{m}{n}{\color[rgb]{0,0,0}$\neg P$}%
}}}
\put(5926,-2911){\makebox(0,0)[b]{\smash{\fontsize{9}{10.8}
\usefont{T1}{cmr}{m}{n}{\color[rgb]{0,0,0}$PA$}%
}}}
\put(6376,-2911){\makebox(0,0)[b]{\smash{\fontsize{9}{10.8}
\usefont{T1}{cmr}{m}{n}{\color[rgb]{0,0,0}$P\neg A$}%
}}}
\put(5476,-3961){\makebox(0,0)[b]{\smash{\fontsize{9}{10.8}
\usefont{T1}{cmr}{m}{n}{\color[rgb]{0,0,0}$X$}%
}}}
\put(5926,-3961){\makebox(0,0)[b]{\smash{\fontsize{9}{10.8}
\usefont{T1}{cmr}{m}{n}{\color[rgb]{0,0,0}$.$}%
}}}
\put(5926,-4261){\makebox(0,0)[b]{\smash{\fontsize{9}{10.8}
\usefont{T1}{cmr}{m}{n}{\color[rgb]{0,0,0}$.$}%
}}}
\put(5926,-4561){\makebox(0,0)[b]{\smash{\fontsize{9}{10.8}
\usefont{T1}{cmr}{m}{n}{\color[rgb]{0,0,0}$.$}%
}}}
\put(5476,-4261){\makebox(0,0)[b]{\smash{\fontsize{9}{10.8}
\usefont{T1}{cmr}{m}{n}{\color[rgb]{0,0,0}$.$}%
}}}
\put(5476,-4561){\makebox(0,0)[b]{\smash{\fontsize{9}{10.8}
\usefont{T1}{cmr}{m}{n}{\color[rgb]{0,0,0}$.$}%
}}}
\put(6376,-4561){\makebox(0,0)[b]{\smash{\fontsize{9}{10.8}
\usefont{T1}{cmr}{m}{n}{\color[rgb]{0,0,0}$Z$}%
}}}
\put(6376,-3961){\makebox(0,0)[b]{\smash{\fontsize{9}{10.8}
\usefont{T1}{cmr}{m}{n}{\color[rgb]{0,0,0}$.$}%
}}}
\put(6376,-4261){\makebox(0,0)[b]{\smash{\fontsize{9}{10.8}
\usefont{T1}{cmr}{m}{n}{\color[rgb]{0,0,0}$.$}%
}}}
\end{picture}%
\nop{
     C'
------------
                                    .
           |
X    .    .|
.    .    .|
.    .    .v
.    .    Z
}
\end{hfigure}

The connected preorder $C''$ sorts $X$ and $Z$ like $C$ and all other subsets
like $C'$.

\begin{hfigure}
%
%
\setlength{\unitlength}{3750sp}%
\begin{picture}(1374,2418)(5239,-4876)
\thinlines
{\color[rgb]{0,0,0}\put(5251,-3061){\line( 1, 0){1350}}
}%
{\color[rgb]{0,0,0}\put(6451,-4636){\vector( 0, 1){600}}
}%
\put(5926,-2611){\makebox(0,0)[b]{\smash{\fontsize{9}{10.8}
\usefont{T1}{cmr}{m}{n}{\color[rgb]{0,0,0}$C''$}%
}}}
\put(5476,-2911){\makebox(0,0)[b]{\smash{\fontsize{9}{10.8}
\usefont{T1}{cmr}{m}{n}{\color[rgb]{0,0,0}$\neg P$}%
}}}
\put(5926,-2911){\makebox(0,0)[b]{\smash{\fontsize{9}{10.8}
\usefont{T1}{cmr}{m}{n}{\color[rgb]{0,0,0}$PA$}%
}}}
\put(6376,-2911){\makebox(0,0)[b]{\smash{\fontsize{9}{10.8}
\usefont{T1}{cmr}{m}{n}{\color[rgb]{0,0,0}$P\neg A$}%
}}}
\put(5926,-3961){\makebox(0,0)[b]{\smash{\fontsize{9}{10.8}
\usefont{T1}{cmr}{m}{n}{\color[rgb]{0,0,0}$.$}%
}}}
\put(5926,-4261){\makebox(0,0)[b]{\smash{\fontsize{9}{10.8}
\usefont{T1}{cmr}{m}{n}{\color[rgb]{0,0,0}$.$}%
}}}
\put(6376,-3961){\makebox(0,0)[b]{\smash{\fontsize{9}{10.8}
\usefont{T1}{cmr}{m}{n}{\color[rgb]{0,0,0}$Z$}%
}}}
\put(6376,-4261){\makebox(0,0)[b]{\smash{\fontsize{9}{10.8}
\usefont{T1}{cmr}{m}{n}{\color[rgb]{0,0,0}$.$}%
}}}
\put(5476,-4861){\makebox(0,0)[b]{\smash{\fontsize{9}{10.8}
\usefont{T1}{cmr}{m}{n}{\color[rgb]{0,0,0}$.$}%
}}}
\put(5926,-4861){\makebox(0,0)[b]{\smash{\fontsize{9}{10.8}
\usefont{T1}{cmr}{m}{n}{\color[rgb]{0,0,0}$.$}%
}}}
\put(6376,-4861){\makebox(0,0)[b]{\smash{\fontsize{9}{10.8}
\usefont{T1}{cmr}{m}{n}{\color[rgb]{0,0,0}$.$}%
}}}
\put(5476,-4561){\makebox(0,0)[b]{\smash{\fontsize{9}{10.8}
\usefont{T1}{cmr}{m}{n}{\color[rgb]{0,0,0}$.$}%
}}}
\put(5476,-4261){\makebox(0,0)[b]{\smash{\fontsize{9}{10.8}
\usefont{T1}{cmr}{m}{n}{\color[rgb]{0,0,0}$X$}%
}}}
\put(5476,-3961){\makebox(0,0)[b]{\smash{\fontsize{9}{10.8}
\usefont{T1}{cmr}{m}{n}{\color[rgb]{0,0,0}$.$}%
}}}
\put(5926,-4561){\makebox(0,0)[b]{\smash{\fontsize{9}{10.8}
\usefont{T1}{cmr}{m}{n}{\color[rgb]{0,0,0}$.$}%
}}}
\put(6376,-4561){\makebox(0,0)[b]{\smash{\fontsize{9}{10.8}
\usefont{T1}{cmr}{m}{n}{\color[rgb]{0,0,0}$.$}%
}}}
\end{picture}%
\nop{
     C''
------------
                                    .
                                    .
                                    .
.    .    Z
X    .    .^
.    .    .|
.    .    .|
.    .    .
}
\end{hfigure}

Since $C''$ sorts all models like $C'$ except for $X$ and $Z$, which it sorts
like $C$, it is strictly closer to $C$ than $C'$.

Since $C''$ sorts all $Y$ like $C'$, it sorts all $Y$ to all $Z$ like $C'$.
Therefore, it respects $P>A$ like $C'$ does.

\

The only remaining changes are some $Z$ that increases over some $Y$ and
possibly over some $X$.

\[
Z \mbox{ \bf increases over some } Y \mbox{ \bf and possibly over some } X
\]

These $Y$ and $Z$ may not be equivalent to each other.

\item No $Z$ increases compared to $Y \not\in \min(PA)$;

Proof is by contradiction: $Z$ increases compared to $Y \not\in \min(PA)$. By the
previous points, it may also increase compared to some $X$, which maintain
their order with $Y$. These $X$ may not compare the same as $Y$ like in the
example figure.

\begin{hfigure}
\setlength{\unitlength}{3750sp}%
\begin{picture}(1374,1890)(5239,-4348)
\thinlines
{\color[rgb]{0,0,0}\put(5251,-3061){\line( 1, 0){1350}}
}%
{\color[rgb]{0,0,0}\put(6376,-3436){\vector( 0,-1){900}}
}%
\put(5926,-2611){\makebox(0,0)[b]{\smash{\fontsize{9}{10.8}
\usefont{T1}{cmr}{m}{n}{\color[rgb]{0,0,0}$C$}%
}}}
\put(5476,-2911){\makebox(0,0)[b]{\smash{\fontsize{9}{10.8}
\usefont{T1}{cmr}{m}{n}{\color[rgb]{0,0,0}$\neg P$}%
}}}
\put(5926,-2911){\makebox(0,0)[b]{\smash{\fontsize{9}{10.8}
\usefont{T1}{cmr}{m}{n}{\color[rgb]{0,0,0}$PA$}%
}}}
\put(6376,-2911){\makebox(0,0)[b]{\smash{\fontsize{9}{10.8}
\usefont{T1}{cmr}{m}{n}{\color[rgb]{0,0,0}$P\neg A$}%
}}}
\put(5476,-3961){\makebox(0,0)[b]{\smash{\fontsize{9}{10.8}
\usefont{T1}{cmr}{m}{n}{\color[rgb]{0,0,0}$X$}%
}}}
\put(6376,-3361){\makebox(0,0)[b]{\smash{\fontsize{9}{10.8}
\usefont{T1}{cmr}{m}{n}{\color[rgb]{0,0,0}$Z$}%
}}}
\put(5926,-3961){\makebox(0,0)[b]{\smash{\fontsize{9}{10.8}
\usefont{T1}{cmr}{m}{n}{\color[rgb]{0,0,0}$Y$}%
}}}
\end{picture}%
\nop{
     C
------------
          Z
          |
X    Y    |
          |
          v
                                    .
}
\end{hfigure}

A pair $Z/X$ or $Z/Y$ of minimal interdistance in $C'$ is selected, two in case
of a tie on the same $Z$ like in the example depiction. This ensures that all
subsets in between are empty.

\begin{hfigure}
%
%
\setlength{\unitlength}{3750sp}%
\begin{picture}(1374,2115)(5239,-4576)
\thinlines
{\color[rgb]{0,0,0}\put(5251,-3061){\line( 1, 0){1350}}
}%
{\color[rgb]{0,0,0}\put(6451,-3436){\vector( 0,-1){900}}
}%
\put(5926,-2611){\makebox(0,0)[b]{\smash{\fontsize{9}{10.8}
\usefont{T1}{cmr}{m}{n}{\color[rgb]{0,0,0}$C'$}%
}}}
\put(5476,-2911){\makebox(0,0)[b]{\smash{\fontsize{9}{10.8}
\usefont{T1}{cmr}{m}{n}{\color[rgb]{0,0,0}$\neg P$}%
}}}
\put(5926,-2911){\makebox(0,0)[b]{\smash{\fontsize{9}{10.8}
\usefont{T1}{cmr}{m}{n}{\color[rgb]{0,0,0}$PA$}%
}}}
\put(6376,-2911){\makebox(0,0)[b]{\smash{\fontsize{9}{10.8}
\usefont{T1}{cmr}{m}{n}{\color[rgb]{0,0,0}$P\neg A$}%
}}}
\put(5476,-3961){\makebox(0,0)[b]{\smash{\fontsize{9}{10.8}
\usefont{T1}{cmr}{m}{n}{\color[rgb]{0,0,0}$X$}%
}}}
\put(6376,-4561){\makebox(0,0)[b]{\smash{\fontsize{9}{10.8}
\usefont{T1}{cmr}{m}{n}{\color[rgb]{0,0,0}$Z$}%
}}}
\put(5476,-4261){\makebox(0,0)[b]{\smash{\fontsize{9}{10.8}
\usefont{T1}{cmr}{m}{n}{\color[rgb]{0,0,0}$.$}%
}}}
\put(5476,-4561){\makebox(0,0)[b]{\smash{\fontsize{9}{10.8}
\usefont{T1}{cmr}{m}{n}{\color[rgb]{0,0,0}$.$}%
}}}
\put(5926,-4561){\makebox(0,0)[b]{\smash{\fontsize{9}{10.8}
\usefont{T1}{cmr}{m}{n}{\color[rgb]{0,0,0}$.$}%
}}}
\put(5926,-4261){\makebox(0,0)[b]{\smash{\fontsize{9}{10.8}
\usefont{T1}{cmr}{m}{n}{\color[rgb]{0,0,0}$.$}%
}}}
\put(5926,-3961){\makebox(0,0)[b]{\smash{\fontsize{9}{10.8}
\usefont{T1}{cmr}{m}{n}{\color[rgb]{0,0,0}$Y$}%
}}}
\put(6376,-4261){\makebox(0,0)[b]{\smash{\fontsize{9}{10.8}
\usefont{T1}{cmr}{m}{n}{\color[rgb]{0,0,0}$.$}%
}}}
\put(6376,-3961){\makebox(0,0)[b]{\smash{\fontsize{9}{10.8}
\usefont{T1}{cmr}{m}{n}{\color[rgb]{0,0,0}$.$}%
}}}
\end{picture}%
\nop{
     C'
------------
                                    .
           |
X    Y    .|
.    .    .|
.    .    .v
          Z
}
\end{hfigure}

The connected preorder $C''$ sorts all models like $C'$ except for $Z$ with $Y$
or with $Y$ and $X$, which it sorts like $C$.

\begin{hfigure}
%
%
\setlength{\unitlength}{3750sp}%
\begin{picture}(1374,2418)(5239,-4876)
\thinlines
{\color[rgb]{0,0,0}\put(5251,-3061){\line( 1, 0){1350}}
}%
{\color[rgb]{0,0,0}\put(6451,-4636){\vector( 0, 1){600}}
}%
\put(5476,-2911){\makebox(0,0)[b]{\smash{\fontsize{9}{10.8}
\usefont{T1}{cmr}{m}{n}{\color[rgb]{0,0,0}$\neg P$}%
}}}
\put(5926,-2911){\makebox(0,0)[b]{\smash{\fontsize{9}{10.8}
\usefont{T1}{cmr}{m}{n}{\color[rgb]{0,0,0}$PA$}%
}}}
\put(6376,-2911){\makebox(0,0)[b]{\smash{\fontsize{9}{10.8}
\usefont{T1}{cmr}{m}{n}{\color[rgb]{0,0,0}$P\neg A$}%
}}}
\put(6376,-3961){\makebox(0,0)[b]{\smash{\fontsize{9}{10.8}
\usefont{T1}{cmr}{m}{n}{\color[rgb]{0,0,0}$Z$}%
}}}
\put(6376,-4261){\makebox(0,0)[b]{\smash{\fontsize{9}{10.8}
\usefont{T1}{cmr}{m}{n}{\color[rgb]{0,0,0}$.$}%
}}}
\put(6376,-4561){\makebox(0,0)[b]{\smash{\fontsize{9}{10.8}
\usefont{T1}{cmr}{m}{n}{\color[rgb]{0,0,0}$.$}%
}}}
\put(5476,-4861){\makebox(0,0)[b]{\smash{\fontsize{9}{10.8}
\usefont{T1}{cmr}{m}{n}{\color[rgb]{0,0,0}$.$}%
}}}
\put(5476,-4561){\makebox(0,0)[b]{\smash{\fontsize{9}{10.8}
\usefont{T1}{cmr}{m}{n}{\color[rgb]{0,0,0}$.$}%
}}}
\put(5476,-4261){\makebox(0,0)[b]{\smash{\fontsize{9}{10.8}
\usefont{T1}{cmr}{m}{n}{\color[rgb]{0,0,0}$X$}%
}}}
\put(5926,-4861){\makebox(0,0)[b]{\smash{\fontsize{9}{10.8}
\usefont{T1}{cmr}{m}{n}{\color[rgb]{0,0,0}$.$}%
}}}
\put(5926,-4561){\makebox(0,0)[b]{\smash{\fontsize{9}{10.8}
\usefont{T1}{cmr}{m}{n}{\color[rgb]{0,0,0}$.$}%
}}}
\put(5926,-4261){\makebox(0,0)[b]{\smash{\fontsize{9}{10.8}
\usefont{T1}{cmr}{m}{n}{\color[rgb]{0,0,0}$Y$}%
}}}
\put(6376,-4861){\makebox(0,0)[b]{\smash{\fontsize{9}{10.8}
\usefont{T1}{cmr}{m}{n}{\color[rgb]{0,0,0}$.$}%
}}}
\put(5926,-3961){\makebox(0,0)[b]{\smash{\fontsize{9}{10.8}
\usefont{T1}{cmr}{m}{n}{\color[rgb]{0,0,0}$.$}%
}}}
\put(5476,-3961){\makebox(0,0)[b]{\smash{\fontsize{9}{10.8}
\usefont{T1}{cmr}{m}{n}{\color[rgb]{0,0,0}$.$}%
}}}
\put(5926,-2611){\makebox(0,0)[b]{\smash{\fontsize{9}{10.8}
\usefont{T1}{cmr}{m}{n}{\color[rgb]{0,0,0}$C''$}%
}}}
\end{picture}%
\nop{
     C''
------------
                                    .
                                    .
.    .    Z
X    Y    .^
.    .    .|
.    .    .|
                                    .
}
\end{hfigure}

Since $C''$ sorts all models like $C'$ except for $Z$ with $Y$ or with $Y$ and
$X$, which it sorts like $C$, it is strictly closer to $C$ than $C'$.

Since $Y$ is not in $\min(PA)$, its order with all $Z$ is unchanged from $C'$
to $C''$. Since $C'$ respects $P>A$, so does $C''$.

\

Overall, the only remaining change is $Z$ increasing compared to $\min(PA)$ and
possibly some $X$.

\[
Z \mbox{ \bf increases compared to } \min(PA)
\mbox{ \bf and possibly to some } X
\]

\item No $Z$ increases compared to some $X$ that is strictly greater than
$\min(PA)$.

Proof is by contradiction: some $Z$ increase compared to some $X$ that is
strictly greater than $\min(PA)$.

\begin{hfigure}
%
%
\setlength{\unitlength}{3750sp}%
\begin{picture}(1374,2487)(5239,-4948)
\thinlines
{\color[rgb]{0,0,0}\put(5251,-3061){\line( 1, 0){1350}}
}%
{\color[rgb]{0,0,0}\put(6376,-3436){\vector( 0,-1){1500}}
}%
\put(5926,-2611){\makebox(0,0)[b]{\smash{\fontsize{9}{10.8}
\usefont{T1}{cmr}{m}{n}{\color[rgb]{0,0,0}$C$}%
}}}
\put(5476,-2911){\makebox(0,0)[b]{\smash{\fontsize{9}{10.8}
\usefont{T1}{cmr}{m}{n}{\color[rgb]{0,0,0}$\neg P$}%
}}}
\put(5926,-2911){\makebox(0,0)[b]{\smash{\fontsize{9}{10.8}
\usefont{T1}{cmr}{m}{n}{\color[rgb]{0,0,0}$PA$}%
}}}
\put(6376,-2911){\makebox(0,0)[b]{\smash{\fontsize{9}{10.8}
\usefont{T1}{cmr}{m}{n}{\color[rgb]{0,0,0}$P\neg A$}%
}}}
\put(6376,-3361){\makebox(0,0)[b]{\smash{\fontsize{9}{10.8}
\usefont{T1}{cmr}{m}{n}{\color[rgb]{0,0,0}$Z$}%
}}}
\put(5476,-4561){\makebox(0,0)[b]{\smash{\fontsize{9}{10.8}
\usefont{T1}{cmr}{m}{n}{\color[rgb]{0,0,0}$X$}%
}}}
\put(5926,-3961){\makebox(0,0)[b]{\smash{\fontsize{9}{10.8}
\usefont{T1}{cmr}{m}{n}{\color[rgb]{0,0,0}$\min(PA)$}%
}}}
\end{picture}%
\nop{
     C
------------
          Z
          |
  min(PA) |
          |
X         |
          v
                                    .
}
\end{hfigure}

The previous point proves that no $Z$ increases compred to any $Y \not\in
\min(PA)$. Consequently, all $Y$ between $\min(PA)$ and $Z$ in $C'$ are empty.

\begin{hfigure}
%
%
\setlength{\unitlength}{3750sp}%
\begin{picture}(1374,2715)(5239,-5176)
\thinlines
{\color[rgb]{0,0,0}\put(5251,-3061){\line( 1, 0){1350}}
}%
{\color[rgb]{0,0,0}\put(6451,-3436){\vector( 0,-1){1500}}
}%
\put(5926,-2611){\makebox(0,0)[b]{\smash{\fontsize{9}{10.8}
\usefont{T1}{cmr}{m}{n}{\color[rgb]{0,0,0}$C'$}%
}}}
\put(5476,-2911){\makebox(0,0)[b]{\smash{\fontsize{9}{10.8}
\usefont{T1}{cmr}{m}{n}{\color[rgb]{0,0,0}$\neg P$}%
}}}
\put(5926,-2911){\makebox(0,0)[b]{\smash{\fontsize{9}{10.8}
\usefont{T1}{cmr}{m}{n}{\color[rgb]{0,0,0}$PA$}%
}}}
\put(6376,-2911){\makebox(0,0)[b]{\smash{\fontsize{9}{10.8}
\usefont{T1}{cmr}{m}{n}{\color[rgb]{0,0,0}$P\neg A$}%
}}}
\put(5476,-4561){\makebox(0,0)[b]{\smash{\fontsize{9}{10.8}
\usefont{T1}{cmr}{m}{n}{\color[rgb]{0,0,0}$X$}%
}}}
\put(6376,-5161){\makebox(0,0)[b]{\smash{\fontsize{9}{10.8}
\usefont{T1}{cmr}{m}{n}{\color[rgb]{0,0,0}$Z$}%
}}}
\put(5926,-4561){\makebox(0,0)[b]{\smash{\fontsize{9}{10.8}
\usefont{T1}{cmr}{m}{n}{\color[rgb]{0,0,0}$.$}%
}}}
\put(5926,-4861){\makebox(0,0)[b]{\smash{\fontsize{9}{10.8}
\usefont{T1}{cmr}{m}{n}{\color[rgb]{0,0,0}$.$}%
}}}
\put(5926,-5161){\makebox(0,0)[b]{\smash{\fontsize{9}{10.8}
\usefont{T1}{cmr}{m}{n}{\color[rgb]{0,0,0}$.$}%
}}}
\put(5476,-5161){\makebox(0,0)[b]{\smash{\fontsize{9}{10.8}
\usefont{T1}{cmr}{m}{n}{\color[rgb]{0,0,0}$.$}%
}}}
\put(5476,-4861){\makebox(0,0)[b]{\smash{\fontsize{9}{10.8}
\usefont{T1}{cmr}{m}{n}{\color[rgb]{0,0,0}$.$}%
}}}
\put(5926,-4261){\makebox(0,0)[b]{\smash{\fontsize{9}{10.8}
\usefont{T1}{cmr}{m}{n}{\color[rgb]{0,0,0}$.$}%
}}}
\put(5926,-3961){\makebox(0,0)[b]{\smash{\fontsize{9}{10.8}
\usefont{T1}{cmr}{m}{n}{\color[rgb]{0,0,0}$\min(PA)$}%
}}}
\put(6376,-4261){\makebox(0,0)[b]{\smash{\fontsize{9}{10.8}
\usefont{T1}{cmr}{m}{n}{\color[rgb]{0,0,0}$.$}%
}}}
\put(6376,-4561){\makebox(0,0)[b]{\smash{\fontsize{9}{10.8}
\usefont{T1}{cmr}{m}{n}{\color[rgb]{0,0,0}$.$}%
}}}
\put(6376,-4861){\makebox(0,0)[b]{\smash{\fontsize{9}{10.8}
\usefont{T1}{cmr}{m}{n}{\color[rgb]{0,0,0}$.$}%
}}}
\end{picture}%
\nop{
     C'
------------
                                    .
          |
  min(PA) |
     .    |
X    .    |
.    .    v
.    .    Z
}
\end{hfigure}

Among all $X/Z$ in such conditions, a pair of minimal interdistance in $C$' is
selected. This ensures that all $X'$ and $Z'$ in between are empty.

\begin{hfigure}
%
%
\setlength{\unitlength}{3750sp}%
\begin{picture}(1374,3015)(5239,-5476)
\thinlines
{\color[rgb]{0,0,0}\put(5251,-3061){\line( 1, 0){1350}}
}%
{\color[rgb]{0,0,0}\put(6451,-5236){\vector( 0, 1){600}}
}%
\put(5926,-2611){\makebox(0,0)[b]{\smash{\fontsize{9}{10.8}
\usefont{T1}{cmr}{m}{n}{\color[rgb]{0,0,0}$C''$}%
}}}
\put(5476,-2911){\makebox(0,0)[b]{\smash{\fontsize{9}{10.8}
\usefont{T1}{cmr}{m}{n}{\color[rgb]{0,0,0}$\neg P$}%
}}}
\put(5926,-2911){\makebox(0,0)[b]{\smash{\fontsize{9}{10.8}
\usefont{T1}{cmr}{m}{n}{\color[rgb]{0,0,0}$PA$}%
}}}
\put(6376,-2911){\makebox(0,0)[b]{\smash{\fontsize{9}{10.8}
\usefont{T1}{cmr}{m}{n}{\color[rgb]{0,0,0}$P\neg A$}%
}}}
\put(5926,-4261){\makebox(0,0)[b]{\smash{\fontsize{9}{10.8}
\usefont{T1}{cmr}{m}{n}{\color[rgb]{0,0,0}$.$}%
}}}
\put(5926,-3961){\makebox(0,0)[b]{\smash{\fontsize{9}{10.8}
\usefont{T1}{cmr}{m}{n}{\color[rgb]{0,0,0}$\min(PA)$}%
}}}
\put(6376,-4861){\makebox(0,0)[b]{\smash{\fontsize{9}{10.8}
\usefont{T1}{cmr}{m}{n}{\color[rgb]{0,0,0}$.$}%
}}}
\put(5476,-5461){\makebox(0,0)[b]{\smash{\fontsize{9}{10.8}
\usefont{T1}{cmr}{m}{n}{\color[rgb]{0,0,0}$.$}%
}}}
\put(5926,-5461){\makebox(0,0)[b]{\smash{\fontsize{9}{10.8}
\usefont{T1}{cmr}{m}{n}{\color[rgb]{0,0,0}$.$}%
}}}
\put(5926,-5161){\makebox(0,0)[b]{\smash{\fontsize{9}{10.8}
\usefont{T1}{cmr}{m}{n}{\color[rgb]{0,0,0}$.$}%
}}}
\put(5476,-5161){\makebox(0,0)[b]{\smash{\fontsize{9}{10.8}
\usefont{T1}{cmr}{m}{n}{\color[rgb]{0,0,0}$.$}%
}}}
\put(5476,-4861){\makebox(0,0)[b]{\smash{\fontsize{9}{10.8}
\usefont{T1}{cmr}{m}{n}{\color[rgb]{0,0,0}$X$}%
}}}
\put(5926,-4861){\makebox(0,0)[b]{\smash{\fontsize{9}{10.8}
\usefont{T1}{cmr}{m}{n}{\color[rgb]{0,0,0}$.$}%
}}}
\put(6376,-4561){\makebox(0,0)[b]{\smash{\fontsize{9}{10.8}
\usefont{T1}{cmr}{m}{n}{\color[rgb]{0,0,0}$Z$}%
}}}
\put(6376,-5161){\makebox(0,0)[b]{\smash{\fontsize{9}{10.8}
\usefont{T1}{cmr}{m}{n}{\color[rgb]{0,0,0}$.$}%
}}}
\put(5926,-4561){\makebox(0,0)[b]{\smash{\fontsize{9}{10.8}
\usefont{T1}{cmr}{m}{n}{\color[rgb]{0,0,0}$.$}%
}}}
\put(5476,-4561){\makebox(0,0)[b]{\smash{\fontsize{9}{10.8}
\usefont{T1}{cmr}{m}{n}{\color[rgb]{0,0,0}$.$}%
}}}
\end{picture}%
\nop{
     C''
------------
                                    .
                                    .
  min(PA)  
     .     
.    .    Z
X    .    .^
.    .    .|
.    .    .
}
\end{hfigure}

The preorder $C''$ swaps $X$ and $Z$. It sorts $Z$ and $X$ like $C$ and all
other models like $C'$. It is therefore closer to $C$ than $C''$. Since it
sorts $\min(PA)$ and any other model like $C'$, it respects $P>A$ like $C'$
does.

The only remaining changes are $Z$ increasing over $\min(PA)$ or over some $X$
between it and $\min(PA)$ in $C$. Since $C'$ respects $P>A$, it sorts $Z$
strictly greater than $\min(PA)$.

\[
\begin{array}{c}
Z \mbox{ \bf increases strictly over } \min(PA) \\
\mbox{ \bf and possibly to some } X
\mbox{ \bf less or equal than } \min(PA) \mbox{ \bf in } C
\end{array}
\]

\end{enumerate}

By definition, $Z$ increasing compared to $X$ implies that $X$ is greater than
or equal to $Z$ in $C$.

\begin{hfigure}
%
%
\setlength{\unitlength}{3750sp}%
\begin{picture}(1374,2487)(5239,-4948)
\thinlines
{\color[rgb]{0,0,0}\put(5251,-3061){\line( 1, 0){1350}}
}%
{\color[rgb]{0,0,0}\put(6376,-3436){\vector( 0,-1){1500}}
}%
\put(5926,-2611){\makebox(0,0)[b]{\smash{\fontsize{9}{10.8}
\usefont{T1}{cmr}{m}{n}{\color[rgb]{0,0,0}$C'$}%
}}}
\put(5476,-2911){\makebox(0,0)[b]{\smash{\fontsize{9}{10.8}
\usefont{T1}{cmr}{m}{n}{\color[rgb]{0,0,0}$\neg P$}%
}}}
\put(5926,-2911){\makebox(0,0)[b]{\smash{\fontsize{9}{10.8}
\usefont{T1}{cmr}{m}{n}{\color[rgb]{0,0,0}$PA$}%
}}}
\put(6376,-2911){\makebox(0,0)[b]{\smash{\fontsize{9}{10.8}
\usefont{T1}{cmr}{m}{n}{\color[rgb]{0,0,0}$P\neg A$}%
}}}
\put(6376,-3361){\makebox(0,0)[b]{\smash{\fontsize{9}{10.8}
\usefont{T1}{cmr}{m}{n}{\color[rgb]{0,0,0}$Z$}%
}}}
\put(5926,-4261){\makebox(0,0)[b]{\smash{\fontsize{9}{10.8}
\usefont{T1}{cmr}{m}{n}{\color[rgb]{0,0,0}$\min(PA)$}%
}}}
\put(5476,-3661){\makebox(0,0)[b]{\smash{\fontsize{9}{10.8}
\usefont{T1}{cmr}{m}{n}{\color[rgb]{0,0,0}$X$}%
}}}
\put(5476,-3961){\makebox(0,0)[b]{\smash{\fontsize{9}{10.8}
\usefont{T1}{cmr}{m}{n}{\color[rgb]{0,0,0}$X'$}%
}}}
\put(5476,-4561){\makebox(0,0)[b]{\smash{\fontsize{9}{10.8}
\usefont{T1}{cmr}{m}{n}{\color[rgb]{0,0,0}$.$}%
}}}
\put(5926,-4561){\makebox(0,0)[b]{\smash{\fontsize{9}{10.8}
\usefont{T1}{cmr}{m}{n}{\color[rgb]{0,0,0}$.$}%
}}}
\put(5926,-4861){\makebox(0,0)[b]{\smash{\fontsize{9}{10.8}
\usefont{T1}{cmr}{m}{n}{\color[rgb]{0,0,0}$.$}%
}}}
\put(5476,-4861){\makebox(0,0)[b]{\smash{\fontsize{9}{10.8}
\usefont{T1}{cmr}{m}{n}{\color[rgb]{0,0,0}$.$}%
}}}
\end{picture}%
\nop{
     C
------------
          Z
          |
X         |
X'        |
          |
  min(PA) |
.    .    |
.    .    v
                                    .
}
\end{hfigure}

Natural revision does these changes as well.

It increases $Z$ strictly over $\min(PA)$ because it realizes $\min(PA) < Z$
for every $Z$.

It increases $Z$ strictly over $X$ as well. Since $Z$ contains only models
satisfying $P \neg A$, its models satisfy $P$ as well. Since $X$ is greater
than or equal to $Z$ in $C$, its models are greater than or equal to $\min(P)$.
Since $X$ is less than or equal to $\min(PA)$, they are less than or equal to
$\min(PA)$. Therefore, natural revision realizes $X < Z$.

The conclusion is that $C'$ only changes what natural revision changes. It does
not change $C$ more than natural revision. It does not change it less since
natural revision is a minimal distance revision. Therefore, $C'$ is natural
revision.

\qed

\section{Uncontingent revision}
\label{uncontingent-section}

\draft

\begin{itemize}

\item definition 

\item uncontingent revision on the example

\item natural changes the order strictly less than uncontingent

\item uncontingent revision changes the order minimally to satisfy all
subconditionals

\item reduction natural -> uncontingent:
the contingent context of natural revision

\end{itemize}

\enddraft

%

Uncontingent revision believes every scenario of $PA$ more than every scenario
of $P \neg A$.

\begin{definition}
\label{uncontingent}

The {\em uncontingent revision} of the order $[C(0),\ldots,C(m)]$ by a
conditional formula $P>A$ is as follows.

\long\def\ttytex#1#2{#1}
\ttytex{
\begin{eqnarray*}
\lefteqn{[C(0) \ldots C(m)] \unc(P>A) = } \\
&& [
     C(0)                                \ldots C(\minidx(P)-1) \\
&&   C(\minidx(P)) \backslash (P \neg A) \ldots C(\maxidx(PA)) \backslash (P \neg A) \\
&&   C(\minidx(P))(P \neg A)             \ldots C(\maxidx(PA))(P \neg A) \\
&&   C(\maxidx(PA)+1)                    \ldots C(m)
   ]
\end{eqnarray*}
}{
[C(0) .. C(m)] unc(P>A) = 
[
        C(0)            .. C(min(P)-1)
        C(min(P))-(P-A) .. C(max(PA))-(P-A)
        C(min(P))(P-A)  .. C(max(PA))(P-A)
        C(max(PA)+1)    .. C(m)
]
}

\end{definition}

All models of $P \neg A$ move below all models of $PA$.

The models of $\neg P$ between $\minidx(P)$ and $\maxidx(PA)$ line with some
models of $PA$ and some models of $P \neg A$; since $PA$ moves above $P \neg
A$, they stay with $PA$ by naivety. The models between $\maxidx(PA)+1$ and
$\maxidx(P)$ line with no models of $PA$; they remains aligned with $P \neg A$
by minimal change.

Uncontingent revision only changes the ordering between $\minidx(P)$ and
$\maxidx(PA)$, between the models of
{} $L = (\equal\min(P)) \cup \cdots \cup (\equal\max(PA))$.
Among these models, it sets $J<K$ and $I<K$ for all models $I$, $J$ and $K$ of
respectively $\neg PL$, $PAL$ and $P \neg AL$ since the classes of $P \neg A$
are after those of $\neg(P \neg A) = \neg P \vee PA$.

\

%

Uncontingent revision is correct on $\true > outside$, $\true > meow$ and
$\true > \neg outside$.

\begin{theorem}
\label{uncontingent-general}

Uncontingently revising
{} $[
{}	\{\{x,y\},\{x,\neg y\}\},
{}	\{\{\neg x,y\},\{\neg x,\neg y\}\}
{} ]$
by $\true > y$ gives
{} $[
{}	\{\{x,y\}\},
{}	\{\{\neg x,y\}\},
{}	\{\{x,\neg y\}\},
{}	\{\{\neg x,\neg y\}\}
{} ]$.

\end{theorem}

\proof The order that is revised is $C_x$, depicted in
Figure~\ref{figure-variable-1}. Its equivalence classes are:

\begin{itemize}

\item $C_x(0) = \{\{x,y\},\{x,\neg y\}\}$;

\item $C_x(1) = \{\{\neg x,y\},\{\neg x,\neg y\}\}$.

\end{itemize}

\begin{hfigure}
\setlength{\unitlength}{3750sp}%
\begin{picture}(1374,774)(4489,-1573)
\thinlines
{\color[rgb]{0,0,0}\put(4501,-1186){\line( 1, 0){1350}}
}%
{\color[rgb]{0,0,0}\put(4501,-1561){\framebox(1350,750){}}
}%
\put(4876,-1036){\makebox(0,0)[b]{\smash{\fontsize{9}{10.8}
\usefont{T1}{cmr}{m}{n}{\color[rgb]{0,0,0}$xy$}%
}}}
\put(5476,-1036){\makebox(0,0)[b]{\smash{\fontsize{9}{10.8}
\usefont{T1}{cmr}{m}{n}{\color[rgb]{0,0,0}$x\neg y$}%
}}}
\put(4876,-1411){\makebox(0,0)[b]{\smash{\fontsize{9}{10.8}
\usefont{T1}{cmr}{m}{n}{\color[rgb]{0,0,0}$\neg xy$}%
}}}
\put(5476,-1411){\makebox(0,0)[b]{\smash{\fontsize{9}{10.8}
\usefont{T1}{cmr}{m}{n}{\color[rgb]{0,0,0}$\neg x\neg y$}%
}}}
\end{picture}%
\nop{
+------------+
|  xy   x-y  |
+------------+
| -xy  -x-y  |
+------------+
}
\label{figure-variable-1}
\hcaption{The order $C_x$.}
\end{hfigure}

Uncontingent revision depends on $\minidx(P)$ and $\maxidx(PA)$. Since
$P=\true$ and $A=y$, these numbers are:

\begin{itemize}

\item $\minidx(P) = \minidx(\true) = 0$;

\item $\maxidx(PA) = \maxidx(\true \wedge y) = \maxidx(y) = 1$.

\end{itemize}

The result of uncontingent revision is:

\long\def\ttytex#1#2{#1}
\ttytex{
\begin{eqnarray*}
\lefteqn{C \unc(P>A) = } \\
&& [
     C_x(0)                             \ldots C_x(\minidx(P)-1) \\
&&   C_x(\minidx(P)) \backslash (P \neg A) \ldots C_x(\maxidx(PA)) \backslash (P \neg A) \\
&&   C_x(\minidx(P))(P \neg A)             \ldots C_x(\maxidx(PA))(P \neg A) \\
&&   C_x(\maxidx(PA)+1)                    \ldots C_x(1)
   ]
= \\
&& [
     C_x(0)                       \ldots C_x(0-1) \\
&&   C_x(0) \backslash (P \neg A) \ldots C_x(1) \backslash (P \neg A) \\
&&   C_x(0)(P \neg A)             \ldots C_x(1)(P \neg A) \\
&&   C_x(1+1)                     \ldots C_x(1)
   ]
= \\
&& [
     C_x(0) \backslash (P \neg A), C_x(1) \backslash (P \neg A) \\
&&   C_x(0)(P \neg A), C_x(1)(P \neg A)
   ]
= \\
&& [
     \{\{x,y\},\{x,\neg y\}\} \backslash (T \wedge \neg y),
     \{\{\neg x,y\},\{\neg x,\neg y\}\} \backslash (T \wedge \neg y), \\
&&   \{\{x,y\},\{x,\neg y\}\} \cap (T \wedge \neg y),
     \{\{\neg x,y\},\{\neg x,\neg y\}\} \cap (T \wedge \neg y)
   ]
= \\
&& [
     \{\{x,y\},\{x,\neg y\}\} \backslash \neg y,
     \{\{\neg x,y\},\{\neg x,\neg y\}\} \backslash \neg y, \\
&&   \{\{x,y\},\{x,\neg y\}\} \cap \neg y,
     \{\{\neg x,y\},\{\neg x,\neg y\}\} \cap \neg y
   ]
= \\
&& [
     \{\{x,y\}\},
     \{\{\neg x,y\}\},
     \{\{x,\neg y\}\},
     \{\{\neg x,\neg y\}\}
   ]
\end{eqnarray*}
}{
[Cx(0) .. Cx(m)] unc(P>A) =
[
        Cx(0)            .. Cx(min(P)-1)
        Cx(min(P))-(P-A) .. Cx(max(PA))-(P-A)
        Cx(min(P))(P-A)  .. Cx(max(PA))(P-A)
        Cx(max(PA)+1)    .. Cx(m)
]
=
[
        Cx(0)       .. Cx(0-1)
        Cx(0)-(P-A) .. Cx(1)-(P-A)
        Cx(0)(P-A)  .. Cx(1)(P-A)
        Cx(1+1)     .. Cx(1)
]
=
[
        Cx(0)-(P-A),
        Cx(1)-(P-A),
        Cx(0)(P-A),
        Cx(1)(P-A)
]
=
[
        {{x,y},{x,-y}}-(T-y),
        {{-x,y},{-x,-y}}-(T-y),
        {{x,y},{x,-y}}(T-y),
        {{-x,y},{-x,-y}}(T-y)
]
=
[
        {{x,y},{x,-y}}-(-y),
        {{-x,y},{-x,-y}}-(-y),
        {{x,y},{x,-y}}(-y),
        {{-x,y},{-x,-y}}(-y)
]
=
[
        {{x,y}},
        {{-x,y}},
        {{x,-y}},
        {{-x,-y}}
]
}

\begin{hfigure}
\setlength{\unitlength}{3750sp}%
\begin{picture}(1374,1524)(4489,-2323)
\thinlines
{\color[rgb]{0,0,0}\put(4501,-1186){\line( 1, 0){1350}}
}%
{\color[rgb]{0,0,0}\put(4501,-1561){\line( 1, 0){1350}}
}%
{\color[rgb]{0,0,0}\put(4501,-1936){\line( 1, 0){1350}}
}%
{\color[rgb]{0,0,0}\put(4501,-2311){\framebox(1350,1500){}}
}%
\put(4876,-1036){\makebox(0,0)[b]{\smash{\fontsize{9}{10.8}
\usefont{T1}{cmr}{m}{n}{\color[rgb]{0,0,0}$xy$}%
}}}
\put(4876,-1411){\makebox(0,0)[b]{\smash{\fontsize{9}{10.8}
\usefont{T1}{cmr}{m}{n}{\color[rgb]{0,0,0}$\neg xy$}%
}}}
\put(5476,-1786){\makebox(0,0)[b]{\smash{\fontsize{9}{10.8}
\usefont{T1}{cmr}{m}{n}{\color[rgb]{0,0,0}$x\neg y$}%
}}}
\put(5476,-2161){\makebox(0,0)[b]{\smash{\fontsize{9}{10.8}
\usefont{T1}{cmr}{m}{n}{\color[rgb]{0,0,0}$\neg x\neg y$}%
}}}
\end{picture}%
\nop{
+------------+
|  xy        |
+------------+
| -xy        |
+------------+
|       x-y  |
+------------+
|      -x-y  |
+------------+
}
\label{figure-linear-1}
\hcaption{The order $C_x \unc(\true>y)$.}
\end{hfigure}

This order is shown in Figure~\ref{figure-linear-1}, and is the order of the
claim.~\qed

Uncontingent revision is also correct on $\true > outside$, $outside > cat$,
$\true > \neg outside$.

\begin{theorem}
\label{uncontingent-context}

Uncontingently revising
{} $[
{}      \{\{x,y\},\{x,\neg y\}\},
{}      \{\{\neg x,y\},\{\neg x,\neg y\}\}
{} ]$
by $x>y$ gives
{} $[
{}      \{\{x,y\}\},
{}      \{\{x,\neg y\}\},
{}      \{\{\neg x,y\},\{\neg x,\neg y\}\}
{} ]$.

\end{theorem}

\proof The order to be revised is $C_x$, shown in
Figure~\ref{figure-variable-2}. Its equivalence classes are:

\begin{itemize}

\item $C_x(0) = \{\{x,y\},\{x,\neg y\}\}$;

\item $C_x(1) = \{\{\neg x,y\},\{\neg x,\neg y\}\}$.

\end{itemize}

\begin{hfigure}
\setlength{\unitlength}{3750sp}%
\begin{picture}(1374,774)(4489,-1573)
\thinlines
{\color[rgb]{0,0,0}\put(4501,-1186){\line( 1, 0){1350}}
}%
{\color[rgb]{0,0,0}\put(4501,-1561){\framebox(1350,750){}}
}%
\put(4876,-1036){\makebox(0,0)[b]{\smash{\fontsize{9}{10.8}
\usefont{T1}{cmr}{m}{n}{\color[rgb]{0,0,0}$xy$}%
}}}
\put(5476,-1036){\makebox(0,0)[b]{\smash{\fontsize{9}{10.8}
\usefont{T1}{cmr}{m}{n}{\color[rgb]{0,0,0}$x\neg y$}%
}}}
\put(4876,-1411){\makebox(0,0)[b]{\smash{\fontsize{9}{10.8}
\usefont{T1}{cmr}{m}{n}{\color[rgb]{0,0,0}$\neg xy$}%
}}}
\put(5476,-1411){\makebox(0,0)[b]{\smash{\fontsize{9}{10.8}
\usefont{T1}{cmr}{m}{n}{\color[rgb]{0,0,0}$\neg x\neg y$}%
}}}
\end{picture}%
\nop{
+------------+
|  xy   x-y  |
+------------+
| -xy  -x-y  |
+------------+
}
\label{figure-variable-2}
\hcaption{The order $C_x$.}
\end{hfigure}

Uncontingent revision depends on $\minidx(P)$ and $\maxidx(PA)$. Since $P=x$
and $A=y$, these numbers are:

\begin{itemize}

\item $\minidx(P) = \minidx(x) = 0$;

\item $\maxidx(PA) = \maxidx(x \wedge y) = 0$.

\end{itemize}

The result of uncontingent revision is:

\long\def\ttytex#1#2{#1}
\ttytex{
\begin{eqnarray*}
\lefteqn{[C_x(0) \ldots C_x(m)] \unc(P>A) = } \\
&& [
     C_x(0)                             \ldots C_x(\minidx(P)-1) \\
&&   C_x(\minidx(P)) \backslash (P \neg A) \ldots
				C_x(\maxidx(PA)) \backslash (P \neg A) \\
&&   C_x(\minidx(P))(P \neg A)          \ldots C_x(\maxidx(PA))(P \neg A) \\
&&   C_x(\maxidx(PA)+1)                 \ldots C_x(m)
   ] = \\
&& [
     C_x(0)                       \ldots C_x(0-1) \\
&&   C_x(0) \backslash (P \neg A) \ldots C_x(0) \backslash (P \neg A) \\
&&   C_x(0)(P \neg A)             \ldots C_x(0)(P \neg A) \\
&&   C_x(0+1)                     \ldots C_x(1)
   ] = \\
&& [
     C_x(0) \backslash (P \neg A) \\
&&   C_x(0)(P \neg A) \\
&&   C_x(1)
   ] = \\
&& [
     \{\{x,y\},\{x,\neg y\}\} \backslash (x \wedge \neg y) \\
&&   \{\{x,y\},\{x,\neg y\}\} \cap (x \wedge \neg y) \\
&&   \{\{\neg x,y\},\{\neg x,\neg y\}\}
   ] = \\
&& [
     \{\{x,y\}\} \\
&&   \{\{x,\neg y\}\} \\
&&   \{\{\neg x,y\},\{\neg x,\neg y\}\}
   ]
\end{eqnarray*}
}{
[Cx(0) .. Cx(m)] unc(P>A) = 
[
        Cx(0)            .. Cx(min(P)-1)
        Cx(min(P))-(P-A) .. Cx(max(PA))-(P-A)
        Cx(min(P))(P-A)  .. Cx(max(PA))(P-A)
        Cx(max(PA)+1)    .. Cx(m)
]
=
[
        Cx(0)       .. Cx(0-1)
        Cx(0)-(P-A) .. Cx(0)-(P-A)
        Cx(0)(P-A)  .. Cx(0)(P-A)
        Cx(0+1)     .. Cx(1)
]
=
[
        Cx(0)-(P-A)
        Cx(0)(P-A)
        Cx(1)
]
=
[
        {{x,y},{x,-y}}-(x-y)
        {{x,y},{x,-y}}(x-y)
        {{-x,y},{-x,-y}}
]
=
[
        {{x,y}}
        {{x,-y}}
        {{-x,y},{-x,-y}}
]
}

\begin{hfigure}
\setlength{\unitlength}{3750sp}%
\begin{picture}(1374,1149)(4489,-1948)
\thinlines
{\color[rgb]{0,0,0}\put(4501,-1186){\line( 1, 0){1350}}
}%
{\color[rgb]{0,0,0}\put(4501,-1561){\line( 1, 0){1350}}
}%
{\color[rgb]{0,0,0}\put(4501,-1936){\framebox(1350,1125){}}
}%
\put(4876,-1036){\makebox(0,0)[b]{\smash{\fontsize{9}{10.8}
\usefont{T1}{cmr}{m}{n}{\color[rgb]{0,0,0}$xy$}%
}}}
\put(5476,-1411){\makebox(0,0)[b]{\smash{\fontsize{9}{10.8}
\usefont{T1}{cmr}{m}{n}{\color[rgb]{0,0,0}$x\neg y$}%
}}}
\put(4876,-1786){\makebox(0,0)[b]{\smash{\fontsize{9}{10.8}
\usefont{T1}{cmr}{m}{n}{\color[rgb]{0,0,0}$\neg xy$}%
}}}
\put(5476,-1786){\makebox(0,0)[b]{\smash{\fontsize{9}{10.8}
\usefont{T1}{cmr}{m}{n}{\color[rgb]{0,0,0}$\neg x\neg y$}%
}}}
\end{picture}%
\nop{
+------------+
|  xy        |
+------------+
|       x-y  |
+------------+
| -xy  -x-y  |
+------------+
}
\label{figure-top-bis}
\hcaption{The order $C_x \unc(x > y)$.}
\end{hfigure}

This order is shown in Figure~\ref{figure-top-bis} and is the order of the
claim.~\qed

\

%

Uncontingent revision is not a revision of minimal change. Natural revision
changes less.

Theorem~\ref{natural-minimal} proves natural revision minimal, but does not
prove it uniquely so. It does not exclude distance incomparability with
uncontingent revision. The following theorem does.

\begin{theorem}
\label{natural-lesschange-uncontingent}

Natural revision is closer or equally distant to the revised order than
uncontingent revision.

\end{theorem}

\proof Natural revision only moves models between $\minidx(P)$ and
$\minidx(PA)$. Uncontingent revision only moves models between $\minidx(P)$ and
$\maxidx(PA)$. These are the models of the following formulae.

\begin{eqnarray*}
L	&=&	(\equal\min(P)) \cup \cdots \cup (\equal\min(PA))	\\
L'	&=&	(\equal\min(P)) \cup \cdots \cup (\equal\max(PA))
\end{eqnarray*}

Every change of natural revision is also done by uncontingent revision.

Natural revision only drops all models $K \in P \neg AL$ below all models $I
\in \neg PL$ and $J \in PAL$. It realizes $J<K$ and $I<K$. Since $L \subseteq
L'$, the models $I$, $J$ and $K$ are also respectively models of $\neg PL'$,
$PAL'$ and $P \neg AL'$. Uncontingent revision drops all models of $P \neg
AL'$, including $K$, below all models of $\neg PL'$ and $PAL'$, including $J$
and $I$. It realizes all comparisons $J<K$ and $I<K$ realized by natural
revision.~\qed

Most revisions coincide on very simple cases, like the flat ordering revised by
$\true>x$. None changes the order strictly less than another. None of the
theorem cat state that a revision change the order strictly less than another.
The strict version of above theorem does not holds for all orderings and
revisions. Yet, it holds for some: natural revision changes the ordering less
than or the same as uncontingent revision, in some cases strictly so.

\begin{theorem}
\label{natural-strictly-lesschange-uncontingent}

Natural revision $C_x \nat(\true>y)$ is strictly closer to $C_x$ than
$C_x \unc(\true>y)$.

\end{theorem}

\proof The equivalence classes of $C_x$, shown in Figure~\ref{figure-variable},
are:

\begin{eqnarray*}
C_x(0)			&=&	\{\{     x,y\}, \{     x,\neg y\}\}	\\
C_x(1)			&=&	\{\{\neg x,y\}, \{\neg x,\neg y\}\}
\end{eqnarray*}

\begin{hfigure}
\setlength{\unitlength}{3750sp}%
\begin{picture}(1374,774)(4489,-1573)
\thinlines
{\color[rgb]{0,0,0}\put(4501,-1186){\line( 1, 0){1350}}
}%
{\color[rgb]{0,0,0}\put(4501,-1561){\framebox(1350,750){}}
}%
\put(4876,-1036){\makebox(0,0)[b]{\smash{\fontsize{9}{10.8}
\usefont{T1}{cmr}{m}{n}{\color[rgb]{0,0,0}$xy$}%
}}}
\put(5476,-1036){\makebox(0,0)[b]{\smash{\fontsize{9}{10.8}
\usefont{T1}{cmr}{m}{n}{\color[rgb]{0,0,0}$x\neg y$}%
}}}
\put(4876,-1411){\makebox(0,0)[b]{\smash{\fontsize{9}{10.8}
\usefont{T1}{cmr}{m}{n}{\color[rgb]{0,0,0}$\neg xy$}%
}}}
\put(5476,-1411){\makebox(0,0)[b]{\smash{\fontsize{9}{10.8}
\usefont{T1}{cmr}{m}{n}{\color[rgb]{0,0,0}$\neg x\neg y$}%
}}}
\end{picture}%
\nop{
+------------+
|  xy   x-y  |
+------------+
| -xy  -x-y  |
+------------+
}
\label{figure-variable}
\hcaption{The order $C_x$.}
\end{hfigure}

Theorem~\ref{natural-nocondition} tells the result of natural revision: is the
order in Figure~\ref{figure-top}.

\begin{eqnarray*}
C_x \nat(\true>y) (0)	& = &	\{\{     x,y\}                   \}	\\
C_x \nat(\true>y) (1)	& = &	\{              \{     x,\neg y\}\}	\\
C_x \nat(\true>y) (2)	& = &	\{\{\neg x,y\}, \{\neg x,\neg y\}\}	\\
\end{eqnarray*}

\begin{hfigure}
\setlength{\unitlength}{3750sp}%
\begin{picture}(1374,1149)(4489,-1948)
\thinlines
{\color[rgb]{0,0,0}\put(4501,-1186){\line( 1, 0){1350}}
}%
{\color[rgb]{0,0,0}\put(4501,-1561){\line( 1, 0){1350}}
}%
{\color[rgb]{0,0,0}\put(4501,-1936){\framebox(1350,1125){}}
}%
\put(4876,-1036){\makebox(0,0)[b]{\smash{\fontsize{9}{10.8}
\usefont{T1}{cmr}{m}{n}{\color[rgb]{0,0,0}$xy$}%
}}}
\put(5476,-1411){\makebox(0,0)[b]{\smash{\fontsize{9}{10.8}
\usefont{T1}{cmr}{m}{n}{\color[rgb]{0,0,0}$x\neg y$}%
}}}
\put(4876,-1786){\makebox(0,0)[b]{\smash{\fontsize{9}{10.8}
\usefont{T1}{cmr}{m}{n}{\color[rgb]{0,0,0}$\neg xy$}%
}}}
\put(5476,-1786){\makebox(0,0)[b]{\smash{\fontsize{9}{10.8}
\usefont{T1}{cmr}{m}{n}{\color[rgb]{0,0,0}$\neg x\neg y$}%
}}}
\end{picture}%
\nop{
+------------+
|  xy        |
+------------+
|       x-y  |
+------------+
| -xy  -x-y  |
+------------+
}
\label{figure-top}
\hcaption{The order $C_x \nat(\true>y)$.}
\end{hfigure}

Natural revision turns
{} $\{x,y\} \equiv \{x,\neg y\}$
into
{} $\{x,y\} < \{x,\neg y\}$.
In terms of sets, it removes the pair
{} $\{x,\neg y\} \leq \{x,y\}$.
This is its only change:
$\diff(C_x, C_x \nat(\true>y)) = \{\{x,\neg y\} \leq \{x,y\}\}$.

Theorem~\ref{uncontingent-general} tells the result of uncontingent revision.
It is shown in Figure~\ref{figure-linear-2}.

\begin{eqnarray*}
C_x \unc(\true>y) (0)	& = &	\{\{     x,y\}                   \}	\\
C_x \unc(\true>y) (1)	& = &	\{\{\neg x,y\}                   \}	\\
C_x \unc(\true>y) (2)	& = &	\{              \{     x,\neg y\}\}	\\
C_x \unc(\true>y) (3)	& = &	\{              \{\neg x,\neg y\}\}
\end{eqnarray*}

\begin{hfigure}
\setlength{\unitlength}{3750sp}%
\begin{picture}(1374,1524)(4489,-2323)
\thinlines
{\color[rgb]{0,0,0}\put(4501,-1186){\line( 1, 0){1350}}
}%
{\color[rgb]{0,0,0}\put(4501,-1561){\line( 1, 0){1350}}
}%
{\color[rgb]{0,0,0}\put(4501,-1936){\line( 1, 0){1350}}
}%
{\color[rgb]{0,0,0}\put(4501,-2311){\framebox(1350,1500){}}
}%
\put(4876,-1036){\makebox(0,0)[b]{\smash{\fontsize{9}{10.8}
\usefont{T1}{cmr}{m}{n}{\color[rgb]{0,0,0}$xy$}%
}}}
\put(4876,-1411){\makebox(0,0)[b]{\smash{\fontsize{9}{10.8}
\usefont{T1}{cmr}{m}{n}{\color[rgb]{0,0,0}$\neg xy$}%
}}}
\put(5476,-1786){\makebox(0,0)[b]{\smash{\fontsize{9}{10.8}
\usefont{T1}{cmr}{m}{n}{\color[rgb]{0,0,0}$x\neg y$}%
}}}
\put(5476,-2161){\makebox(0,0)[b]{\smash{\fontsize{9}{10.8}
\usefont{T1}{cmr}{m}{n}{\color[rgb]{0,0,0}$\neg x\neg y$}%
}}}
\end{picture}%
\nop{
+------------+
|  xy        |
+------------+
|       x-y  |
+------------+
| -xy        |
+------------+
|      -x-y  |
+------------+
}
\label{figure-linear-2}
\hcaption{The order $C_\epsilon \unc(\true>y)$.}
\end{hfigure}

Uncontingent revision turns
{} $\{x,y\} \equiv \{x,\neg y\}$
into
{} $\{x,y\} < \{x,\neg y\}$,
like natural revision does; it removes the pair
{} $\{x,\neg y\} \leq \{x,y\}$.
Additionally, it turns
{} $\{\neg x,y\} \equiv \{\neg x,\neg y\}$
into
{} $\{\neg x,y\} < \{\neg x,\neg y\}$.
It removes the pair
{} $\{\neg x,\neg y\} \leq \{\neg x,y\}$.
The difference $\diff(C_x, C_x \unc(\true>y))$ contains
{} $\{\neg x,\neg y\} \leq \{\neg x,y\}$,
in addition to
{} $\{x,\neg y\} \leq \{x,y\}$.

The containment
{} $\diff(C_x, C_x \nat(\true>y)) \subset \diff(C_x, C_x \unc(\true>y))$
defines natural revision strictly closer to $C_x$ than uncontingent
revision.~\qed

Uncontingent revision is incomparable with line-down revision on their amount
of change.

\

%

Uncontingent revision is not a minimal change to satisfy $P>A$, it is not the
minimal change to make some models of $PA$ more believed than all models of $P
\neg A$. The following theorem shows that it is the minimal change to believe
all models of $PA$ more than all models of $P \neg A$. In short, it believes
$A$ in all conditions that are consistent with $PA$. The theorem that proves it
calls $R$ a formula identifying any such condition.

\begin{theorem}
\label{uncontingent-minimal}

No connected preorder satisfying $RP>A$ for every $R$ consistent with $PA$
is strictly closer to $C$ than uncontingent revision $C \unc(P>A)$.

\end{theorem}

\draft

\[
\forall R ~.~
R \wedge P \wedge A \not\models \bot 
\Rightarrow
\not\exists C' ~.~
	C' \mbox{ verifies } RP>A ,~
	\diff(C',C) \subset \diff(C \unc(RP>A),C)
\]

\enddraft

\proof The claim is proved by deriving a contradition from the assumptions that
a connected preorder $C'$:

\begin{enumerate}

\item \label{uncontingent-minimal-conditional}

compares some model of $RPA$ strictly less than every model of $RP\neg A$ for
every $R$ that is consistent with $PA$;

\item \label{uncontingent-minimal-same}

compares every pair of models as $C$ if uncontingent revision does the same;

\item \label{uncontingent-minimal-differ}

differs from uncontingent revision.

\end{enumerate}

Since $C'$ differs from uncontingent revision by
Assumption~\ref{uncontingent-minimal-differ}, it compares two models
differently from uncontingent revision. If uncontingent revision compares them
the same as $C$, Assumption~\ref{uncontingent-minimal-same} is violated. The
other case is that this comparison differs between $C$ and $C \unc(P>A)$. Since
it also differs between $C \unc(P>A)$ and $C'$, it is the same between $C$ and
$C'$.

For each comparison changed by uncontingent revision, $C'$ is assumed to agree
with $C$ on it, and contradiction is derived.

Uncontingent revision only moves models between $\minidx(P)$ and $\maxidx(PA)$.
A formula satisfied by these models is named $L$.

\[
L = (\equal\min(P)) \cup \cdots \cup (\equal\max(PA))
\]

Among the models of $L$, uncontingent revision only changes the order between
the models of $P \neg A$ and the models of $PA$ and $\neg P$. Arbitrary such
models are named as follows.

\begin{eqnarray*}
I	& \in &		\neg P L			\\
J	& \in &		PA				\\
K	& \in &		P \neg A L
\end{eqnarray*}

Uncontingent revision only changes the order by turning all models $I$ and $J$
strictly less than all models $K$: it realizes $J<K$ and $I<K$. It adds $I \leq
K$ and $J \leq K$ and removes $K \leq I$ and $K \leq J$.

Each such a comparison is assumed to be agreed by $C'$ and $C$ in turn.

\begin{description}

\item[$K \leq J \in C'$:]

%
%
the condition $R = \{J, K\}$ is consistent with both $P$ and $A$, since its
first model $J$ satisfies both;
%
%
since $J$ is a model of $PA$, it is also a model of $RPA$; since $K$ is a model
of $P \neg A$, it is not a model of $PA$ and therefore not of $RPA$ either; as
a result, $J$ is the only model of $RPA$;
%
%
since $K$ is a model of $P \neg A$, it is also a model of $RP\neg A$;
%
%
by Assumption~\ref{uncontingent-minimal-conditional}, $C'$ satisfies $RP>A$,
which requires it to imply $J < K$ since $J$ is the only model of $RPA$ and $K$
is a model of $RP\neg A$;
%
%
this contradicts the assumption $K \leq J \in C'$.

\item[$J \leq K \not\in C'$:]

since $C'$ does not contain $K \leq J$ by the point above, if it does not
contain $J \leq K$ either it is not a connected preorder, violating the
assumption of the lemma;

the conclusion of the first two points is that $C'$ sorts $J$ strictly less
than $K$;

\item[$K \leq I \in C'$:]

%
%
by definition, $I$ is a model of $\neg P L$, where
{} $L = (\equal\min(P)) \cup \cdots \cup (\equal\max(PA))$:
the index of its equivalence class in $C$ is less than or equal to some model
$J$ of $PA$; uncontingent revision does
not change the comparisons between the models of $\neg P$ and the models of
$PA$; therefore, $I \leq J$ is in $C \unc(P>A)$;
%
%
since $I \leq J$ is in both $C$ and $C \unc(P>A)$, it is also in $C'$
by Assumption~\ref{uncontingent-minimal-same};
%
%
the previous point proves that $J < K$ holds in $C'$;
%
%
transitivity between $I \leq J$ and $J < K$ implies $I < K$, contradicting
the assumption $K \leq I \in C'$

\item[$I \leq K \not\in C'$:]

since $C'$ does not contain $K \leq I$ either by the point above, it is not a
connected preorder.

\end{description}~\qed

\

%

Uncontingent revision include natural revision as the subcase of the current
conditions. These current conditions are not however only what currently deemed
the case $C(0)$, but may include also situations that are progressively less
and less likely until some satisfy $\min(PA)$.

\begin{theorem}
\label{contingent-context}

The equality
{} $C \nat(P>A) = C \unc(Q>A)$ with $Q = P \leq \min(PA)$
holds for every connected preorder $C$ and formulae $P$ and $A$.

\end{theorem}

%
%

\proof The definition of $C \unc(Q>A)$ is as follows.

\long\def\ttytex#1#2{#1}
\ttytex{
\begin{eqnarray*}
\lefteqn{C \unc(Q>A) =} \\
&& [
     C(0) \backslash (Q \neg A)  \ldots C(\maxidx(QA)) \backslash (Q \neg A)   \\
&&   C(0) (Q \neg A)             \ldots C(\maxidx(QA)) (Q \neg A)		    \\
&&   C(\maxidx(QA)+1)               \ldots C(m)
   ]
\end{eqnarray*}
}{
C unc(Q>A) =
[
    C(0) - (Q-A)      .. C(max(QA)) - (Q-A)
    C(0) (Q-A)        .. C(max(QA)) (Q-A)
    C(max(QA)+1)      .. C(m)
]
}

Since $QA$ is the intersection of $P$, $(\lequal\min(PA))$ and $A$, it is also
the intersection of $PA$ and $(\lequal\min(PA))$, which equals $\min(PA)$.
Therefore, $\max(QA) = \min(PA)$.


The definition of $C \unc(Q>A)$ rewrites as follows.

\long\def\ttytex#1#2{#1}
\ttytex{
\begin{eqnarray*}
\lefteqn{C \unc(Q>A) =} \\
&& =                                                                        \\
&& [
     C(0) \backslash (Q \neg A)  \ldots C(\minidx(PA)) \backslash (Q \neg A)   \\
&&   C(0) (Q \neg A)             \ldots C(\minidx(PA)) (Q \neg A)		    \\
&&   C(\minidx(PA)+1)            \ldots C(m)
   ]                                                                        \\
\end{eqnarray*}
}{
=
[
    C(0) - (Q-A)      .. C(min(PA)) - (Q-A)
    C(0) (Q-A)        .. C(min(PA)) (Q-A)
    C(min(PA)+1)      .. C(m)
]
}

Since $Q \neg A$ is $P (\lequal\min(PA)) \neg A$, it is also $P \neg A
(\lequal\min(PA))$.

\long\def\ttytex#1#2{#1}
\ttytex{
\begin{eqnarray*}
\lefteqn{C \unc(Q>A) = } \\
&& [
     C(0) \backslash (P\neg A (\lequal\min(PA))) \ldots
         C(\minidx(PA)) \backslash (P\neg A (\lequal\min(PA))) \\
&&   C(0) (P\neg A (\lequal\min(PA)))  \ldots
	C(\minidx(PA)) (P\neg A) (\lequal\min(PA)) \\
&&   C(\minidx(PA)+1)                \ldots C(m)
   ]
\end{eqnarray*}
}{
C unc(Q>A) =
[
    C(0) - (P-A <=min(PA))  .. C(min(PA)) - (P-A <=min(PA))
    C(0) (P-A <=min(PA))    .. C(min(PA)) (P-A <=min(PA))
    C(min(PA)+1)            .. C(m)
]
}

The classes from $C(0)$ to $\minidx(PA)$ are all contained in
$(\lequal\min(PA))$ by definition. Their intersection with $P \neg A
(\lequal\min(PA))$ is the same as their intersection with $P \neg A$ only.

\long\def\ttytex#1#2{#1}
\ttytex{
\begin{eqnarray*}
\lefteqn{C \unc(Q>A) = } \\
&& [
     C(0) \backslash (P\neg A (\lequal\min(PA))) \ldots
         C(\minidx(PA)) \backslash (P\neg A (\lequal\min(PA))) \\
&&   C(0) (P\neg A)                         \ldots C(\minidx(PA)) (P\neg A) \\
&&   C(\minidx(PA)+1)                       \ldots C(m)
   ]
\end{eqnarray*}
}{
C unc(Q>A) =
[
    C(0) - (P-A <=min(PA))  .. C(min(PA)) - (P-A <=min(PA))
    C(0) (P-A)              .. C(min(PA)) (P-A)
    C(min(PA)+1)            .. C(m)
]
}

Removing the models of $P \neg A (\lequal\min(PA))$ is the same as removing the
models of $P \neg A$, since the models of $P \neg A$ that are not in
$(\lequal\min(PA))$ are not in $C(0),\ldots,C(\minidx(PA))$ either.

\long\def\ttytex#1#2{#1}
\ttytex{
\begin{eqnarray*}
\lefteqn{C \unc(Q>A) = } \\
&& [
     C(0) \backslash (P\neg A)  \ldots C(\minidx(PA)) \backslash (P\neg A)) \\
&&   C(0) (P\neg A)             \ldots C(\minidx(PA)) (P\neg A) \\
&&   C(\minidx(PA)+1)           \ldots C(m)
   ]
\end{eqnarray*}
}{
C unc(Q>A) =
[
    C(0) - (P-A)  .. C(min(PA)) - (P-A)
    C(0) (P-A)    .. C(min(PA)) (P-A)
    C(min(PA)+1)  .. C(m)
]
=
C nat(P>A)
}

This is the definition of $C \nat(P>A)$.~\qed

%

Uncontingent revision satisfies the Kern-Isberner postulates for iterated
belief revision~\cite{kern-99} except CR2.

\begin{theorem}
\label{uncontingent-postulates}

Uncontingent revision satisfies postulates CR0, CR1 and CR3-CR7~\cite{kern-99}
and falsified CR2.

\end{theorem}

\proof
\begin{description}

\item[CR0] $C \unc(P>A)$ is a connected preorder.

Holds by definition.

\item[CR1] $C \unc(P>A) \models P>A$.

Holds because uncontingent revision makes all models of $PAL$ less than all
models of $P \neg A L$ and leaves them less than or equal to all models of $P
\neg A \neg L$. Since $PAL$ has all models of $\min(PA)$, the claim follows:
all models of $\min(PA)$ are less than all models of $P \neg A$.

\item[CR2] $C \unc(P>A) = C$ if and only if $C \models P>A$.

This postulate is only true in a direction: if $C \unc(P>A) = C$ then $C
\models P>A$; the premise $C \unc(P>A) = C$ with $C \unc(P>A) \models P>A$,
proved as CR1, implies $C \models P>A$.

The converse is disproved by a counterexample: $C \unc(\true>x)$ where
{} $C = [\{\{x,y\}\}, \{\{\neg x,y\}, \{\neg x,\neg y\}\}, \{\{x,\neg y\}\}]$.

\begin{hfigure}
\setlength{\unitlength}{3750sp}%
\begin{picture}(1374,1149)(4489,-1948)
\thinlines
{\color[rgb]{0,0,0}\put(4501,-1186){\line( 1, 0){1350}}
}%
{\color[rgb]{0,0,0}\put(4501,-1936){\framebox(1350,1125){}}
}%
{\color[rgb]{0,0,0}\put(4501,-1561){\line( 1, 0){1350}}
}%
\put(4876,-1411){\makebox(0,0)[b]{\smash{\fontsize{9}{10.8}
\usefont{T1}{cmr}{m}{n}{\color[rgb]{0,0,0}$\neg xy$}%
}}}
\put(5476,-1411){\makebox(0,0)[b]{\smash{\fontsize{9}{10.8}
\usefont{T1}{cmr}{m}{n}{\color[rgb]{0,0,0}$\neg x\neg y$}%
}}}
\put(5176,-1786){\makebox(0,0)[b]{\smash{\fontsize{9}{10.8}
\usefont{T1}{cmr}{m}{n}{\color[rgb]{0,0,0}$x\neg y$}%
}}}
\put(5176,-1036){\makebox(0,0)[b]{\smash{\fontsize{9}{10.8}
\usefont{T1}{cmr}{m}{n}{\color[rgb]{0,0,0}$xy$}%
}}}
\end{picture}%
\nop{
+-----------+
|    xy     |
+-----------+
| -xy  -x-y |
+-----------+
|    x-y    |
+-----------+
}
\end{hfigure}

The entailment $C \models \true>x$ holds because the minimal model $\{x,y\}$ of
$x$ is strictly less than all models of $\neg x$. Yet, uncontingent revision
reverses the order between the models of $\neg x$ and $\{x,\neg y\}$ because it
establishes $\{x,\neg y\} < \{\neg x,y\}$ and $\{x,\neg y\} < \{\neg x,\neg
y\}$. It changes $C$ in spite of $C \models \true>A$.

\begin{hfigure}
\setlength{\unitlength}{3750sp}%
\begin{picture}(1374,1149)(4489,-1948)
\thinlines
{\color[rgb]{0,0,0}\put(4501,-1186){\line( 1, 0){1350}}
}%
{\color[rgb]{0,0,0}\put(4501,-1936){\framebox(1350,1125){}}
}%
{\color[rgb]{0,0,0}\put(4501,-1561){\line( 1, 0){1350}}
}%
\put(5176,-1036){\makebox(0,0)[b]{\smash{\fontsize{9}{10.8}
\usefont{T1}{cmr}{m}{n}{\color[rgb]{0,0,0}$xy$}%
}}}
\put(5176,-1411){\makebox(0,0)[b]{\smash{\fontsize{9}{10.8}
\usefont{T1}{cmr}{m}{n}{\color[rgb]{0,0,0}$x\neg y$}%
}}}
\put(5476,-1786){\makebox(0,0)[b]{\smash{\fontsize{9}{10.8}
\usefont{T1}{cmr}{m}{n}{\color[rgb]{0,0,0}$\neg x\neg y$}%
}}}
\put(4876,-1786){\makebox(0,0)[b]{\smash{\fontsize{9}{10.8}
\usefont{T1}{cmr}{m}{n}{\color[rgb]{0,0,0}$\neg xy$}%
}}}
\end{picture}%
\nop{
+-----------+
|    xy     |
+-----------+
|    x-y    |
+-----------+
| -xy  -x-y |
+-----------+
}
\end{hfigure}

\item[CR3] $C \unc(\true>A)$ is an AGM revision operator.

Consequence of Theorem~\ref{lexicographic-uncontingent}: $C \unc(\true>A) = C
\lex(A)$, and lexicographic revision is an AGM revision operator.

\item[CR4] If $P>A \equiv Q>B$ then $C \unc(P>A) = C \unc(Q>B)$.

Equivalence $P>A \equiv Q>B$ is $P \equiv Q$ and $PA \equiv QB$. It implies
$\neg P \equiv \neg Q$ and $P \neg A \equiv Q \neg B$. uncontingent revision is
defined from these formulae only, beside $C$.

\item[CR5] If $Q \subset PA$
then $C \models Q>B$ if and only if $C \unc(P>A) \models Q>B$.

Theorem~8 by Gabriele Kern--Isberner~\cite{kern-99} prove that this postulates
is the same as conditional preservation, which uncontingent revision obeys
because it does not change the order between models of $\neg P$, models of $PA$
and models of $P \neg A$.

\item[CR6]
If $QB \subseteq PA$, $Q \neg B \subseteq P \neg A$ and $C \models Q>B$
then $C \unc(P>A) \models Q>B$.

Theorem~11 by Gabriele Kern--Isberner~\cite{kern-99} gives an equivalent
formulation: $J <_C K$ implies $J <_{C \unc(P>A)} K$ for all $J \in PA$ and $K
\in P \neg A$. This is the case because uncontingent revision makes all models
of $PAL = PA$ less than all models of $P \neg AL$ and leaves them less than all
models of $P \neg A \neg L$. All models $J$ of $PA$ are therefore less than all
models $K$ of $P \neg A$. The comparison $J <_{C \unc(P>A)} K$ holds regardless
of $J <_C K$.

\item[CR7]
If $QB \subseteq P\neg A$, $Q \neg B \subseteq PA$ and $C \models Q>B$
then $C \unc(P>A) \models Q>B$.

Theorem~11 by Gabriele Kern--Isberner~\cite{kern-99} gives an equivalent
formulation: $K <_{C \unc(P>A)} J$ implies $K <_C J$ for all $J \in PA$ and $K
\in P \neg A$. This implication holds because its premise $K <_{C \unc(P>A)} J$
never does since uncontingent revision makes all models $J$ less than all
models $K$, as shown above for CR6.

\end{description}
\qed

Uncontingent revision satisfies most of the postulates. The only exception is
one direction of CR2: uncontingent revision may change the order even if the
revising conditional is already true. The counterexample is based on a
conditional where the premise $P$ is void, a case where uncontingent revision
reduces to lexicographic revision, which may indeed change the order even if
the formula is satisfied.

\section{Lexicographic revision}
\label{lesschange}

\draft

\begin{itemize}

\item lexicographic revision

\item uncontingent revision reduces to lexicographic revision

\item uncontingent and natural revision change the order less than
lexicographic revision

\item line-down revision change the order incomparably to lexicographic,
but also uncontingent and natural

\item lexicographic revision is strictly more naive than natural revision

\end{itemize}

\enddraft

%

Lexicographic revision by $A$ drops the models of $\neg A$ under all models of
$A$. This is the same as duplicating all classes $C(0),\ldots,C(m)$
intersecting the first copy with $A$ and the second with $\neg A$, modulo empty
classes.

\begin{definition}
\label{lexicographic}

The {\em lexicographic revision} of the order $[C(0),\ldots,C(m)]$ by a
propositional formula $F$ is as follows.

\long\def\ttytex#1#2{#1}
\ttytex{
\begin{eqnarray*}
\lefteqn{C \lex(F) = } \\
&& [
     C(0) F,		\ldots, C(m) F \\
&&   C(0) \neg F,	\ldots, C(m) \neg F
   ]
\end{eqnarray*}
}{
C lex(F) =
[
    C(0) F  .. C(m) F
    C(0) -F .. C(m) -F
]
}

\end{definition}

It extends to conditionals by $C \lex(P>A) = C \lex(P \rightarrow A)$.
Its equivalence classes are
{} $
{} C \lex(P>A) =
{} [
{}     C(0) (P \rightarrow A)       \ldots  C(m) (P \rightarrow A)     \\
{}     C(0) \neg P (\rightarrow A)  \ldots  C(m) \neg (P \rightarrow A)
{} ]
{} $.

Lexicographic revision sets $J<K$ and $I<K$ for all models $I$, $J$ and $K$ of
respectively $\neg PL$, $PAL$ and $P \neg AL$ with $L=\true$ since the classes
of $\neg (P \rightarrow A) = P \neg A$ are after those of $P \rightarrow A = 
\neg P \vee PA$.

\

%

Uncontingent revision extends lexicographic revision to conditionals.

\begin{theorem}
\label{lexicographic-uncontingent}

The equality
{} $C \unc(\true>A) = C \lex(A)$
holds for every connected preorder $C$ and formula $A$.

\end{theorem}

\proof The claim is proved by the following chain of equations.

\long\def\ttytex#1#2{#1}
\ttytex{
\begin{eqnarray*}
\lefteqn{[C(0) \ldots C(m)] unc(true>A) = }				\\
&& [
        C(0)                     \ldots C(\minidx(true)-1)		\\
&&      C(\minidx(true)) \backslash (true \neg A) \ldots
        			 C(\maxidx(true A)) \backslash (true \neg A)\\
&&      C(\minidx(true))(true \neg A)
                                 \ldots C(\maxidx(true A))(true \neg A)	\\
&&      C(max(true A)+1)         \ldots C(m)
] =									\\
&& [
        C(0)                     \ldots C(0-1)				\\
&&      C(0) \backslash (\neg A) \ldots C(\maxidx(A)) \backslash (\neg A) \\
&&      C(0) (\neg A)            \ldots C(\maxidx(A)) (\neg A)		\\
&&      C(\maxidx(A)+1)          \ldots C(m)
] =									\\
&& [
        C(0) A                   \ldots C(\maxidx(A)) A			\\
&&      C(\maxidx(A)+1) A        \ldots C(m) A				\\
&&      C(0) \neg A              \ldots C(\maxidx(A)) \neg A		\\
&&      C(\maxidx(A)+1) \neg A   \ldots C(m) \neg A
] =									\\
&& [
        C(0) A      \ldots C(m) \neg A					\\
&&      C(0) \neg A \ldots C(m) \neg A
] =									\\
&& [C(0) \ldots C(m)] lex(A)
\end{eqnarray*}
}{
[C(0) .. C(m)] unc(true>A) = 
[
        C(0)                  .. C(min(true)-1)
        C(min(true))-(true-A) .. C(max(true A))-(true-A)
        C(min(true))(true-A)  .. C(max(true A))(true-A)
        C(max(true A)+1)      .. C(m)
] =
[
        C(0)        .. C(0-1)
        C(0)-(-A)   .. C(max(A))-(-A)
        C(0)(-A)    .. C(max(A))(-A)
        C(max(A)+1) .. C(m)
] =
[
        C(0)A         .. C(max(A))A
        C(max(A)+1)A  .. C(m)A                           no A > max(A)
        C(0)-A        .. C(max(A))-A
        C(max(A)+1)-A .. C(m)-A                          all -A > max(A)
= [
        C(0)A  .. C(m)A
        C(0)-A .. C(m)-A
] =
] =
[C(0) .. C(m)] lex(A)
}

\qed

\

%

The following theorem shows that uncontingent revision reduces to lexicographic
revision. The reduction highlights their difference.

\begin{theorem}
\label{uncontingent-lexicographic}

The equality
{} $C \unc(P>A) = C \lex((\lequal\max(PA)) (P \rightarrow A))$
holds for every connected preorder $C$ and formulae $P$ and $A$.

\end{theorem}

\proof The definition of lexicographic revision is applied to
{} $F = (\lequal\max(PA)) (P \rightarrow A)$.

\long\def\ttytex#1#2{#1}
\ttytex{
\begin{eqnarray*}
\lefteqn{C \lex(F) = } \\
&& [
     C(0) F  \ldots  C(m) F \\
&&   C(0) \neg F \ldots  C(m) \neg F
   ] = \\
&& [
     C(0) ((\lequal\max(PA))(P \rightarrow A))
        \ldots
     C(m) ((\lequal\max(PA))(P \rightarrow A)) \\
&&   C(0) \neg ((\lequal\max(PA))(P \rightarrow A))
        \ldots
     C(m) \neg ((\lequal\max(PA))(P \rightarrow A))
   ]
\end{eqnarray*}
}{
C lex(F) =
[
    C(0) F  .. C(m) F
    C(0) -F .. C(m) -F
]
=
[
    C(0) (<=max(PA)(P->A))   .. C(m) (<=max(PA)(P->A))
    C(0) \neg (<=max(PA)(P->A)) .. C(m) \neg (<=max(PA)(P->A))
]
}

The negation of a conjunction is the same as the disjunction of negations. The
negation of $(\lequal\max(PA))(P \rightarrow A)$ equals $\neg (\lequal\max(PA))
\vee \neg (P \rightarrow A)$. The negation of $\leq$ is $>$ and the negation of
$P \rightarrow A$ is $P \neg A$. Their disjunction is therefore%
{} $(\greater\min(PA)) \vee (P \neg A)$.

\long\def\ttytex#1#2{#1}
\ttytex{
\begin{eqnarray*}
\lefteqn{C \lex((\lequal\max(PA))(P \rightarrow A)) = } \\
&& [
     C(0) (\lequal\max(PA))(P \rightarrow A)
        \ldots
     C(m) (\lequal\max(PA))(P \rightarrow A) \\
&&   C(0) ((\greater\max(PA)) \vee (P \neg A))
        \ldots
     C(m) ((\greater\max(PA)) \vee (P \neg A))
   ]
\end{eqnarray*}
}{
C lex(P<=max(PA)->A)
=
[
    C(0) <=max(PA)(P->A)   .. C(m) <=max(PA)(P->A)
    C(0) (>max(PA)v(P-A))  .. C(m) (>max(PA)v(P-A))
]
}

The sequence $C(0) \ldots C(m)$ is
{} $C(0) \ldots C(\maxidx(PA))$
followed by
{} $C(\maxidx(PA)+1) \ldots C(m)$.

\long\def\ttytex#1#2{#1}
\ttytex{
\begin{eqnarray*}
\lefteqn{C \lex((\lequal\max(PA))(P \rightarrow A)) = } \\
&& [
     C(0) (\lequal\max(PA))(P \rightarrow A)
        \ldots
     C(\maxidx(PA)) (\lequal\max(PA)()P \rightarrow A) \\
&&   C(\maxidx(PA)+1) (\lequal\max(PA))(P \rightarrow A)
        \ldots
     C(m) (\lequal\max(PA))(P \rightarrow A) \\
&&   C(0) ((\greater\max(PA)) \vee (P \neg A))
        \ldots
     C(\maxidx(PA)) ((\greater\max(PA)( \vee (P \neg A)) \\
&&   C(\maxidx(PA)+1) ((\greater\max(PA)( \vee (P \neg A))
        \ldots
     C(m) ((\greater\max(PA)) \vee (P \neg A))
   ]
\end{eqnarray*}
}{
C lex(P<=max(PA)->A)
=
[
    C(0) <=max(PA)(P->A)           .. C(max(PA)) <=max(PA)(P->A)
    C(max(PA)+1) <=max(PA)(P->A)   .. C(m) <=max(PA)(P->A)
    C(0) (>max(PA)v(P-A))          .. C(max(PA)) (>max(PA)v(P-A))
    C(max(PA)+1) (>max(PA)v(P-A))  .. C(m) (>max(PA)v(P-A))
]
}

All classes from $C(0)$ to $C(\maxidx(PA))$ are contained in $\leq\max(PA)$.
The remaining classes $C(\maxidx(PA)+1) \ldots C(m)$ do not intersect $\leq
\max(PA)$.

\long\def\ttytex#1#2{#1}
\ttytex{
\begin{eqnarray*}
\lefteqn{C \lex((\lequal\max(PA))(P \rightarrow A)) = } \\
&& [
     C(0) (P \rightarrow A)
        \ldots
     C(\maxidx(PA)) (P \rightarrow A) \\
&&   \emptyset
	\ldots 
     \emptyset \\
&&   C(0) ((\greater\max(PA)) \vee (P \neg A))
        \ldots
     C(\maxidx(PA)) ((\greater\max(PA)) \vee (P \neg A)) \\
&&   C(\maxidx(PA)+1) ((\greater\max(PA)) \vee (P \neg A))
        \ldots
     C(m) ((\greater\max(PA)) \vee (P \neg A))
   ]
\end{eqnarray*}
}{
C lex(P<=max(PA)->A)
=
[
    C(0) (P->A)                    .. C(max(PA)) (P->A)
    0                              .. 0
    C(0) (>max(PA)v(P-A))          .. C(max(PA)) (>max(PA)v(P-A))
    C(max(PA)+1) (>max(PA)v(P-A))  .. C(m) (>max(PA)v(P-A))
]
}

The classes
{} $C(i) ((\greater\max(PA)) \vee (P \neg A))$
coincide with
{} $(C(i) (\greater\max(PA))) \cup (C(i) P \neg A)$.

The first part $C(i) (\greater\max(PA))$ is empty when $i$ is between $0$ and
$\maxidx(PA)$, leaving only the second part
{} $C(i) P \neg A$.

The first part $C(i) (\greater\max(PA))$ is $C(i)$ when $i$ is between
$\maxidx(PA)+1$ and $m$, turning
{} $(C(i) (\greater\max(PA))) \cup (C(i) P \neg A)$
into
{} $(C(i) \cup (C(i) P \neg A)$,
which is the same as
{} $C(i)$.

\long\def\ttytex#1#2{#1}
\ttytex{
\begin{eqnarray*}
\lefteqn{C \lex((\lequal\max(PA))(P \rightarrow A)) = } \\
&& [
     C(0) (P \rightarrow A)   \ldots  C(\maxidx(PA)) (P \rightarrow A)	\\
&&   \emptyset                \ldots  \emptyset				\\
&&   C(0) (P \neg A)          \ldots  C(\maxidx(PA)) (P \neg A)		\\
&&   C(\maxidx(PA)+1)         \ldots  C(m)
   ]
\end{eqnarray*}
}{
=
[
    C(0) (P->A)          .. C(max(PA)) (P->A)
    0                    .. 0
    C(0) (P-A)           .. C(max(PA)) (P-A)
    C(max(PA)+1)         .. C(m)
]
=
C unc(P>A)
}

The empty classes can be removed.

\long\def\ttytex#1#2{#1}
\ttytex{
\begin{eqnarray*}
\lefteqn{C \lex((\lequal\max(PA))(P \rightarrow A)) = } \\
&& [
     C(0) (P \rightarrow A)   \ldots  C(\maxidx(PA)) (P \rightarrow A)	\\
&&   C(0) (P \neg A)          \ldots  C(\maxidx(PA)) (P \neg A)		\\
&&   C(\maxidx(PA)+1)         \ldots  C(m)
   ]
\end{eqnarray*}
}{
=
[
    C(0) (P->A)          .. C(max(PA)) (P->A)
    C(0) (P-A)           .. C(max(PA)) (P-A)
    C(max(PA)+1)         .. C(m)
]
=
C unc(P>A)
}

Intersecting with $P \rightarrow A$ is the same as removing its negation $P
\neg A$.

\long\def\ttytex#1#2{#1}
\ttytex{
\begin{eqnarray*}
\lefteqn{C \lex((\lequal\max(PA))(P \rightarrow A)) = } \\
&& [
     C(0) \backslash (P \neg A)  \ldots  C(\maxidx(PA)) \backslash (P \neg A) \\
&&   C(0) (P \neg A)             \ldots  C(\maxidx(PA)) (P \neg A)	\\
&&   C(\maxidx(PA)+1)            \ldots  C(m)
   ]
\end{eqnarray*}
}{
=
[
    C(0) - (P-A)         .. C(max(PA)) - (P-A)
    C(0) (P-A)           .. C(max(PA)) (P-A)
    C(max(PA)+1)         .. C(m)
]
=
C unc(P>A)
}
This is the definition of $C \unc(P>A)$.~\qed

\

%

Lexicographic revision does not minimally change the order: uncontingent
revision changes it less.

\begin{theorem}
\label{uncontingent-lesschange-lexicographic}

Uncontingent revision is closer or equally distant to the revised order than
lexicographic revision.

\end{theorem}

\proof Uncontingent revision only moves models between $\minidx(P)$ and
$\maxidx(PA)$, the models of
{} $L = (\equal\min(P)) \cup \cdots \cup (\equal\max(PA))$.
It drops all models $K \in P \neg AL$ under all models $J \in PAL$ and $I \in
\neg PL$. It realizes $J<K$ and $I<K$. These models $K$, $J$ and $I$ are also
respectively models of $P \neg A$, $PA$ and $\neg P$. Lexicographic revision
drops all models of $P \neg A$, including $K$, under all models of $PA$ of
$\neg P$, including respectively $J$ and $I$. It realizes $J<K$ and $I<K$, like
uncontingent revision does.~\qed

Most revisions coincide on very simple cases, like the flat ordering and a
simple conditional $\true>x$; none changes the order strictly less than
another. Yet, revisions differ in most other cases. Some of them show that
natural revision adheres to the principle of minimal change strictly more than
lexicographic revision.

\begin{theorem}
\label{uncontingent-strictly-lesschange-lexicographic}

Uncontingent revision $C_x \unc(x>y)$ is strictly closer to $C_x$ than
lexicographic revision $C_x \lex(x \rightarrow y)$.

\end{theorem}

\proof The following are the equivalence classes of $C_x$, shown in
Figure~\ref{figure-variable-4}.

\begin{eqnarray*}
C_x(0)	& = &	\{ \{      x,y \}, \{ x,     \neg y\} \}	\\
C_x(1)	& = &	\{ \{ \neg x,y \}, \{ \neg x,\neg y\} \}
\end{eqnarray*}

\begin{hfigure}
\setlength{\unitlength}{3750sp}%
\begin{picture}(1374,774)(4489,-1573)
\thinlines
{\color[rgb]{0,0,0}\put(4501,-1186){\line( 1, 0){1350}}
}%
{\color[rgb]{0,0,0}\put(4501,-1561){\framebox(1350,750){}}
}%
\put(4876,-1036){\makebox(0,0)[b]{\smash{\fontsize{9}{10.8}
\usefont{T1}{cmr}{m}{n}{\color[rgb]{0,0,0}$xy$}%
}}}
\put(5476,-1036){\makebox(0,0)[b]{\smash{\fontsize{9}{10.8}
\usefont{T1}{cmr}{m}{n}{\color[rgb]{0,0,0}$x\neg y$}%
}}}
\put(4876,-1411){\makebox(0,0)[b]{\smash{\fontsize{9}{10.8}
\usefont{T1}{cmr}{m}{n}{\color[rgb]{0,0,0}$\neg xy$}%
}}}
\put(5476,-1411){\makebox(0,0)[b]{\smash{\fontsize{9}{10.8}
\usefont{T1}{cmr}{m}{n}{\color[rgb]{0,0,0}$\neg x\neg y$}%
}}}
\end{picture}%
\nop{
+------------+
|  xy   x-y  |
+------------+
| -xy  -x-y  |
+------------+
}
\label{figure-variable-4}
\hcaption{The order $C_\epsilon$.}
\end{hfigure}

The minimal models of $P = x$ are $\{x,y\}$ and $\{x, \neg y\}$, in class $0$;
therefore, $\minidx(P)=0$. The maximal model of $PA = x \wedge y$ is $\{x,y\}$, in
class $0$; therefore, $\maxidx(PA)=0$.

Uncontingent revision only changes the comparisons between the models of
{} $L = (\equal\min(P)) \cup \cdots \cup (\equal\max(PA))
{}    = C_x(0) \cup \cdots \cup C_x(0)
{}    = C_x(0)
{}    = \{\{x,y\}, \{x,\neg y\}\} = 
{}    = (x \wedge y) \vee (x \wedge \neg y)$.

It only lowers the models of $P \neg AL$ under the models of $\neg PL$ and
$PA$:

\begin{eqnarray*}
\neg PL
	&=& \neg x \wedge (x \wedge y) \vee (x \wedge \neg y)		\\
	&=& \bot							\\
PA
	&=& x \wedge y \wedge (x \wedge y) \vee (x \wedge \neg y)	\\
	&=& x \wedge y							\\
P \neg AL
	&=& x \wedge \neg y \wedge (x \wedge y) \vee (x \wedge \neg y)	\\
	&=& x \wedge \neg y 						\\
\end{eqnarray*}

It only lowers $\{x,\neg y\}$ under $\{x,y\}$, as shown in
Figure~\ref{figure-variable-natural}.

\begin{hfigure}
\long\def\ttytex#1#2{#1}
\ttytex{
\begin{tabular}{ccc}
\setlength{\unitlength}{3750sp}%
\begin{picture}(1374,774)(4489,-1573)
\thinlines
{\color[rgb]{0,0,0}\put(4501,-1186){\line( 1, 0){1350}}
}%
{\color[rgb]{0,0,0}\put(4501,-1561){\framebox(1350,750){}}
}%
\put(4876,-1036){\makebox(0,0)[b]{\smash{\fontsize{9}{10.8}
\usefont{T1}{cmr}{m}{n}{\color[rgb]{0,0,0}$xy$}%
}}}
\put(5476,-1036){\makebox(0,0)[b]{\smash{\fontsize{9}{10.8}
\usefont{T1}{cmr}{m}{n}{\color[rgb]{0,0,0}$x\neg y$}%
}}}
\put(4876,-1411){\makebox(0,0)[b]{\smash{\fontsize{9}{10.8}
\usefont{T1}{cmr}{m}{n}{\color[rgb]{0,0,0}$\neg xy$}%
}}}
\put(5476,-1411){\makebox(0,0)[b]{\smash{\fontsize{9}{10.8}
\usefont{T1}{cmr}{m}{n}{\color[rgb]{0,0,0}$\neg x\neg y$}%
}}}
\end{picture}%
&
\setlength{\unitlength}{3750sp}%
\begin{picture}(1104,744)(5659,-3943)
\thinlines
{\color[rgb]{0,0,0}\multiput(6391,-3391)(-9.47368,4.73684){20}{\makebox(2.1167,14.8167){\tiny.}}
\put(6211,-3301){\line( 0,-1){ 45}}
\put(6211,-3346){\line(-1, 0){180}}
\put(6031,-3346){\line( 0,-1){ 90}}
\put(6031,-3436){\line( 1, 0){180}}
\put(6211,-3436){\line( 0,-1){ 45}}
\multiput(6211,-3481)(9.47368,4.73684){20}{\makebox(2.1167,14.8167){\tiny.}}
}%
\end{picture}%
&
\setlength{\unitlength}{3750sp}%
\begin{picture}(1374,1824)(4489,-2023)
\thinlines
{\color[rgb]{0,0,0}\put(4501,-1411){\line( 1, 0){1350}}
}%
{\color[rgb]{0,0,0}\put(4501,-2011){\framebox(1350,1800){}}
}%
{\color[rgb]{0,0,0}\put(4501,-811){\line( 1, 0){1350}}
}%
\put(4876,-586){\makebox(0,0)[b]{\smash{\fontsize{9}{10.8}
\usefont{T1}{cmr}{m}{n}{\color[rgb]{0,0,0}$xy$}%
}}}
\put(4876,-1711){\makebox(0,0)[b]{\smash{\fontsize{9}{10.8}
\usefont{T1}{cmr}{m}{n}{\color[rgb]{0,0,0}$\neg xy$}%
}}}
\put(5401,-1711){\makebox(0,0)[b]{\smash{\fontsize{9}{10.8}
\usefont{T1}{cmr}{m}{n}{\color[rgb]{0,0,0}$\neg x\neg yz$}%
}}}
\put(5401,-1186){\makebox(0,0)[b]{\smash{\fontsize{9}{10.8}
\usefont{T1}{cmr}{m}{n}{\color[rgb]{0,0,0}$x\neg y$}%
}}}
\end{picture}%
\end{tabular}
}{
+------------+                   +------------+
|  xy   x-y  |     unc(x>y)      |  xy        |
+------------+       ===>        +------------+
| -xy  -x-y  |                   |       x-y  |
+------------+                   +------------+
                                 | -xy  -x-y  |
                                 +------------+
}
\label{figure-variable-natural}
\hcaption{The order $C_x \unc(x>y)$.}
\end{hfigure}

Lexicographic revision lowers all models $P\neg A = x \neg y$ under all models
$\neg P = \neg x$ and $PA = x \wedge y$, as shown in
Figure~\ref{figure-variable-lexicographic}. It lowers $\{x,\neg y\}$ under all
other models. It replicates the only change by uncontingent revision by
lowering $x \neg y$ under $xy$. It additionally lowers $x \neg y$ under $\neg x
y$. It is further from $C_x$ than uncontingent revision.

\begin{hfigure}
\long\def\ttytex#1#2{#1}
\ttytex{
\begin{tabular}{ccc}
\setlength{\unitlength}{3750sp}%
\begin{picture}(1374,774)(4489,-1573)
\thinlines
{\color[rgb]{0,0,0}\put(4501,-1186){\line( 1, 0){1350}}
}%
{\color[rgb]{0,0,0}\put(4501,-1561){\framebox(1350,750){}}
}%
\put(4876,-1036){\makebox(0,0)[b]{\smash{\fontsize{9}{10.8}
\usefont{T1}{cmr}{m}{n}{\color[rgb]{0,0,0}$xy$}%
}}}
\put(5476,-1036){\makebox(0,0)[b]{\smash{\fontsize{9}{10.8}
\usefont{T1}{cmr}{m}{n}{\color[rgb]{0,0,0}$x\neg y$}%
}}}
\put(4876,-1411){\makebox(0,0)[b]{\smash{\fontsize{9}{10.8}
\usefont{T1}{cmr}{m}{n}{\color[rgb]{0,0,0}$\neg xy$}%
}}}
\put(5476,-1411){\makebox(0,0)[b]{\smash{\fontsize{9}{10.8}
\usefont{T1}{cmr}{m}{n}{\color[rgb]{0,0,0}$\neg x\neg y$}%
}}}
\end{picture}%
&
\setlength{\unitlength}{3750sp}%
\begin{picture}(1104,744)(5659,-3943)
\thinlines
{\color[rgb]{0,0,0}\multiput(6391,-3391)(-9.47368,4.73684){20}{\makebox(2.1167,14.8167){\tiny.}}
\put(6211,-3301){\line( 0,-1){ 45}}
\put(6211,-3346){\line(-1, 0){180}}
\put(6031,-3346){\line( 0,-1){ 90}}
\put(6031,-3436){\line( 1, 0){180}}
\put(6211,-3436){\line( 0,-1){ 45}}
\multiput(6211,-3481)(9.47368,4.73684){20}{\makebox(2.1167,14.8167){\tiny.}}
}%
\end{picture}%
&
\setlength{\unitlength}{3750sp}%
\begin{picture}(1374,1824)(4489,-2023)
\thinlines
{\color[rgb]{0,0,0}\put(4501,-1411){\line( 1, 0){1350}}
}%
{\color[rgb]{0,0,0}\put(4501,-2011){\framebox(1350,1800){}}
}%
{\color[rgb]{0,0,0}\put(4501,-811){\line( 1, 0){1350}}
}%
\put(4876,-586){\makebox(0,0)[b]{\smash{\fontsize{9}{10.8}
\usefont{T1}{cmr}{m}{n}{\color[rgb]{0,0,0}$xy$}%
}}}
\put(5401,-1186){\makebox(0,0)[b]{\smash{\fontsize{9}{10.8}
\usefont{T1}{cmr}{m}{n}{\color[rgb]{0,0,0}$\neg x\neg yz$}%
}}}
\put(4876,-1186){\makebox(0,0)[b]{\smash{\fontsize{9}{10.8}
\usefont{T1}{cmr}{m}{n}{\color[rgb]{0,0,0}$\neg xy$}%
}}}
\put(5401,-1786){\makebox(0,0)[b]{\smash{\fontsize{9}{10.8}
\usefont{T1}{cmr}{m}{n}{\color[rgb]{0,0,0}$x\neg y$}%
}}}
\end{picture}%
\end{tabular}
}{
+------------+                   +------------+
|  xy   x-y  |    lex(x->y)      |  xy        |
+------------+       ===>        +------------+
| -xy  -x-y  |                   | -xy  -x-y  |
+------------+                   +------------+
                                 |       x-y  |
                                 +------------+
}
\label{figure-variable-lexicographic}
\hcaption{The order $C_x \lex(x \rightarrow y)$.}
\end{hfigure}
~\qed

These two theorems show prove that lexicographic revision does not change the
order minimally: uncontingent revision changes it less, in some cases strictly
less. In turns, uncontingent revision changes the order less than natural
revision by Theorem~\ref{natural-lesschange-uncontingent}, in some cases
strictly less by Theorem~\ref{natural-strictly-lesschange-uncontingent}.
Transitivity implies the same between lexicographic and natural revision.

\begin{theorem}
\label{natural-strictly-lesschange-lexicographic}

Natural revision is closer or equally distant to the revised order than
lexicographic revision, and strictly closer in some cases.

\end{theorem}

\proof Vicinity is proved by combining
Theorem~\ref{natural-lesschange-uncontingent} and
Theorem~\ref{uncontingent-lesschange-lexicographic}: natural revision is closer
to the revised order than uncontingent revision, which is closer to that than
lexicographic revision.

Strict vicinity is proved by combining
Theorem~\ref{natural-strictly-lesschange-uncontingent} and
Theorem~\ref{uncontingent-lesschange-lexicographic}: natural revision is
strictly closer to some ordering than uncontingent revision, which is closer
or equally distant to lexicographic revision for all orderings.~\qed

\

%

The changes by line-down revision are incomparable with the ones by
lexicographic, uncontingent and natural revision.

\begin{theorem}
\label{down-incomparablechange-others}

Line-down revision
{} $C_\epsilon \dow(x>y)$
is incomparably distant to $C_\epsilon$ than
natural revision
$C_\epsilon \nat(x>y)$,
uncontingent revision
$C_\epsilon \unc(x>y)$ and
lexicographic revision
$C_\epsilon \lex(x>y)$.

\end{theorem}

\proof The flat ordering $C_\epsilon$ has all models in class $0$:

\[
C_\epsilon(0) = \{
	\{     x,     y\},
	\{     x,\neg y\},
	\{\neg x,     y\},
	\{\neg x,\neg y\}
\}
\]

The models of the relevant formulae are as follows.

\begin{eqnarray*}
PA = x \wedge y			&=&	\{\{x,y\}\}			\\
\min(PA) = \min(x \wedge y)	&=&	\{\{x,y\}\}			\\
P \neg A = x \wedge \neg y	&=&	\{\{x,\neg y\}			\\
\neg P = \neg x			&=&	\{\{\neg x,y\}, \{\neg x, \neg y\}\}
\end{eqnarray*}

The line-down revision of $C_x$ is $C_{\{x \wedge y\}}$; its natural,
uncontingent and lexicographic revisions are $C_{\{x \vee \neg y\}}$. They are
depicted in Figure~\ref{figure-incomparable}.

\begin{hfigure}
\begin{tabular}{ccc}
\setlength{\unitlength}{3750sp}%
\begin{picture}(1899,399)(4489,-1348)
\thinlines
{\color[rgb]{0,0,0}\put(4501,-1336){\framebox(1875,375){}}
}%
\put(4726,-1186){\makebox(0,0)[b]{\smash{\fontsize{9}{10.8}
\usefont{T1}{cmr}{m}{n}{\color[rgb]{0,0,0}$xy$}%
}}}
\put(5176,-1186){\makebox(0,0)[b]{\smash{\fontsize{9}{10.8}
\usefont{T1}{cmr}{m}{n}{\color[rgb]{0,0,0}$x\neg y$}%
}}}
\put(5626,-1186){\makebox(0,0)[b]{\smash{\fontsize{9}{10.8}
\usefont{T1}{cmr}{m}{n}{\color[rgb]{0,0,0}$\neg xy$}%
}}}
\put(6076,-1186){\makebox(0,0)[b]{\smash{\fontsize{9}{10.8}
\usefont{T1}{cmr}{m}{n}{\color[rgb]{0,0,0}$\neg x\neg y$}%
}}}
\end{picture}%
\nop{
 +--------------------+
 | xy  x-y  -xy  -x-y |
 +--------------------+
}
&
\setlength{\unitlength}{3750sp}%
\begin{picture}(1899,774)(4489,-1723)
\thinlines
{\color[rgb]{0,0,0}\put(4501,-1711){\framebox(1875,750){}}
}%
{\color[rgb]{0,0,0}\put(4501,-1336){\line( 1, 0){1875}}
}%
\put(4726,-1186){\makebox(0,0)[b]{\smash{\fontsize{9}{10.8}
\usefont{T1}{cmr}{m}{n}{\color[rgb]{0,0,0}$xy$}%
}}}
\put(5176,-1561){\makebox(0,0)[b]{\smash{\fontsize{9}{10.8}
\usefont{T1}{cmr}{m}{n}{\color[rgb]{0,0,0}$x\neg y$}%
}}}
\put(5626,-1561){\makebox(0,0)[b]{\smash{\fontsize{9}{10.8}
\usefont{T1}{cmr}{m}{n}{\color[rgb]{0,0,0}$\neg xy$}%
}}}
\put(6076,-1561){\makebox(0,0)[b]{\smash{\fontsize{9}{10.8}
\usefont{T1}{cmr}{m}{n}{\color[rgb]{0,0,0}$\neg x\neg yz$}%
}}}
\end{picture}%
\nop{
 +--------------------+
 | xy                 |
 +--------------------+
 |     x-y  -xy  -x-y |
 +--------------------+
}
&
\setlength{\unitlength}{3750sp}%
\begin{picture}(1899,774)(4489,-1723)
\thinlines
{\color[rgb]{0,0,0}\put(4501,-1711){\framebox(1875,750){}}
}%
{\color[rgb]{0,0,0}\put(4501,-1336){\line( 1, 0){1875}}
}%
\put(4726,-1186){\makebox(0,0)[b]{\smash{\fontsize{9}{10.8}
\usefont{T1}{cmr}{m}{n}{\color[rgb]{0,0,0}$xy$}%
}}}
\put(5626,-1186){\makebox(0,0)[b]{\smash{\fontsize{9}{10.8}
\usefont{T1}{cmr}{m}{n}{\color[rgb]{0,0,0}$\neg xy$}%
}}}
\put(6076,-1186){\makebox(0,0)[b]{\smash{\fontsize{9}{10.8}
\usefont{T1}{cmr}{m}{n}{\color[rgb]{0,0,0}$\neg x\neg yz$}%
}}}
\put(5176,-1561){\makebox(0,0)[b]{\smash{\fontsize{9}{10.8}
\usefont{T1}{cmr}{m}{n}{\color[rgb]{0,0,0}$x\neg y$}%
}}}
\end{picture}%
\nop{
 +--------------------+
 | xy       -xy  -x-y |
 +--------------------+
 |     x-y            |
 +--------------------+
} 
\\
$C_\epsilon$
&
$C_\epsilon \dow(x>y) = C_{x \wedge y}$
&
$C_\epsilon \nat(x>y) = C_\epsilon \unc(x>y) =$
\\
&&
$C_\epsilon \lex(x>y) = C_{\{x \vee \neg y\}}$
\end{tabular}
\label{figure-incomparable}
\hcaption{Incomparable orders.}
\end{hfigure}

Line-down revision
{} realizes $\min(PA) < P \neg A$:
{} it realizes $\{x,y\} < \{x, \neg y\}$.
{} It maintains the order between $P \neg A$ and $\neg P$:
{} it maintains $\{x,\neg y\} \leq \{\neg x,\neg y\}$.
By transitivity,
{} it realizes $\{x, y\} < \{\neg x,\neg y\}$.
{}	It removes $\{\neg x,\neg y\} \leq \{x,y\}$.

The other revisions
{} maintain the order between the models of $\neg P$ and $PA$:
{}	they do not remove $\{\neg x,\neg y\} \leq \{x,y\}$.

The other revisions
{} realize $\min(PA) < P \neg A$;
{} they realize $\{x,y\} < \{x,\neg y\}$.
{} They maintain the order between $\neg P$ and $PA$:
{} they maintain $\{\neg x,\neg y\} \leq \{x,y\}$;
By transitivity,
{} they realize $\{\neg x,\neg y\} < \{x,\neg y\}$.
{}	They remove $\{x,\neg y\} \leq \{\neg x,\neg y\}$.

Line down revision
{} maintains the order between $P \neg A$ and $\neg P$:
{}	it does not remove $\{x,\neg y\} \leq \{\neg x,\neg y\}$.

Line-down revision removes a comparison that the others do not and vice versa.
This is difference incomparability.~\qed

\

%

The following theorem shows how revisions compare on naivety.

\begin{theorem}
\label{naive-chain}

Lexicographic revision is strictly more naive than uncontingent revision, which
is strictly more naive than natural revision, which is strictly more naive than
line-down revision.

\end{theorem}

\proof Natural revision only moves models between $\minidx(P)$ and
$\minidx(PA)$. Uncontingent revision only moves models between $\minidx(P)$ and
$\maxidx(PA)$. These are the models of the following formulae.

\begin{eqnarray*}
L	&=&	(\equal\min(P)) \cup \cdots \cup (\equal\min(PA))	\\
L'	&=&	(\equal\min(P)) \cup \cdots \cup (\equal\max(PA))
\end{eqnarray*}

The non-strict part of the claim is proved first.

Naivety is defined from the strength of the models of $\neg P$, which is
defined from the models that are greater than or equal to the models of $\neg
P$.

Natural revision changes such comparisons only by realizing $I < K$ for
every $I \in \neg PL$ and $K \in P\neg AL$. Uncontingent revision does the same
with $L'$ in place of $L$. Lexicographic revision does the same with $\true$ in
place of $L'$. Since $L \subseteq L' \subseteq \true$, the increases of
strength are contained each in the following.

Natural revision realizes $I < K$ for every $I \in \neg PL$ and $K \in P\neg
AL$, which increases the strength of $I$. Line-down revision changes the
comparisons only by realizing $J < I$ for every $J \in PAL$ and $I \in \neg
PL$; it removes $I \leq J$ and $I \leq K$, which decreases the strength of $I$
instead of increasing it.

The strict part of the claim is proved as follows.

The first three revisions increase the strength of the models $I$ of
respectively
{} $\neg PL$, $\neg PL'$ or $\neg P\true$.
They increase them by the models $K$ of respectively
{} $P \neg A L$, $P \neg A L'$ or $P \neg A \true$.
These increases are properly contained each in the next if
{} $\neg PL \subset \neg PL' \subset P\true$,
and
{} $P\neg AL = P\neg AL' = P\neg A\true$,
and some models of the latter three formulae compare the same to some of $\neg
P$. Such conditions are realized by the order in
Figure~\ref{figure-strict-naive}, defined from a conditional $P>A$ such that
$\neg P$, $PA$ and $P \neg A$ contain at least respectively three, two and one
model.

\begin{hfigure}
%
%
\setlength{\unitlength}{3750sp}%
\begin{picture}(3477,1374)(4489,-2173)
\thinlines
{\color[rgb]{0,0,0}\put(4501,-1261){\line( 1, 0){1800}}
}%
{\color[rgb]{0,0,0}\put(4501,-1711){\line( 1, 0){1800}}
}%
{\color[rgb]{0,0,0}\put(4501,-2161){\framebox(1800,1350){}}
}%
{\color[rgb]{0,0,0}\multiput(6601,-1261)(7.89474,-7.89474){20}{\makebox(2.2222,15.5556){\tiny.}}
\put(6751,-1411){\line( 0,-1){150}}
\multiput(6751,-1561)(-7.89474,-7.89474){20}{\makebox(2.2222,15.5556){\tiny.}}
}%
{\color[rgb]{0,0,0}\multiput(7201,-1261)(7.89474,-7.89474){20}{\makebox(2.2222,15.5556){\tiny.}}
\put(7351,-1411){\line( 0,-1){600}}
\multiput(7351,-2011)(-7.89474,-7.89474){20}{\makebox(2.2222,15.5556){\tiny.}}
}%
{\color[rgb]{0,0,0}\multiput(7801,-811)(7.89474,-7.89474){20}{\makebox(2.2222,15.5556){\tiny.}}
\put(7951,-961){\line( 0,-1){1050}}
\multiput(7951,-2011)(-7.89474,-7.89474){20}{\makebox(2.2222,15.5556){\tiny.}}
}%
\put(4876,-1111){\makebox(0,0)[b]{\smash{\fontsize{9}{10.8}
\usefont{T1}{cmr}{m}{n}{\color[rgb]{0,0,0}$\neg P$}%
}}}
\put(4876,-1561){\makebox(0,0)[b]{\smash{\fontsize{9}{10.8}
\usefont{T1}{cmr}{m}{n}{\color[rgb]{0,0,0}$\neg P$}%
}}}
\put(5401,-1561){\makebox(0,0)[b]{\smash{\fontsize{9}{10.8}
\usefont{T1}{cmr}{m}{n}{\color[rgb]{0,0,0}$PA$}%
}}}
\put(5926,-1561){\makebox(0,0)[b]{\smash{\fontsize{9}{10.8}
\usefont{T1}{cmr}{m}{n}{\color[rgb]{0,0,0}$P\neg A$}%
}}}
\put(4876,-2011){\makebox(0,0)[b]{\smash{\fontsize{9}{10.8}
\usefont{T1}{cmr}{m}{n}{\color[rgb]{0,0,0}$\neg P$}%
}}}
\put(5401,-2011){\makebox(0,0)[b]{\smash{\fontsize{9}{10.8}
\usefont{T1}{cmr}{m}{n}{\color[rgb]{0,0,0}$PA$}%
}}}
\put(6751,-1561){\makebox(0,0)[lb]{\smash{\fontsize{9}{10.8}
\usefont{T1}{cmr}{m}{n}{\color[rgb]{0,0,0}$L$}%
}}}
\put(7351,-1711){\makebox(0,0)[lb]{\smash{\fontsize{9}{10.8}
\usefont{T1}{cmr}{m}{n}{\color[rgb]{0,0,0}$L'$}%
}}}
\put(7951,-1561){\makebox(0,0)[lb]{\smash{\fontsize{9}{10.8}
\usefont{T1}{cmr}{m}{n}{\color[rgb]{0,0,0}$\true$}%
}}}
\end{picture}%
\nop{
+-----------+               \                        .
| -P        |               |
+-----------+  \     \      |
| -P PA P-A |  | L   |      | true
+-----------+  /     | L'   |
| -P PA     |        |      |
+-----------+        /      /
}
\label{figure-strict-naive}
\hcaption
{Strenght of models in natural, uncontingent and lexicographic revision.}
\end{hfigure}

Natural revision does not decrease strengths. Line-down revision only changes
strengths by decreasing that of models $I$ of $\neg PL$, and decrease them by
models $J$ of $\min(PA)$; the result is in Figure~\ref{figure-strict-down}.
Suffice a case where these formulae contain at least a model each. This implies
that $PA$ contains a model as well since otherwise $L$ is empty.

\begin{hfigure}
%
%
\setlength{\unitlength}{3750sp}%
\begin{picture}(2277,474)(4489,-1723)
\thinlines
{\color[rgb]{0,0,0}\multiput(6601,-1261)(7.89474,-7.89474){20}{\makebox(2.2222,15.5556){\tiny.}}
\put(6751,-1411){\line( 0,-1){150}}
\multiput(6751,-1561)(-7.89474,-7.89474){20}{\makebox(2.2222,15.5556){\tiny.}}
}%
{\color[rgb]{0,0,0}\put(4501,-1711){\framebox(1800,450){}}
}%
\put(4876,-1561){\makebox(0,0)[b]{\smash{\fontsize{9}{10.8}
\usefont{T1}{cmr}{m}{n}{\color[rgb]{0,0,0}$\neg P$}%
}}}
\put(5401,-1561){\makebox(0,0)[b]{\smash{\fontsize{9}{10.8}
\usefont{T1}{cmr}{m}{n}{\color[rgb]{0,0,0}$PA$}%
}}}
\put(5926,-1561){\makebox(0,0)[b]{\smash{\fontsize{9}{10.8}
\usefont{T1}{cmr}{m}{n}{\color[rgb]{0,0,0}$P\neg A$}%
}}}
\put(6751,-1561){\makebox(0,0)[lb]{\smash{\fontsize{9}{10.8}
\usefont{T1}{cmr}{m}{n}{\color[rgb]{0,0,0}$L$}%
}}}
\end{picture}%
\nop{
+-----------+  \                                      .
| -P PA P-A |  | L
+-----------+  /
}
\label{figure-strict-down}
\hcaption{Strength of models in line-down revision.}
\end{hfigure}~\qed

\begin{figure}[t]
\begin{center}
\begin{tabular}{ccc}
Amount of change
& \hspace{3cm} &
Naivety
\\
\setlength{\unitlength}{3750sp}%
\begin{picture}(1557,2024)(4909,-5305)
\thinlines
{\color[rgb]{0,0,0}\put(4951,-5011){\vector( 0, 1){600}}
}%
{\color[rgb]{0,0,0}\put(4951,-4111){\vector( 0, 1){600}}
}%
\put(4951,-4336){\makebox(0,0)[b]{\smash{\fontsize{9}{10.8}
\usefont{T1}{cmr}{m}{n}{\color[rgb]{0,0,0}$uncontingent$}%
}}}
\put(4951,-5236){\makebox(0,0)[b]{\smash{\fontsize{9}{10.8}
\usefont{T1}{cmr}{m}{n}{\color[rgb]{0,0,0}$lexicographic$}%
}}}
\put(4951,-3436){\makebox(0,0)[b]{\smash{\fontsize{9}{10.8}
\usefont{T1}{cmr}{m}{n}{\color[rgb]{0,0,0}$natural$}%
}}}
\put(6451,-3436){\makebox(0,0)[b]{\smash{\fontsize{9}{10.8}
\usefont{T1}{cmr}{m}{n}{\color[rgb]{0,0,0}$down$}%
}}}
\end{picture}%
\nop{
  natural       line-down
     ^
     |
     |
uncontingent
     ^
     |
     |
lexicographic
}
& &
\setlength{\unitlength}{3750sp}%
\begin{picture}(84,2875)(4909,-6151)
\thinlines
{\color[rgb]{0,0,0}\put(4951,-5011){\vector( 0, 1){600}}
}%
{\color[rgb]{0,0,0}\put(4951,-4111){\vector( 0, 1){600}}
}%
{\color[rgb]{0,0,0}\put(4951,-5911){\vector( 0, 1){600}}
}%
\put(4951,-4336){\makebox(0,0)[b]{\smash{\fontsize{9}{10.8}
\usefont{T1}{cmr}{m}{n}{\color[rgb]{0,0,0}$uncontingent$}%
}}}
\put(4951,-3436){\makebox(0,0)[b]{\smash{\fontsize{9}{10.8}
\usefont{T1}{cmr}{m}{n}{\color[rgb]{0,0,0}$lexicographic$}%
}}}
\put(4951,-5236){\makebox(0,0)[b]{\smash{\fontsize{9}{10.8}
\usefont{T1}{cmr}{m}{n}{\color[rgb]{0,0,0}$natural$}%
}}}
\put(4951,-6136){\makebox(0,0)[b]{\smash{\fontsize{9}{10.8}
\usefont{T1}{cmr}{m}{n}{\color[rgb]{0,0,0}$down$}%
}}}
\end{picture}%
\nop{
lexicographic
      ^
      |
      |
uncontingent
      ^
      |
      |
   natural
      ^
      |
      |
  line-down
}
\end{tabular}
\caption{How revisions compare on amount of change and naivety}
\label{comparisons}
\end{center}
\end{figure}

\bibliographystyle{alpha}

\begin{thebibliography}{BCD{\etalchar{+}}93}

\bibitem[AGM85]{alch-gard-maki-85}
C.~E. Alchourr\'on, P.~G{\"a}rdenfors, and D.~Makinson.
\newblock On the logic of theory change: Partial meet contraction and revision
  functions.
\newblock {\em Journal of Symbolic Logic}, 50:510--530, 1985.

\bibitem[APW19]{arav-etal-19}
T.I. Aravanis, P.~Peppas, and M.-A. Williams.
\newblock Observations on darwiche and pearl's approach for iterated belief
  revision.
\newblock In Sarit Kraus, editor, {\em Proceedings of the Twenty-Eighth
  International Joint Conference on Artificial Intelligence (IJCAI~2019)},
  pages 1509--1515. ijcai.org, 2019.

\bibitem[BCD{\etalchar{+}}93]{benf-etal-93}
S.~Benferhat, C.~Cayrol, D.~Dubois, J.~Lang, and H.~Prade.
\newblock Inconsistency management and prioritized syntax-based entailment.
\newblock In {\em Proceedings of the Thirteenth International Joint Conference
  on Artificial Intelligence (IJCAI'93)}, pages 640--647. Morgan Kaufmann, Los
  Altos, 1993.

\bibitem[BG93]{bout-gold-93}
C.~Boutilier and M.~Goldszmidt.
\newblock Revision by conditional beliefs.
\newblock In {\em Proceedings of the Eleventh National Conference on Artificial
  Intelligence (AAAI'93)}, pages 649--654. {AAAI} Press/The {MIT} Press, 1993.

\bibitem[Bou96]{bout-96-a}
C.~Boutilier.
\newblock Iterated revision and minimal change of conditional beliefs.
\newblock {\em Journal of Philosophical Logic}, 25(3):263--305, 1996.

\bibitem[CB20]{chan-boot-20}
J.~Chandler and R.~Booth.
\newblock Revision by conditionals: From hook to arrow.
\newblock In {\em Proceedings of the Seventeenth International Conference on
  Principles of Knowledge Representation and Reasoning, (KR~2020)}, pages
  233--242, 2020.

\bibitem[DP97]{darw-pear-97}
A.~Darwiche and J.~Pearl.
\newblock On the logic of iterated belief revision.
\newblock {\em Artificial Intelligence Journal}, 89(1--2):1--29, 1997.

\bibitem[DTHS13]{delg-etal-13}
J.P. Delgrande, Schaub T., Tompits H., and Woltran S.
\newblock A model-theoretic approach to belief change in answer set
  programming.
\newblock {\em {ACM} Transactions on Computational Logic}, 14(2):14:1--14:46,
  2013.

\bibitem[EG96]{eite-gott-96}
T.~Eiter and G.~Gottlob.
\newblock The complexity of nested counterfactuals and iterated knowledge base
  revisions.
\newblock {\em Journal of Computer and System Sciences}, 53(3):497--512, 1996.

\bibitem[FR98]{ferm-rodr-98}
E.~Ferm{\'e} and R.~Rodriguez.
\newblock A brief note about {R}ott contraction.
\newblock {\em Journal of the Interest Group in Pure and Applied Logic},
  6(6):835--842, 1998.

\bibitem[Han92]{hans-92}
S.O. Hansson.
\newblock In defense of the {R}amsey test.
\newblock {\em Journal of Philosophy}, 89:522--540, 1992.

\bibitem[Han99]{hans-99}
S.O. Hansson.
\newblock A survey of non-prioritized belief revision.
\newblock {\em Erkenntnis}, 50:413--427, 1999.

\bibitem[JT07]{jin-thie-07}
Y.~Jin and M.~Thielscher.
\newblock Iterated belief revision, revised.
\newblock {\em Artificial Intelligence}, 171(1):1--18, 2007.

\bibitem[KB20]{komo-beie-20}
C.~Komo and C.~Beierle.
\newblock Nonmonotonic inferences with qualitative conditionals based on
  preferred structures on worlds.
\newblock In {\em KI-2020: Advances in Artificial Intelligence - Forty-Third
  German Conference on AI}, volume 12325, pages 102--115. Springer, 2020.

\bibitem[Ker99]{kern-99}
G.~Kern{-}Isberner.
\newblock Postulates for conditional belief revision.
\newblock In {\em Proceedings of the Sixteenth International Joint Conference
  on Artificial Intelligence (IJCAI'99)}, pages 186--191, 1999.

\bibitem[Ker02a]{kern-02}
G.~Kern{-}Isberner.
\newblock Handling conditionals adequately in uncertain reasoning and belief
  revision.
\newblock {\em Journal of Applied Non-Classical Logics}, 12(2):215--237, 2002.

\bibitem[Ker02b]{kern-02-a}
G.~Kern{-}Isberner.
\newblock The principle of conditional preservation in belief revision.
\newblock In {\em Proceedings of the Second International Symposium on
  Foundations of Information and Knowledge Systems (FoIKS~2002)}, volume 2284,
  pages 105--129. Springer, 2002.

\bibitem[Ker04]{kern-04}
G.~Kern{-}Isberner.
\newblock A thorough axiomatization of a principle of conditional preservation
  in belief revision.
\newblock {\em Annals of Mathematics and Artificial Intelligence},
  40(1-2):127--164, 2004.

\bibitem[Key21]{keyn-21}
J.M. Keynes.
\newblock {\em A Treatise on Probability}.
\newblock Macmillan and Company, 1921.

\bibitem[KR10]{kern-ritt-10}
G.~Kern{-}Isberner and M.~Ritterskamp.
\newblock Preference fusion for default reasoning beyond {S}ystem {Z}.
\newblock {\em Journal of Automated Reasoning}, 45(1):3--19, 2010.

\bibitem[KSB23]{kern-etal-23}
G.~Kern{-}Isberner, M.~Sezgin, and C.~Beierle.
\newblock A kinematics principle for iterated revision.
\newblock {\em Artificial Intelligence}, 314, 2023.

\bibitem[Lib97]{libe-97-c}
P.~Liberatore.
\newblock The complexity of iterated belief revision.
\newblock In {\em Proceedings of the Sixth International Conference on Database
  Theory (ICDT'97)}, pages 276--290, 1997.

\bibitem[Lib22]{libe-22}
P.~Liberatore.
\newblock Belief merging in absence of reliability information.
\newblock {\em Synthese}, 200(4), 2022.

\bibitem[Lib23]{libe-23}
P.~Liberatore.
\newblock Mixed iterated revisions: Rationale, algorithms and complexity.
\newblock {\em {ACM} Transactions on Computational Logic}, 24(3), 2023.

\bibitem[LM92]{lehm-magi-92}
D.~Lehmann and M.~Magidor.
\newblock What does a conditional knowledge base entail?
\newblock {\em Artificial Intelligence}, 55:1--60, 1992.

\bibitem[Mak87]{maki-87}
D.~Makinson.
\newblock On the status of the postulate of recovery in the logic of theory
  change.
\newblock {\em Journal of Philosophical Logic}, 16(4):383--394, 1987.

\bibitem[NPFP96]{naya-etal-96}
A.C. Nayak, M.~Pagnucco, N.Y. Foo, and P.~Peppas.
\newblock Learning from conditionals: {J}udy {B}enjamin's other problems.
\newblock In {\em Proceedings of the Twelfth European Conference on Artificial
  Intelligence (ECAI'96)}, pages 75--79. John Wiley \& Sons, 1996.

\bibitem[NPP03]{naya-etal-03}
A.C. Nayak, M.~Pagnucco, and Peppas P.
\newblock Dynamic belief revision operators.
\newblock {\em Artificial Intelligence}, 146(2):193--228, 2003.

\bibitem[Par99]{pari-99}
R.~Parikh.
\newblock Beliefs, belief revision, and splitting languages.
\newblock {\em Logic, language and computation}, 2(96):266--268, 1999.

\bibitem[Pea90]{pear-90}
J.~Pearl.
\newblock System {Z:} {A} natural ordering of defaults with tractable
  applications to nonmonotonic reasoning.
\newblock In {\em Proceedings of the Third Conference on Theoretical Aspects of
  Reasoning about Knowledge (TARK'90)}, pages 121--135. Morgan Kaufmann, Los
  Altos, 1990.

\bibitem[RP99]{rott-pagn-99}
H.~Rott and M.~Pagnucco.
\newblock Severe withdrawal (and recovery).
\newblock {\em Journal of Philosophical Logic}, 28(5):501--547, 1999.

\bibitem[Sat88]{sato-88}
K.~Satoh.
\newblock Nonmonotonic reasoning by minimal belief revision.
\newblock In {\em Proceedings of the International Conference on Fifth
  Generation Computer Systems (FGCS'88)}, pages 455--462, 1988.

\bibitem[Sha07]{shac-07}
N.~Shackel.
\newblock Bertrand’s paradox and the principle of indifference.
\newblock {\em Philosophy of Science}, 74(2):150--175, 2007.

\bibitem[SKPP22]{schw-etal-22}
N.~Schwind, S.~Konieczny, and R.~Pino~P{\'{e}}rez.
\newblock On the representation of darwiche and pearl's epistemic states for
  iterated belief revision.
\newblock In Gabriele Kern{-}Isberner, Gerhard Lakemeyer, and Thomas Meyer,
  editors, {\em Proceedings of the Nineteenth International Conference on
  Principles of Knowledge Representation and Reasoning, (KR~2022)}, 2022.

\bibitem[Spo88]{spoh-88}
W.~Spohn.
\newblock Ordinal conditional functions: A dynamic theory of epistemic states.
\newblock In {\em Causation in Decision, Belief Change, and Statistics}, pages
  105--134. Kluwer Academics, 1988.

\end{thebibliography}
\newcommand{\etalchar}[1]{$^{#1}$}

\end{document}